\documentclass[3p,review]{elsarticle}

\usepackage{times}

\usepackage{soul}
\usepackage{url}
\usepackage[utf8]{inputenc}
\usepackage[small]{caption}
\usepackage{subcaption}
\usepackage{graphicx}
\usepackage{amsmath}
\usepackage{booktabs}
\usepackage{listings}
\usepackage{xcolor}
\usepackage{todonotes}
\usepackage{adjustbox}
\usepackage{float}
\usepackage{multirow}
\usepackage{setspace}

\usepackage[toc,page,header]{appendix}
\usepackage{minitoc}

\journal{ }

\begin{document}

\doparttoc 
\faketableofcontents

\begin{frontmatter}

\title{Current Trends in Deep Learning for Earth Observation:\\An Open-source Benchmark Arena for Image Classification}






\author{Ivica Dimitrovski$^{a,b}$, Ivan Kitanovski$^{a,b}$, Dragi Kocev$^{a,c}$, Nikola Simidjievski$^{a,c,d}$\\
\vspace{5pt}
$^{a}${\small \textit{Bias Variance Labs, d.o.o., Ljubljana, Slovenia} }\\
$^{b}${\small \textit{Faculty of Computer Science and Engineering, University Ss Cyril and Methodius, Skopje, N. Macedonia}}\\
$^{c}${\small \textit{Department of Knowledge Technologies, Jo\v{z}ef Stefan Institute, Ljubljana, Slovenia}}\\
$^{d}${\small \textit{Department of Computer Science and Technology, University of Cambridge, Cambridge,United Kingdom}}\\
$^{ }$ {\footnotesize Correspondence to: [ivica,ivan,dragi,nikola]@bvlabs.ai}
}




\begin{abstract}
We present \emph{AiTLAS: Benchmark Arena} -- an open-source benchmark suite for evaluating state-of-the-art deep learning approaches for image classification in Earth Observation (EO). To this end, we present a comprehensive comparative analysis of more than 500 models derived from ten different state-of-the-art architectures and compare them to a variety of multi-class and multi-label classification tasks from 22 datasets with different sizes and properties. In addition to models trained entirely on these datasets, we benchmark models trained in the context of transfer learning, leveraging pre-trained model variants, as it is typically performed in practice. All presented approaches are general and can be easily extended to many other remote sensing image classification tasks not considered in this study. To ensure reproducibility and facilitate better usability and further developments, \emph{all of the experimental resources} including the trained models, model configurations, and processing details of the datasets (with their corresponding splits used for training and evaluating the models) are \textit{publicly available on the repository}: \url{https://github.com/biasvariancelabs/aitlas-arena}.

\end{abstract}


\end{frontmatter}


\section{Introduction}

Recent trends in machine learning (ML) have ushered in a new era of image-data analyses, repeatedly achieving great performance across a variety of computer-vision tasks in different domains~\citep{Khan2020,KhanTransformers2022}. Deep learning (DL) approaches have been at the forefront of these efforts -- leveraging novel, modular and scalable deep neural network (DNN) architectures able to process large amounts of data. The inherent capabilities of these approaches also extend to various areas of remote sensing, in particular Earth Observation (EO), employed for analyzing different types of large-scale satellite data \citep{Ball17:jrnl}. Many of these contributions are instances of image-scene classification, such as land-use and/or land-cover (LULC) identification tasks, focusing on image-scene analyses, characterizations, and classifications of changes in the landscape caused either by human activities or by the elements.

Historically, from the perspective of ML, many of these tasks have been addressed mostly through the paradigms of either pixel-level \citep{Tuia2009,Li2014} or object-level classification tasks \citep{Blaschke2010ObjectBI}. The former refers to classification tasks focusing on each pixel in the image, associating it with the appropriate semantic label. Such approaches typically do not scale well on high-resolution images, but more importantly, many times struggle to capture more high-level patterns in the image that can span over many pixels \citep{Blaschke2001WhatsWW}. The latter, object-level classification methods, focus on analyzing distinguishable and meaningful objects in the image (as a collection of pixels) rather than independent pixels. This generally allows for better scalability and performance; however, such approaches may struggle with images containing more diverse and hardly-distinguishable objects, which prevail in most high-resolution remote-sensing data. Methods based on pixel-level and object-level paradigms have shown decent performance and are still actively researched, mostly as instances of image segmentation and object detection tasks. More recently, however, methods based on a new paradigm of scene-level classification ~\citep{Cheng2017_RSSurvey,yang2010uc_merced} have shown significant performance improvements, focusing on learning semantically meaningful representations of more sophisticated patterns in an image by leveraging the capabilities of deep learning.

Deep learning approaches have been successfully applied in various remote-sensing scenarios, be it learning models from scratch or via transfer learning\citep{Marmanis2016,Chen2019}, in a fully supervised or self-supervised setting \citep{Castillo-Navarro2022,Wang2022SSL}, exploiting the heterogeneity~\citep{Neumann2020} and temporal properties~\citep{Ienco2017} of the available data. As a result, this synergy of accurate DL approaches, on the one hand, and accessible high-resolution aerial/satellite imagery, on the other, has led to important contributions in various domains ranging from agriculture~\citep{CHLINGARYAN201861,JOHNSON201674,XU2021_crops}, ecology~\citep{Ayhan2020,Jo2018}, geology~\citep{Shirmard2022} and meteorology~\citep{ijerph15051032,Mojtaba,Chen2019} to urban mapping/planning\citep{Longbotham,Lv,HUANG201873} and archaeology~\citep{Somrak2020}. 

Nevertheless, most of these efforts typically focus on very narrow tasks stemming from domain-specific and/or spatially constrained datasets. As a result, models have been evaluated in different settings and under different conditions~\citep{Cheng2020} -- hardly reproducible and comparable. These persistent challenges, akin to a lack of standardized and consistent validation and evaluation of novel approaches, have also been identified by the community~\citep{Schneider2022}. Citing the lack of available documentation on the design and evaluation of the employed machine learning approaches, the community highlights the urgent need for standardized benchmarks that will not only enable proper and fair model comparison across datasets and similar tasks but will also facilitate faster progress in designing better and more accurate modeling approaches.

Motivated by this, in this work, we introduce \emph{AiTLAS: Benchmark Arena} -- an \emph{open-source EO benchmark suite} for evaluating state-of-the-art DL approaches for EO image classification. To this end, we present extensive comparative analyses of models derived from ten different state-of-the-art architectures, comparing them on a variety of multi-class and multi-label classification tasks from 22 datasets with different sizes and properties. We benchmark models trained from scratch and in the context of transfer learning, leveraging pre-trained model variants as it is typically performed in practice. While in this work, we mainly focus on EO-image classification tasks, such as LULC, all presented approaches are general and easily extendable to other remote-sensing image classification tasks. More importantly, to ensure reproducibility, facilitate better usability, and further exploitation of the results from our work, we provide \emph{all of the experimental resources} - freely available on our repository\footnote{\url{https://github.com/biasvariancelabs/aitlas-arena}}. The repository includes the complete study details, such as the trained models, model parameters, train/evaluation configurations, and measured performance scores, as well as the details on all of the datasets and their prepossessed versions (with the appropriate train/validation/test splits) used for training and evaluating the models.


To our knowledge, we present a unique systematic review and evaluation of different state-of-the-art DL methods in the context of EO image classification across many classification problems -- benchmarked in the same conditions and using the same hardware. Related efforts, while relevant, have mainly focused on evaluating approaches on particular datasets~\citep{Cheng2017_RSSurvey,Cheng2020,papoutsis2022efficient,xia2017aid}; evaluating different aspects of method-design~\citep{Zhai2019ALS,Neumann2020} relevant to remote-sensing classification tasks; or providing a more general overview of the common tasks at hand~\citep{Zhang2016,Zhu2017}. In particular, \citet{Cheng2017_RSSurvey} introduce a dataset and surveys several ML representation-learning approaches commonly used for remote-sensing classification tasks, comparing their performance when combined with traditional convolutional neural network (CNN) architectures. \citet{xia2017aid} also introduce a benchmark dataset for aerial-image classification, providing a comparison similar to \citet{Cheng2017_RSSurvey} of representation-learning approaches combined with three deep networks. Another, more recent study~\citep{Cheng2020}, discusses and compares more recent DL approaches and surveys several applications on three different datasets. In particular, the authors showcase the performance of the different methods for each dataset, as reported in the respective papers. The underlying, persistent conclusions from these studies show that model performances are associated with a particular dataset and study design, presenting difficulties for fair and general model comparisons. This is expected, but in our work, we seek to remedy this issue by training and evaluating all models under the same conditions.

In this context, our work is related to one of \citet{Zhai2019ALS}, which presents a large-scale study on more recent representation-learning approaches, benchmarking different aspects of method design and model parameters. However, \citet{Zhai2019ALS} considers a relatively broad scope of different datasets with only a few relevant to remote-sensing and LULC classification. \citet{Neumann2020} present a large-scale study on five different benchmark datasets; however, they investigate the effect of transfer learning on these tasks. More specifically, they evaluate different variants of the same model architecture, trained under different circumstances, rather than comparing different model architectures. Another related study by \citet{Stewart2021} reports on the comparison of different variants of ResNets on EO-image classification tasks from four datasets. More recently, and arguably most related to our work in terms of the number of evaluated models, \citet{papoutsis2022efficient} present an extensive empirical evaluation of different state-of-the-art DL architectures suitable for EO-image classification tasks, specifically LULC tasks, focusing exclusively on the BigEarthNet \citep{Sumbul2021BigearthnetAL} dataset. Namely, the authors benchmark different classes of model architectures across different criteria and introduce an efficient and well-performing model tailored specifically for BigEarthNet. \looseness=-1

\begin{figure*}[!tb]
    \centering
    \begin{subfigure}[t]{0.50\linewidth}
    \includegraphics[width=\linewidth]{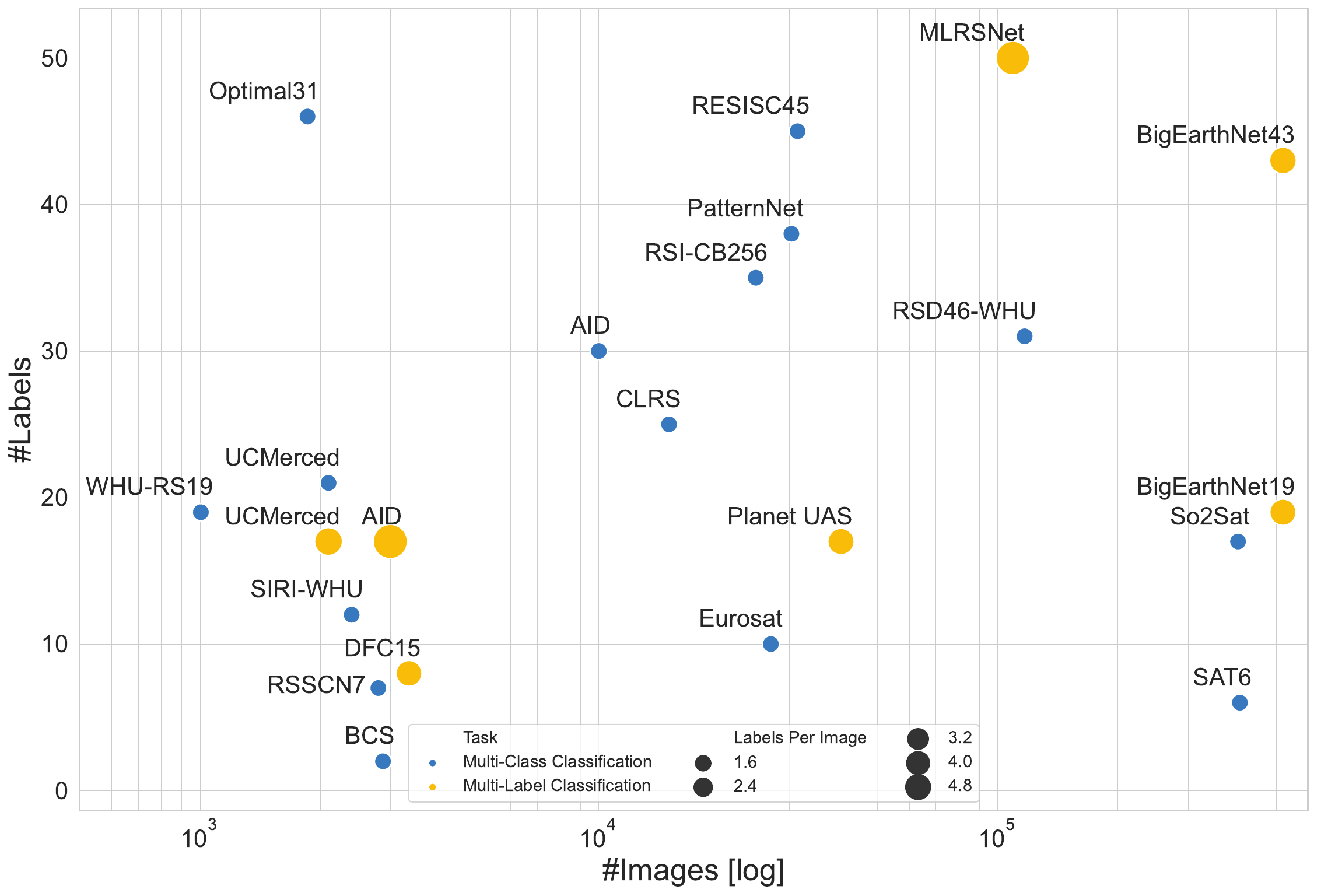}
    \subcaption{}
    \label{fig:datasets_statistics}
    \end{subfigure}
    \hfill
    \begin{subfigure}[t]{0.49\linewidth}
    \includegraphics[width=\linewidth]{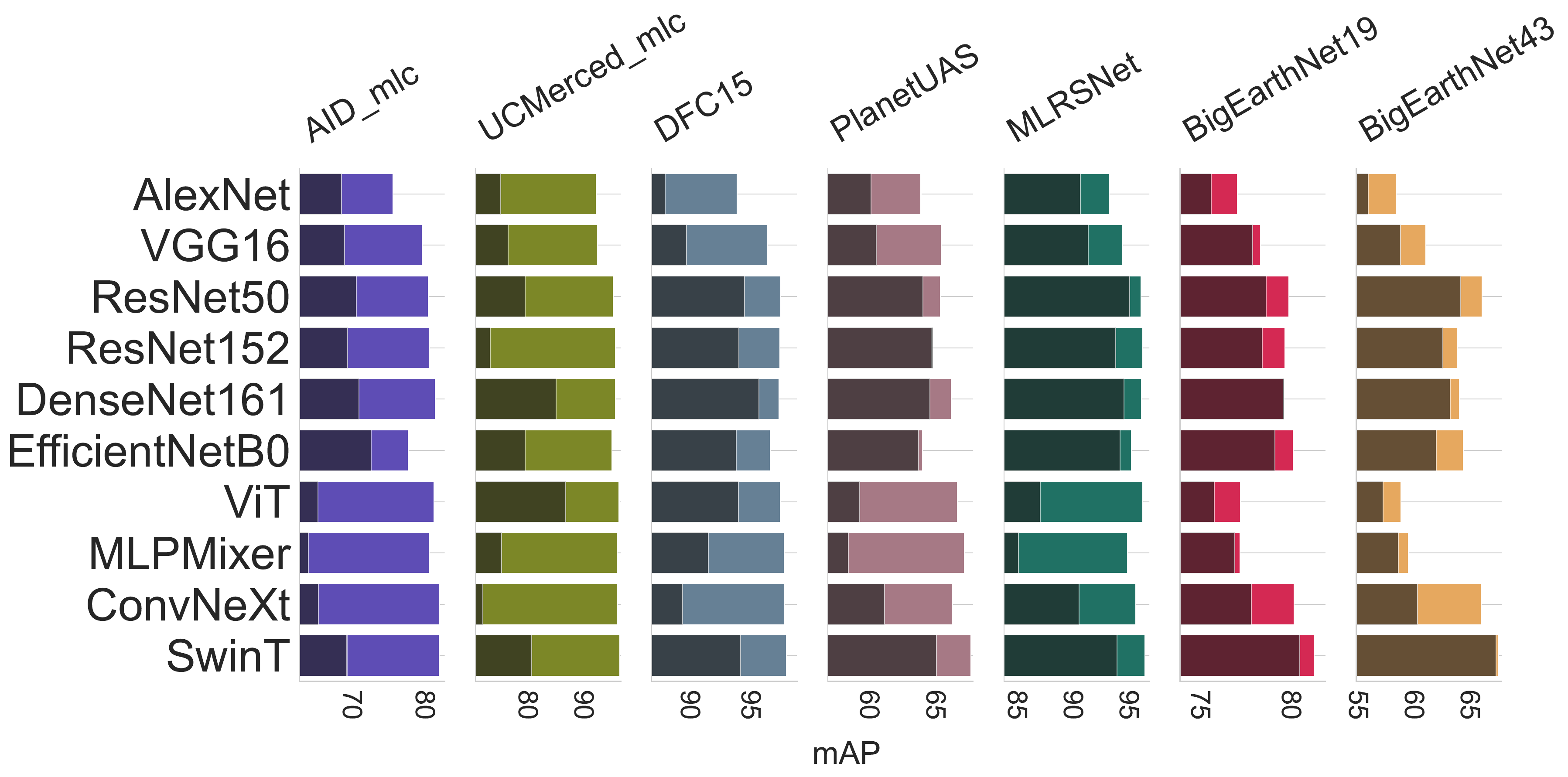}
    \subcaption{}
    \label{fig:comparisonMLC}
    \end{subfigure}
    \begin{subfigure}[t]{\linewidth}
    \includegraphics[width=\linewidth]{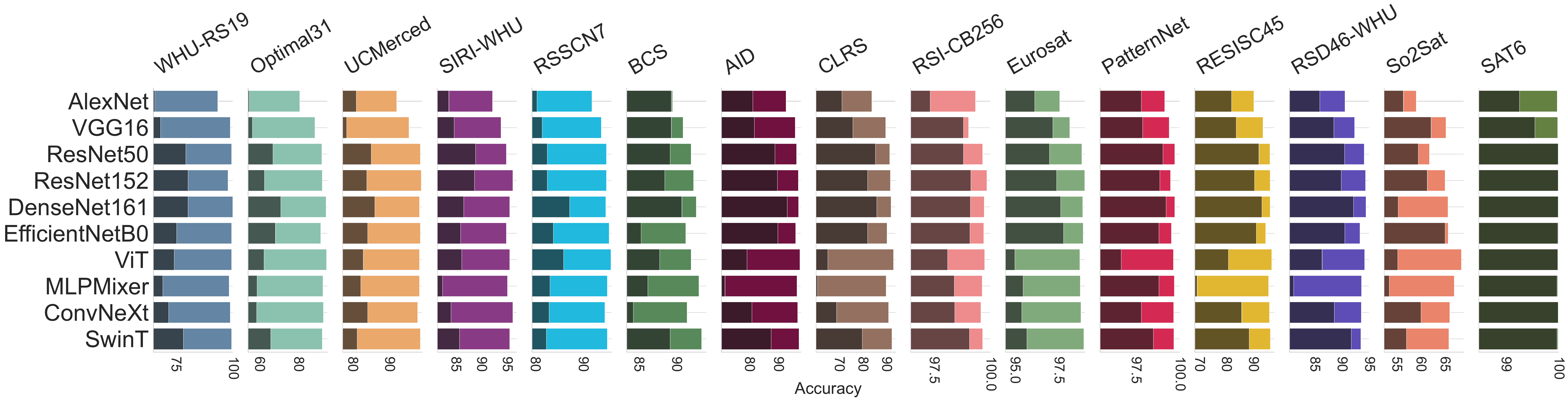}
    \subcaption{}
    \label{fig:comparisonMCC}
    \end{subfigure}
     
    \caption{\textbf{Overview of the study}: We benchmarked more than 500 models from 10 different model architectures on tasks from \textbf{(a)} 22 datasets with different sizes and properties; comparing them on \textbf{(b)} multi-label and \textbf{(c)} multi-class classification tasks. We evaluate two versions of each model architecture: (i) trained from scratch (denoted with \textit{darker shading}) and (ii) pre-trained on ImageNet-1K (denoted with \textit{lighter shading}). Note the varying scales in (b) and (c), made purposely for better visibility. Detailed results are presented in Section~\ref{section:results} and Appendices ~\ref{appendix:times},~\ref{appendix:model_generalization_results} and ~\ref{appendix:data}.}
    \label{fig:overivew}
\end{figure*}

In this work, we go beyond all the aforementioned studies, significantly extending the scope of research in two directions: the number of model architectures (and model variants) being evaluated and the datasets being considered. This results in assessing more than 500 different models with different architectures, varying designs, and learning paradigms across 22 datasets. We provide essential study-design principles and model training details that will aid in more systematic and rigorous experiments in future work. The proposed \emph{AiTLAS: Benchmark Arena} builds on the AiTLAS toolbox \citep{AitlasArxiv}\footnote{\url{https://aitlas.bvlabs.ai}} -- a recent open-source library for exploratory and predictive analysis of satellite imaginary pertaining to different remote-sensing tasks. AiTLAS implements various methods and libraries for data handling, processing, and analysis, with PyTorch \citep{pytorch} as a backbone for constructing and learning DL models. By having all of the methods and datasets under the same umbrella, we provide the means for a fair, unbiased, and reproducible comparison of approaches across different criteria that include: overall model performance, data- and task-dependent model performance, model size, and learning efficiency as well as the effect of transfer learning via model pre-training.  

The results, summarized in Figure~\ref{fig:overivew}, show that many of the current state-of-the-art architectures for vision tasks can lead to decent predictive performance when applied to EO image classification tasks. While, in some cases, training models from scratch can lead to satisfactory performance, using pre-trained models and fine-tuning them on each dataset leads to the best performance overall. We observed this in all cases, regardless of the dataset properties, the type of classification tasks, or the model architecture. We found more considerable performance gains on tasks from smaller datasets, which, as expected, benefited more from the pre-training process than models trained on larger datasets. In terms of model architectures, our experiments showed that pre-trained Transformer models, i.e. both Vision Transformer~\citep{dosovitskiy2020image} and Swin Transformer~\citep{ze2022swinv2} models, were, in general, able to achieve the best performance. Specifically, Vision Transformer models showed the best performance on various multi-classification tasks, while Swin Transformer models led to much better performance on multi-label tasks, albeit at the cost of much longer training time. Throughout the paper, we further evidence and discuss these findings. 

In summary, in this paper, we make several contributions. Specifically, we:
\begin{itemize}
 \itemsep0em 
	\item Introduce \emph{AiTLAS: Benchmark Arena} -- an open-source benchmark suite that enables standardized evaluation of machine learning models for Earth Observation (EO) applications;
	\item Provide study-design principles for training and evaluating state-of-the-art deep learning models on various supervised EO image classification tasks from 22 datasets with different sizes and properties;
	\item Implement and benchmark more than 500 models stemming from 10 state-of-the-art architectures, including models trained from scratch and their pre-trained variants;
    \item Investigate models' generalization abilities to unseen in-domain datasets;
	\item Evaluate different pre-training strategies that relate to pre-training models from in-domain EO datasets and investigate their effect on the downstream predictive performance;
	\item Discuss common issues that typically affect the models' performance, specifically in the context of EO tasks.
	\item Provide open-source access to all experimental details, including trained models, dataset details, train/evaluation configurations, and detailed performance scores.
\end{itemize}

\section{Data \& models}

\subsection{Data description} 

With the ever-growing availability of remote sensing data, there has been a significant effort by many research groups to prepare, label, and provide proper datasets that will support the development and evaluation of sophisticated machine learning methods. While there are many such datasets, both proprietary and publicly available, in this work, we focus on the latter -- open-access publicly available dataset. Given this criterion, we select 22 open-access datasets usually considered in different EO studies for benchmarking DL approaches. The selected datasets have varying sizes (number of images), varying image types, image sizes, and formats, and, more importantly, related to different classification tasks. 

Namely, we consider datasets related to multi-class and multi-label classification tasks, mainly addressing LULC applications. The objective of \emph{multi-class classification} tasks is to predict one (and only one) class (label) from a set of predefined classes for each image in a dataset. \emph{Multi-label classification}, on the other hand, refers to predicting multiple labels from a predefined set of labels for each image in the dataset \citep{tsoumakas2009} (e.g., an image can belong to more than one class simultaneously). In our experimental study, we consider 15 multi-class and seven multi-label datasets.

\begin{table*}[!ht]
\centering
\caption{Summary of the multi-class classification (MCC) datasets.}
\begin{adjustbox}{width=0.89\linewidth}
\begin{tabular}{lllllllll}
\hline
\textbf{Name} & \textbf{\rotatebox[origin=c]{90}{Image type}} & \textbf{\rotatebox[origin=c]{90}{\#Images}} & \textbf{\rotatebox[origin=c]{90}{Image size}} & \textbf{\rotatebox[origin=c]{90}{Spatial resolution}} & \textbf{\rotatebox[origin=c]{90}{\#Labels}} & \textbf{\rotatebox[origin=c]{90}{Predefined splits}} & \textbf{\rotatebox[origin=c]{90}{Image format}} &  \\
\hline
UC Merced \citep{yang2010uc_merced} & Aerial RGB & 2100    & 256×256 & 0.3m    & 21  & No      & tif     &  \\
\hline
WHU-RS19 \citep{xia2010whurs19}  & Aerial RGB & 1005    & 600×600 & 0.5m    & 19  & No      & jpg     &  \\
\hline
AID \citep{xia2017aid}  & Aerial RGB & 10000   & 600×600 & 0.5m - 8m    & 30  & No      & jpg     &  \\
\hline
Eurosat \citep{helber2019eurosat}   & Sat. Multispectral      & 27000   & 64×64 & 10m      & 10  & No      & jpg/tif      &  \\
\hline
PatternNet \citep{zhou2018patternnet} & Aerial RGB & 30400   & 256×256 & 0.06m - 4.69m   & 38  & No      & jpg     &  \\
\hline
Resisc45 \citep{Cheng2017_RSSurvey}  & Aerial RGB & 31500   & 256×256 & 0.2m - 30m    & 45  & No      & jpg     &  \\
\hline
RSI-CB256 \citep{haifeng2020rsicb256} & Aerial RGB & 24747   & 256×256 & 0.3 - 3m    & 35  & No      & tif     &  \\
\hline
RSSCN7 \citep{Zou2015RSSCN7}    & Aerial RGB & 2800    & 400×400 & n/a   & 7   & No      & jpg     &  \\
\hline
SAT6 \citep{Basu2015sat6} & RGB + NIR & 405000  & 28×28  & 1m    & 6   & Yes     & mat     &  \\
\hline
Siri-Whu \citep{zhu2016siriwhu}  & Aerial RGB & 2400    & 200×200 & 2m    & 12  & No      & tif     &  \\
\hline
CLRS \citep{haifeng2020clrs} & Aerial RGB & 15000   & 256×256  & 0.26m - 8.85m   & 25  & No      & tif     &  \\
\hline
RSD46-WHU \citep{Yang2017rsd46whu} & Aerial RGB & 116893  & 256×256  & 0.5m - 2m   & 46  & Yes     & jpg     & \\
\hline
Optimal 31 \citep{qi2019optimal31} & Aerial RGB & 1860  & 256×256   & n/a & 31  & No     & jpg     & \\
\hline
Brazilian Coffee Scenes (BSC) \citep{penatti2015deep} & Aerial RGB & 2876  & 64×64 & 10m   & 2  & No     & jpg     & \\
\hline
So2Sat \citep{Zhu2020So2Sat} & Sat. Multispectral & 400673  & 32×32  & 10m  & 17  & Yes     & h5     & \\
\hline
\end{tabular}
\end{adjustbox}
\label{table:multiclass}
\end{table*}

\begin{table*}[!ht]
\centering
\caption{Summary of the multi-label classification (MLC) datasets.}
\begin{adjustbox}{width=0.89\linewidth}
\begin{tabular}{lllllllll}
\hline
\textbf{Name} & \textbf{\rotatebox[origin=c]{90}{Image type}}     & \textbf{\rotatebox[origin=c]{90}{\#Images}} & \textbf{\rotatebox[origin=c]{90}{Image size}} & \textbf{\rotatebox[origin=c]{90}{Spatial resolution}} &  \textbf{\rotatebox[origin=c]{90}{\#Labels}} & \textbf{\rotatebox[origin=c]{90}{\#Labels per image}}                               & \textbf{\rotatebox[origin=c]{90}{Predefined splits}}                                  & \textbf{\rotatebox[origin=c]{90}{Image format}}                              \\
\hline
UC Merced (mlc) \citep{chaudhuri2018ucmerced}   & Aerial RGB  & 2100         & 256×256   & 0.3m       & 17    & 3.3    & No      & tif        \\
\hline
MLRSNet \citep{qi2020mlrsnet}     & Aerial RGB  & 109161       & 256×256 & 0.1m - 10m        & 60         & 5.0       & No      & jpg        \\
\hline
DFC15 \citep{hua2019dfc15}       & Aerial RGB  & 3342         & 600×600 & 0.05m         & 8   & 2.8   & Yes       & png        \\
\hline
AID (mlc)\citep{hua2020aidmultilabel}       & Aerial RGB  & 3000         & 600×600 & 0.5m - 8m        & 17    & 5.2     & Yes       & jpg        \\
\hline
PlanetUAS \citep{webplanetuas}   & Aerial RGB  & 40479       & 256×256 & 3m         & 17    & 2.9   & No & jpg/tiff     \\
BigEarthNet 19 \citep{Sumbul2021BigearthnetAL}   & Sat. Multispectral & 519284 & \shortstack[l]{20×20 \\ 60x60 \\ 120x120} & \shortstack[l]{60m \\ 20m \\ 10m} & 19         & 2.9 & Yes  & tif, json \\
\hline
BigEarthNet 43 \citep{Sumbul2019BigearthnetAL}   & Sat. Multispectral & 519284       & \shortstack[l]{20×20 \\ 60x60 \\ 120x120} & \shortstack[l]{60m \\ 20m \\ 10m} & 43         & 3.0 & Yes & tif, json \\
\hline
\end{tabular}
\end{adjustbox}
\label{table:multilabel}
\end{table*}

Tables \ref{table:multiclass} and \ref{table:multilabel} summarizes the properties of the considered multi-class (MCC) and multi-label (MLC) classification datasets, respectively. The number of images across datasets is quite diverse, ranging from datasets with $\sim2K$ images to datasets with $\sim500K$ images. This also extends toward the number of labels per image, ranging from $2$ to $60$. Figure~\ref{fig:datasets_statistics} visualizes the datasets with respect to their sizes, with the x-axis denoting the number of images (on a log scale) and the y-axis indicating the number of labels (with marker size denoting the number of labels per image) for each of the different datasets. Most of the datasets consist of Aerial RGB images (with only a few comprised of satellite multi-spectral data) that are different in spatial resolution, size, and format. Finally, we note the datasets that include predefined splits (for training, validation, and testing) given by the original authors and provide the splits for the ones that are missing, as further discussed in Section \ref{section:evaluation_protocol}. An extended description of each dataset is given in Appendix~\ref{appendix:data}. \looseness=-1

\subsection{Model architectures}\label{sec:model_arch}

Current trends in EO image classification leverage the capabilities of DL architectures for computer vision, learning data representations that often lead to superior predictive performance. We recognize that there are many different approaches stemming from different model architectures and model variants. These can differ in various 'finer' details (e.g., number and width of layers, hyper-parameter values, and learning regimes), often developed for a particular task. Rather than seeking a state-of-the-art performance for each EO problem/dataset, in this study, we are interested in providing a more general evaluation framework and benchmarking models by analyzing their characteristics and unique properties through the lens of their predictive performance and learning efficiency across all datasets. 

Therefore, our model-architecture (and parameter) choices are motivated by different architecture 'classes', such as the traditional convolutional architectures and the more recent attentional and multilayer-perceptron (MLP) architectures. This renders models with different sizes, training/inference time, different abilities in a transfer-learning setting, etc. More specifically, we investigate several architectures which have been traditionally used for EO image classification tasks, such as: AlexNet \citep{krizhevsky2012imagenet}, VGG16 \citep{simonyan2014very}, ResNet \citep{he2016deep} and DenseNet \citep{huang2017densely}. Moreover, we investigate more recent architectures, which include EfficientNet \citep{tan2019efficientnet}, ConvNeXt \citep{liu2022convnet}, Vision Transformer \citep{dosovitskiy2020image}, Swin Transformer \citep{ze2022swinv2} and MLPMixer \citep{tolstikhin2021mlp}, that have shown state-of-the-art performance in various vision tasks. In the following, we provide a brief overview of these architectures, highlighting their properties in Table \ref{table:models}.

\begin{table}[ht]
\caption{Summary of the representative model architectures considered in this study.}
\centering
\begin{adjustbox}{width=1\linewidth}
\begin{tabular}{lrrrrl}
\hline
\textbf{Model} & \textbf{Year}     & \textbf{\#Layers} & \textbf{\#Parameters} & \textbf{FLOPS} & \textbf{Based on} \\
\hline
AlexNet \citep{krizhevsky2012imagenet}     & 2012              & 8             & $\sim$ 57$\cdot10^{6}$ & 0.72~G &\citep{marcel2010torchvision}\\

VGG16 \citep{simonyan2014very}     & 2014              & 16             & $\sim$ 134.2$\cdot10^{6}$  & 15.47~G &\citep{marcel2010torchvision}\\

ResNet50 \citep{he2016deep}     & 2015              & 50             & $\sim$ 23.5$\cdot10^{6}$  &  	

4.09~G &\citep{marcel2010torchvision}\\

ResNet152 \citep{he2016deep}     & 2015              & 152             & $\sim$ 58.1$\cdot10^{6}$  & 11.52~G &\citep{marcel2010torchvision}\\

DenseNet161 \citep{huang2017densely}     & 2017              & 161             & $\sim$ 26.4$\cdot10^{6}$  & 7.73~G&\citep{marcel2010torchvision}\\

EfficientNet B0 \citep{tan2019efficientnet}     & 2019              &      237        & $\sim$ 5.2$\cdot10^{6}$  & 0.39~G &\citep{marcel2010torchvision} version: B0\\

Vision Transformer (ViT) \citep{dosovitskiy2020image}     & 2020              & 12             & $\sim$ 86.5$\cdot10^{6}$  & 17.57~G &\citep{rw2019timm} version: b\_16\_224 \\
MLPMixer \citep{tolstikhin2021mlp}     & 2021              & 12              & $\sim$ 59.8$\cdot10^{6}$  & 12.61~G &\citep{rw2019timm} version: b\_16\_224 \\
ConvNeXt \citep{liu2022convnet}     & 2022              &      174         & $\sim$ 28$\cdot10^{6}$  & 4.46~G &\citep{marcel2010torchvision} version: tiny\\
Swin Transformer \citep{ze2022swinv2}  & 2022              &     24          & $\sim$ 49.7$\cdot10^{6}$  & 11.55~G &\citep{marcel2010torchvision} version: v2 small \\
\hline
\end{tabular}
\end{adjustbox}
\label{table:models}
\end{table}

The first class of models we consider relies on convolutional architectures, which, in recent years, have driven many of the advances in computer vision. The architecture of convolutional neural networks (CNN) consists of many (hidden) layers stacked together, designed to process (image) data in the form of multiple arrays. Most typically, CNNs consist of a series of convolutional layers, which apply convolution operation (passing the data through a kernel/filter), forwarding the output to the next layer. This serves as a mechanism for constructing feature maps, with former layers typically learning low-level features (such as edges and contours), subsequently increasing the complexity of the learned features with deeper layers in the network. Convolutional layers are typically followed by pooling operations, which serve as a downsampling mechanism by aggregating the feature maps through local non-linear operations. In turn, these feature maps are fed to fully-connected layers, which perform the ML task at hand -- in this case, classification. All the layers in a network employ an activation function. In practice, the intermediate, hidden layers employ a non-linear function such as rectified linear unit (ReLU) or Gaussian Error Linear Unit (GELU) as common choices. The choice of activation function in the final layer relates to the tasks at hand, typically a sigmoid function in the case of classification. CNN architectures can also include different normalization and/or dropout operators embedded among the different layers, which can further improve the network's performance.

CNN architectures have been widely researched, with models applied in many contexts of remote sensing, and in particular EO image classification~\citep{Chen2019,Weng2017,Castelluccio2015, papoutsis2022efficient}. This includes \textit{AlexNet}~\citep{krizhevsky2012imagenet}, a pioneering architecture that introduced and successfully demonstrated the utility of the CNN blueprint, mentioned earlier, for computer vision tasks. Namely, even though the architecture of AlexNet has a modest depth (relative to more recent architectures) consisting of eight layers, it remains an efficient baseline approach for a variety of EO tasks~\citep{Cheng2017_RSSurvey,Marmanis2016}, leading to decent performance, especially when pre-trained with large image datasets~\citep{Han2017}. We also consider the more sophisticated \textit{VGG}~\citep{simonyan2014very}, which employs a deeper architecture inspired by AlexNet. VGG has shown great performance in a variety of vision tasks, including EO-image classification problems~\citep{kang2018building,Hu2015,zhou2018patternnet}. There are two variants of VGG in practice, VGG16 and VGG19; both extend AlexNet mainly by increasing the depth of the network with 13 and 16 convolutional layers, respectively. In this study, we evaluate the performance of the former \textit{VGG16}. VGGs employ kernels with smaller sizes than the ones typically used in AlexNet, demonstrating that stacking multiple smaller kernels are better able to extract more complex representations than one larger filter. While, in general, increasing the network depth by adding convolutional layers helps for learning more complex and more informative representations thereof, in practice, this can lead to several issues, such as the vanishing gradient problem~\citep{Goodfellow-et-al-2016}, which impairs the network training.

The Residual neural networks (\textit{ResNets}) \citep{he2016deep,Zagoruyko2016} tackle this issue explicitly by employing skip connections between blocks, therefore enabling better backprop gradient flow, better training, and, in general, better predictive performance. ResNet architecture follows a typical CNN blueprint: Stacking residual blocks (typically same-size CNN layers) and convolutional blocks (typically introducing a bottleneck via different-size CNN layers) together, followed by fully-connected layers. By employing skip connections, the ResNet architecture allows stacking multiple layers in a block, therefore training models with much deeper architectures. Here we investigate two variants with varying depths, \textit{ResNet50} and \textit{ResNet152}, with 50 and 152 layers, respectively. Since their inception, ResNets have been a prevalent choice in practice. This also extends towards their utility for EO tasks, applied in the context of image classification and semantic segmentation \citep{Cheng2017_RSSurvey,audebert2018beyond,Stewart2021,papoutsis2022efficient}. Dense Convolutional Networks (\textit{DenseNets})~\citep{huang2017densely} are another well-performing architecture variant of ResNets that has demonstrated state-of-the-art results on many classification tasks, including applications in the domain of remote sensing~\citep{Zhang2019,Tong2020,Chen2021}. As the name suggests, DenseNets consist of dense blocks, where each layer is connected to every preceding layer, taking an additional (channel-wise) concatenated input of the feature maps learned in the former layers. This differs from the ResNets, which propagate (element-wise) aggregated feature maps through the network layers. The architecture of DenseNets encourages feature reuse throughout the network, leading to well-performing and more compact models (with fewer trainable parameters than a ResNet of equivalent size), albeit at the cost of increased memory during training.

\textit{EfficientNets}~\citep{tan2019efficientnet} are a recent class of lightweight architecture that alleviate such common computational difficulties, typical when scaling deep architectures on larger and/or harder problems. Namely, rather than scaling the architecture in one aspect of increasing the depth (number of layers)~\citep{he2016deep}, width (number of channels)~\citep{Zagoruyko2016} or (input image) resolution~\citep{Lin2017}; EfficientNets implement compound scaling, that uniformly scales the architecture along the three dimensions simultaneously. Compound scaling seeks an optimal balance between these three dimensions, given the available resources and the task at hand. In turn, such an approach leads to substantially smaller models (than CNN variants of equivalent performance) while retaining state-of-the-art predictive performance. In the context of EO tasks, (variants of) EfficentNets have been successfully applied in different settings~\citep{liu2020light,tian2020resolution,Alhichri2021,Chen2021}, and have also been thoroughly investigated in the context of multi-label image classification tasks from BigEarthNet~\citep{papoutsis2022efficient}. While there are eight variants of EfficientNets, differing in the size and complexity of the architectures, here we investigate the performance of the baseline \textit{EfficientB0} architecture with 5.2M parameters, substantially lower than any of the other competing model architectures. Most recently, \citep{liu2022convnet} introduce \textit{ConvNeXt}, a novel class of convolutional architectures that leverage various successful design decisions of preceding convolutional and attentional architectures typically applied for vision tasks. Namely, ConvNeXt models implement various techniques at different levels: from reconfiguring details like activation functions and normalization layers, to redesigning more general architecture details related to residual/convolutional blocks, to modifications in the training strategies. This, in turn, leads to models that achieve good predictive performance, not only better than popular models from the class of convolutional architectures but also better than the more recent attentional architectures, such as transformers, discussed next. While there are several variants of the ConvNeXt architecture that mainly differ in their size, in this study, we evaluate the performance of the smallest variant, \mbox{\textit{ConvNeXt\_tiny}}. Note that, to our knowledge, this is the first application of ConvNeXt on EO-image classification tasks.

We next take the notion of the recent success of the class of attentional network architectures and study the performance of \textit{Vision Transformers} ({ViT})~\citep{dosovitskiy2020image} in the context of EO-image classification tasks. Namely, ViTs inspire by the popular NLP (natural language processing) Transformer architecture~\citep{devlin2018bert}, leveraging an attention mechanism for vision tasks. Much like the original Transformer that seeks to learn implicit relationships in sequences of word-tokens via multi-head self-attention, ViTs focus on learning such relationships between image patches. Typically they employ a standard transformer encoder that takes a lower-dimensional (linear) representation of these image patches together with additional positional embedding from each, in turn, feeding the encoder-output to a standard MLP head. ViTs have shown excellent performance on various vision tasks, particularly when combined with pre-training from large datasets. This also includes several applications in remote sensing~\citep{Bazi2021,papoutsis2022efficient,Gong2022}.

More recent and sophisticated, attentional network architectures such as the \textit{Swin Transformers} ({SwinT})~\citep{Ze2021SwinV1,ze2022swinv2} rely on additional visual inductive biases by introducing hierarchy, translation invariance, and locality in the attention mechanism. Like ViTs, SwinT architectures also attempt to learn relationships between image patches but operate on image windows (a group of neighboring image patches). SwinTs focus on computing attention between patches within a window (locality), in turn shifting these windows to allow learning of cross-window attention (translation invariance). Starting with windows with smaller patches and increasing their size at each subsequent stage, SwinTs also allow for learning representations at different granularity (hierarchy). All this leads to SwinTs performing well in practice on a variety of vision tasks, including in the domain of remote sensing \citep{Scheibenreif_SwinLC,Zhang2022SwinSUNet,Wang2022_emp}, often outperforming ViTs and other convolutional architectures. In this study, we evaluate the 'small' architecture variant of the latest version of Swin Transformers V2 \citep{ze2022swinv2}. \looseness=-1

In the context of vision tasks, an attention mechanism can be achieved differently (e.g., attending over channels and/or spatial information, etc.) and even employed with typically convolutional architectures\citep{liu2020light,Xu2021,Alhichri2021}. One alternative that builds only on the classical MLP architecture is the \textit{MLPMixer}~\citep{tolstikhin2021mlp}. Namely, similar to a transformer architecture, an MLPMixer operates on image patches; and contains two main components: A block of MLP layers for 'mixing' the spatial, patch-level information on every channel; and a block of MLP layers for 'mixing' the channel-information of an image. This renders lightweight models, with performance on par with many much more sophisticated architectures, on a variety of vision problems, both more general as well as specific EO tasks~\citep{Meng2021,papoutsis2022efficient,Gong2022}. We employ an MLPMixer with an input size of 224x224 and a patch resolution of 16×16 pixels. 

From each of the ten highlighted architectures, we evaluate two model versions: trained entirely on a given dataset and fine-tuned models that have been pre-trained on a different image dataset. This results in comparing 20 models on each predictive task, which are available on our repository.


\section{Experimental design}

\subsection{Training and evaluation protocol}
\label{section:evaluation_protocol}

To establish a unified evaluation framework and support the results' reproducibility, we generated train, validation, and test splits using 60\%, 20\%, and 20\% fractions, respectively. All of the data splits were obtained using stratified sampling. This technique ensures that the distribution of the target variable(s) among the different splits remains the same \citep{sechidis2011}. We performed such stratification for all datasets except the ones which include predefined splits provided by the original authors. More specifically, for the \emph{BigEarthNet} and \emph{So2Sat} datasets, we use the train, validation, and test splits as provided in  \citep{Sumbul2019BigearthnetAL,Sumbul2021BigearthnetAL, Zhu2020So2Sat}. Since \emph{SAT6}, \emph{RSD46-WHU}, \emph{DFC15} and \emph{AID} datasets consist only of predefined train and test splits, we further take 20\% from the train part for validation. Finally, note that the PlanetUAS dataset was part of a competition, and as such, the test data is not publicly available. Therefore, we generated train, validation, and test splits from the original train data using the 60\%, 20\%, and 20\% fractions, respectively. 

All the models were trained using the same train splits, with parameters selection/search performed using the same validation splits. Additionally, to overcome over-fitting, we perform early stopping on the validation split for each dataset; the best checkpoint/model found (with the lowest validation loss) is saved and then applied to the original test split to obtain the final assessment of the predictive performance. All the train/validation/test splits for each dataset are available on our repository.

To better assess the generalization capabilities of the trained models, we evaluate their performance on different (in-domain) datasets not used for training. Specifically, we present two schemes of this evaluation: (1) performance measured on a holdout set compiled of test images with the same labels but from different datasets; (2) an exhaustive cross-dataset evaluation between pairs of datasets that contain the same labels. The former variant refers to a new test set consisting of 3216 images from the test splits of seven datasets (\emph{RESISC45, UC Merced, CLRS, PatternNet, AID, RSI-CB256} and \emph{WHU-RS19}) with labels present in all datasets (in our experiments, this results in five common labels: 'Forest', 'Parking', 'River', 'Harbor' and 'Beach'). We employ this evaluation setting only for multi-class classification tasks. In the latter variant, in a pairwise fashion, we evaluate every model on test splits from other datasets not used for training it. We measure the performance only on images with labels shared between the pairs of source (used for training the model) and target (used for evaluating the model) datasets. We employ this setting in both multi-class and multi-label classification scenarios. Note that in all cases, the models are only evaluated on the unseen datasets without additional fine-tuning. These configurations are also available on our repository.

During training, we perform \emph{data augmentation} for each dataset by first resizing all the images to 256x256, followed by selecting a random crop of size 224x224. We then perform random horizontal and/or vertical flips. During evaluation/testing, we first resize the images to 256×256, followed by a central crop of size 224×224. We believe that this, in general, helps our models to generalize better on a given dataset. Also note that in the study, we are using only RGB images. In the case of the multispectral datasets (\emph{Eurosat}, \emph{So2Sat} and \emph{BigEarthNet}), we computed the images in the RGB color space by combining the red (B04), green (B03), and blue (B02) bands. For the \textit{Brazilian Coffee Scenes} dataset, we use images in green, red, and near-infrared spectral bands since these are most useful and representative for distinguishing vegetation areas, as suggested by the authors.

Since we train models on 22 datasets with a different number of classes, different training samples, and class distributions (as shown in Tables~\ref{table:multiclass} and ~\ref{table:multilabel}), we perform a hyperparameters search for each model and each dataset, to account for these variations. Namely, we search over different learning-rate values: $0.01$, $0.001$, and $0.0001$. We use \emph{ReduceLROnPlateau} as a learning scheduler which reduces the learning rate when the loss has stopped improving. Models often benefit from reducing the learning rate by a factor once learning stagnates. This scheduler tracks the values of the loss measure, reducing the learning rate by a given factor when there is no improvement for a certain number of epochs (denoted as `patience’). In our experiments, we track the value of the validation loss with patience set to 5 and a reduction factor set to 0.1 (the new learning rate will be $lr * factor$). The maximum number of epochs is set to $100$. Additionally, we also apply early stop criteria if no improvements in the validation loss are observed over $10$ epochs. We use fixed values for some of the hyperparameters, such as batch size, which we set to $128$. For optimization, we use \emph{RAdam optimizer}~\citep{liu2019radam} without weight decay. RAdam is a variant of the standard Adam~\citep{kingma2014adam}, with a mechanism that rectifies the variance from the adaptive learning rate. This, in turn, allows for an automated warm-up tailored to the particular dataset at hand.

For each model architecture, we train two variants: (i) models trained entirely on a given dataset and (2) fine-tuned models previously trained on a different (and larger) image dataset. The former, which we refer to as models 'trained from scratch', refer to models trained only on the dataset at hand and initialized with random weights in the training procedure. The latter leverages transfer learning via model pre-training. The next section provides further details on how we use and fine-tune these pre-trained models. All models were trained on NVIDIA A100-PCIe GPUs with 40 GB of memory running CUDA version 11.5. We used the AiTLAS toolbox~\footnote{\url{https://github.com/biasvariancelabs/aitlas}} to configure and run the experiments. All configuration files for each experiment are also available in our repository, along with the trained models. We believe this provides a standardized evaluation framework for EO image classification tasks.

\subsection{Transfer learning strategy}\label{section:transfer_learning_strategy}

In this study, we take the notion of \emph{transfer learning} as a strategy that can lead to performance improvements of vision models on image classification tasks~\citep{Zhai2019ALS}, in particular in EO domains~\citep{Risojevic2021}. In our problem setting, transfer learning allows downstream, task-specific models to leverage learned representations from model architectures pre-trained on much larger image datasets. This, in turn, often leads to (fine-tuned) models with much better generalization power using fewer training data (and training iterations), which is especially useful for tasks that stem from smaller datasets. In the case of DL models for image classification, two strategies are often used for performing transfer learning: (1) fine-tuning the model weights only for the last classifier layer or (2) fine-tuning the model weights of all layers in the network. The former approach retains the values of all but the last layer's weights of the model from the pre-training, keeping them 'frozen' during fine-tuning. The latter, on the other hand, allows the weights to change throughout the entire network during fine-tuning. In practice, this can lead to better generalization \citep{yosinski2014,kornblith2019} and higher accuracy.

In our experiments, we implement the latter approach. Starting with a pre-train model, we fine-tune each network entirely (the entire parameter set) for each specific dataset. Note that the choice of the pre-training dataset, and its relation to the domain of the downstream task, may also influence the predictive performance of the fine-tuned model~\citep{Neumann2020}. Since here we are interested in a more general evaluation that considers 22 different datasets, we evaluate a standard approach for transfer learning using pre-trained model architectures on the ImageNet-1K \citep{krizhevsky2012imagenet} dataset (version V1). More specifically, we use implementations from the PyTorch vision catalog~\citep{marcel2010torchvision} for most models, except ViT and MLPMixer, for which we base the implementations on \citep{rw2019timm}. 

Furthermore, to evaluate the effect of the pre-training dataset on the performance of the downstream model, in a set of smaller-scale experiments, we benchmark architectures that have been pre-trained using different 'in-domain' EO datasets. In particular, we evaluate two strategies: (i) models pre-trained entirely on an EO dataset and (ii) models pre-trained on both ImageNet-1K and an EO dataset. The latter relates to a two-stage pre-training strategy, where models are first pre-trained on ImageNet-1K, followed by intermediate tuning on an in-domain EO dataset, and finally, fine-tuning them on the target EO dataset. We evaluate these pre-training strategies by comparing models from two architectures (ViT and DenseNet) using four in-domain EO datasets for pre-training.

\subsection{Evaluation measures}

Evaluating the performance of machine learning models is a non-trivial task that is specific to the learning task at hand and dependent on the general objectives of the model being learned. Different evaluation metrics capture different aspects of the models' behavior and their predictive capabilities measured on image samples not used for training. Since the goal of this study analyzing the predictive performance of different DL models across different datasets on multi-class and multi-label classification tasks -- we examine the experimental work through the lens of evaluation measures most suitable for these two tasks.

More specifically, for multi-class classification tasks, we report the following measures: Accuracy, Macro Precision, Weighted Precision, Macro Recall, Weighted Recall, Macro F1 score, and Weighted F1 score. Note that, since for these tasks, the micro-averaged measures such as F1 score, Micro Precision, and Micro Recall have values equal to accuracy, we do not report them. Note that, for image classification tasks, it is customary to report \emph{top-n accuracy} (typically $n$ is set to 1 or 5)~\citep{krizhevsky2012imagenet}, where the score is computed based on the correct label being among the $n$ most probable labels outputted by the model. In this paper, we report \emph{top-1 accuracy}, denoted as 'Accuracy' unless stated otherwise. For multi-label classification tasks, we report Micro Precision, Macro Precision, Weighted Precision, Micro Recall, Macro Recall, Weighted Recall, Micro F1 score, Macro F1 score, Weighted F1 score, and mean average precision (mAP). Since all measures, but mAP, require setting a threshold on the predictions, we choose a threshold value of $0.5$ for all models and settings. Further details and definitions of the evaluation measures used in the study are given in Appendix~\ref{sec:app:measures}. We also provide additional performance details in terms of confusion matrices of each experiment, allowing for a more detailed (per class/label) analysis of model performance (reported in Appendix~\ref{appendix:data}). \looseness=-1

\section{Results}
\label{section:results}

We present the results of a large-scale study comparing different DL models for multi-class (MCC) and multi-label classification (MLC) tasks from 22 datasets. To this end, we evaluate models from 10 architectures: AlexNet, VGG16, ResNet50, ResNet152, DenseNet162, EfficientNetB0, ConvNeXt, Vision Transformer (ViT), Swin Transformer (SwinT) and MLPMixer. For each model architecture, we evaluate two variants: (i) models trained from scratch and (2) fine-tuned models previously trained on the ImageNet-1K dataset. We additionally assess the performance of models pre-trained using in-domain EO datasets. In the remainder, we outline and discuss the following:

\begin{itemize}
 \itemsep0em 
    \item Performance of models trained from scratch with respect to the two types of tasks
    \item Benefits of pre-training models of different architectures and their effect in view of the dataset properties 
    \item Models' ability to generalize on unseen in-domain datasets
    \item The choice of the pre-trained dataset and its effect on the performance of the downstream model
    \item The 'performance vs. cost of model training' trade-off between the considered modeling approaches
    \item Common issues that affect the models' predictive performance in the context of EO applications.
\end{itemize}

Detailed results of each experiment, with additional performance measures, are given in Appendices~\ref{appendix:times}, \ref{appendix:model_generalization_results} and \ref{appendix:data}.

\subsection{Training models from scratch}

We begin by analyzing the performance of models trained from scratch, i.e., models initialized with random weights during training. Tables \ref{tab:multiclass_summary_scratch} and \ref{tab:multilabel_summary_scratch} present these results for the MCC and MLC tasks, respectively. Table~\ref{tab:multiclass_summary_scratch} reports the accuracy (\%) of the models learned from scratch for the 15 MCC datasets. It also reports the rank of the models, estimated based on their performance and averaged over the 15 datasets. The results show that, in general, convolutional architectures, especially the DenseNet, the EfficientNet, and the two ResNets, consistently perform well. This is even more evident for datasets such as PatternNet, RSI-CB256, and SAT6, where the DenseNet (and the other top-ranked models) lead to near-perfect results (accuracy greater than 99\%). More specifically, DenseNet is the best-performing model in more than half of the tasks (9 out of 15) and achieves accuracy greater than 90\% in 8 tasks. These performances are generally much lower for smaller datasets, such as WHU-RS19, Optimal31, UC Merced, SIRI-WHU, RSSCN7, and CLRS. However, the most challenging task is \textit{So2SAT}, where EfficientNetB0 achieves the highest accuracy of 65.17\%, while many of the models trail behind with a performance of 55-60\%. These results are consistent with previous findings~\citep{Stewart2021}, suggesting clear signs of over-fitting, influenced by the quality and size of the images in the dataset. The two transformer architectures (SwinT and ViT), the MLPMixer, and the latest ConvNeXt models are ranked at the bottom (only better than AlexNet), with lower, but, in many cases, still practically comparable performance to the leading DenseNets.

\begin{table}[!t]\centering
\caption{Accuracy (\%) of models trained from scratch on multi-class classification datasets. Bold indicates best performing model for a given dataset. We report the \emph{average rank} of a model (lower is better), ranked based on the performance and averaged across the 15 datasets.}\label{tab:multiclass_summary_scratch}
\scriptsize
\begin{tabular}{l|rrrrrrrrrr}\toprule
Dataset \textbackslash Model &AlexNet &VGG16 &ResNet50 &ResNet152 &DenseNet161 &EfficientNetB0 &ViT &MLPMixer  &ConvNeXt &SwinT \\\midrule
    WHU-RS19 & 66.169 & 68.657 & 79.602 & 80.597 & \textbf{80.597} & 75.622 & 74.627 & 69.652 & 72.139 & 78.607\\
    Optimal31 & 55.108 & 56.720 & 67.204 & 62.903 & \textbf{71.237} & 68.548 & 62.634 & 59.140 & 58.871 & 66.129\\
    UC Merced & 81.190 & 78.571 & 85.238 & 84.048 & \textbf{86.190} & 84.286 & 83.095 & 82.381 & 84.286 & 81.429\\
    SIRI-WHU & 83.750 & 84.792 & \textbf{88.958} & 88.750 & 86.667 & 86.042 & 86.250 & 82.5 & 84.167 & 85.833\\
    RSSCN7 & 80.536 & 81.607 & 82.679 & 82.679 & \textbf{87.321} & 83.929 & 86.071 & 83.214 & 83.036 & 82.5\\
    BCS   & 89.410 & 89.410 & 89.236 & 88.542 & \textbf{90.799} & 85.417 & 87.847 & 86.285 & 84.375 & 89.236\\
    AID   & 81.350 & 81.950 & 89.050 & 89.9  & \textbf{93.300} & 90.050 & 79.350 & 71.750 & 81.1 & 87.700\\
    CLRS  & 71.4 & 76.067 & 85.567 & 82.3 & \textbf{86.167} & 82.267 & 65.467 & 61.133 & 69.167 & 80\\
    RSI-CB256 & 97.354 & 98.828 & 98.828 & \textbf{99.152} & 99.131 & 99.111 & 98.121 & 98.424 & 98.444 & 99.091\\
    Eurosat & 96.167 & 97.185 & 97   & 97.407 & 97.630 & \textbf{97.796} & 95.037 & 95.5 & 95.426 & 95.722\\
    PatternNet & 97.829 & 97.911 & 99.063 & 98.882 & \textbf{99.243} & 98.832 & 96.694 & 98.832 & 97.829 & 98.520\\
    RESISC45 & 82.159 & 83.889 & 92.333 & 90.683 & \textbf{93.460} & 91.365 & 81.016 & 69.413 & 85.937 & 88.730\\
    RSD46-WHU & 86.032 & 88.625 & 90.549 & 89.944 & 92.211 & 90.612 & 86.466 & 81.253 & 88.693 & 91.806\\
    So2Sat & 56.511 & 62.271 & 59.587 & 61.477 & 55.428 & \textbf{65.173} & 55.333 & 53.580 & 60.154 & 57.128\\
    SAT6  & 99.272 & 99.564 & \textbf{100} & 99.998 & 99.995 & 99.998 & 99.985 & 99.984 & 99.998 & 99.980\\
\midrule
    \textit{Avg. Rank} & 8.13 & 6.60  & 3.27 & 3.47 & \textbf{2.00} & 3.33 & 7.33 & 8.07 & 6.60 & 5.47 \\
\bottomrule
\end{tabular}
\end{table}

\begin{table}[!t]\centering
\caption{Mean average precision (mAP \%) of models trained from scratch on multi-label classification datasets. Bold indicates best performing model for a given dataset. We report the \emph{average rank} of a model (lower is better), ranked based on the performance and averaged across the 7 datasets.}\label{tab:multilabel_summary_scratch}
\scriptsize
\begin{tabular}{l|rrrrrrrrrr}\toprule
Dataset \textbackslash Model &AlexNet &VGG16 &ResNet50 &ResNet152 &DenseNet161 &EfficientNetB0 &ViT &MLPMixer &ConvNeXt &SwinT \\\midrule
AID (mlc) & 68.780 & 69.206 & 70.867 & 69.646 & 71.218 & \textbf{72.889} & 65.581 & 64.235 & 65.595 & 69.548\\
UC Merced (mlc) & 75.516 & 76.797 & 79.867 & 73.657 & 85.414 & 79.874 & \textbf{87.142} & 75.677 & 72.271 & 81.071\\
DFC15 & 88.099 & 89.871 & 94.675 & 94.188 & \textbf{95.848} & 93.973 & 94.164 & 91.663 & 89.564 & 94.349\\
Planet UAS & 60.282 & 60.682 & 64.192 & 64.956 & 64.738 & 63.868 & 59.414 & 58.550 & 61.277 & \textbf{65.229}\\
MLRSNet & 90.850 & 91.524 & \textbf{95.259} & 93.982 & 94.745 & 94.395 & 87.250 & 85.281 & 90.710 & 94.099\\
BigEarthNet 19 & 75.711 & 77.989 & 78.726 & 78.519 & 79.725 & 79.211 & 75.871 & 77.005 & 77.909 & \textbf{80.586}\\
BigEarthNet 43 & 56.082 & 58.969 & 64.343 & 62.736 & 63.390 & 62.173 & 57.410 & 58.772 & 60.472 & \textbf{67.487}\\
\midrule
\textit{Avg. Rank} & 8.57 & 6.57 & 3.00 & 4.71 & \textbf{2.14} & 3.86 & 7.29 & 8.57 & 7.71 & 2.57 \\
\bottomrule
\end{tabular}
\end{table}

Next, we shift our focus to MLC tasks. Table~\ref{tab:multilabel_summary_scratch} reports the mean average precision (\%) of the models learned from scratch across the 7 MLC datasets. While DenseNets rank the best, they achieve the best result in only 1 out of 7 tasks. The second-ranked SwinT models achieve the best performance in 3 tasks with comparable performance in the remaining 4. Unlike the MCC tasks, the performance difference to other convolutional models (i.e., the two ResNets and the EfficientNetB0) here is much smaller. Moreover, most models were only able to achieve high performance (above 90\%) on two tasks, \textit{DFC15} and \textit{MLRSNet}, with DenseNet and ResnNet50 achieving the best results. However, this is an expected result, as MLC tasks are generally more challenging than MCC tasks. This can be attributed to two things in particular: First, in many cases, the semantic labels can be very similar, which makes many of the models struggle. Second, MLC datasets tend to have a more significant class/label imbalance, in contrast to MCC datasets' more uniform class distribution. In this context, the most challenging MLC tasks overall are \textit{PlanetUAS} and \textit{BigEarthNet43}, where the best performing SwinT models achieve mAP od 65.229\% and 67.487\%, respectively. Finally, similar to the previous MCC analysis, ViT, MLPMixer, and ConvNeXt remain only better ranked than AlexNet. Nevertheless, their performance on these MLC tasks is much more competitive, for instance, in the case of ViT, which is the best model on the \textit{ UC Merced} task. 

\subsection{The benefits of model pre-training}

While training models from scratch leads to decent performance, in practice, leveraging pre-trained models can lead to significant performance improvements on image classification tasks~\citep{Zhai2019ALS}, and in particular on tasks in EO domains~\citep{Risojevic2021}. 

\begin{figure}
    \centering
    \includegraphics[width=0.9\linewidth]{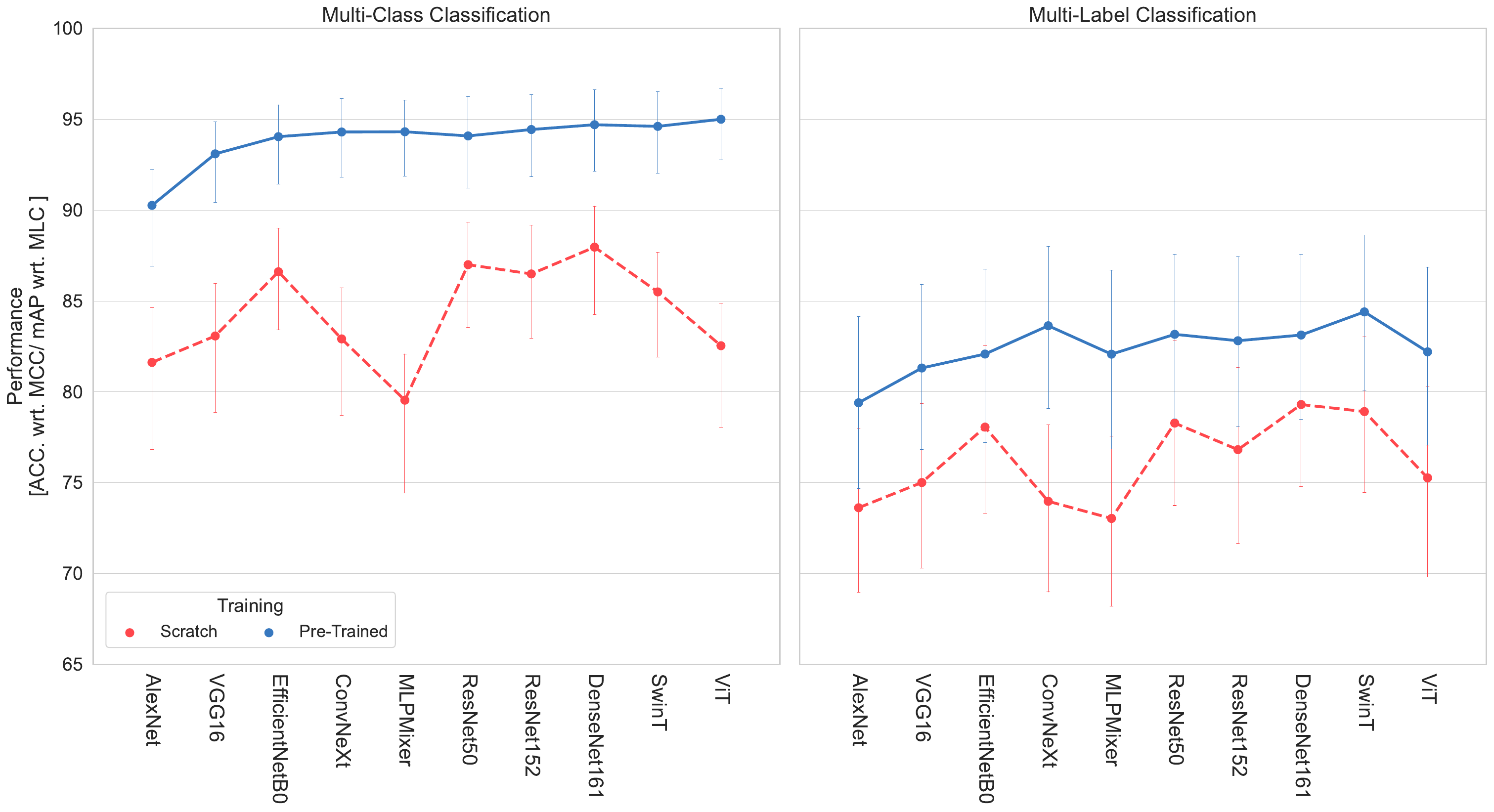}
    \caption{\textbf{Comparison of average performance improvement} of models from the 10 different architectures when trained from scratch \textbf{(red)}  and employing pre-trained models \textbf{(blue)} across \textbf{(left)} MCC and \textbf{(right)} MLC datasets. Error bars indicate confidence interval of 68\%. Models are ordered (worst to best) based on the average performance-rank of the pre-trained variants across all of the 22 datasets. Model pre-training leads to substantial performance improvements.}
    \label{fig:scratchVspre-train}
\end{figure}

This is also the general conclusion from our analysis. When using models that were first pre-trained on ImageNet-1K and then fine-tuned on the specific datasets, we found that: \emph{Pre-trained models lead to substantial performance improvements compared to models trained from scratch.} Figure~\ref{fig:scratchVspre-train} illustrates this performance-improvement trend for different models across the 22 MCC and MLC tasks. We find that pre-training significantly improves the performance of all the evaluated models. Notably, we observe that the transformer models,  based on either ViT or SwinT architectures, benefit the most from pre-training, followed by MLPMixer and ConvNeXt models. This is a significant improvement over the models trained from scratch. These results, especially for the case of ViT, are consistent with previously reported findings~\citep{dosovitskiy2020image,papoutsis2022efficient}.

\begin{table}[!th]\centering
\caption{Accuracy (\%) of models pre-trained on ImageNet-1K on multi-class classification datasets. Bold indicates best performing model for a given dataset. We report the \emph{average rank} of a model (lower is better), ranked based on the performance and averaged across the 15 datasets.}\label{tab:multiclass_summary_pre-trained}
\scriptsize
\begin{tabular}{l|rrrrrrrrrr}\toprule

    Dataset\textbackslash{}Model & AlexNet &VGG16 &ResNet50 &ResNet152 &DenseNet161 &EfficientNetB0 &ViT &MLPMixer &ConvNeXt &SwinT \\
    \midrule
    WHU-RS19 & 93.532 & 99.005 & 99.502 & 98.01& \textbf{100} & 99.502 & 99.502 & 98.507 & 99.005 & 99.502\\
    Optimal31 & 80.914 & 88.71& 92.204 & 92.473 & 94.355 & 91.667 & \textbf{94.624} & 92.742 & 93.011 & 92.473\\
    UC Merced & 92.143 & 95.476 & 98.571 & \textbf{98.810} & 98.333 & 98.571 & 98.333 & 98.333 & 97.857 & 98.571\\
    SIRI-WHU & 92.292 & 93.958 & 95   & \textbf{96.25} & 95.625 & 95   & 95.625 & 95.208 & \textbf{96.25} & 95.625\\
    RSSCN7 & 91.964 & 93.929 & 95   & 95   & 94.821 & 95.536 & \textbf{95.893} & 95.179 & 94.643 & 95.179\\
    BCS   & 89.583 & 90.972 & 92.014 & 92.361 & 92.708 & 91.319 & 92.014 & 93.056 & 91.493 & \textbf{93.403}\\
    AID   & 92.9 & 96.1 & 96.55& 97.2 & 97.25& 96.25& \textbf{97.750} & 96.7 & 96.95 & 97.4\\
    CLRS  & 84.1 & 89.9 & 91.567 & 91.9 & 92.2 & 90.5 & \textbf{93.200} & 90.1 & 91.1 & 92.533\\
    RSI-CB256 & 99.354 & 99.051 & 99.677 & \textbf{99.859} & 99.737 & 99.717 & 99.758 & 99.657 & 99.596 & 99.677\\
    Eurosat & 97.574 & 98.148 & 98.833 & \textbf{99} & 98.889 & 98.907 & 98.722 & 98.741 & 98.778 & 98.944\\
    PatternNet & 99.161 & 99.424 & \textbf{99.737} & 99.49& \textbf{99.737} & 99.539 & 99.655 & 99.704 & 99.671 & 99.688\\
    RESISC45 & 90.492 & 93.905 & 96.46& 96.54 & 96.508 & 94.873 & \textbf{97.079} & 95.952 & 96.27 & 96.587\\
    RSD46-WHU & 90.646 & 92.422 & 94.158 & 94.404 & \textbf{94.507} & 93.387 & 94.238 & 93.673 & 93.627 & 93.536\\
    So2Sat & 59.203 & 65.375 & 61.903 & 65.169 & 65.756 & 65.801 & \textbf{68.551} & 67.066 & 66.169 & 65.950\\
    SAT6  & 99.98& 99.993 & \textbf{100} & \textbf{100} & \textbf{100} & 99.988 & 99.998 & 99.995 & 99.999 & 99.999\\
\midrule
    \textit{Avg. Rank} & 9.93	& 8.67	& 4.67	& 3.80	& 3.13 & 5.87	& \textbf{3.07}	& 5.33 & 5.47 & 3.20 \\
\bottomrule
\end{tabular}
\end{table}

Tables \ref{tab:multiclass_summary_pre-trained} and \ref{tab:multilabel_summary_pre-trained} present the detailed results of these analyses for MCC and MLC tasks, respectively. Similar to the analyses in the previous section, we report model accuracy (\%) in the case of MCC tasks and mean average precision (\%) in the case of MLC tasks. We also report the rank of the models, averaged over the respective datasets. Considering MCC tasks (Table \ref{tab:multiclass_summary_pre-trained}), most models achieve very good performance (accuracy over 90\%) on 14 (out of 15) tasks, with (almost) perfect results in five of those. Notably, we observed significant performance improvements, compared to model counterparts trained from scratch, on smaller datasets (such as \textit{WHU-RS19, Optimal31, UC Merced, SIRI-WHU, RSSCN7, and CLRS}), reaffirming the utility of transfer learning from large datasets in the context of EO image classification tasks. In terms of model architectures, the ViT ranks at the top among the model architectures, achieving the best performance in 6 out of 15 cases, followed by DenseNet161, SwinT, and ResNet152 with lower but comparable performance. Transformer architectures, and ViTs in particular, typically require large amounts of training data \cite{dosovitskiy2020image,Paul_Chen_2022} for learning robust, good performing models. As a result, using pre-trained models and fine-tuning them leads to substantial performance improvements, compared to training them from scratch. The performance of ViTs is further highlighted for the case of the challenging \textit{So2SAT} task, where the ViT model leads to an accuracy of 68.55\%, in contrast to the next ranked DenseNet and SwinT with an accuracy of 65.75\% and 65.95\%, respectively. In this specific case of \textit{So2SAT}, we observed that over-fitting remains an issue, even for pre-trained models. Our further investigation of the train/validation loss trends showed that, regardless of the model at hand, with the training loss decreasing, the validation errors increase almost instantly (after 1-2 epochs) - a typical trend observed in over-fitting models (see Figure~\ref{fig:so2sat_learning_curve} that illustrates such behavior in a ViT model). This, fortunately, is not the case for the remaining tasks, where we observed a decent performance overall. Most models, especially the top half ranked, achieved stable and mostly comparable performance.

\begin{table}[!t]\centering
\caption{Mean average precision (mAP \%) of models pre-trained on ImageNet-1K on multi-label classification datasets. Bold indicates best performing model for a given dataset. We report the \emph{average rank} of a model (lower is better), ranked based on the performance and averaged across the 7 datasets.}\label{tab:multilabel_summary_pre-trained}
\scriptsize
\begin{tabular}{l|rrrrrrrrrr}\toprule
Dataset \textbackslash Model &AlexNet &VGG16 &ResNet50 &ResNet152 &DenseNet161 &EfficientNetB0 &ViT &MLPMixer &ConvNeXt &SwinT \\\midrule
    AID (mlc)   & 75.906 & 79.893 & 80.758 & 80.942 & 81.708 & 78.002 & 81.539 & 80.879 & \textbf{82.298} & 82.254\\
    UC Merced (mlc) & 92.638 & 92.848 & 95.665 & 96.01& 96.056 & 95.384 & 96.699 & 96.34 & 96.431 & \textbf{96.831}\\
    DFC15 & 94.057 & 96.566 & 97.662 & 97.6 & 97.529 & 96.787 & 97.617 & 97.941 & 97.994 & \textbf{98.111} \\
    Planet UAS & 64.048 & 65.584 & 65.528 & 64.825 & 66.339 & 64.157 & 66.804 & 67.330 & 66.447 & \textbf{67.837} \\
    MLRSNet & 93.399 & 94.633 & 96.272 & 96.432 & 96.306 & 95.391 & 96.41& 95.049 & 95.807 & \textbf{96.620}\\
    BigEarthNet 19 & 77.147 & 78.418 & 79.983 & 79.776 & 79.686 & 80.221 & 77.31& 77.288 & 80.283 & \textbf{81.384} \\
    BigEarthNet 43 & 58.554 & 61.205 & 66.256 & 64.066 & 64.229 & 64.589 & 58.997 & 59.648 & 66.166 & \textbf{67.733} \\
\midrule
\textit{Avg. Rank} & 10.00  & 7.86  & 5.14  & 5.43  & 5.00  & 6.86  & 4.86  & 5.71  & 3.00 & \textbf{1.14} \\
\bottomrule
\end{tabular}
\end{table}

The benefits of pre-training models also extend to MLC tasks (Table~\ref{tab:multilabel_summary_pre-trained}), in several cases with significant performance gains, compared to model counterparts trained from scratch. In particular, we found that pre-training can lead to minor improvements (1\%-2\%) on challenging tasks such as \textit{PlanetUAS} and \textit{BigEarthNet43} (mAP of 67.837\% and 67.733\% achieved by SwinTs); to more considerable improvements (up to 15\%) in some cases such as \textit{AID} and \textit{UCMerced} (mAP of 82.298\% and 96.83\% obtained by ConvNeXt and SwinT, respectively). Also, in this case, we found that the transformer models benefited the most from pre-training. This is in line with studies\cite{Ze2021SwinV1,ze2022swinv2} that highlight the significance of pre-training to the generalization performance of these types of models. Notably, SwinT models ranked the best overall and achieved the best performance on 6 (out of the 7) tasks. They are followed by ViT and ConvNeXt, with comparable performance on most tasks.

\subsection{Generalization capabilities to unseen data}

We further investigate the generalization ability of the trained models by evaluating their performance across datasets not used during training. In particular, we present results from two evaluation settings: (1) performance measured on a holdout set compiled of test images with shared labels and (2) an exhaustive cross-dataset evaluation between pairs of datasets with overlapping labels. First, we analyze the predictive performance of all models when applied to the same holdout set with 3216 images sampled from the test splits from seven MCC datasets (\emph{RESISC45, UC Merced, CLRS, PatternNet, AID, RSI-CB256} and \emph{WHU-RS19}) using only images with labels shared among the seven datasets: 'Forest', 'Parking', 'River', 'Harbor', and 'Beach'. Figure~\ref{fig:multiclass_common_labels_aux} (in Appendix~\ref{appendix:model_generalization_results}) presents further details of the distribution of images in the holdout set w.r.t. source datasets and labels.  We evaluate and report the predictive performance of pre-trained models from all ten architectures. Note that here we only evaluate the models on the holdout set without additional fine-tuning. Table~\ref{tab:multiclass_common_labels_results} reports the predictive performance assessed using accuracy (\%) as an evaluation measure.

\begin{table}[!b]\centering
\caption{Accuracy (\%) of models pre-trained on ImageNet-1K and fine-tuned on a specific source dataset and evaluated on the common test dataset with shared labels. Bold indicates best performing model for a given source dataset.}\label{tab:multiclass_common_labels_results}
\scriptsize
\begin{tabular}{l|rrrrrrrrrr}\toprule
Dataset \textbackslash Model &AlexNet &VGG16 &ResNet50 &ResNet152 &DenseNet161 &EfficientNetB0 &ViT &MLPMixer &ConvNeXt &SwinT \\\midrule
RESISC45 & 66.853 & 78.514 & 81.063 & 84.08 & 84.111 & 77.985 & \textbf{86.007} & 82.121 & 84.422 & 83.706 \\
UC Merced & 63.371 & 67.04 & 76.057 & 73.01 & 74.254 & 74.44 & 75.995 & \textbf{79.478} & 75.902 & 72.326 \\
CLRS & 80.037 & 83.427 & 89.801 & 88.557 & 89.024 & 86.07 & \textbf{92.6} & 89.646 & 89.303 & 90.299 \\ 
PatternNet & 43.501 & 52.332 & 56.965 & 54.54 & 56.716 & 60.044 & 64.739 & 62.687 & 59.391 & \textbf{65.205} \\
AID & 71.393 & 69.714 & 79.384 & 80.1 & 66.169 & 77.892 & \textbf{83.862} & 77.954 & 79.851 & 79.789 \\
RSI-CB256 & 56.872 & 61.412 & 58.893 & 63.65 & 64.832 & 61.723 & 66.014 & \textbf{66.791} & 64.677 & 66.294 \\
WHU-RS19 & 61.101 & 62.624 & 71.953 & 73.321 & 72.388 & 68.284 & 72.917 & 74.036 & \textbf{74.876} & 71.144 \\
\midrule
\textit{Avg. Rank} & 9.71	& 8.71	& 5.43	& 5.29	& 5.86	& 6.86	& \textbf{2.14}	& 3.29	& 3.57	& 4.14
 \\
\bottomrule
\end{tabular}
\end{table}

The results show that ViT models are able to generalize well to unseen images from other in-domain datasets. Namely, in many cases, ViT models perform better than the competitors, further supporting previous results regarding their performance on MCC tasks. The performance of ViTs is followed by models based on more recent architectures, such as SwinT, MLPMixer, and ConvNeXt, which show worse but, in many cases, practically comparable performance. With respect to specific datasets, our experiments show that models fine-tuned on the \emph{CLRS} and \emph{RESISC45} datasets were able to achieve much better performance than the others (with ViT models achieving 92.6\% in the case of \emph{CLRS}). We hypothesize that such performance may be related to the particular properties of these datasets: Both \emph{CLRS} and \emph{RESISC45} are multi-resolution datasets (containing images at different spatial resolutions) with a large number of diverse labels. However, this is not the case for models fine-tuned on \emph{PatternNet} and \emph{RSI-CB256}. While models trained and evaluated on these datasets separately show great performance (~99\% accuracy), this performance decreases significantly when evaluated on a holdout set (down to 66.79\% and 65.2\% for \emph{RSI-CB256} and \emph{PatternNet}, respectively). These results, along with results from models learned from scratch (Table~\ref{tab:multiclass_summary_scratch}), are indicative of both datasets being easily learned, producing models that are not able to generalize well to other unseen images and classification tasks. \looseness=-1

\begin{figure}[t]
    \centering
    \begin{subfigure}[]{0.7\textwidth}
    \includegraphics[width=1\linewidth]{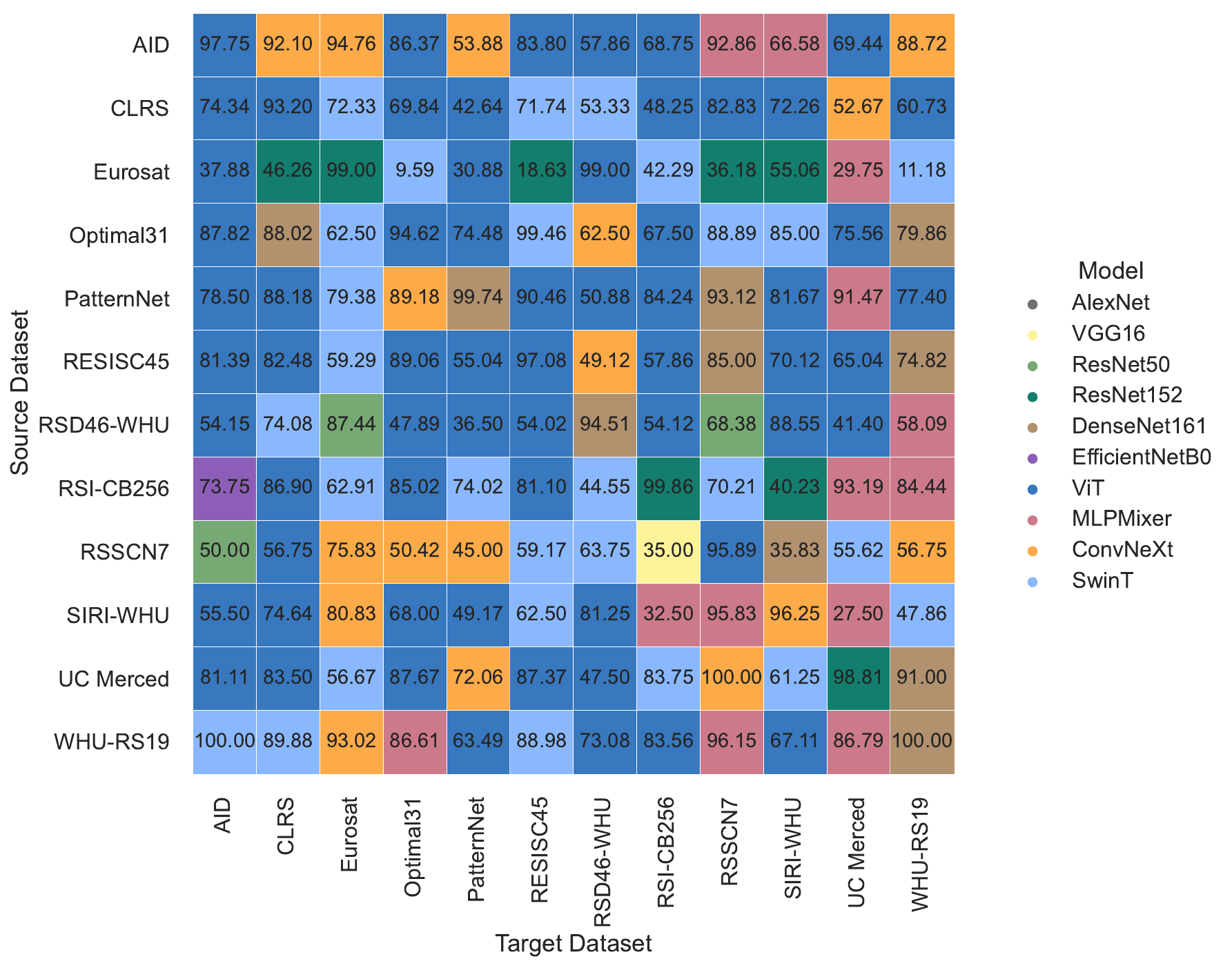}
    \end{subfigure}
    \hfill
    \begin{subfigure}[t]{0.29\textwidth}
    \includegraphics[width=1\linewidth]{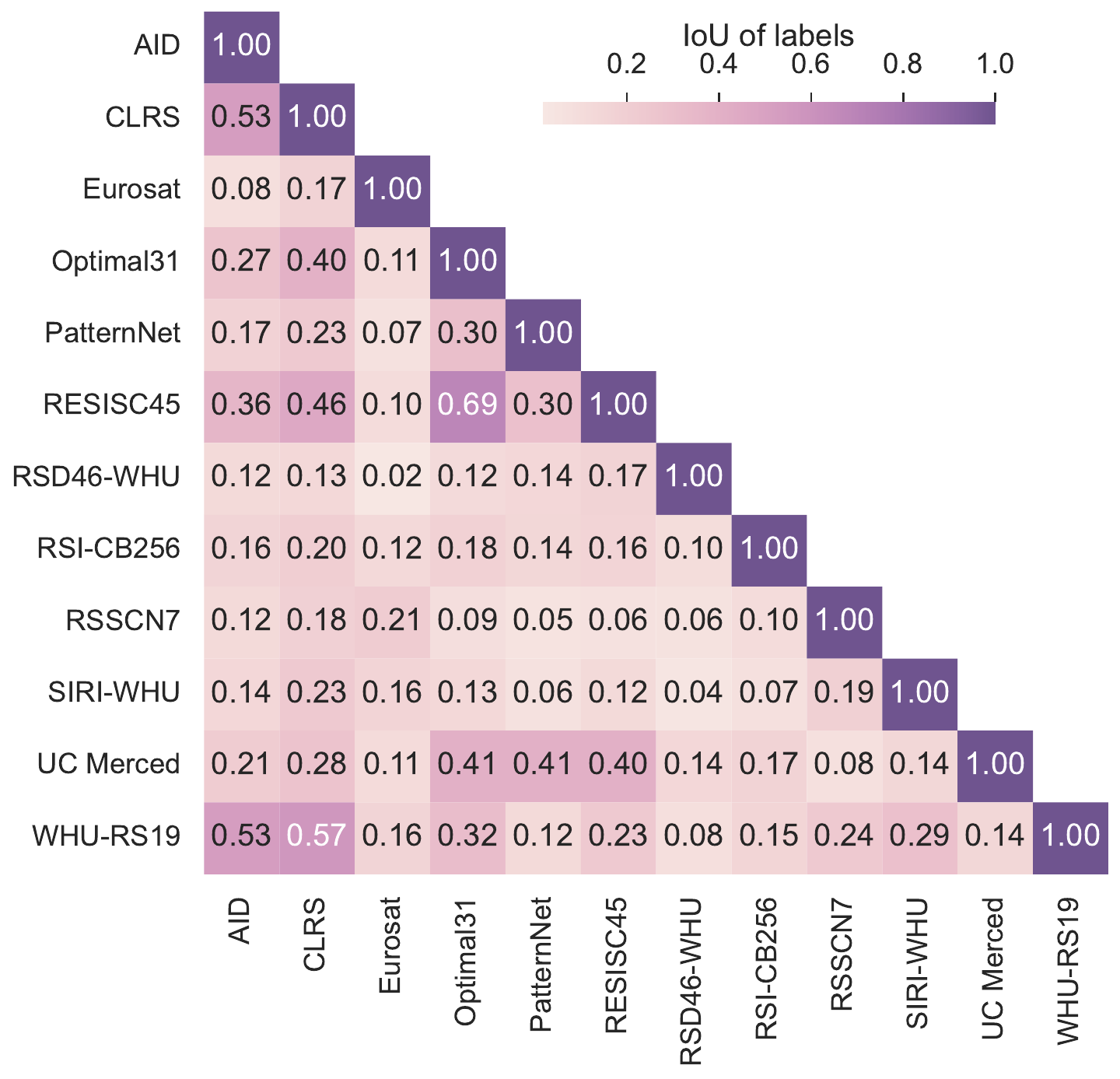}
    \end{subfigure}
    \caption{\textbf{Model generalization on multi-class classification tasks:} Comparison of the best performing pre-trained models (\textbf{left}) from the 10 different architectures (\textbf{color-coded}) in terms of \textbf{accuracy} (\% acc. is indicated in each field); the models are fine-tuned on \textit{source} dataset and evaluated on images with common/overlapping labels in \textit{target} dataset. The heatmap (\textbf{right}) reports the label overlap between each pair of datasets, in terms of IoU. Transformer-based models, in particular the ViT models, perform the best when evaluated on other in-domain MCC datasets.}
    \label{fig:generalization_results_mcc}
\end{figure}

\begin{figure}[t]
       \begin{subfigure}[]{0.55\linewidth}
        \centering
    \includegraphics[width=1\linewidth]{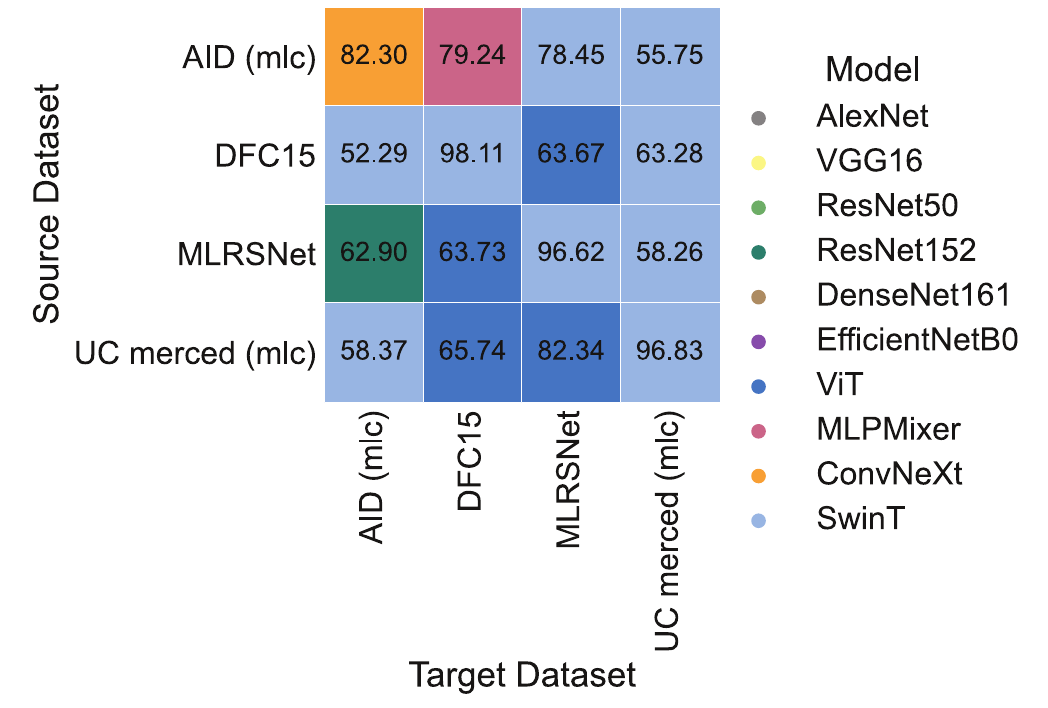}
    \end{subfigure}\hspace{1em}
     \begin{subfigure}[]{0.25\linewidth}
    \includegraphics[width=1\linewidth]{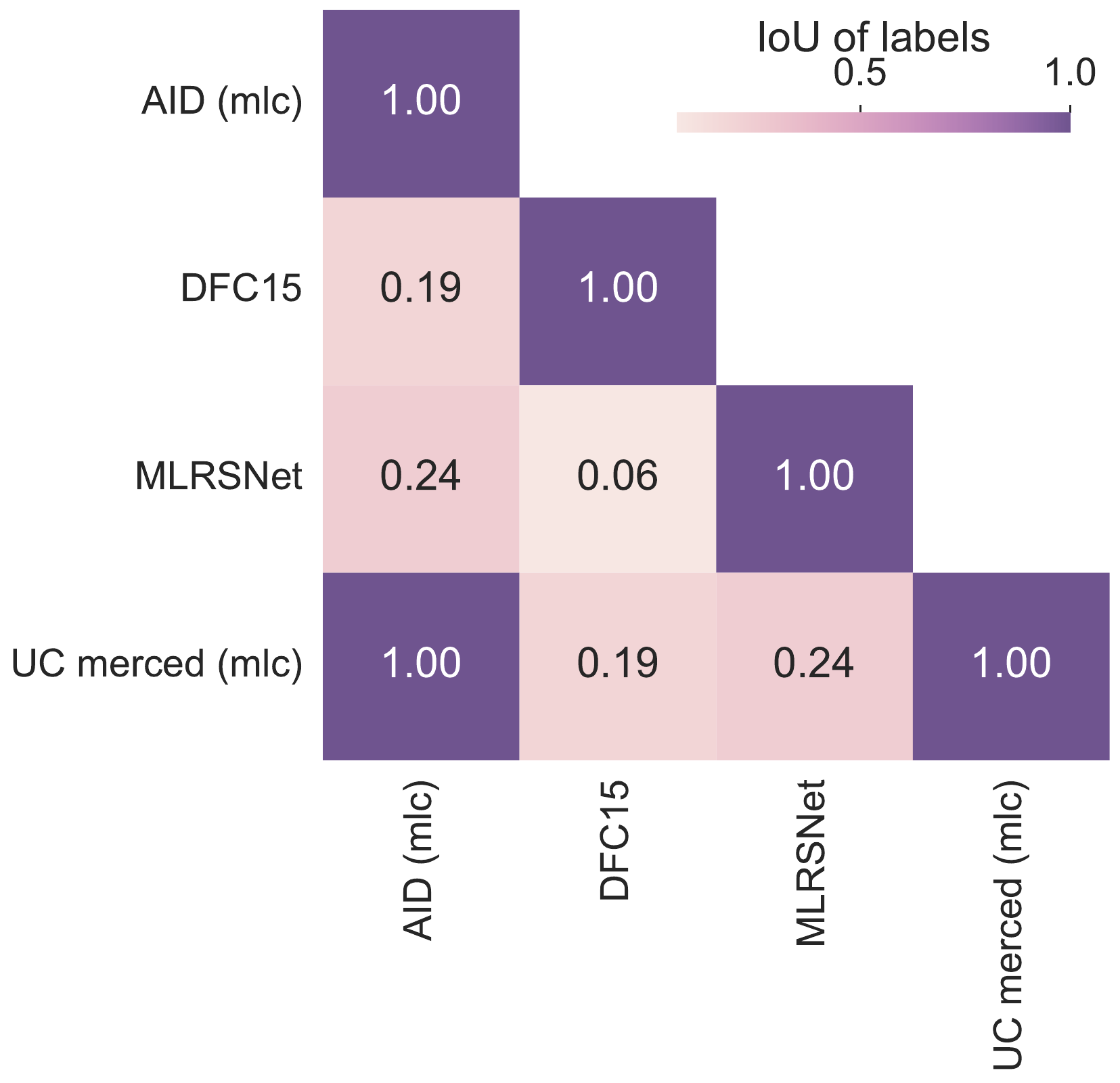}
    \end{subfigure}\hfill
    \caption{\textbf{Model generalization on multi-label classification tasks:} Comparison of the best performing pre-trained models (\textbf{left}) from the 10 different architectures (\textbf{color-coded}) in terms of \textbf{mean average precision} (\% mAP is indicated in each field); the models are fine-tuned on \textit{source} dataset and evaluated on images with common/overlapping labels in \textit{target} dataset. The heatmap (\textit{right}) reports the label overlap between each pair of datasets, in terms of IoU. In general, transformer-based models, in particular the SwinT models, lead to the best performance on MLC tasks.}
    \label{fig:generalization_results_mlc}
\end{figure}

In the second experimental setup, we employ the following pairwise evaluation scheme. We consider pairs of \textit{source} and \textit{target} datasets: We take pre-trained models that have been fine-tuned on a \textit{source} dataset and evaluate them on test images from a \textit{target} dataset. Note that we only \emph{evaluate} the models on the target dataset without additional fine-tuning. We measure the performance only on a subset of images with shared labels between the source and target datasets. Therefore, for this experiment, we selected datasets with at least 0.15 IoU\footnote{Intersection over Union (IoU), measures the overlap between two sets. Values range from 0 to 1, where 0 indicates no overlap and 1 indicates complete overlap between the sets} overlap of labels with at least one other dataset. This resulted in pairs from 12 (out of 15) MCC datasets and 4 (out of 7) MLC datasets, yielding 256 comparisons of pre-trained models from each of the ten considered architectures. Figures ~\ref{fig:generalization_results_mcc} and ~\ref{fig:generalization_results_mlc} present the performance of the best model for each MCC and MLC comparison in terms of accuracy (\%) and mAP (\%), respectively. They also provide a summary of the overlap between each pair of datasets in terms of IoU. Detailed results of all comparisons, per architecture, are given in Appendix~\ref{appendix:model_generalization_results}.

    

The results support our earlier findings that the \emph{transformer}-based models, in particular the ViT models (on MCC tasks) and the SwinT models (on MLC tasks), perform best when applied to other in-domain datasets. More specifically, when considering MCC tasks, the transformer-based models perform best in almost 2/3 of the comparisons, with the ViT models alone performing best in $\sim$40\% of them. ViTs are followed by SwinT, ConvNeXt, and MLPMixer models that, in many cases, showed practically comparable performance. We observed that convolutional models such as DenseNets, which exhibited good performance in our previous analyses (when evaluated on test images from the same dataset), generally lead to worse performance than models from more recent architectures. The dominance of the transformer-based models also extends to MLC tasks, with SwinT models producing the best overall performance, followed closely by ViT models. Note that these empirical results are also consistent with other studies~\citep{bhojanapalli2021,Paul_Chen_2022,Zhang2022DelvingDI}, that highlight the robustness and good generalization capabilities of transformer-based models for general-domain images.

\subsection{Domain-adaptive transfer learning}

Having demonstrated the practical benefits and generalization capabilities of using pre-trained models, we further investigate the impact of the pre-trained dataset on the performance of the downstream model. As we focus on particular domains of interest that leverage satellite imagery, we evaluate whether and how choosing more appropriate in-domain EO pre-training datasets (and strategies) affects downstream predictive performance. Our experimental setup aims to investigate two different strategies for such in-domain pre-training: (i) in-domain only, where models are pre-trained entirely on an EO dataset (ii) two-stage pre-training, where models are pre-trained on a combination of ImageNet-1K and an EO dataset. The former strategy is analogous to the ImageNet-1K pre-training strategy but uses a different EO dataset. In the second strategy, on the other hand, the models are first pre-trained on ImageNet-1K, followed by intermediate tuning on an in-domain EO dataset, before fine-tuning the models on the target EO dataset.

\begin{table}[!t]
 \caption{Comparison of pre-training strategies for (a) Vision Transformers (ViT) and (b) DenseNets161 using 4 in-domain EO datasets (SAT6, RSD46-WHU, MLRSNet, RESISC45) and ImageNet-1K. We report their performance on 3 multi-class and 3 multi-labels classification tasks, in terms of accuracy (\% Acc.) and mean average precision (\% mAP), respectively.}
  \begin{subtable}[t]{1\textwidth}
  \centering
 \caption{}
    \begin{adjustbox}{width=.8\linewidth}
    \begin{tabular}{l|l|rrrrrr}
    \toprule
    \multicolumn{1}{c|}{\multirow{2}[0]{*}{\shortstack{In-domain\\dataset}}} & \multicolumn{1}{c|}{\multirow{2}[0]{*}{\shortstack{Pre-training\\strategy}}} & \multicolumn{6}{c}{Target dataset} \\
          &       &  \shortstack{CLRS\\$[$\%Acc.$]$} & \shortstack{Optimal31\\$[$\%Acc.$]$}  & \shortstack{So2Sat\\$[$\%Acc.$]$} & \shortstack{AID (mlc)\\$[$\%mAP$]$} & \shortstack{Planet UAS\\$[$\%mAP$]$} & \shortstack{BigEarthNet 19\\$[$\%mAP$]$} \\\midrule
 
    \multirow{2}[0]{*}{SAT6} & In-domain only & 60.767 & 58.065 & 54.672 & 62.2  & 59.538 & 74.618 \\
          & ImageNet-1K + In-domain & 71.2  & 68.011 & 64.284 & 67.595 & 62.356 & 76.009 \\
    \multirow{2}[0]{*}{RSD46-WHU} & In-domain only & 72.267 & 75.269 & 57.859 & 71.209 & 61.015 & 75.529 \\
          & ImageNet-1K + In-domain & 91.067 & 92.204 & 65.322 & 80.102 & 66.46 & 76.809 \\
    \multirow{2}[0]{*}{MLRSNet} & In-domain only & 71.7  & 77.688 & 54.746 & 72.915 & 60.985 & 74.827 \\
          & ImageNet-1K + In-domain & 91.033 & 95.430 & 64.321 & 83.069 &64.574 & 77.308 \\
     \multirow{2}[0]{*}{RESISC45} & In-domain only & 68.7 & 86.022 & 57.446 & 69.552 & 61.457 & 75.345 \\
          & ImageNet-1K + In-domain & 92.533 & 98.925 & 66.876 & 82.888 &66.654 & 76.682 \\\midrule      
              /     & ImageNet-1K only  & 93.2 & 94.624 & 68.551 & 81.539 & 66.804 & 77.31 \\   
    \bottomrule
    \end{tabular}%
    \end{adjustbox}
 \end{subtable}
  \begin{subtable}[t]{1\textwidth}
  \centering
 \caption{}
   \begin{adjustbox}{width=.8\linewidth}
    \begin{tabular}{l|l|rrrrrr}
    \toprule
\multicolumn{1}{c|}{\multirow{2}[0]{*}{\shortstack{In-domain\\dataset}}} & \multicolumn{1}{c|}{\multirow{2}[0]{*}{\shortstack{Pre-training\\strategy}}} & \multicolumn{6}{c}{Target dataset} \\
               &       &  \shortstack{CLRS\\$[$\%Acc.$]$}  & \shortstack{Optimal31\\$[$\%Acc.$]$} & \shortstack{So2Sat\\$[$\%Acc.$]$} & \shortstack{AID (mlc)\\$[$\%mAP$]$} & \shortstack{Planet UAS\\$[$\%mAP$]$} & \shortstack{BigEarthNet 19\\$[$\%mAP$]$} \\\midrule
\multirow{2}[0]{*}{SAT6} & In-domain & 65.467 & 55.914 & 58.455 & 59.363 & 59.419 & 76.745 \\
      & ImageNet-1K + In-domain & 89.467 & 85.215 & 65.334 & 74.653 & 64.918 & 79.773 \\
\multirow{2}[0]{*}{RSD46-WHU} & In-domain & 89.267 & 86.559 & 60.89 & 77.056 & 65.101 & 79.374 \\
      & ImageNet-1K + In-domain & 91.8 & 93.280 & 65.152 & 82.339 & 66.161 & 79.867 \\
\multirow{2}[0]{*}{MLRSNet} & In-domain & 89.7 & 92.742 & 61.03 & 80.144 & 64.53 & 79.646 \\
      & ImageNet-1K + In-domain & 91.367 & 96.505 & 62.808 & 84.07 & 64.859 & 79.945 \\
\multirow{2}[0]{*}{RESISC45} & In-domain & 86.433 & 93.011 & 60.009 & 73.199 & 63.532 & 78.309 \\
      & ImageNet-1K + In-domain & 91.267 & 98.387 & 64.011 & 82.936 & 66.276 & 79.695 \\\midrule
      /     & ImageNet-1K only  & 92.2 & 94.355 & 65.756 & 81.708 & 66.339 & 79.686 \\
    \bottomrule
    \end{tabular}%
    \end{adjustbox}
    
 \end{subtable}
 
  \label{tab:indomain}%
\end{table}%

Rather than evaluating all architectures, in this set of experiments, we evaluate two types of architectures: a ViT and a DenseNet161, as representatives of transformer and convolutional architectures that have shown overall good performance in our previous experiments. Specifically, we analyze their performance on six tasks (3 MCC and 3 MLC) that proved somewhat challenging for these models: \textit{CLRS, Optimal31, So2SAT, AID (mlc), PlanetUAS, and BigEarthNet 19}. We select four different in-domain datasets for our pre-training: \textit{SAT6, RSD46-WHU, MLRSNet} and \textit{RESISC45}; based on the overall performance achieved in the previous analyses, their size (number of images), and their heterogeneity (in terms of semantic labels). Table \ref{tab:indomain} reports the results of these experiments.

Our general conclusion regarding pre-training remains: Pre-trained models based entirely on EO datasets can still outperform their counterparts trained from scratch. However, we find that the choice of the pre-training dataset has a significant impact on the downstream performance and is not necessarily related to the quality of the pre-training dataset (measured as stand-alone performance) or solely to its size. For instance, we found that models pre-trained entirely using \textit{SAT6} (a dataset on which most models performed very well) performed much worse than the other pre-trained counterparts and, in some cases, even worse than models trained from scratch. This is not the case when pre-traning models on \textit{RSD46-WHU}, \textit{MLRSNet}, and \textit{RESISC45}, which led to better performance, compared to their counterparts trained from scratch (in both cases of ViT and DenseNets), albeit worse than models pre-trained on ImageNet-1K. \looseness=-1

Importantly, we found that using a combined pre-training procedure, with ImageNet-1K followed by an in-domain dataset, can lead to improvements (up to 5\%), especially when combined with \textit{MLRSNet} or \textit{RESISC45} datasets. This is specifically the case for \textit{Optimal31} and \textit{AID (mlc)}, where models from both ViT and DenseNet161 architectures were able to outperform their counterparts pre-trained only on ImageNet-1K. These results suggest that using datasets for intermediate fine-tuning that contain images at different resolutions with heterogeneous (but potentially semantically similar) labels, in addition to ImageNet-1K, can lead to performance improvements. However, in most cases, we did not observe neither practical nor significant benefits for using a combined pre-training procedure with an additional in-domain dataset that would justify the additional computational overhead for training such models.

\subsection{The 'performance vs. training cost' trade-off}

\begin{figure}[t]
    \centering
   \captionsetup[subfigure]{justification=centering}
    \begin{subfigure}[t]{0.48\linewidth}
    \includegraphics[width=\linewidth]{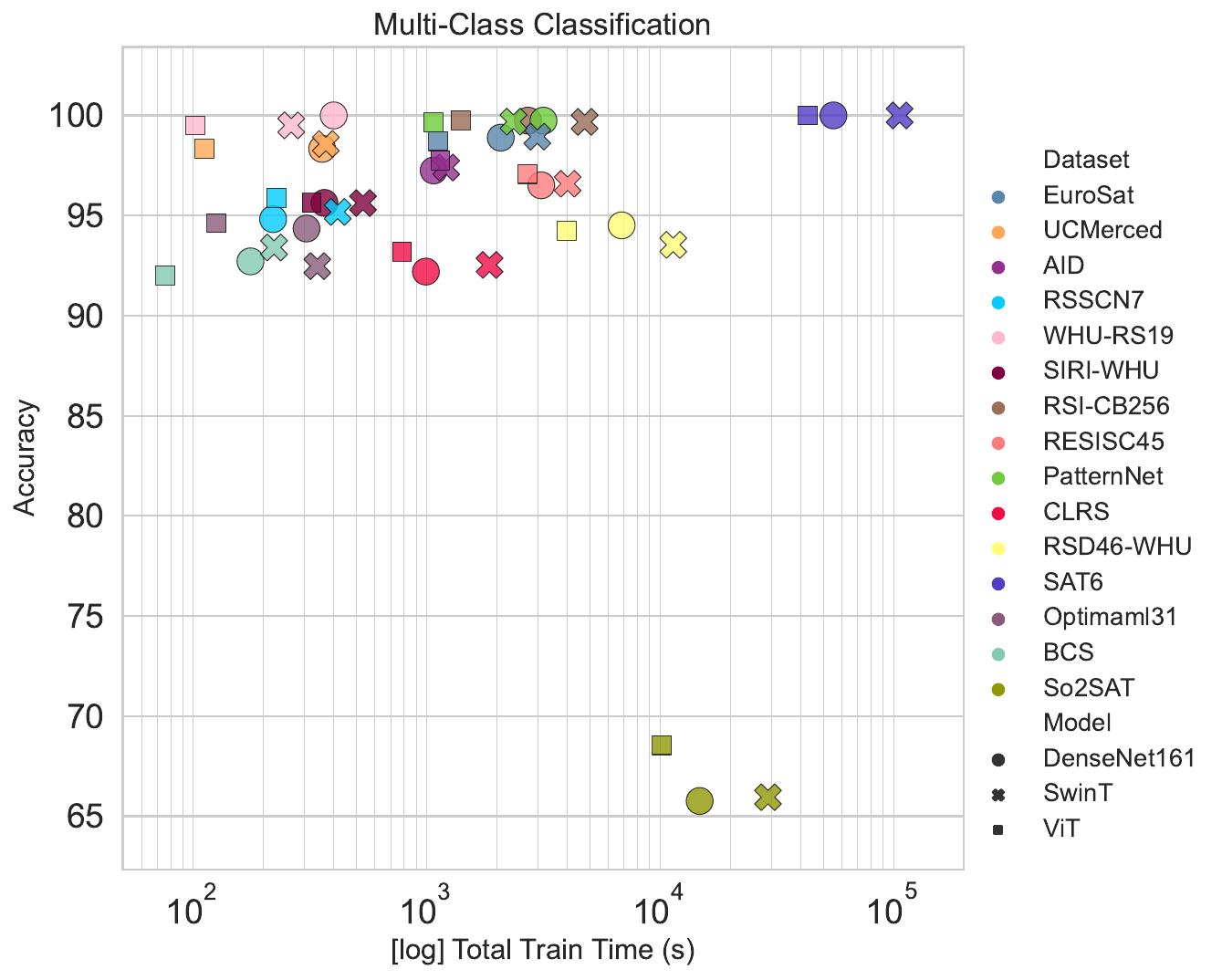}
   \subcaption{MCC tasks}
    \end{subfigure}
    \hfill
    \begin{subfigure}[t]{0.48\linewidth}
    \includegraphics[width=\linewidth]{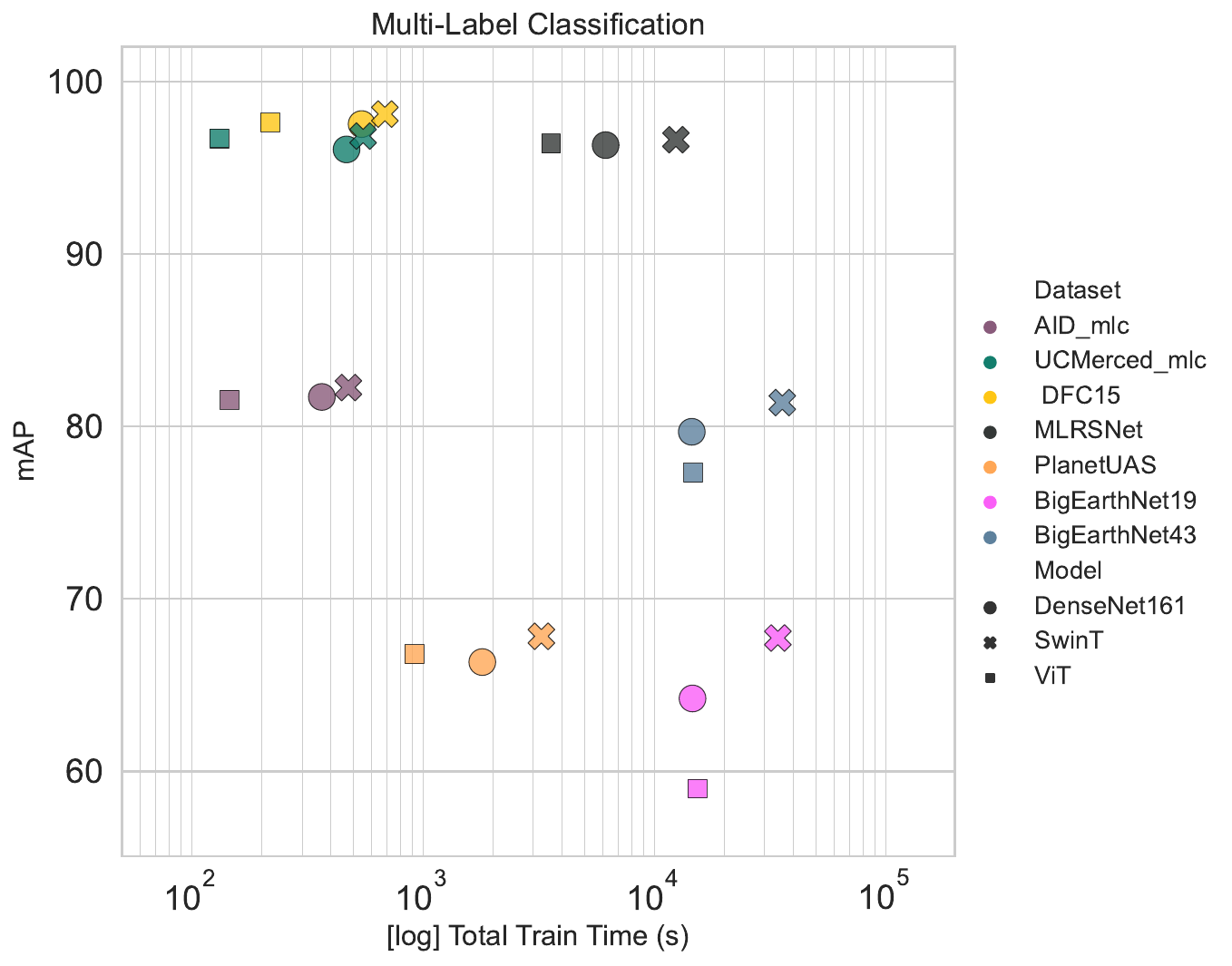}
   \subcaption{MLC tasks}
    \end{subfigure}
    \hfill
    \caption{\textbf{Performance vs. total training time} comparison of the overall top-3 performing \textit{pre-trained} model architectures, ViT, DenseNet161 and SwinT (denoted with different markers); evaluated on (\textbf{left}) MCC and \textbf{(right)} MLC datasets (color-coded). Performance is reported as accuracy (\%) and mean average precision (mAP \%) for MCC and MLC tasks, respectively. Note the log scale of the total training time (seconds).}
    \label{fig:performance_cost}
\end{figure}

Having established the performance of our evaluated models and demonstrated the clear benefits of using pre-trained models, we focus here on another line of comparison - the cost of model training. Recall from Section~\ref{sec:model_arch}, and in particular Table~\ref{table:models}, that we study model architectures that differ significantly in the number of learnable parameters. Typically, larger models require more computing resources and much more training time than smaller models. In our experimental setup, we train all models on the same computing infrastructure, under the same conditions, and with the same training/evaluation setup (in terms of hyperparameters and data partitioning). Therefore we can directly analyze the 'performance vs. training cost' (in terms of total training time) trade-off for each model variant from the ten different architectures (either pre-trained or trained from scratch) across the 22 datasets. This way, we can explicitly measure the benefits of each model and make further modeling decisions based on the performance of the models and the 'cost' of training them.

\begin{figure*}[!t]
    \centering
   \captionsetup[subfigure]{justification=centering}
    \begin{subfigure}[t]{0.48\linewidth}
    \includegraphics[width=\linewidth]{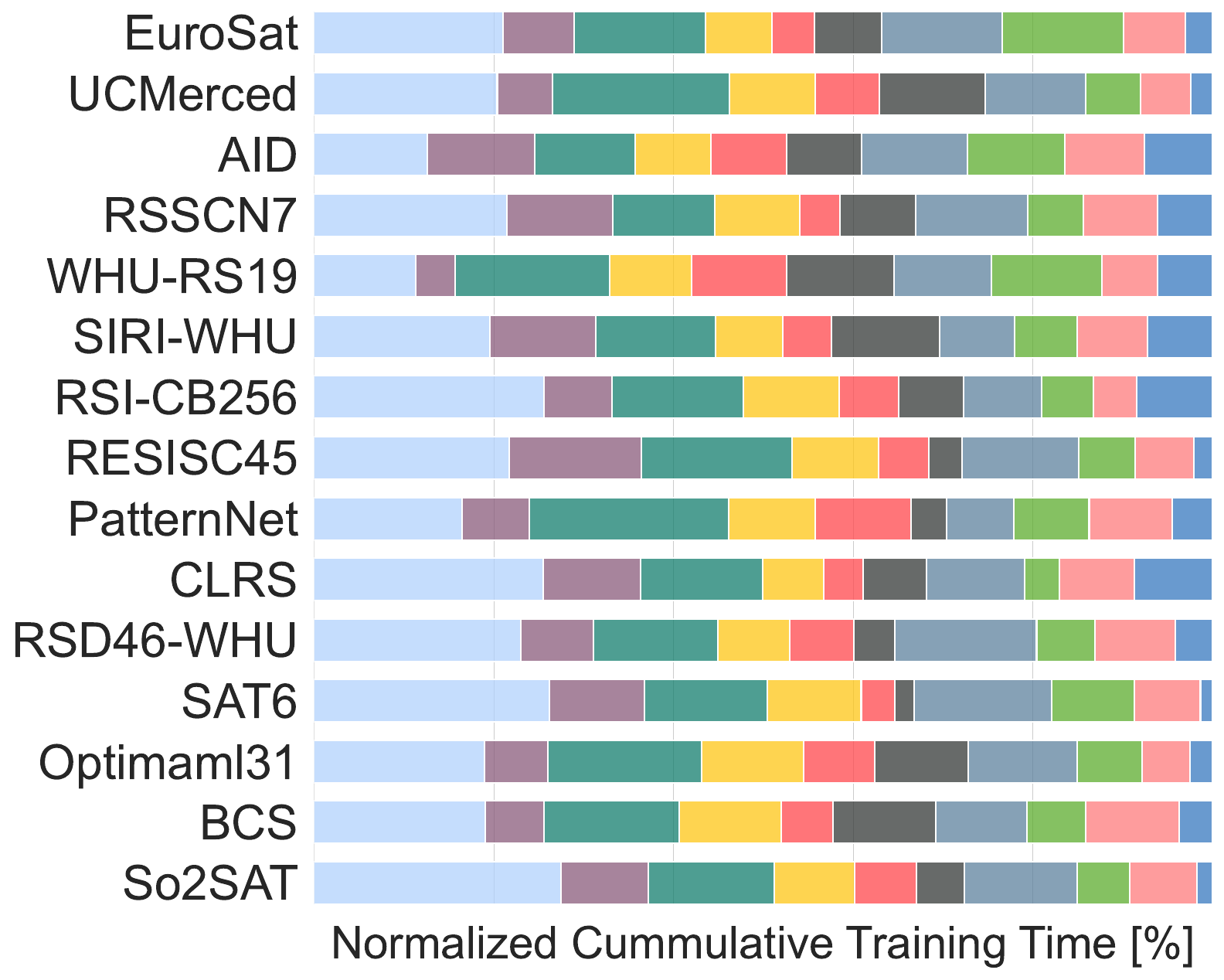}
   \subcaption{MCC tasks (total time)}
    \end{subfigure}
    \hfill
    \begin{subfigure}[t]{0.4\linewidth}
    \includegraphics[width=\linewidth]{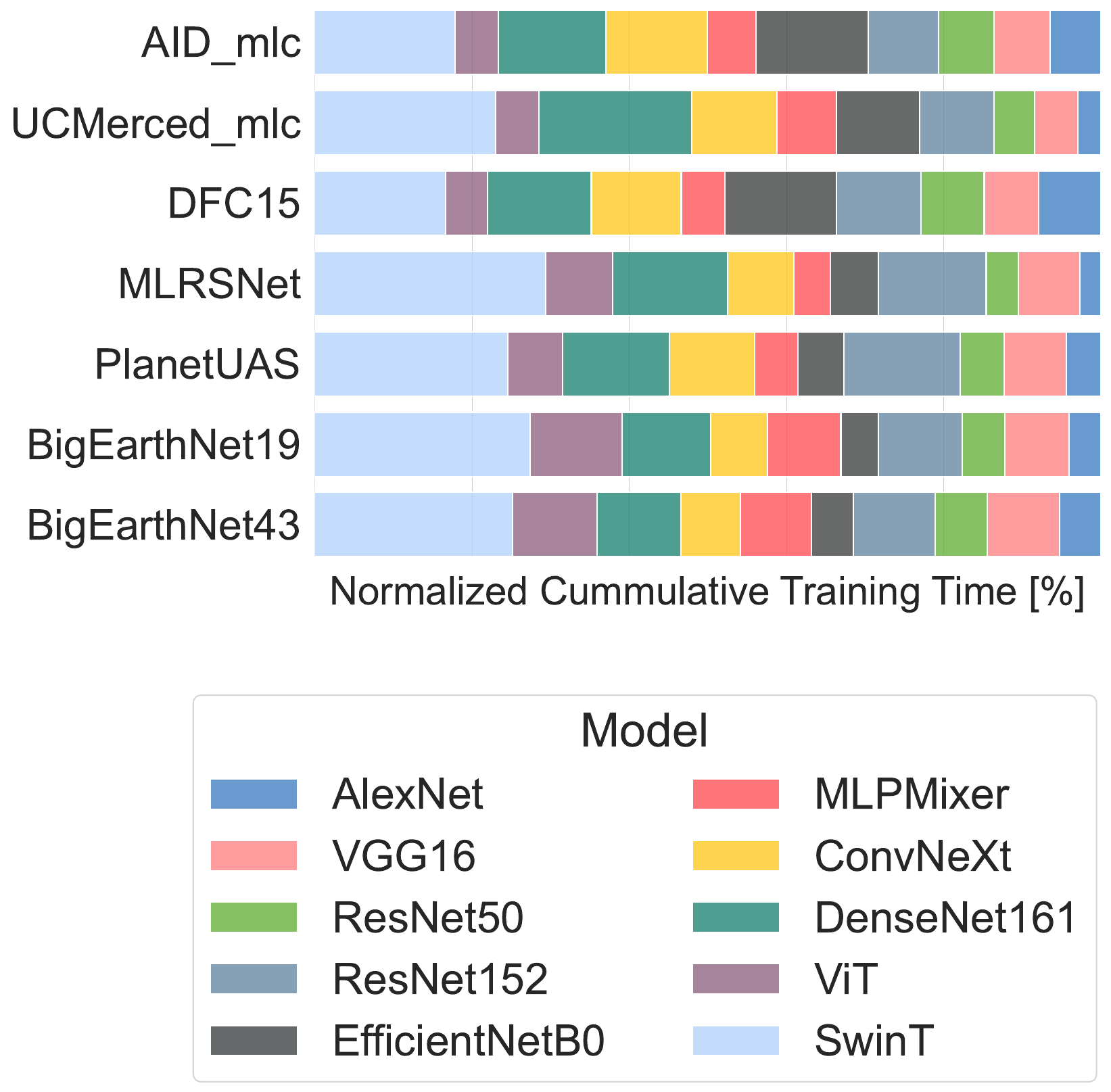}
   \subcaption{MLC tasks (total time)}
    \end{subfigure}
    \hfill
    \caption{\textbf{Total training time} of pre-trained models for each of the \textbf{(a)} MCC and \textbf{(b)} MLC datasts. The training time of each model architecture (denoted with different colors) is depicted as a fraction (\%) of the cumulative training time for each dataset. Furthermore, (c) and (d) illustrate the average time per epoch of each model variant on \textbf{(c)} MCC  and \textbf{(d)} MLC tasks, comparing the \textbf{(red)} pre-trained model variants (from (a) and (b)) to their counterparts \textbf{(blue) }trained from scratch. }
    \label{fig:times_overview_pre-train}
\end{figure*}

Figure~\ref{fig:performance_cost} illustrates the trade-off for the top-3 best performing model architectures overall (as shown in Tables \ref{tab:multiclass_summary_pre-trained} and \ref{tab:multilabel_summary_pre-trained}), DenseNet, ViT, and SwinT; applied to the 22 MCC and MLC tasks. While the performance analyses showed many similarities between these models, the difference between them in terms of training times is much more pronounced. In general, ViT requires less training time than both DenseNets and SwinTs. DenseNets have nearly a quarter of the number of parameters of ViT but achieve almost half fewer FLOPS (floating-point operations per second) than them. For MCC tasks, ViT models generally result in comparable/better predictive performance than DenseNet models and, in many cases, require half the training time. SwinT models, on the other hand, are much more demanding. In almost all cases, training SwinT models takes up to 2-3 times longer than training ViTs and DenseNets. This is also true for MLC tasks, where SwinT models perform the best performance but at the cost of significant training time. These findings further support previous results~\citep{Ze2021SwinV1}, which point out that Swin transformers (the 'small' variant) have slower training and inference performance than Vision Transformers, which have significantly more parameters but achieve considerably more FLOPS. For an extended illustration of these trade-offs, covering all 10 model architectures, see Figure~\ref{fig:full_perf_time} in the Appendix~\ref{appendix:times}.

We can further analyze these trends in training time trends for each model and dataset, as presented in Figure~\ref{fig:times_overview_pre-train}. In particular, Figure~\ref{fig:times_overview_pre-train} illustrates the training (fine-tuning) times of each pre-trained model as a fraction of the cumulative training time of all models summed across all (a) multi-class and (b) multi-label datasets. This shows that, in many cases, ViT models can be trained almost twice as fast as the models of the other best-performing architectures, such as DenseNet and SwinT. The training cost of ViT models is similar to that of EfficientNetB0, ConvNeXt, and MLPMixer, which are efficient but generally perform worse on these tasks. We can also observe that these variants of SwinT models are the slowest to train on all 22 tasks compared to the other architectures. This is also evident when comparing the time for each epoch (see Appendix~\ref{appendix:times}), with SwinT models taking twice longer to train compared to DenseNet161 models, the next slowest architecture. We also observed that fine-tuning pre-trained models almost halves the training time compared to training models from scratch, even though they take about the same time per epoch. Note, however, that we have not accounted for the time required to pre-train each model, which certainly increases the overall training times significantly. This is generally expected behavior but may help in the design and planning of DL pipelines for similar EO. Additional results presenting models' training costs can be found in Appendix~\ref{appendix:times}.

\subsection{A closer look on several tasks}

To better understand the performance of the learned models on the various MCC and MLC tasks, we examine the model decisions in detail, focusing on datasets (and classes) where the models tend to perform poorly. We hypothesize that these cases are related to several overarching issues that often affect the performance of the models: 

\begin{itemize}
 \itemsep0em 
    \item High inter-class similarity between images from different classes;
    \item Many EO image-classification tasks, which are formulated as MCC, are, in fact, MLC problems. In many cases, an image has a single label, but there are more than one classes/concepts present;
    \item Presence of abstract/complex/compound classes within the datasets, can cause many difficulties in detecting useful and consistent patterns; 
    \item Absence of additional spatio-temporal data which captures the dynamics of land-cover changes
\end{itemize}

To investigate these issues, we simultaneously analyze the models' confusion matrices and visualizations of localized activation maps that highlight the distinguishing parts of the image responsible for the model decision. To generate such visualizations, we use Gradient-weighted Class Activation Mapping (GradCAM) \citep{selvaraju2017}, which is typically used to diagnose model predictions for various deep learning architectures \citep{jacobgilpytorchcam}, including Earth Observation applications \citep{papoutsis2022efficient,li2020rs}. GradCAM uses the gradients of the target classes from the last convolutional layer and produces a coarse localization map highlighting important regions in the image for class prediction. In this set of analyses, we select several cases from the datasets considered datasets, especially those containing classes/land types for which the models perform poorly (based on the various evaluation scores, as reported in Appendix~\ref{appendix:data}), and calculate/visualize the corresponding GradCAM maps.

\begin{figure}[ht]
  \centering
  \includegraphics[width=1.0\linewidth]{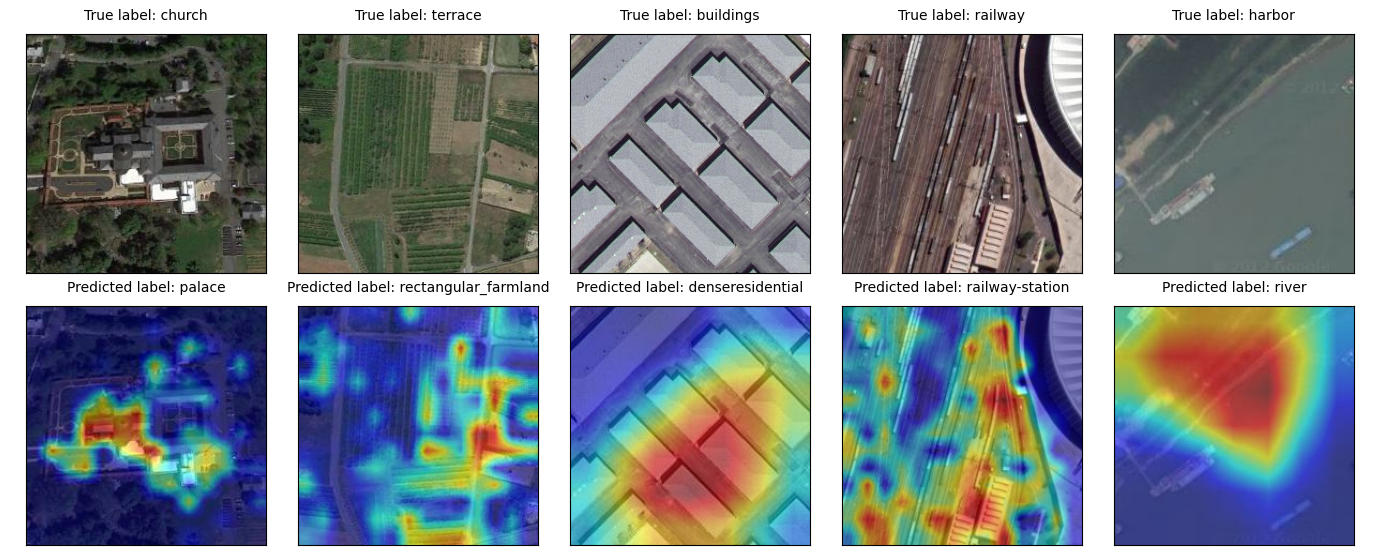}
  \caption{GradCAM visualizations calculated for example images with high inter-class similarity. The input images with their ground-truth label are shown in the first row, while the corresponding activation maps with predicted labels are shown below in the second row. The datasets for the images and the models used to predict the labels are as follows, from left to right: (1) Resisc45, ViT model (2) Resisc45, ViT model (3) UC Merced,  ResNet152 model (4) CLRS, ViT model and (5) SIRI-WHU, ResNet152 model}
  \label{fig:g1_gradcam}
\end{figure}

We start by investigating the inter-class similarities between images assigned to different classes. This is a common problem in practice in many similar EO applications, caused by the presence of visually similar (often indistinguishable) objects in an image. Figure~\ref{fig:g1_gradcam} illustrates this problem using GradCAM activation maps of some sample images with their respective classes/labels from the different datasets.

Our qualitative analyses show that the predictive models are generally able to focus on the correct parts of the images (with distinguishable patterns) but cannot identify the correct object. This is the case, for example, when distinguishing between a 'church' and a 'palace' or a 'terrace' and a 'rectangular farmland', which are visually very similar but semantically different. As expected, the models also struggle with cases where the image labels are also semantically similar, such as in the distinction between 'railway' and 'railway station' or 'river' and 'harbor,' which even a human expert would have difficulty classifying. Similar cases can be further analyzed by examining the confusion matrices. For example, the most challenging dataset, \textit{So2Sat}, contains many such examples (see Figure~\ref{fig:so2sat_confusionmatrix} in Appendix~\ref{app:so2Sat}), which are the reason for the poor overall performance of the models. 

\begin{figure}[ht]
  \centering
  \includegraphics[width=1.0\linewidth]{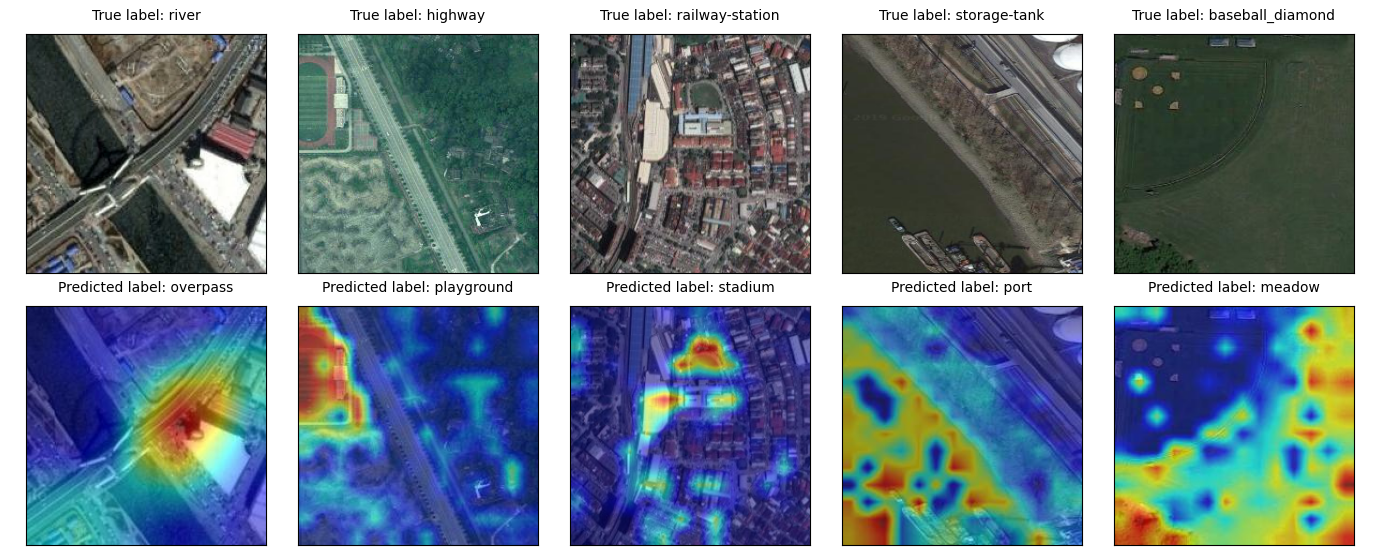}
  \caption{GradCAM visualizations that illustrate the MCC/MLC issues. The input images with their ground-truth label are shown in the first row, while the corresponding activation maps with predicted labels are shown below in the second row. The datasets for the images and the models used to predict the labels are as follows, from left to right: (1) Resisc45, ViT model (2) Resisc45, ViT model (3) UC Merced,  ResNet152 model (4) CLRS, ViT model and (5) SIRI-WHU, ResNet152 model}
  \label{fig:g2_gradcam}
\end{figure}

The second issue that we highlight is related to the fact that, in many cases, multiple land-cover classes/concepts are present in a single image, but the image itself is assigned to only one class - making it a multi-class instead of a multi-label problem. Figure~\ref{fig:g2_gradcam} shows several activation maps illustrating this issue. For example, consider the image-pair on the far left: The image is labeled only as 'river', but we can also see an 'overpass' (a label also present in the dataset) that causes the model to make an 'incorrect' prediction, albeit with a probability of 0.54. Similar situations can be observed for the remaining images: Objects from other classes that are substantially present in an image are detected, thus confusing the models. This, however, shows that the models have been trained well and are performing as expected, but instead of outputting multiple labels (as in a typical MLC setting), they have to choose a single one - which can lead to errors and lower performance. 

To evaluate the third issue, which relates to complex/compound classes, we examine samples with lower F1 scores. Complex/compound classes refer to classes that consist of objects with different physical properties and spatial distribution, making it very difficult to detect useful and consistent patterns. This is also true for abstract classes, where the semantic gap (in terms of labels) is challenging to overcome, which is typically the case when the features learned from the models differ from human interpretation. 

\begin{figure}[ht]
  \centering
  \includegraphics[width=1.0\linewidth]{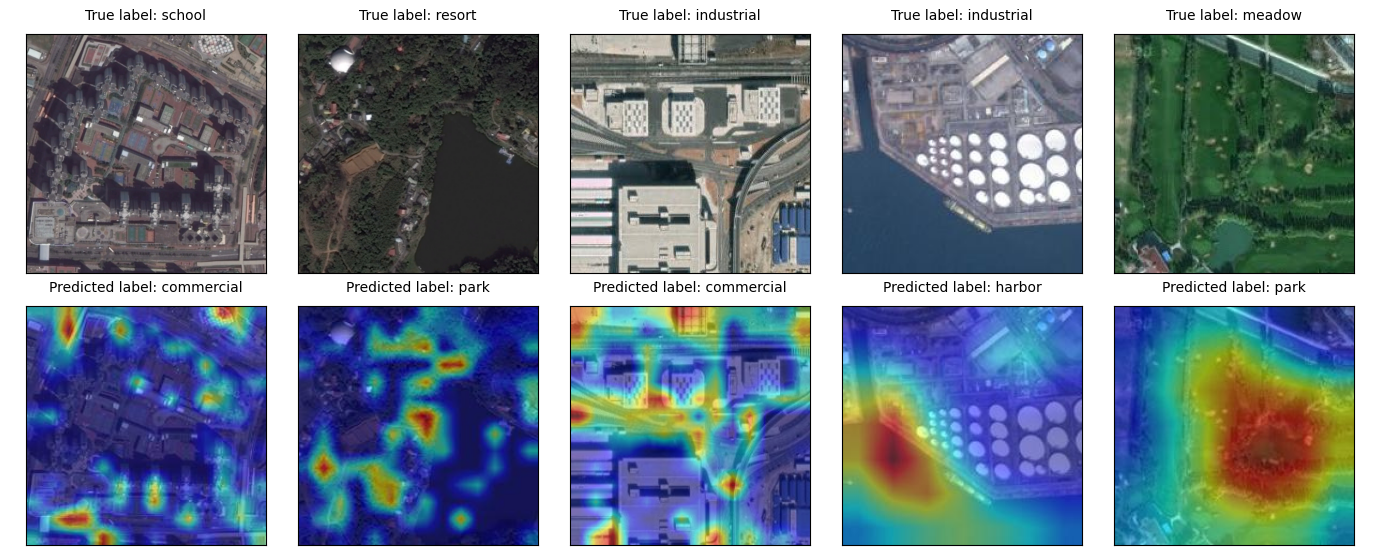}
  \caption{GradCAM visualizations for images with complex/compound classes. The input images with their ground-truth label are shown in the first row, while the corresponding activation maps with predicted labels are shown below in the second row. The datasets for the images and the models used to predict the labels are as follows, from left to right: (1) AID, ViT model (2) AID, ViT model (3) CLRS, ViT model (4) SIRI-WHU, ResNet152 model and (5) SIRI-WHU, ResNet152 model}
  \label{fig:g3_gradcam}
\end{figure}

Figure~\ref{fig:g3_gradcam} illustrates these problems using the respective activation maps. In particular, in the case of \textit{AID} (the two pairs of images on the far left), the model confuses 'school' with 'commercial', the latter being quite vague, for which the semantic gap is not easily dealt with. In the second case, the model has difficulty distinguishing between 'park' and 'resort' (which is also evident in the confusion matrix in Figure~\ref{fig:aid_confusionmatrix} in Appendix~\ref{app:aid}). This could be because these classes consist of common objects but have different spatial distributions. Similar problems can be seen in the cases of \textit{CLRS} and \textit{SIRI-WHU} (the last three image pairs), where labels such as 'industrial' or 'meadow' are confused with labels such as 'commercial/residential/park', which are visually and semantically almost indistinguishable from the ground truth. Similar problems exist in MLC datasets, such as \textit{BigEarthNet}, that contain multiple complex/compound classes. From the evaluation details (see Appendix \ref{app:bigearth}), we can see that complex/compound classes such as 'Complex cultivation patterns', 'Land principally occupied by agriculture, with significant areas of natural vegetation', and 'Industrial and commercial units' have lower F1 scores.

Finally, our analysis shows that for some tasks (such as \textit{So2Sat}), one needs additional and more sophisticated (spatio-temporal) data to improve the performance of the predictive models. For example, the \textit{So2Sat} dataset is very challenging, not only because of the high inter-class similarity but also because of the relatively low spatial resolution of the images. Images labeled 'Open high rise' or 'Compact low rise' are often confused with 'Open middle rise' or 'Lightweight low rise', respectively, which is hardly surprising without additional data that can capture such subtle and often subjective differences. Moreover, in the case of \textit{BigEarthNet}, classes such as 'Permanent crops', 'Coastal wetlands', and 'Natural grassland and sparsely vegetated areas' require additional spatio-temporal data that capture the dynamics caused by frequent land cover changes, making the process of classification more reliable and thus more accurate. \looseness=-1

\section{Conclusions}

We present a systematic review and evaluation of several modern DL architectures applied in Earth Observation. Specifically, we introduce \emph{AiTLAS: Benchmark Arena} -- an \emph{open-source EO benchmark suite} and demonstrate its utility with a comprehensive comparative analysis of models from ten different state-of-the-art DL architectures, comparing them to a variety of multi-class and multi-label image classification tasks from 22 datasets. We compare models trained from scratch and pre-trained models under the same conditions and with the same hardware. We evaluate more than 500 models with different architectures and learning paradigms across tasks from 22 datasets with different sizes and properties. To our knowledge, the evaluation of these different setups (in terms of machine learning tasks, model setups, model architectures, and datasets) makes this the largest and most comprehensive empirical study of deep learning methods applied to EO datasets to date. All of the important details about the study design, the results, and the trained models are freely available. This will contribute to more systematic and rigorous experiments in future work and, more importantly, will enable better usability and faster development of novel approaches. We believe that both this study and the associated repository can serve as a starting point and a guiding design principle for evaluating and documenting machine learning approaches in the different domains of EO. More importantly, we hope that with further involvement from the community, AiTLAS: Benchmark Arena can become a reference point for further studies in this highly active research area.

More broadly, we believe that this work, along with the developed resources, will strongly impact the AI and EO research communities. First, such ready-to-use resources containing trained models, clear experimental designs, and detailed results will facilitate better adoption of sophisticated modeling approaches in the EO community - bringing the EO and AI communities closer together. Second, it demonstrates the FAIRification process of AI4EO resources, i.e., making resources adhere to the FAIR principles (Findable, Accessible, Interoperable, and Reusable \citep{wilkinson2016fair}). Finally, it contributes to the 'Green AI' initiative by saving additional computational overhead. Since all experimental details, especially the trained models, are publicly available -- other experts and researchers can compare, reproduce, and reuse these resources - reducing the need to (repeatedly) run unnecessary experiments.

\section*{Reproducibility}
All the necessary details, in terms of the trained models, model parameters and implementations as well as details on all of the used datasets and their prepossessed versions are available at \url{https://github.com/biasvariancelabs/aitlas-arena}. All the models were trained/fine-tuned on NVIDIA A100-PCIE-40GB GPUs, running CUDA Version 11.5 (\url{www.nvidia.com/en-gb/data-center/a100/}). Note that, we do not host the datasets. To obtain them, please refer to each of the respective studies (referenced in Tables~\ref{table:multiclass} and ~\ref{table:multilabel}) or follow the links provided in our repository. The study was performed using the AiTLAS Toolbox \citep{AitlasArxiv}, a library for exploratory and predictive analysis of satellite imaginary pertaining to different remote-sensing tasks, available at \url{https://aitlas.bvlabs.ai}.

\section*{Acknowledgements}
We acknowledge the support of the European Space Agency ESA through the activity AiTLAS - AI4EO rapid prototyping environment. We thank Sofija Dimitrovska for her thoughtful feedback.

\newpage
{\small
\bibliography{arxiv_main}}

\begin{thebibliography}{105}
\expandafter\ifx\csname natexlab\endcsname\relax\def\natexlab#1{#1}\fi
\providecommand{\url}[1]{\texttt{#1}}
\providecommand{\href}[2]{#2}
\providecommand{\path}[1]{#1}
\providecommand{\DOIprefix}{doi:}
\providecommand{\ArXivprefix}{arXiv:}
\providecommand{\URLprefix}{URL: }
\providecommand{\Pubmedprefix}{pmid:}
\providecommand{\doi}[1]{\href{http://dx.doi.org/#1}{\path{#1}}}
\providecommand{\Pubmed}[1]{\href{pmid:#1}{\path{#1}}}
\providecommand{\bibinfo}[2]{#2}
\ifx\xfnm\relax \def\xfnm[#1]{\unskip,\space#1}\fi
\bibitem[{Khan et~al.(2020)Khan, Sohail, Zahoora, and Qureshi}]{Khan2020}
\bibinfo{author}{A.~Khan}, \bibinfo{author}{A.~Sohail},
  \bibinfo{author}{U.~Zahoora}, \bibinfo{author}{A.~S. Qureshi},
\newblock \bibinfo{title}{A survey of the recent architectures of deep
  convolutional neural networks},
\newblock \bibinfo{journal}{Artificial Intelligence Review}
  \bibinfo{volume}{53} (\bibinfo{year}{2020}) \bibinfo{pages}{5455--5516}.
\bibitem[{Khan et~al.(2021)Khan, Naseer, Hayat, Zamir, Khan, and
  Shah}]{KhanTransformers2022}
\bibinfo{author}{S.~Khan}, \bibinfo{author}{M.~Naseer},
  \bibinfo{author}{M.~Hayat}, \bibinfo{author}{S.~W. Zamir},
  \bibinfo{author}{F.~S. Khan}, \bibinfo{author}{M.~Shah},
\newblock \bibinfo{title}{Transformers in vision: A survey},
\newblock \bibinfo{journal}{ACM Comput. Surv.}  (\bibinfo{year}{2021}).
  \DOIprefix\doi{10.1145/3505244}.
\bibitem[{Ball et~al.(2017)Ball, Anderson, and Chan~Sr.}]{Ball17:jrnl}
\bibinfo{author}{J.~E. Ball}, \bibinfo{author}{D.~T. Anderson},
  \bibinfo{author}{C.~S. Chan~Sr.},
\newblock \bibinfo{title}{{Comprehensive survey of deep learning in remote
  sensing: theories, tools, and challenges for the community}},
\newblock \bibinfo{journal}{Journal of Applied Remote Sensing}
  \bibinfo{volume}{11} (\bibinfo{year}{2017}) \bibinfo{pages}{1 -- 54}.
\bibitem[{Tuia et~al.(2009)Tuia, Ratle, Pacifici, Kanevski, and
  Emery}]{Tuia2009}
\bibinfo{author}{D.~Tuia}, \bibinfo{author}{F.~Ratle},
  \bibinfo{author}{F.~Pacifici}, \bibinfo{author}{M.~F. Kanevski},
  \bibinfo{author}{W.~J. Emery},
\newblock \bibinfo{title}{Active learning methods for remote sensing image
  classification},
\newblock \bibinfo{journal}{IEEE Transactions on Geoscience and Remote Sensing}
  \bibinfo{volume}{47} (\bibinfo{year}{2009}) \bibinfo{pages}{2218--2232}.
  \DOIprefix\doi{10.1109/TGRS.2008.2010404}.
\bibitem[{Li et~al.(2014)Li, Zang, Zhang, Li, and Wu}]{Li2014}
\bibinfo{author}{M.~Li}, \bibinfo{author}{S.~Zang}, \bibinfo{author}{B.~Zhang},
  \bibinfo{author}{S.~Li}, \bibinfo{author}{C.~Wu},
\newblock \bibinfo{title}{A review of remote sensing image classification
  techniques: the role of spatio-contextual information},
\newblock \bibinfo{journal}{European Journal of Remote Sensing}
  \bibinfo{volume}{47} (\bibinfo{year}{2014}) \bibinfo{pages}{389--411}.
  \DOIprefix\doi{10.5721/EuJRS20144723}.
\bibitem[{Blaschke(2010)}]{Blaschke2010ObjectBI}
\bibinfo{author}{T.~Blaschke},
\newblock \bibinfo{title}{Object based image analysis for remote sensing},
\newblock \bibinfo{journal}{Isprs Journal of Photogrammetry and Remote Sensing}
  \bibinfo{volume}{65} (\bibinfo{year}{2010}) \bibinfo{pages}{2--16}.
\bibitem[{Blaschke and Strobl(2001)}]{Blaschke2001WhatsWW}
\bibinfo{author}{T.~Blaschke}, \bibinfo{author}{J.~Strobl},
\newblock \bibinfo{title}{What’s wrong with pixels? some recent developments
  interfacing remote sensing and gis},
\newblock in: \bibinfo{booktitle}{GIS – Zeitschrift für
  Geoinformationssysteme}, \bibinfo{year}{2001}.
\bibitem[{Cheng et~al.(2017)Cheng, Han, and Lu}]{Cheng2017_RSSurvey}
\bibinfo{author}{G.~Cheng}, \bibinfo{author}{J.~Han}, \bibinfo{author}{X.~Lu},
\newblock \bibinfo{title}{Remote sensing image scene classification: Benchmark
  and state of the art},
\newblock \bibinfo{journal}{Proceedings of the IEEE} \bibinfo{volume}{105}
  (\bibinfo{year}{2017}) \bibinfo{pages}{1865--1883}.
  \DOIprefix\doi{10.1109/JPROC.2017.2675998}.
\bibitem[{Yang and Newsam(2010)}]{yang2010uc_merced}
\bibinfo{author}{Y.~Yang}, \bibinfo{author}{S.~Newsam},
\newblock \bibinfo{title}{Bag-of-visual-words and spatial extensions for
  land-use classification},
\newblock in: \bibinfo{booktitle}{Proceedings of the 18th SIGSPATIAL
  International Conference on Advances in Geographic Information Systems},
  \bibinfo{publisher}{Association for Computing Machinery},
  \bibinfo{year}{2010}, p. \bibinfo{pages}{270–279}.
\bibitem[{Marmanis et~al.(2016)Marmanis, Datcu, Esch, and
  Stilla}]{Marmanis2016}
\bibinfo{author}{D.~Marmanis}, \bibinfo{author}{M.~Datcu},
  \bibinfo{author}{T.~Esch}, \bibinfo{author}{U.~Stilla},
\newblock \bibinfo{title}{Deep learning earth observation classification using
  imagenet pretrained networks},
\newblock \bibinfo{journal}{IEEE Geoscience and Remote Sensing Letters}
  \bibinfo{volume}{13} (\bibinfo{year}{2016}) \bibinfo{pages}{105--109}.
  \DOIprefix\doi{10.1109/LGRS.2015.2499239}.
\bibitem[{Chen et~al.(2019)Chen, Chandrasekar, Tan, and Cifelli}]{Chen2019}
\bibinfo{author}{H.~Chen}, \bibinfo{author}{V.~Chandrasekar},
  \bibinfo{author}{H.~Tan}, \bibinfo{author}{R.~Cifelli},
\newblock \bibinfo{title}{Rainfall estimation from ground radar and trmm
  precipitation radar using hybrid deep neural networks},
\newblock \bibinfo{journal}{Geophysical Research Letters} \bibinfo{volume}{46}
  (\bibinfo{year}{2019}) \bibinfo{pages}{10669--10678}.
  \DOIprefix\doi{https://doi.org/10.1029/2019GL084771}.
\bibitem[{Castillo-Navarro et~al.(2022)Castillo-Navarro, Le~Saux, Boulch, and
  Lefèvre}]{Castillo-Navarro2022}
\bibinfo{author}{J.~Castillo-Navarro}, \bibinfo{author}{B.~Le~Saux},
  \bibinfo{author}{A.~Boulch}, \bibinfo{author}{S.~Lefèvre},
\newblock \bibinfo{title}{Energy-based models in earth observation: From
  generation to semisupervised learning},
\newblock \bibinfo{journal}{IEEE Transactions on Geoscience and Remote Sensing}
  \bibinfo{volume}{60} (\bibinfo{year}{2022}) \bibinfo{pages}{1--11}.
  \DOIprefix\doi{10.1109/TGRS.2021.3126428}.
\bibitem[{Wang et~al.(2022)Wang, Albrecht, Braham, Mou, and Zhu}]{Wang2022SSL}
\bibinfo{author}{Y.~Wang}, \bibinfo{author}{C.~M. Albrecht},
  \bibinfo{author}{N.~A.~A. Braham}, \bibinfo{author}{L.~Mou},
  \bibinfo{author}{X.~X. Zhu},
\newblock \bibinfo{title}{Self-supervised learning in remote sensing: A
  review},
\newblock \bibinfo{journal}{CoRR}  (\bibinfo{year}{2022}).
  \DOIprefix\doi{10.48550/ARXIV.2206.13188}.
\bibitem[{Neumann et~al.(2020)Neumann, Pinto, Zhai, and Houlsby}]{Neumann2020}
\bibinfo{author}{M.~Neumann}, \bibinfo{author}{A.~S. Pinto},
  \bibinfo{author}{X.~Zhai}, \bibinfo{author}{N.~Houlsby},
\newblock \bibinfo{title}{Training general representations for remote sensing
  using in-domain knowledge},
\newblock in: \bibinfo{booktitle}{IGARSS 2020 - 2020 IEEE International
  Geoscience and Remote Sensing Symposium}, \bibinfo{year}{2020}, pp.
  \bibinfo{pages}{6730--6733}.
  \DOIprefix\doi{10.1109/IGARSS39084.2020.9324501}.
\bibitem[{Ienco et~al.(2017)Ienco, Gaetano, Dupaquier, and Maurel}]{Ienco2017}
\bibinfo{author}{D.~Ienco}, \bibinfo{author}{R.~Gaetano},
  \bibinfo{author}{C.~Dupaquier}, \bibinfo{author}{P.~Maurel},
\newblock \bibinfo{title}{Land cover classification via multitemporal spatial
  data by deep recurrent neural networks},
\newblock \bibinfo{journal}{IEEE Geoscience and Remote Sensing Letters}
  \bibinfo{volume}{14} (\bibinfo{year}{2017}) \bibinfo{pages}{1685--1689}.
  \DOIprefix\doi{10.1109/LGRS.2017.2728698}.
\bibitem[{Chlingaryan et~al.(2018)Chlingaryan, Sukkarieh, and
  Whelan}]{CHLINGARYAN201861}
\bibinfo{author}{A.~Chlingaryan}, \bibinfo{author}{S.~Sukkarieh},
  \bibinfo{author}{B.~Whelan},
\newblock \bibinfo{title}{Machine learning approaches for crop yield prediction
  and nitrogen status estimation in precision agriculture: A review},
\newblock \bibinfo{journal}{Computers and Electronics in Agriculture}
  \bibinfo{volume}{151} (\bibinfo{year}{2018}) \bibinfo{pages}{61--69}.
  \DOIprefix\doi{https://doi.org/10.1016/j.compag.2018.05.012}.
\bibitem[{Johnson et~al.(2016)Johnson, Hsieh, Cannon, Davidson, and
  Bédard}]{JOHNSON201674}
\bibinfo{author}{M.~D. Johnson}, \bibinfo{author}{W.~W. Hsieh},
  \bibinfo{author}{A.~J. Cannon}, \bibinfo{author}{A.~Davidson},
  \bibinfo{author}{F.~Bédard},
\newblock \bibinfo{title}{Crop yield forecasting on the canadian prairies by
  remotely sensed vegetation indices and machine learning methods},
\newblock \bibinfo{journal}{Agricultural and Forest Meteorology}
  \bibinfo{volume}{218-219} (\bibinfo{year}{2016}) \bibinfo{pages}{74--84}.
  \DOIprefix\doi{https://doi.org/10.1016/j.agrformet.2015.11.003}.
\bibitem[{Xu et~al.(2021)Xu, Yang, Xiong, Li, Huang, Ting, Ying, and
  Lin}]{XU2021_crops}
\bibinfo{author}{J.~Xu}, \bibinfo{author}{J.~Yang}, \bibinfo{author}{X.~Xiong},
  \bibinfo{author}{H.~Li}, \bibinfo{author}{J.~Huang},
  \bibinfo{author}{K.~Ting}, \bibinfo{author}{Y.~Ying},
  \bibinfo{author}{T.~Lin},
\newblock \bibinfo{title}{Towards interpreting multi-temporal deep learning
  models in crop mapping},
\newblock \bibinfo{journal}{Remote Sensing of Environment}
  \bibinfo{volume}{264} (\bibinfo{year}{2021}) \bibinfo{pages}{112599}.
  \DOIprefix\doi{https://doi.org/10.1016/j.rse.2021.112599}.
\bibitem[{Ayhan et~al.(2020)Ayhan, Kwan, Budavari, Kwan, Lu, Perez, Li,
  Skarlatos, and Vlachos}]{Ayhan2020}
\bibinfo{author}{B.~Ayhan}, \bibinfo{author}{C.~Kwan},
  \bibinfo{author}{B.~Budavari}, \bibinfo{author}{L.~Kwan},
  \bibinfo{author}{Y.~Lu}, \bibinfo{author}{D.~Perez}, \bibinfo{author}{J.~Li},
  \bibinfo{author}{D.~Skarlatos}, \bibinfo{author}{M.~Vlachos},
\newblock \bibinfo{title}{Vegetation detection using deep learning and
  conventional methods},
\newblock \bibinfo{journal}{Remote Sensing} \bibinfo{volume}{12}
  (\bibinfo{year}{2020}). \DOIprefix\doi{10.3390/rs12152502}.
\bibitem[{Jo et~al.(2018)Jo, Kim, and Kim}]{Jo2018}
\bibinfo{author}{Y.-H. Jo}, \bibinfo{author}{D.-W. Kim},
  \bibinfo{author}{H.~Kim},
\newblock \bibinfo{title}{Chlorophyll concentration derived from microwave
  remote sensing measurements using artificial neural network algorithm},
\newblock \bibinfo{journal}{Journal of Marine Science and Technology}
  \bibinfo{volume}{26} (\bibinfo{year}{2018}).
  \DOIprefix\doi{10.6119/JMST.2018.02_(1).0004}.
\bibitem[{Shirmard et~al.(2022)Shirmard, Farahbakhsh, Müller, and
  Chandra}]{Shirmard2022}
\bibinfo{author}{H.~Shirmard}, \bibinfo{author}{E.~Farahbakhsh},
  \bibinfo{author}{R.~D. Müller}, \bibinfo{author}{R.~Chandra},
\newblock \bibinfo{title}{A review of machine learning in processing remote
  sensing data for mineral exploration},
\newblock \bibinfo{journal}{Remote Sensing of Environment}
  \bibinfo{volume}{268} (\bibinfo{year}{2022}) \bibinfo{pages}{112750}.
  \DOIprefix\doi{https://doi.org/10.1016/j.rse.2021.112750}.
\bibitem[{Zhang et~al.(2018)Zhang, Zhang, Zhang, Nie, Gui, and
  Que}]{ijerph15051032}
\bibinfo{author}{X.~Zhang}, \bibinfo{author}{Q.~Zhang},
  \bibinfo{author}{G.~Zhang}, \bibinfo{author}{Z.~Nie},
  \bibinfo{author}{Z.~Gui}, \bibinfo{author}{H.~Que},
\newblock \bibinfo{title}{A novel hybrid data-driven model for daily land
  surface temperature forecasting using long short-term memory neural network
  based on ensemble empirical mode decomposition},
\newblock \bibinfo{journal}{International Journal of Environmental Research and
  Public Health} \bibinfo{volume}{15} (\bibinfo{year}{2018}).
\bibitem[{Sadeghi et~al.(2019)Sadeghi, Asanjan, Faridzad, Nguyen, Hsu,
  Sorooshian, and Braithwaite}]{Mojtaba}
\bibinfo{author}{M.~Sadeghi}, \bibinfo{author}{A.~A. Asanjan},
  \bibinfo{author}{M.~Faridzad}, \bibinfo{author}{P.~Nguyen},
  \bibinfo{author}{K.~Hsu}, \bibinfo{author}{S.~Sorooshian},
  \bibinfo{author}{D.~Braithwaite},
\newblock \bibinfo{title}{Persiann-cnn: Precipitation estimation from remotely
  sensed information using artificial neural networks–convolutional neural
  networks},
\newblock \bibinfo{journal}{Journal of Hydrometeorology} \bibinfo{volume}{20}
  (\bibinfo{year}{2019}) \bibinfo{pages}{2273 -- 2289}.
  \DOIprefix\doi{10.1175/JHM-D-19-0110.1}.
\bibitem[{Longbotham et~al.(2012)Longbotham, Chaapel, Bleiler, Padwick, Emery,
  and Pacifici}]{Longbotham}
\bibinfo{author}{N.~Longbotham}, \bibinfo{author}{C.~Chaapel},
  \bibinfo{author}{L.~Bleiler}, \bibinfo{author}{C.~Padwick},
  \bibinfo{author}{W.~J. Emery}, \bibinfo{author}{F.~Pacifici},
\newblock \bibinfo{title}{Very high resolution multiangle urban classification
  analysis},
\newblock \bibinfo{journal}{IEEE Transactions on Geoscience and Remote Sensing}
  \bibinfo{volume}{50} (\bibinfo{year}{2012}) \bibinfo{pages}{1155--1170}.
  \DOIprefix\doi{10.1109/TGRS.2011.2165548}.
\bibitem[{Lv et~al.(2022)Lv, Liu, Benediktsson, and Falco}]{Lv}
\bibinfo{author}{Z.~Lv}, \bibinfo{author}{T.~Liu}, \bibinfo{author}{J.~A.
  Benediktsson}, \bibinfo{author}{N.~Falco},
\newblock \bibinfo{title}{Land cover change detection techniques:
  Very-high-resolution optical images: A review},
\newblock \bibinfo{journal}{IEEE Geoscience and Remote Sensing Magazine}
  \bibinfo{volume}{10} (\bibinfo{year}{2022}) \bibinfo{pages}{44--63}.
  \DOIprefix\doi{10.1109/MGRS.2021.3088865}.
\bibitem[{Huang et~al.(2018)Huang, Zhao, and Song}]{HUANG201873}
\bibinfo{author}{B.~Huang}, \bibinfo{author}{B.~Zhao},
  \bibinfo{author}{Y.~Song},
\newblock \bibinfo{title}{Urban land-use mapping using a deep convolutional
  neural network with high spatial resolution multispectral remote sensing
  imagery},
\newblock \bibinfo{journal}{Remote Sensing of Environment}
  \bibinfo{volume}{214} (\bibinfo{year}{2018}) \bibinfo{pages}{73--86}.
  \DOIprefix\doi{https://doi.org/10.1016/j.rse.2018.04.050}.
\bibitem[{Somrak et~al.(2020)Somrak, Dzeroski, and Kokalj}]{Somrak2020}
\bibinfo{author}{M.~Somrak}, \bibinfo{author}{S.~Dzeroski},
  \bibinfo{author}{Z.~Kokalj},
\newblock \bibinfo{title}{Learning to classify structures in als-derived
  visualizations of ancient maya settlements with {CNN}},
\newblock \bibinfo{journal}{Remote. Sens.} \bibinfo{volume}{12}
  (\bibinfo{year}{2020}) \bibinfo{pages}{2215}.
  \DOIprefix\doi{10.3390/rs12142215}.
\bibitem[{Cheng et~al.(2020)Cheng, Xie, Han, Guo, and Xia}]{Cheng2020}
\bibinfo{author}{G.~Cheng}, \bibinfo{author}{X.~Xie}, \bibinfo{author}{J.~Han},
  \bibinfo{author}{L.~Guo}, \bibinfo{author}{G.-S. Xia},
\newblock \bibinfo{title}{Remote sensing image scene classification meets deep
  learning: Challenges, methods, benchmarks, and opportunities},
\newblock \bibinfo{journal}{IEEE Journal of Selected Topics in Applied Earth
  Observations and Remote Sensing} \bibinfo{volume}{13} (\bibinfo{year}{2020})
  \bibinfo{pages}{3735--3756}. \DOIprefix\doi{10.1109/JSTARS.2020.3005403}.
\bibitem[{Schneider et~al.(2022)Schneider, Bonavita, Geer, Arcucci, Dueben,
  Vitolo, Le~Saux, Demir, and Mathieu}]{Schneider2022}
\bibinfo{author}{R.~Schneider}, \bibinfo{author}{M.~Bonavita},
  \bibinfo{author}{A.~Geer}, \bibinfo{author}{R.~Arcucci},
  \bibinfo{author}{P.~Dueben}, \bibinfo{author}{C.~Vitolo},
  \bibinfo{author}{B.~Le~Saux}, \bibinfo{author}{B.~Demir},
  \bibinfo{author}{P.-P. Mathieu},
\newblock \bibinfo{title}{Esa-ecmwf report on recent progress and research
  directions in machine learning for earth system observation and prediction},
\newblock \bibinfo{journal}{npj Climate and Atmospheric Science}
  \bibinfo{volume}{5} (\bibinfo{year}{2022}) \bibinfo{pages}{51}.
  \DOIprefix\doi{10.1038/s41612-022-00269-z}.
\bibitem[{Papoutsis et~al.(2022)Papoutsis, Bountos, Zavras, Michail, and
  Tryfonopoulos}]{papoutsis2022efficient}
\bibinfo{author}{I.~Papoutsis}, \bibinfo{author}{N.-I. Bountos},
  \bibinfo{author}{A.~Zavras}, \bibinfo{author}{D.~Michail},
  \bibinfo{author}{C.~Tryfonopoulos},
\newblock \bibinfo{title}{Efficient deep learning models for land cover image
  classification},
\newblock \bibinfo{journal}{arXiv:2111.09451}  (\bibinfo{year}{2022}).
\bibitem[{Xia et~al.(2017)Xia, Hu, Hu, Shi, Bai, Zhong, Zhang, and
  Lu}]{xia2017aid}
\bibinfo{author}{G.-S. Xia}, \bibinfo{author}{J.~Hu}, \bibinfo{author}{F.~Hu},
  \bibinfo{author}{B.~Shi}, \bibinfo{author}{X.~Bai},
  \bibinfo{author}{Y.~Zhong}, \bibinfo{author}{L.~Zhang},
  \bibinfo{author}{X.~Lu},
\newblock \bibinfo{title}{{AID}: A benchmark data set for performance
  evaluation of aerial scene classification},
\newblock \bibinfo{journal}{{IEEE} Transactions on Geoscience and Remote
  Sensing} \bibinfo{volume}{55} (\bibinfo{year}{2017})
  \bibinfo{pages}{3965--3981}.
\bibitem[{Zhai et~al.(2019)Zhai, Puigcerver, Kolesnikov, Ruyssen, Riquelme,
  Lucic, Djolonga, Pinto, Neumann, Dosovitskiy, Beyer, Bachem, Tschannen,
  Michalski, Bousquet, Gelly, and Houlsby}]{Zhai2019ALS}
\bibinfo{author}{X.~Zhai}, \bibinfo{author}{J.~Puigcerver},
  \bibinfo{author}{A.~Kolesnikov}, \bibinfo{author}{P.~Ruyssen},
  \bibinfo{author}{C.~Riquelme}, \bibinfo{author}{M.~Lucic},
  \bibinfo{author}{J.~Djolonga}, \bibinfo{author}{A.~S. Pinto},
  \bibinfo{author}{M.~Neumann}, \bibinfo{author}{A.~Dosovitskiy},
  \bibinfo{author}{L.~Beyer}, \bibinfo{author}{O.~Bachem},
  \bibinfo{author}{M.~Tschannen}, \bibinfo{author}{M.~Michalski},
  \bibinfo{author}{O.~Bousquet}, \bibinfo{author}{S.~Gelly},
  \bibinfo{author}{N.~Houlsby},
\newblock \bibinfo{title}{A large-scale study of representation learning with
  the visual task adaptation benchmark},
\newblock \bibinfo{journal}{arXiv:1910.04867}  (\bibinfo{year}{2019}).
\bibitem[{Zhang et~al.(2016)Zhang, Zhang, and Du}]{Zhang2016}
\bibinfo{author}{L.~Zhang}, \bibinfo{author}{L.~Zhang},
  \bibinfo{author}{B.~Du},
\newblock \bibinfo{title}{Deep learning for remote sensing data: A technical
  tutorial on the state of the art},
\newblock \bibinfo{journal}{IEEE Geoscience and Remote Sensing Magazine}
  \bibinfo{volume}{4} (\bibinfo{year}{2016}) \bibinfo{pages}{22--40}.
  \DOIprefix\doi{10.1109/MGRS.2016.2540798}.
\bibitem[{Zhu et~al.(2017)Zhu, Tuia, Mou, Xia, Zhang, Xu, and
  Fraundorfer}]{Zhu2017}
\bibinfo{author}{X.~X. Zhu}, \bibinfo{author}{D.~Tuia},
  \bibinfo{author}{L.~Mou}, \bibinfo{author}{G.-S. Xia},
  \bibinfo{author}{L.~Zhang}, \bibinfo{author}{F.~Xu},
  \bibinfo{author}{F.~Fraundorfer},
\newblock \bibinfo{title}{Deep learning in remote sensing: A comprehensive
  review and list of resources},
\newblock \bibinfo{journal}{IEEE Geoscience and Remote Sensing Magazine}
  \bibinfo{volume}{5} (\bibinfo{year}{2017}) \bibinfo{pages}{8--36}.
  \DOIprefix\doi{10.1109/MGRS.2017.2762307}.
\bibitem[{Stewart et~al.(2021)Stewart, Robinson, Corley, Ortiz, Ferres, and
  Banerjee}]{Stewart2021}
\bibinfo{author}{A.~J. Stewart}, \bibinfo{author}{C.~Robinson},
  \bibinfo{author}{I.~A. Corley}, \bibinfo{author}{A.~Ortiz},
  \bibinfo{author}{J.~M.~L. Ferres}, \bibinfo{author}{A.~Banerjee},
\newblock \bibinfo{title}{Torchgeo: deep learning with geospatial data},
\newblock \bibinfo{journal}{CoRR} \bibinfo{volume}{abs/2111.08872}
  (\bibinfo{year}{2021}). \href{http://arxiv.org/abs/2111.08872}{{\tt
  arXiv:2111.08872}}.
\bibitem[{Sumbul et~al.(2021)Sumbul, de~Wall, Kreuziger, Marcelino, Costa,
  Benevides, Caetano, Demir, and Markl}]{Sumbul2021BigearthnetAL}
\bibinfo{author}{G.~Sumbul}, \bibinfo{author}{A.~de~Wall},
  \bibinfo{author}{T.~Kreuziger}, \bibinfo{author}{F.~Marcelino},
  \bibinfo{author}{H.~Costa}, \bibinfo{author}{P.~Benevides},
  \bibinfo{author}{M.~Caetano}, \bibinfo{author}{B.~Demir},
  \bibinfo{author}{V.~Markl},
\newblock \bibinfo{title}{{BigEarthNet}-{MM}: A large-scale, multimodal,
  multilabel benchmark archive for remote sensing image classification and
  retrieval [software and data sets]},
\newblock \bibinfo{journal}{{IEEE} Geoscience and Remote Sensing Magazine}
  \bibinfo{volume}{9} (\bibinfo{year}{2021}) \bibinfo{pages}{174--180}.
\bibitem[{Dimitrovski et~al.(2022)Dimitrovski, Kitanovski, Panov, Simidjievski,
  and Kocev}]{AitlasArxiv}
\bibinfo{author}{I.~Dimitrovski}, \bibinfo{author}{I.~Kitanovski},
  \bibinfo{author}{P.~Panov}, \bibinfo{author}{N.~Simidjievski},
  \bibinfo{author}{D.~Kocev},
\newblock \bibinfo{title}{Aitlas: Artificial intelligence toolbox for earth
  observation},
\newblock \bibinfo{journal}{CoRR} \bibinfo{volume}{abs/2201.08789}
  (\bibinfo{year}{2022}). \href{http://arxiv.org/abs/2201.08789}{{\tt
  arXiv:2201.08789}}.
\bibitem[{Paszke et~al.(2019)Paszke, Gross, Massa, Lerer, Bradbury, Chanan,
  Killeen, Lin, Gimelshein, Antiga, Desmaison, Kopf, Yang, DeVito, Raison,
  Tejani, Chilamkurthy, Steiner, Fang, Bai, and Chintala}]{pytorch}
\bibinfo{author}{A.~Paszke}, \bibinfo{author}{S.~Gross},
  \bibinfo{author}{F.~Massa}, \bibinfo{author}{A.~Lerer},
  \bibinfo{author}{J.~Bradbury}, \bibinfo{author}{G.~Chanan},
  \bibinfo{author}{T.~Killeen}, \bibinfo{author}{Z.~Lin},
  \bibinfo{author}{N.~Gimelshein}, \bibinfo{author}{L.~Antiga},
  \bibinfo{author}{A.~Desmaison}, \bibinfo{author}{A.~Kopf},
  \bibinfo{author}{E.~Yang}, \bibinfo{author}{Z.~DeVito},
  \bibinfo{author}{M.~Raison}, \bibinfo{author}{A.~Tejani},
  \bibinfo{author}{S.~Chilamkurthy}, \bibinfo{author}{B.~Steiner},
  \bibinfo{author}{L.~Fang}, \bibinfo{author}{J.~Bai},
  \bibinfo{author}{S.~Chintala},
\newblock \bibinfo{title}{Pytorch: An imperative style, high-performance deep
  learning library},
\newblock in: \bibinfo{booktitle}{Advances in Neural Information Processing
  Systems 32}, \bibinfo{publisher}{Curran Associates, Inc.},
  \bibinfo{year}{2019}, pp. \bibinfo{pages}{8024--8035}.
\bibitem[{Dosovitskiy et~al.(2020)Dosovitskiy, Beyer, Kolesnikov, Weissenborn,
  Zhai, Unterthiner, Dehghani, Minderer, Heigold, Gelly
  et~al.}]{dosovitskiy2020image}
\bibinfo{author}{A.~Dosovitskiy}, \bibinfo{author}{L.~Beyer},
  \bibinfo{author}{A.~Kolesnikov}, \bibinfo{author}{D.~Weissenborn},
  \bibinfo{author}{X.~Zhai}, \bibinfo{author}{T.~Unterthiner},
  \bibinfo{author}{M.~Dehghani}, \bibinfo{author}{M.~Minderer},
  \bibinfo{author}{G.~Heigold}, \bibinfo{author}{S.~Gelly}, et~al.,
\newblock \bibinfo{title}{An image is worth 16x16 words: Transformers for image
  recognition at scale},
\newblock \bibinfo{journal}{arXiv preprint arXiv:2010.11929}
  (\bibinfo{year}{2020}).
\bibitem[{Liu et~al.(2022)Liu, Hu, Lin, Yao, Xie, Wei, Ning, Cao, Zhang, Dong,
  Wei, and Guo}]{ze2022swinv2}
\bibinfo{author}{Z.~Liu}, \bibinfo{author}{H.~Hu}, \bibinfo{author}{Y.~Lin},
  \bibinfo{author}{Z.~Yao}, \bibinfo{author}{Z.~Xie}, \bibinfo{author}{Y.~Wei},
  \bibinfo{author}{J.~Ning}, \bibinfo{author}{Y.~Cao},
  \bibinfo{author}{Z.~Zhang}, \bibinfo{author}{L.~Dong},
  \bibinfo{author}{F.~Wei}, \bibinfo{author}{B.~Guo},
\newblock \bibinfo{title}{Swin transformer v2: Scaling up capacity and
  resolution},
\newblock in: \bibinfo{booktitle}{2022 IEEE/CVF Conference on Computer Vision
  and Pattern Recognition (CVPR)}, \bibinfo{year}{2022}, pp.
  \bibinfo{pages}{11999--12009}. \DOIprefix\doi{10.1109/CVPR52688.2022.01170}.
\bibitem[{Tsoumakas and Katakis(2009)}]{tsoumakas2009}
\bibinfo{author}{G.~Tsoumakas}, \bibinfo{author}{I.~Katakis},
\newblock \bibinfo{title}{Multi-label classification: An overview},
\newblock \bibinfo{journal}{International Journal of Data Warehousing and
  Mining} \bibinfo{volume}{3} (\bibinfo{year}{2009}) \bibinfo{pages}{1--13}.
\bibitem[{Xia et~al.(2010)Xia, Yang, Delon, Gousseau, Sun, and
  Maître}]{xia2010whurs19}
\bibinfo{author}{G.-S. Xia}, \bibinfo{author}{W.~Yang},
  \bibinfo{author}{J.~Delon}, \bibinfo{author}{Y.~Gousseau},
  \bibinfo{author}{H.~Sun}, \bibinfo{author}{H.~Maître},
\newblock \bibinfo{title}{Structural high-resolution satellite image indexing},
\newblock \bibinfo{journal}{International Archives of the Photogrammetry,
  Remote Sensing and Spatial Information Sciences - ISPRS Archives}
  \bibinfo{volume}{38} (\bibinfo{year}{2010}).
\bibitem[{Helber et~al.(2019)Helber, Bischke, Dengel, and
  Borth}]{helber2019eurosat}
\bibinfo{author}{P.~Helber}, \bibinfo{author}{B.~Bischke},
  \bibinfo{author}{A.~Dengel}, \bibinfo{author}{D.~Borth},
\newblock \bibinfo{title}{Eurosat: A novel dataset and deep learning benchmark
  for land use and land cover classification},
\newblock \bibinfo{journal}{IEEE Journal of Selected Topics in Applied Earth
  Observations and Remote Sensing}  (\bibinfo{year}{2019}).
\bibitem[{Zhou et~al.(2018)Zhou, Newsam, Li, and Shao}]{zhou2018patternnet}
\bibinfo{author}{W.~Zhou}, \bibinfo{author}{S.~Newsam},
  \bibinfo{author}{C.~Li}, \bibinfo{author}{Z.~Shao},
\newblock \bibinfo{title}{Patternnet: A benchmark dataset for performance
  evaluation of remote sensing image retrieval},
\newblock \bibinfo{journal}{ISPRS journal of photogrammetry and remote sensing}
  \bibinfo{volume}{145} (\bibinfo{year}{2018}) \bibinfo{pages}{197--209}.
\bibitem[{Li et~al.(2020)Li, Dou, Tao, Wu, Chen, Peng, Deng, and
  Zhao}]{haifeng2020rsicb256}
\bibinfo{author}{H.~Li}, \bibinfo{author}{X.~Dou}, \bibinfo{author}{C.~Tao},
  \bibinfo{author}{Z.~Wu}, \bibinfo{author}{J.~Chen},
  \bibinfo{author}{J.~Peng}, \bibinfo{author}{M.~Deng},
  \bibinfo{author}{L.~Zhao},
\newblock \bibinfo{title}{Rsi-cb: A large-scale remote sensing image
  classification benchmark using crowdsourced data},
\newblock \bibinfo{journal}{Sensors} \bibinfo{volume}{20}
  (\bibinfo{year}{2020}) \bibinfo{pages}{1594}.
  \DOIprefix\doi{doi.org/10.3390/s20061594}.
\bibitem[{Zou et~al.(2015)Zou, Ni, Zhang, and Wang}]{Zou2015RSSCN7}
\bibinfo{author}{Q.~Zou}, \bibinfo{author}{L.~Ni}, \bibinfo{author}{T.~Zhang},
  \bibinfo{author}{Q.~Wang},
\newblock \bibinfo{title}{Deep learning based feature selection for remote
  sensing scene classification},
\newblock \bibinfo{journal}{IEEE Geoscience and Remote Sensing Letters}
  \bibinfo{volume}{12} (\bibinfo{year}{2015}) \bibinfo{pages}{2321--2325}.
  \DOIprefix\doi{10.1109/LGRS.2015.2475299}.
\bibitem[{Basu et~al.(2015)Basu, Ganguly, Mukhopadhyay, DiBiano, Karki, and
  Nemani}]{Basu2015sat6}
\bibinfo{author}{S.~Basu}, \bibinfo{author}{S.~Ganguly},
  \bibinfo{author}{S.~Mukhopadhyay}, \bibinfo{author}{R.~DiBiano},
  \bibinfo{author}{M.~Karki}, \bibinfo{author}{R.~Nemani},
\newblock \bibinfo{title}{Deepsat: A learning framework for satellite imagery},
\newblock in: \bibinfo{booktitle}{Proceedings of the 23rd SIGSPATIAL
  International Conference on Advances in Geographic Information Systems},
  SIGSPATIAL '15, \bibinfo{publisher}{Association for Computing Machinery},
  \bibinfo{year}{2015}.
\bibitem[{Zhu et~al.(2016)Zhu, Zhong, Zhao, Xia, and Zhang}]{zhu2016siriwhu}
\bibinfo{author}{Q.~Zhu}, \bibinfo{author}{Y.~Zhong},
  \bibinfo{author}{B.~Zhao}, \bibinfo{author}{G.-S. Xia},
  \bibinfo{author}{L.~Zhang},
\newblock \bibinfo{title}{Bag-of-visual-words scene classifier with local and
  global features for high spatial resolution remote sensing imagery},
\newblock \bibinfo{journal}{IEEE Geoscience and Remote Sensing Letters}
  \bibinfo{volume}{13} (\bibinfo{year}{2016}) \bibinfo{pages}{747--751}.
\bibitem[{Li et~al.(2020)Li, Jiang, Gu, Peng, Li, Hong, and
  Tao}]{haifeng2020clrs}
\bibinfo{author}{H.~Li}, \bibinfo{author}{H.~Jiang}, \bibinfo{author}{X.~Gu},
  \bibinfo{author}{J.~Peng}, \bibinfo{author}{W.~Li},
  \bibinfo{author}{L.~Hong}, \bibinfo{author}{C.~Tao},
\newblock \bibinfo{title}{Clrs: Continual learning benchmark for remote sensing
  image scene classification},
\newblock \bibinfo{journal}{Sensors} \bibinfo{volume}{20}
  (\bibinfo{year}{2020}).
\bibitem[{Long et~al.(2017)Long, Gong, Xiao, and Liu}]{Yang2017rsd46whu}
\bibinfo{author}{Y.~Long}, \bibinfo{author}{Y.~Gong},
  \bibinfo{author}{Z.~Xiao}, \bibinfo{author}{Q.~Liu},
\newblock \bibinfo{title}{Accurate object localization in remote sensing images
  based on convolutional neural networks},
\newblock \bibinfo{journal}{IEEE Transactions on Geoscience and Remote Sensing}
  \bibinfo{volume}{55} (\bibinfo{year}{2017}) \bibinfo{pages}{2486--2498}.
\bibitem[{Wang et~al.(2019)Wang, Liu, Chanussot, and Li}]{qi2019optimal31}
\bibinfo{author}{Q.~Wang}, \bibinfo{author}{S.~Liu},
  \bibinfo{author}{J.~Chanussot}, \bibinfo{author}{X.~Li},
\newblock \bibinfo{title}{Scene classification with recurrent attention of vhr
  remote sensing images},
\newblock \bibinfo{journal}{IEEE Transactions on Geoscience and Remote Sensing}
  \bibinfo{volume}{57} (\bibinfo{year}{2019}) \bibinfo{pages}{1155--1167}.
\bibitem[{Penatti et~al.(2015)Penatti, Nogueira, and
  Dos~Santos}]{penatti2015deep}
\bibinfo{author}{O.~A. Penatti}, \bibinfo{author}{K.~Nogueira},
  \bibinfo{author}{J.~A. Dos~Santos},
\newblock \bibinfo{title}{Do deep features generalize from everyday objects to
  remote sensing and aerial scenes domains?},
\newblock in: \bibinfo{booktitle}{Proceedings of the IEEE conference on
  computer vision and pattern recognition workshops}, \bibinfo{year}{2015}, pp.
  \bibinfo{pages}{44--51}.
\bibitem[{Zhu et~al.(2020)Zhu, Hu, Qiu, Shi, Kang, Mou, Bagheri, Haberle, Hua,
  Huang, Hughes, Li, Sun, Zhang, Han, Schmitt, and Wang}]{Zhu2020So2Sat}
\bibinfo{author}{X.~X. Zhu}, \bibinfo{author}{J.~Hu}, \bibinfo{author}{C.~Qiu},
  \bibinfo{author}{Y.~Shi}, \bibinfo{author}{J.~Kang},
  \bibinfo{author}{L.~Mou}, \bibinfo{author}{H.~Bagheri},
  \bibinfo{author}{M.~Haberle}, \bibinfo{author}{Y.~Hua},
  \bibinfo{author}{R.~Huang}, \bibinfo{author}{L.~Hughes},
  \bibinfo{author}{H.~Li}, \bibinfo{author}{Y.~Sun},
  \bibinfo{author}{G.~Zhang}, \bibinfo{author}{S.~Han},
  \bibinfo{author}{M.~Schmitt}, \bibinfo{author}{Y.~Wang},
\newblock \bibinfo{title}{So2sat lcz42: A benchmark data set for the
  classification of global local climate zones [software and data sets]},
\newblock \bibinfo{journal}{IEEE Geoscience and Remote Sensing Magazine}
  \bibinfo{volume}{8} (\bibinfo{year}{2020}) \bibinfo{pages}{76--89}.
\bibitem[{Chaudhuri et~al.(2018)Chaudhuri, Demir, Chaudhuri, and
  Bruzzone}]{chaudhuri2018ucmerced}
\bibinfo{author}{B.~Chaudhuri}, \bibinfo{author}{B.~Demir},
  \bibinfo{author}{S.~Chaudhuri}, \bibinfo{author}{L.~Bruzzone},
\newblock \bibinfo{title}{Multilabel remote sensing image retrieval using a
  semisupervised graph-theoretic method},
\newblock \bibinfo{journal}{IEEE Transactions on Geoscience and Remote Sensing}
  \bibinfo{volume}{56} (\bibinfo{year}{2018}) \bibinfo{pages}{1144--1158}.
\bibitem[{Qi et~al.(2020)Qi, Zhu, Wang, Zhang, Peng, Wu, Chen, Zhao, Zang, and
  Mathiopoulos}]{qi2020mlrsnet}
\bibinfo{author}{X.~Qi}, \bibinfo{author}{P.~Zhu}, \bibinfo{author}{Y.~Wang},
  \bibinfo{author}{L.~Zhang}, \bibinfo{author}{J.~Peng},
  \bibinfo{author}{M.~Wu}, \bibinfo{author}{J.~Chen},
  \bibinfo{author}{X.~Zhao}, \bibinfo{author}{N.~Zang}, \bibinfo{author}{P.~T.
  Mathiopoulos},
\newblock \bibinfo{title}{Mlrsnet: A multi-label high spatial resolution remote
  sensing dataset for semantic scene understanding},
\newblock \bibinfo{journal}{ISPRS Journal of Photogrammetry and Remote Sensing}
  \bibinfo{volume}{169} (\bibinfo{year}{2020}) \bibinfo{pages}{337--350}.
\bibitem[{Hua et~al.(2019)Hua, Mou, and Zhu}]{hua2019dfc15}
\bibinfo{author}{Y.~Hua}, \bibinfo{author}{L.~Mou}, \bibinfo{author}{X.~X.
  Zhu},
\newblock \bibinfo{title}{Recurrently exploring class-wise attention in a
  hybrid convolutional and bidirectional {LSTM} network for multi-label aerial
  image classification},
\newblock \bibinfo{journal}{ISPRS Journal of Photogrammetry and Remote Sensing}
  \bibinfo{volume}{149} (\bibinfo{year}{2019}) \bibinfo{pages}{188--199}.
\bibitem[{Hua et~al.(2020)Hua, Mou, and Zhu}]{hua2020aidmultilabel}
\bibinfo{author}{Y.~Hua}, \bibinfo{author}{L.~Mou}, \bibinfo{author}{X.~X.
  Zhu},
\newblock \bibinfo{title}{Relation network for multilabel aerial image
  classification},
\newblock \bibinfo{journal}{IEEE Transactions on Geoscience and Remote Sensing}
  \bibinfo{volume}{58} (\bibinfo{year}{2020}) \bibinfo{pages}{4558--4572}.
\bibitem[{Kaggle(2022)}]{webplanetuas}
\bibinfo{author}{Kaggle}, \bibinfo{title}{Planet: Understanding the amazon from
  space}, \bibinfo{year}{2022}. \URLprefix
  \url{https://www.kaggle.com/competitions/planet-understanding-the-amazon-from-space},
  \bibinfo{note}{last accessed 21 May 2022}.
\bibitem[{Sumbul et~al.(2019)Sumbul, Charfuelan, Demir, and
  Markl}]{Sumbul2019BigearthnetAL}
\bibinfo{author}{G.~Sumbul}, \bibinfo{author}{M.~Charfuelan},
  \bibinfo{author}{B.~Demir}, \bibinfo{author}{V.~Markl},
\newblock \bibinfo{title}{Bigearthnet: A large-scale benchmark archive for
  remote sensing image understanding},
\newblock \bibinfo{journal}{IGARSS 2019 - 2019 IEEE International Geoscience
  and Remote Sensing Symposium}  (\bibinfo{year}{2019})
  \bibinfo{pages}{5901--5904}.
\bibitem[{Krizhevsky et~al.(2012)Krizhevsky, Sutskever, and
  Hinton}]{krizhevsky2012imagenet}
\bibinfo{author}{A.~Krizhevsky}, \bibinfo{author}{I.~Sutskever},
  \bibinfo{author}{G.~E. Hinton},
\newblock \bibinfo{title}{Imagenet classification with deep convolutional
  neural networks},
\newblock \bibinfo{journal}{Advances in neural information processing systems}
  \bibinfo{volume}{25} (\bibinfo{year}{2012}) \bibinfo{pages}{1097--1105}.
\bibitem[{Simonyan and Zisserman(2014)}]{simonyan2014very}
\bibinfo{author}{K.~Simonyan}, \bibinfo{author}{A.~Zisserman},
\newblock \bibinfo{title}{Very deep convolutional networks for large-scale
  image recognition},
\newblock \bibinfo{journal}{arXiv preprint arXiv:1409.1556}
  (\bibinfo{year}{2014}).
\bibitem[{He et~al.(2016)He, Zhang, Ren, and Sun}]{he2016deep}
\bibinfo{author}{K.~He}, \bibinfo{author}{X.~Zhang}, \bibinfo{author}{S.~Ren},
  \bibinfo{author}{J.~Sun},
\newblock \bibinfo{title}{Deep residual learning for image recognition},
\newblock in: \bibinfo{booktitle}{Proceedings of the IEEE conference on
  computer vision and pattern recognition}, \bibinfo{year}{2016}, pp.
  \bibinfo{pages}{770--778}.
\bibitem[{Huang et~al.(2017)Huang, Liu, Van Der~Maaten, and
  Weinberger}]{huang2017densely}
\bibinfo{author}{G.~Huang}, \bibinfo{author}{Z.~Liu}, \bibinfo{author}{L.~Van
  Der~Maaten}, \bibinfo{author}{K.~Q. Weinberger},
\newblock \bibinfo{title}{Densely connected convolutional networks},
\newblock in: \bibinfo{booktitle}{Proceedings of the IEEE conference on
  computer vision and pattern recognition}, \bibinfo{year}{2017}, pp.
  \bibinfo{pages}{4700--4708}.
\bibitem[{Tan and Le(2019)}]{tan2019efficientnet}
\bibinfo{author}{M.~Tan}, \bibinfo{author}{Q.~Le},
\newblock \bibinfo{title}{Efficientnet: Rethinking model scaling for
  convolutional neural networks},
\newblock in: \bibinfo{booktitle}{International conference on machine
  learning}, \bibinfo{organization}{PMLR}, \bibinfo{year}{2019}, pp.
  \bibinfo{pages}{6105--6114}.
\bibitem[{Liu et~al.(2022)Liu, Mao, Wu, Feichtenhofer, Darrell, and
  Xie}]{liu2022convnet}
\bibinfo{author}{Z.~Liu}, \bibinfo{author}{H.~Mao}, \bibinfo{author}{C.-Y. Wu},
  \bibinfo{author}{C.~Feichtenhofer}, \bibinfo{author}{T.~Darrell},
  \bibinfo{author}{S.~Xie},
\newblock \bibinfo{title}{A convnet for the 2020s},
\newblock \bibinfo{journal}{arXiv preprint arXiv:2201.03545}
  (\bibinfo{year}{2022}).
\bibitem[{Tolstikhin et~al.(2021)Tolstikhin, Houlsby, Kolesnikov, Beyer, Zhai,
  Unterthiner, Yung, Steiner, Keysers, Uszkoreit et~al.}]{tolstikhin2021mlp}
\bibinfo{author}{I.~O. Tolstikhin}, \bibinfo{author}{N.~Houlsby},
  \bibinfo{author}{A.~Kolesnikov}, \bibinfo{author}{L.~Beyer},
  \bibinfo{author}{X.~Zhai}, \bibinfo{author}{T.~Unterthiner},
  \bibinfo{author}{J.~Yung}, \bibinfo{author}{A.~Steiner},
  \bibinfo{author}{D.~Keysers}, \bibinfo{author}{J.~Uszkoreit}, et~al.,
\newblock \bibinfo{title}{Mlp-mixer: An all-mlp architecture for vision},
\newblock \bibinfo{journal}{Advances in Neural Information Processing Systems}
  \bibinfo{volume}{34} (\bibinfo{year}{2021}).
\bibitem[{Marcel and Rodriguez(2010)}]{marcel2010torchvision}
\bibinfo{author}{S.~Marcel}, \bibinfo{author}{Y.~Rodriguez},
\newblock \bibinfo{title}{Torchvision the machine-vision package of torch},
\newblock in: \bibinfo{booktitle}{Proceedings of the 18th ACM international
  conference on Multimedia}, \bibinfo{year}{2010}, pp.
  \bibinfo{pages}{1485--1488}.
\bibitem[{Wightman(2019)}]{rw2019timm}
\bibinfo{author}{R.~Wightman}, \bibinfo{title}{Pytorch image models},
  \bibinfo{howpublished}{\url{https://github.com/rwightman/pytorch-image-models}},
  \bibinfo{year}{2019}. \DOIprefix\doi{10.5281/zenodo.4414861}.
\bibitem[{Weng et~al.(2017)Weng, Mao, Lin, and Guo}]{Weng2017}
\bibinfo{author}{Q.~Weng}, \bibinfo{author}{Z.~Mao}, \bibinfo{author}{J.~Lin},
  \bibinfo{author}{W.~Guo},
\newblock \bibinfo{title}{Land-use classification via extreme learning
  classifier based on deep convolutional features},
\newblock \bibinfo{journal}{IEEE Geoscience and Remote Sensing Letters}
  \bibinfo{volume}{14} (\bibinfo{year}{2017}) \bibinfo{pages}{704--708}.
  \DOIprefix\doi{10.1109/LGRS.2017.2672643}.
\bibitem[{Castelluccio et~al.(2015)Castelluccio, Poggi, Sansone, and
  Verdoliva}]{Castelluccio2015}
\bibinfo{author}{M.~Castelluccio}, \bibinfo{author}{G.~Poggi},
  \bibinfo{author}{C.~Sansone}, \bibinfo{author}{L.~Verdoliva},
\newblock \bibinfo{title}{Land use classification in remote sensing images by
  convolutional neural networks},
\newblock \bibinfo{journal}{CoRR}  (\bibinfo{year}{2015}).
  \DOIprefix\doi{10.48550/ARXIV.1508.00092}.
\bibitem[{Han et~al.(2017)Han, Zhong, Cao, and Zhang}]{Han2017}
\bibinfo{author}{X.~Han}, \bibinfo{author}{Y.~Zhong}, \bibinfo{author}{L.~Cao},
  \bibinfo{author}{L.~Zhang},
\newblock \bibinfo{title}{Pre-trained alexnet architecture with pyramid pooling
  and supervision for high spatial resolution remote sensing image scene
  classification},
\newblock \bibinfo{journal}{Remote Sensing} \bibinfo{volume}{9}
  (\bibinfo{year}{2017}). \DOIprefix\doi{10.3390/rs9080848}.
\bibitem[{Kang et~al.(2018)Kang, K{\"o}rner, Wang, Taubenb{\"o}ck, and
  Zhu}]{kang2018building}
\bibinfo{author}{J.~Kang}, \bibinfo{author}{M.~K{\"o}rner},
  \bibinfo{author}{Y.~Wang}, \bibinfo{author}{H.~Taubenb{\"o}ck},
  \bibinfo{author}{X.~X. Zhu},
\newblock \bibinfo{title}{Building instance classification using street view
  images},
\newblock \bibinfo{journal}{ISPRS journal of photogrammetry and remote sensing}
  \bibinfo{volume}{145} (\bibinfo{year}{2018}) \bibinfo{pages}{44--59}.
\bibitem[{Hu et~al.(2015)Hu, Xia, Hu, and Zhang}]{Hu2015}
\bibinfo{author}{F.~Hu}, \bibinfo{author}{G.-S. Xia}, \bibinfo{author}{J.~Hu},
  \bibinfo{author}{L.~Zhang},
\newblock \bibinfo{title}{Transferring deep convolutional neural networks for
  the scene classification of high-resolution remote sensing imagery},
\newblock \bibinfo{journal}{Remote Sensing} \bibinfo{volume}{7}
  (\bibinfo{year}{2015}) \bibinfo{pages}{14680--14707}.
  \DOIprefix\doi{10.3390/rs71114680}.
\bibitem[{Goodfellow et~al.(2016)Goodfellow, Bengio, and
  Courville}]{Goodfellow-et-al-2016}
\bibinfo{author}{I.~Goodfellow}, \bibinfo{author}{Y.~Bengio},
  \bibinfo{author}{A.~Courville}, \bibinfo{title}{Deep Learning},
  \bibinfo{publisher}{MIT Press}, \bibinfo{year}{2016}.
  \bibinfo{note}{\url{http://www.deeplearningbook.org}}.
\bibitem[{Zagoruyko and Komodakis(2016)}]{Zagoruyko2016}
\bibinfo{author}{S.~Zagoruyko}, \bibinfo{author}{N.~Komodakis},
\newblock \bibinfo{title}{Wide residual networks},
\newblock \bibinfo{journal}{CoRR}  (\bibinfo{year}{2016}).
  \DOIprefix\doi{10.48550/ARXIV.1605.07146}.
\bibitem[{Audebert et~al.(2018)Audebert, Le~Saux, and
  Lef{\`e}vre}]{audebert2018beyond}
\bibinfo{author}{N.~Audebert}, \bibinfo{author}{B.~Le~Saux},
  \bibinfo{author}{S.~Lef{\`e}vre},
\newblock \bibinfo{title}{Beyond rgb: Very high resolution urban remote sensing
  with multimodal deep networks},
\newblock \bibinfo{journal}{ISPRS Journal of Photogrammetry and Remote Sensing}
  \bibinfo{volume}{140} (\bibinfo{year}{2018}) \bibinfo{pages}{20--32}.
\bibitem[{Zhang et~al.(2019)Zhang, Lu, Li, Kim, and Wang}]{Zhang2019}
\bibinfo{author}{J.~Zhang}, \bibinfo{author}{C.~Lu}, \bibinfo{author}{X.~Li},
  \bibinfo{author}{H.-J. Kim}, \bibinfo{author}{J.~Wang},
\newblock \bibinfo{title}{A full convolutional network based on densenet for
  remote sensing scene classification},
\newblock \bibinfo{journal}{Mathematical Biosciences and Engineering}
  \bibinfo{volume}{16} (\bibinfo{year}{2019}) \bibinfo{pages}{3345--3367}.
  \DOIprefix\doi{10.3934/mbe.2019167}.
\bibitem[{Tong et~al.(2020)Tong, Chen, Han, Li, and Wang}]{Tong2020}
\bibinfo{author}{W.~Tong}, \bibinfo{author}{W.~Chen}, \bibinfo{author}{W.~Han},
  \bibinfo{author}{X.~Li}, \bibinfo{author}{L.~Wang},
\newblock \bibinfo{title}{Channel-attention-based densenet network for remote
  sensing image scene classification},
\newblock \bibinfo{journal}{IEEE Journal of Selected Topics in Applied Earth
  Observations and Remote Sensing} \bibinfo{volume}{13} (\bibinfo{year}{2020})
  \bibinfo{pages}{4121--4132}. \DOIprefix\doi{10.1109/JSTARS.2020.3009352}.
\bibitem[{Chen and Tsou(2021)}]{Chen2021}
\bibinfo{author}{F.~Chen}, \bibinfo{author}{J.~Y. Tsou},
\newblock \bibinfo{title}{Drsnet: Novel architecture for small patch and
  low-resolution remote sensing image scene classification},
\newblock \bibinfo{journal}{International Journal of Applied Earth Observation
  and Geoinformation} \bibinfo{volume}{104} (\bibinfo{year}{2021})
  \bibinfo{pages}{102577}.
  \DOIprefix\doi{https://doi.org/10.1016/j.jag.2021.102577}.
\bibitem[{Lin et~al.(2017)Lin, Dollár, Girshick, He, Hariharan, and
  Belongie}]{Lin2017}
\bibinfo{author}{T.-Y. Lin}, \bibinfo{author}{P.~Dollár},
  \bibinfo{author}{R.~Girshick}, \bibinfo{author}{K.~He},
  \bibinfo{author}{B.~Hariharan}, \bibinfo{author}{S.~Belongie},
\newblock \bibinfo{title}{Feature pyramid networks for object detection},
\newblock in: \bibinfo{booktitle}{2017 IEEE Conference on Computer Vision and
  Pattern Recognition (CVPR)}, \bibinfo{year}{2017}, pp.
  \bibinfo{pages}{936--944}. \DOIprefix\doi{10.1109/CVPR.2017.106}.
\bibitem[{Liu et~al.(2020)Liu, He, Bai, Zhang, and Cheng}]{liu2020light}
\bibinfo{author}{S.~Liu}, \bibinfo{author}{C.~He}, \bibinfo{author}{H.~Bai},
  \bibinfo{author}{Y.~Zhang}, \bibinfo{author}{J.~Cheng},
\newblock \bibinfo{title}{Light-weight attention semantic segmentation network
  for high-resolution remote sensing images},
\newblock in: \bibinfo{booktitle}{IGARSS 2020-2020 IEEE International
  Geoscience and Remote Sensing Symposium}, \bibinfo{organization}{IEEE},
  \bibinfo{year}{2020}, pp. \bibinfo{pages}{2595--2598}.
\bibitem[{Tian et~al.(2020)Tian, Wang, Tian, Zhan, and
  Zhang}]{tian2020resolution}
\bibinfo{author}{Z.~Tian}, \bibinfo{author}{W.~Wang},
  \bibinfo{author}{B.~Tian}, \bibinfo{author}{R.~Zhan},
  \bibinfo{author}{J.~Zhang},
\newblock \bibinfo{title}{Resolution-aware network with attention mechanisms
  for remote sensing object detection.},
\newblock \bibinfo{journal}{ISPRS Annals of Photogrammetry, Remote Sensing \&
  Spatial Information Sciences} \bibinfo{volume}{5} (\bibinfo{year}{2020}).
\bibitem[{Alhichri et~al.(2021)Alhichri, Alswayed, Bazi, Ammour, and
  Alajlan}]{Alhichri2021}
\bibinfo{author}{H.~Alhichri}, \bibinfo{author}{A.~S. Alswayed},
  \bibinfo{author}{Y.~Bazi}, \bibinfo{author}{N.~Ammour},
  \bibinfo{author}{N.~A. Alajlan},
\newblock \bibinfo{title}{Classification of remote sensing images using
  efficientnet-b3 cnn model with attention},
\newblock \bibinfo{journal}{IEEE Access} \bibinfo{volume}{9}
  (\bibinfo{year}{2021}) \bibinfo{pages}{14078--14094}.
  \DOIprefix\doi{10.1109/ACCESS.2021.3051085}.
\bibitem[{Devlin et~al.(2018)Devlin, Chang, Lee, and
  Toutanova}]{devlin2018bert}
\bibinfo{author}{J.~Devlin}, \bibinfo{author}{M.-W. Chang},
  \bibinfo{author}{K.~Lee}, \bibinfo{author}{K.~Toutanova},
\newblock \bibinfo{title}{Bert: Pre-training of deep bidirectional transformers
  for language understanding},
\newblock \bibinfo{journal}{arXiv preprint arXiv:1810.04805}
  (\bibinfo{year}{2018}).
\bibitem[{Bazi et~al.(2021)Bazi, Bashmal, Rahhal, Dayil, and Ajlan}]{Bazi2021}
\bibinfo{author}{Y.~Bazi}, \bibinfo{author}{L.~Bashmal},
  \bibinfo{author}{M.~M.~A. Rahhal}, \bibinfo{author}{R.~A. Dayil},
  \bibinfo{author}{N.~A. Ajlan},
\newblock \bibinfo{title}{Vision transformers for remote sensing image
  classification},
\newblock \bibinfo{journal}{Remote Sensing} \bibinfo{volume}{13}
  (\bibinfo{year}{2021}). \DOIprefix\doi{10.3390/rs13030516}.
\bibitem[{Gong et~al.(2022)Gong, Zhang, Zhou, Zhang, Wu, and Zhang}]{Gong2022}
\bibinfo{author}{N.~Gong}, \bibinfo{author}{C.~Zhang},
  \bibinfo{author}{H.~Zhou}, \bibinfo{author}{K.~Zhang},
  \bibinfo{author}{Z.~Wu}, \bibinfo{author}{X.~Zhang},
\newblock \bibinfo{title}{Classification of hyperspectral images via improved
  cycle-mlp},
\newblock \bibinfo{journal}{IET Computer Vision} \bibinfo{volume}{16}
  (\bibinfo{year}{2022}) \bibinfo{pages}{468--478}.
  \DOIprefix\doi{https://doi.org/10.1049/cvi2.12104}.
\bibitem[{Liu et~al.(2021)Liu, Lin, Cao, Hu, Wei, Zhang, Lin, and
  Guo}]{Ze2021SwinV1}
\bibinfo{author}{Z.~Liu}, \bibinfo{author}{Y.~Lin}, \bibinfo{author}{Y.~Cao},
  \bibinfo{author}{H.~Hu}, \bibinfo{author}{Y.~Wei},
  \bibinfo{author}{Z.~Zhang}, \bibinfo{author}{S.~Lin},
  \bibinfo{author}{B.~Guo},
\newblock \bibinfo{title}{Swin transformer: Hierarchical vision transformer
  using shifted windows},
\newblock in: \bibinfo{booktitle}{2021 IEEE/CVF International Conference on
  Computer Vision (ICCV)}, \bibinfo{year}{2021}, pp.
  \bibinfo{pages}{9992--10002}. \DOIprefix\doi{10.1109/ICCV48922.2021.00986}.
\bibitem[{Scheibenreif et~al.(2022)Scheibenreif, Hanna, Mommert, and
  Borth}]{Scheibenreif_SwinLC}
\bibinfo{author}{L.~Scheibenreif}, \bibinfo{author}{J.~Hanna},
  \bibinfo{author}{M.~Mommert}, \bibinfo{author}{D.~Borth},
\newblock \bibinfo{title}{Self-supervised vision transformers for land-cover
  segmentation and classification},
\newblock in: \bibinfo{booktitle}{2022 IEEE/CVF Conference on Computer Vision
  and Pattern Recognition Workshops (CVPRW)}, \bibinfo{year}{2022}, pp.
  \bibinfo{pages}{1421--1430}. \DOIprefix\doi{10.1109/CVPRW56347.2022.00148}.
\bibitem[{Zhang et~al.(2022)Zhang, Wang, Cheng, and Li}]{Zhang2022SwinSUNet}
\bibinfo{author}{C.~Zhang}, \bibinfo{author}{L.~Wang},
  \bibinfo{author}{S.~Cheng}, \bibinfo{author}{Y.~Li},
\newblock \bibinfo{title}{Swinsunet: Pure transformer network for remote
  sensing image change detection},
\newblock \bibinfo{journal}{IEEE Transactions on Geoscience and Remote Sensing}
  \bibinfo{volume}{60} (\bibinfo{year}{2022}) \bibinfo{pages}{1--13}.
  \DOIprefix\doi{10.1109/TGRS.2022.3160007}.
\bibitem[{Wang et~al.(2022)Wang, Zhang, Du, Xia, and Tao}]{Wang2022_emp}
\bibinfo{author}{D.~Wang}, \bibinfo{author}{J.~Zhang}, \bibinfo{author}{B.~Du},
  \bibinfo{author}{G.-S. Xia}, \bibinfo{author}{D.~Tao},
\newblock \bibinfo{title}{An empirical study of remote sensing pretraining},
\newblock \bibinfo{journal}{IEEE Transactions on Geoscience and Remote Sensing}
   (\bibinfo{year}{2022}) \bibinfo{pages}{1--1}.
  \DOIprefix\doi{10.1109/TGRS.2022.3176603}.
\bibitem[{Xu et~al.(2021)Xu, Zhang, Zhang, Yang, and Li}]{Xu2021}
\bibinfo{author}{Z.~Xu}, \bibinfo{author}{W.~Zhang},
  \bibinfo{author}{T.~Zhang}, \bibinfo{author}{Z.~Yang},
  \bibinfo{author}{J.~Li},
\newblock \bibinfo{title}{Efficient transformer for remote sensing image
  segmentation},
\newblock \bibinfo{journal}{Remote Sensing} \bibinfo{volume}{13}
  (\bibinfo{year}{2021}). \DOIprefix\doi{10.3390/rs13183585}.
\bibitem[{Meng et~al.(2021)Meng, Zhao, and Liang}]{Meng2021}
\bibinfo{author}{Z.~Meng}, \bibinfo{author}{F.~Zhao},
  \bibinfo{author}{M.~Liang},
\newblock \bibinfo{title}{Ss-mlp: A novel spectral-spatial mlp architecture for
  hyperspectral image classification},
\newblock \bibinfo{journal}{Remote Sensing} \bibinfo{volume}{13}
  (\bibinfo{year}{2021}). \DOIprefix\doi{10.3390/rs13204060}.
\bibitem[{Sechidis et~al.(2011)Sechidis, Tsoumakas, and
  Vlahavas}]{sechidis2011}
\bibinfo{author}{K.~Sechidis}, \bibinfo{author}{G.~Tsoumakas},
  \bibinfo{author}{I.~Vlahavas},
\newblock \bibinfo{title}{On the stratification of multi-label data},
\newblock in: \bibinfo{booktitle}{Proceedings of the 2011 European Conference
  on Machine Learning and Knowledge Discovery in Databases - Volume Part III},
  \bibinfo{publisher}{Springer-Verlag}, \bibinfo{year}{2011}, p.
  \bibinfo{pages}{145–158}.
\bibitem[{Liu et~al.(2020)Liu, Jiang, He, Chen, Liu, Gao, and
  Han}]{liu2019radam}
\bibinfo{author}{L.~Liu}, \bibinfo{author}{H.~Jiang}, \bibinfo{author}{P.~He},
  \bibinfo{author}{W.~Chen}, \bibinfo{author}{X.~Liu},
  \bibinfo{author}{J.~Gao}, \bibinfo{author}{J.~Han},
\newblock \bibinfo{title}{On the variance of the adaptive learning rate and
  beyond},
\newblock in: \bibinfo{booktitle}{Proceedings of the Eighth International
  Conference on Learning Representations (ICLR 2020)}, \bibinfo{year}{2020}.
\bibitem[{Kingma and Ba(2014)}]{kingma2014adam}
\bibinfo{author}{D.~P. Kingma}, \bibinfo{author}{J.~Ba},
\newblock \bibinfo{title}{Adam: A method for stochastic optimization},
\newblock \bibinfo{journal}{arXiv preprint arXiv:1412.6980}
  (\bibinfo{year}{2014}).
\bibitem[{Risojevic and Stojnic(2021)}]{Risojevic2021}
\bibinfo{author}{V.~Risojevic}, \bibinfo{author}{V.~Stojnic},
\newblock \bibinfo{title}{{Do we still need ImageNet pre-training in remote
  sensing scene classification?}},
\newblock \bibinfo{journal}{arXiv} \bibinfo{volume}{abs/2111.03690}
  (\bibinfo{year}{2021}). \href{http://arxiv.org/abs/2111.03690}{{\tt
  arXiv:2111.03690}}.
\bibitem[{Yosinski et~al.(2014)Yosinski, Clune, Bengio, and
  Lipson}]{yosinski2014}
\bibinfo{author}{J.~Yosinski}, \bibinfo{author}{J.~Clune},
  \bibinfo{author}{Y.~Bengio}, \bibinfo{author}{H.~Lipson},
\newblock \bibinfo{title}{How transferable are features in deep neural
  networks?},
\newblock in: \bibinfo{booktitle}{Proceedings of the 27th International
  Conference on Neural Information Processing Systems - Volume 2},
  \bibinfo{year}{2014}, p. \bibinfo{pages}{3320–3328}.
\bibitem[{Kornblith et~al.(2019)Kornblith, Shlens, and Le}]{kornblith2019}
\bibinfo{author}{S.~Kornblith}, \bibinfo{author}{J.~Shlens},
  \bibinfo{author}{Q.~V. Le},
\newblock \bibinfo{title}{Do better imagenet models transfer better?},
\newblock in: \bibinfo{booktitle}{2019 IEEE/CVF Conference on Computer Vision
  and Pattern Recognition (CVPR)}, \bibinfo{year}{2019}, pp.
  \bibinfo{pages}{2656--2666}.
\bibitem[{Paul and Chen(2022)}]{Paul_Chen_2022}
\bibinfo{author}{S.~Paul}, \bibinfo{author}{P.-Y. Chen},
\newblock \bibinfo{title}{Vision transformers are robust learners},
\newblock \bibinfo{journal}{Proceedings of the AAAI Conference on Artificial
  Intelligence} \bibinfo{volume}{36} (\bibinfo{year}{2022})
  \bibinfo{pages}{2071--2081}.
\bibitem[{Bhojanapalli et~al.(2021)Bhojanapalli, Chakrabarti, Glasner, Li,
  Unterthiner, and Veit}]{bhojanapalli2021}
\bibinfo{author}{S.~Bhojanapalli}, \bibinfo{author}{A.~Chakrabarti},
  \bibinfo{author}{D.~Glasner}, \bibinfo{author}{D.~Li},
  \bibinfo{author}{T.~Unterthiner}, \bibinfo{author}{A.~Veit},
\newblock \bibinfo{title}{Understanding robustness of transformers for image
  classification},
\newblock in: \bibinfo{booktitle}{2021 IEEE/CVF International Conference on
  Computer Vision (ICCV)}, \bibinfo{publisher}{IEEE Computer Society},
  \bibinfo{address}{Los Alamitos, CA, USA}, \bibinfo{year}{2021}, pp.
  \bibinfo{pages}{10211--10221}.
\bibitem[{Zhang et~al.(2022)Zhang, Zhang, Zhang, Jin, feng Zhou, Cai, Zhao, Yi,
  Liu, and Liu}]{Zhang2022DelvingDI}
\bibinfo{author}{C.~Zhang}, \bibinfo{author}{M.~Zhang},
  \bibinfo{author}{S.~Zhang}, \bibinfo{author}{D.~Jin},
  \bibinfo{author}{Q.~feng Zhou}, \bibinfo{author}{Z.~Cai},
  \bibinfo{author}{H.~Zhao}, \bibinfo{author}{S.~Yi}, \bibinfo{author}{X.~Liu},
  \bibinfo{author}{Z.~Liu},
\newblock \bibinfo{title}{Delving deep into the generalization of vision
  transformers under distribution shifts},
\newblock \bibinfo{journal}{2022 IEEE/CVF Conference on Computer Vision and
  Pattern Recognition (CVPR)}  (\bibinfo{year}{2022})
  \bibinfo{pages}{7267--7276}.
\bibitem[{Selvaraju et~al.(2017)Selvaraju, Cogswell, Das, Vedantam, Parikh, and
  Batra}]{selvaraju2017}
\bibinfo{author}{R.~R. Selvaraju}, \bibinfo{author}{M.~Cogswell},
  \bibinfo{author}{A.~Das}, \bibinfo{author}{R.~Vedantam},
  \bibinfo{author}{D.~Parikh}, \bibinfo{author}{D.~Batra},
\newblock \bibinfo{title}{Grad-cam: Visual explanations from deep networks via
  gradient-based localization},
\newblock in: \bibinfo{booktitle}{2017 IEEE International Conference on
  Computer Vision (ICCV)}, \bibinfo{year}{2017}, pp. \bibinfo{pages}{618--626}.
\bibitem[{Gildenblat and contributors(2021)}]{jacobgilpytorchcam}
\bibinfo{author}{J.~Gildenblat}, \bibinfo{author}{contributors},
  \bibinfo{title}{Pytorch library for cam methods},
  \bibinfo{howpublished}{\url{https://github.com/jacobgil/pytorch-grad-cam}},
  \bibinfo{year}{2021}.
\bibitem[{Li et~al.(2020)Li, Lin, Wang, Xu, Zhang, Ding, and Zhou}]{li2020rs}
\bibinfo{author}{J.~Li}, \bibinfo{author}{D.~Lin}, \bibinfo{author}{Y.~Wang},
  \bibinfo{author}{G.~Xu}, \bibinfo{author}{Y.~Zhang},
  \bibinfo{author}{C.~Ding}, \bibinfo{author}{Y.~Zhou},
\newblock \bibinfo{title}{Deep discriminative representation learning with
  attention map for scene classification},
\newblock \bibinfo{journal}{Remote Sensing} \bibinfo{volume}{12}
  (\bibinfo{year}{2020}).
\bibitem[{Wilkinson et~al.(2016)Wilkinson, Dumontier, Aalbersberg, Appleton,
  Axton, Baak, Blomberg, Boiten, da~Silva~Santos, Bourne
  et~al.}]{wilkinson2016fair}
\bibinfo{author}{M.~D. Wilkinson}, \bibinfo{author}{M.~Dumontier},
  \bibinfo{author}{I.~J. Aalbersberg}, \bibinfo{author}{G.~Appleton},
  \bibinfo{author}{M.~Axton}, \bibinfo{author}{A.~Baak},
  \bibinfo{author}{N.~Blomberg}, \bibinfo{author}{J.-W. Boiten},
  \bibinfo{author}{L.~B. da~Silva~Santos}, \bibinfo{author}{P.~E. Bourne},
  et~al.,
\newblock \bibinfo{title}{The {FAIR} guiding principles for scientific data
  management and stewardship},
\newblock \bibinfo{journal}{{Scientific Data}} \bibinfo{volume}{3}
  (\bibinfo{year}{2016}) \bibinfo{pages}{1--9}.

\end{thebibliography}


\newpage
\appendix
\renewcommand*{\thesection}{\Alph{section}}
\setcounter{page}{0}
\addcontentsline{toc}{section}{Appendix}
\noindent \Large\textbf{Current Trends in Deep Learning for Earth Observation:\\
An Open-source Benchmark Arena for Image Classification}\\
\large\textit{Ivica Dimitrovski, Ivan Kitanovski, Dragi Kocev, Nikola Simidjievski}

\vspace*{-25pt}
\part{Supplementary Material} 
\begin{spacing}{1}
\parttoc
\end{spacing}

\newpage

\section{Evaluation metrics}\label{sec:app:measures}
\setcounter{table}{0}
\setcounter{figure}{0}

The predictive performance of machine learning models is typically assessed using different evaluation measures that capture different aspects of the models' behavior. Selecting the proper evaluation measures requires knowledge of the task and problem at hand. In order to have an unbiased and fair view of the performance, one needs to consider the models' performance along several measures and then compare their performance. In this study, we assess the performance of the models using a variety of different measures available for the machine learning tasks studied here: multi-class and multi-label classification. 

\textbf{Multi-class classification} refers to the task where a sample can be assigned to exactly one class/label selected from a predefined set of possible classes/labels. Here, we overview several evaluation measures used for this task. Most widely used evaluation measure is \emph{accuracy} due to its intuitive interpretation and straightforward calculation. It denotes the percentage of correctly labeled samples. \emph{Precision} and \emph{Recall} are defined for binary tasks (two classes, often called positive and negative class) by default. To extend the binary measures to multi-class classification tasks, we adopt the One-vs-Rest (One-vs-All) approach which converts a multi-class task into a series of binary tasks for each class/label in the target. Within this approach the sample from given class/label is treated as positive, and the samples from all the other classes/labels are treated as negative.

To calculate most of the evaluation measures, we need to define the following concepts: True Positives (TP), True Negatives (TN), False Positives (FP) and False Negatives (FN). These concepts combined together form the confusion matrix for the performance of a given model over a given dataset. The TP, TN, FP and FN are defined as follows:

\begin{itemize}
\itemsep0em 
\item TP: the label is positive and the prediction is also positive
\item TN: the label is negative and the prediction is also negative
\item FP: the label is negative but the prediction is positive
\item FN: the label is positive but the prediction is negative
\end{itemize}

\emph{Precision} is then calculated as the fraction of correctly predicted positive observations from the total predicted positive observations:

\begin{equation*}
\textstyle Precision = \frac{TP}{TP + FP}
\end{equation*}

\emph{Recall} is calculated as the fraction of correctly predicted positive observations from the available positive observations:

\begin{equation*}
\textstyle Recall = \frac{TP}{TP + FN}
\end{equation*}

\emph{F1 score} is also a common evaluation measure used in machine learning tasks, basically it combines precision and recall through a weighted average. Therefore, this score takes both false positives and false negatives into account and is very useful, especially if we have an imbalanced class/label distribution. The F1 score can be calculated as:

\begin{equation*}
F1 = 2 \cdot \frac{Precision \cdot Recall}{Precision + Recall}
\end{equation*}

These evaluation measures can then be aggregated across multiple classes using three strategies:
\begin{itemize}
\itemsep0em 
\item \emph{Macro averaging}: calculate the evaluation measures for each class/label separately and then average the individual values,
\item \emph{Micro averaging}: calculate the class wise confusion matrices and then aggregate the confusion matrices into a single one (i.e., add together the TP, FP, FN and FP values for each class). The aggregated confusion matrix is then used to calculate the values for the different evaluation measures, and
\item \emph{Weighted averaging}: based on macro averaging but using the frequency of the class/label as a weight in the average calculation.
\end{itemize}

Using these aggregation strategies, we then obtain macro-averaged, micro-averaged and weighted-averaged precision, recall and F1 score. Note that micro F1 score, micro precision and micro recall yield the same values as accuracy for the multi-class classification task. Taking into account this, for the multi-class classification tasks we report the following evaluation measures: Accuracy, Macro Precision, Weighted Precision, Macro Recall, Weighted Recall, Macro F1 score and Weighted F1 score.

\textbf{Multi-label classification} refers to the task where a sample can be assigned to multiple class/label from a predefined set of possible classes/labels. To transform the multi-label classification task to binary classification and apply the same metrics previously defined, we adopt the binary relevance method \citep{tsoumakas2009} that considers each label as an independent binary problem. In our case, in each node from the output layer, we use the sigmoid activation function to obtain a probability of the input image being labeled with each of the classes/labels. To use these probabilities to predict the classes/labels of the image, we need to define a threshold value. The model predicts whether an image contains the classes/labels with a probability that exceed the given threshold. The threshold value controls the rate of false positives v.s false negatives. Increasing the threshold reduces the number of false positives, whereas decreasing it reduces the number of false negatives. In our experiments, we use threshold value of 0.5. Taking into account this transformation, we can apply the formulas from above to calculate the same evaluation measures for multi-label classification tasks. While these evaluation measures are threshold dependent, we additionally use the the \emph{mean average precision} (mAP) - a threshold independent evaluation measure widely used in image classification tasks. mAP is calculated as the mean over the average precision values of the individual labels. Average precision summarizes a precision-recall curve as the weighted mean of the precision values obtained at each threshold, with the increase in recall from the previous threshold used as the weight:
\begin{equation*}
\textstyle AP = \sum_n (R_n - R_{n-1}) P_n
\end{equation*}
Where $P_n$ and $R_n$  are the precision and recall at the n-th threshold. It is a useful metric to compare how well models are ordering the predictions, without considering any specific decision threshold.

For the multi-label classification task, we report the following evaluation measures: Micro Precision, Macro Precision, Weighted Precision, Micro Recall, Macro Recall, Weighted Recall, Micro F1 score, Macro F1 score, Weighted F1 score and mean average precision (mAP). For all measures but mAP, that require a threshold on the predictions, we set it to 0.5 for all the models and settings. \looseness=-1

For both tasks, we provide the means to perform even more detailed analysis of the performance by reporting the confusion matrices as a performance summary of the models. The confusion matrices provide detailed per class/label view of the models' performance.

\clearpage
\section{Training Time Details}\label{appendix:times}
\setcounter{table}{0}
\setcounter{figure}{0}

\begin{figure*}[!ht]
    \centering
    \begin{subfigure}[t]{\linewidth}
    \includegraphics[width=\linewidth]{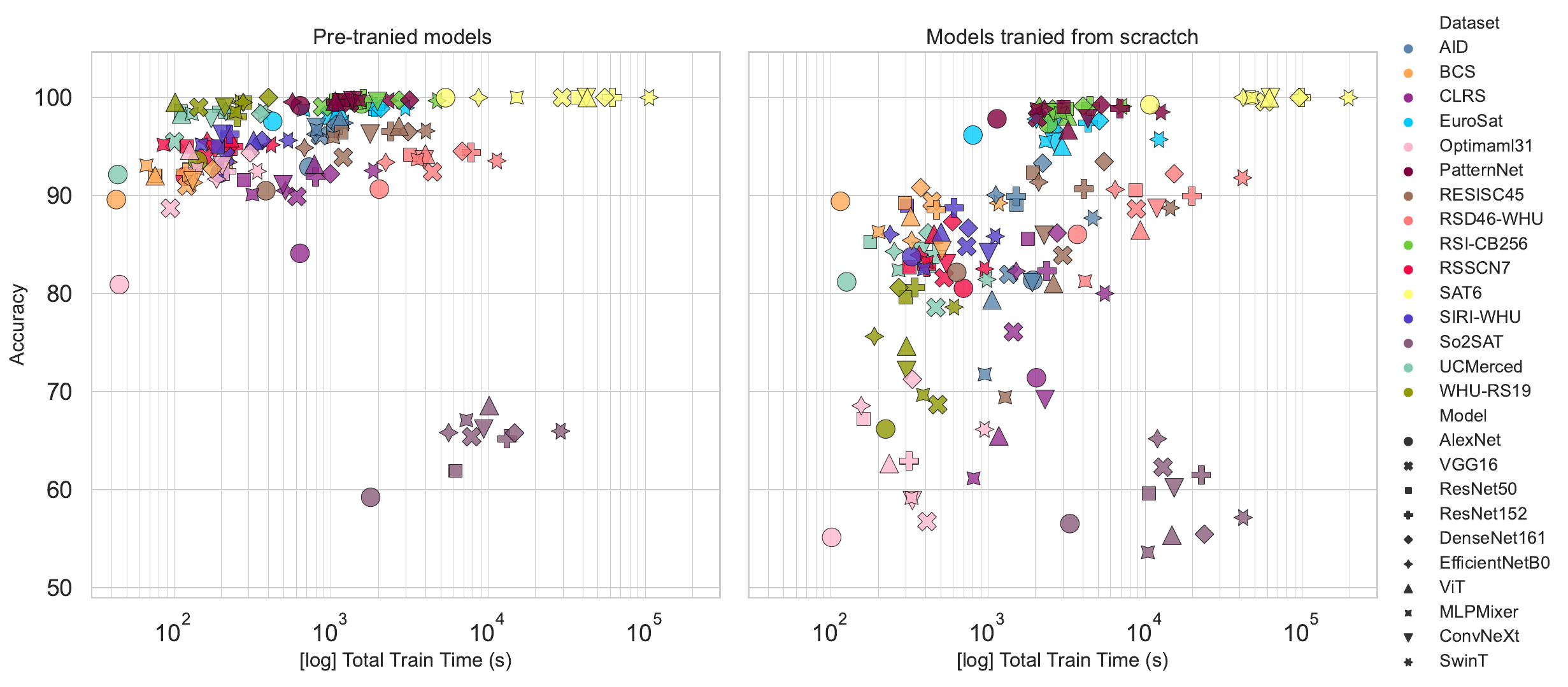}
    \subcaption{Multi-Class classification tasks}
    \end{subfigure}
    \begin{subfigure}[t]{\linewidth}
    \includegraphics[width=\linewidth]{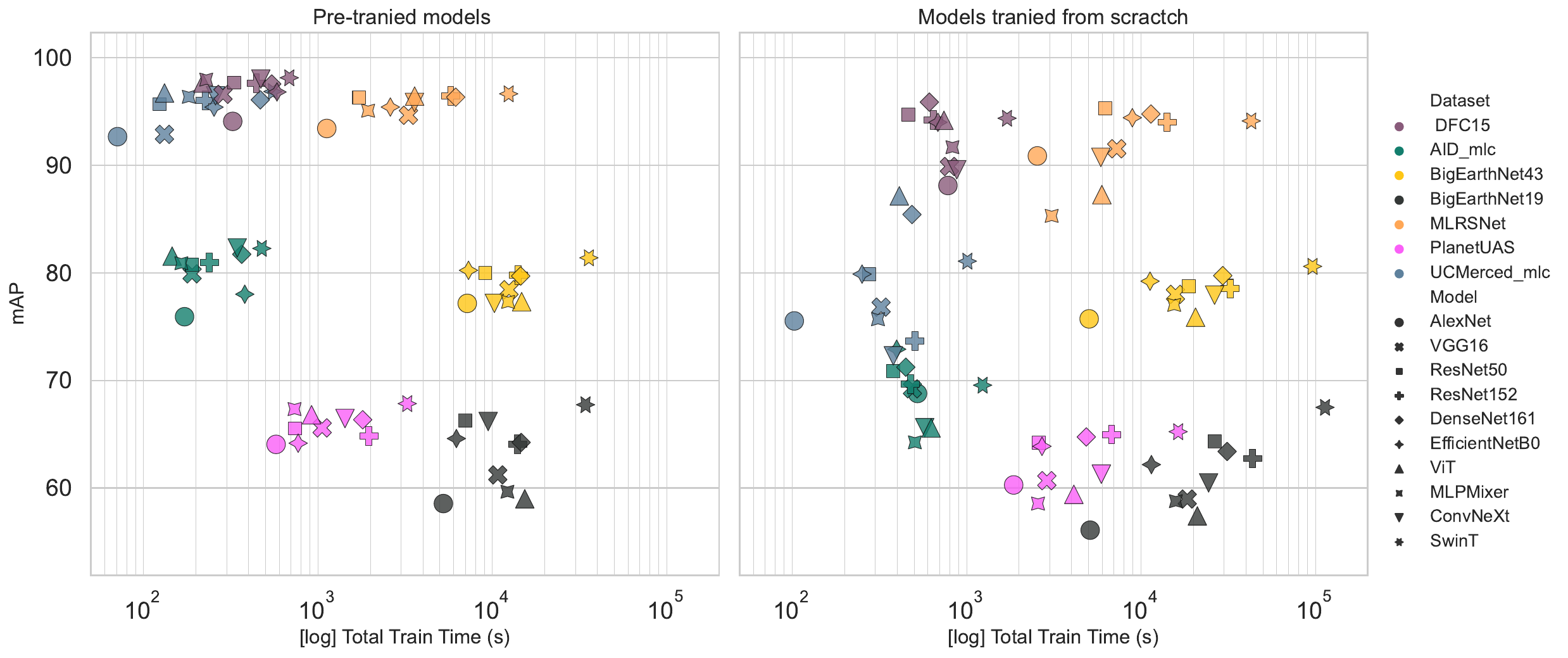}
    \subcaption{Multi-Label classification tasks}
    \end{subfigure}
    \hfill
    \caption{\textbf{Performance vs. total training time} comparison of all model architectures (denoted with different markers); evaluated on (\textbf{top}) MCC and \textbf{(bottom)} MLC datasets (color-coded). We present both pretrained \textbf{(left)} and trained from scratch \textbf{(right)} variants. Performance is reported as accuracy (\%) and mean average precision (mAP \%) for MCC and MLC tasks, respectively. Note the log scale of the total training time (seconds)}
    \label{fig:full_perf_time}
\end{figure*}

\begin{figure*}[!ht]
    \centering
    \begin{subfigure}[t]{0.35\linewidth}
    \includegraphics[width=\linewidth]{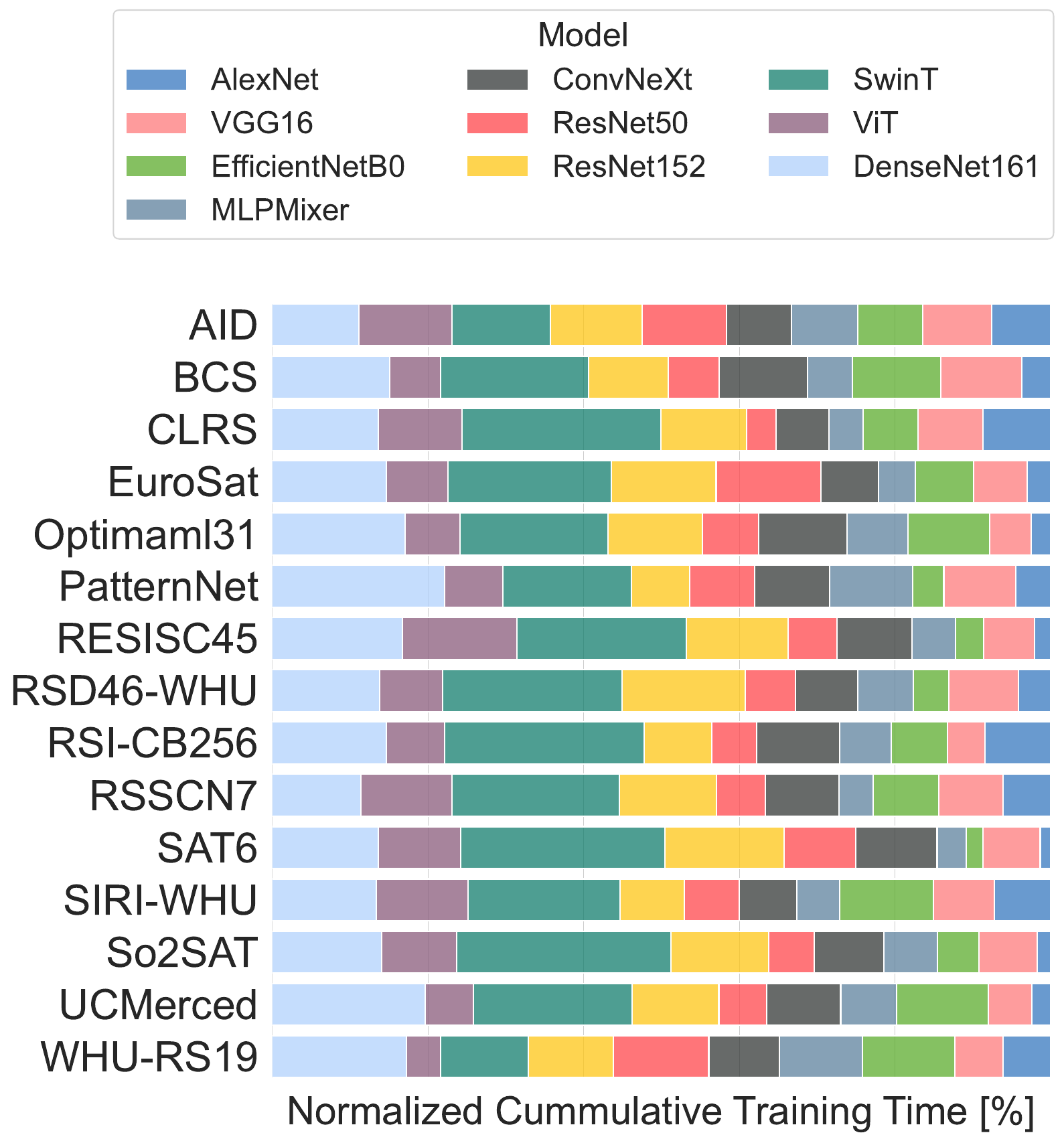}
    \subcaption{MCC: Pre-trained models}
    \end{subfigure}
    \begin{subfigure}[t]{0.35\linewidth}
    \includegraphics[width=\linewidth]{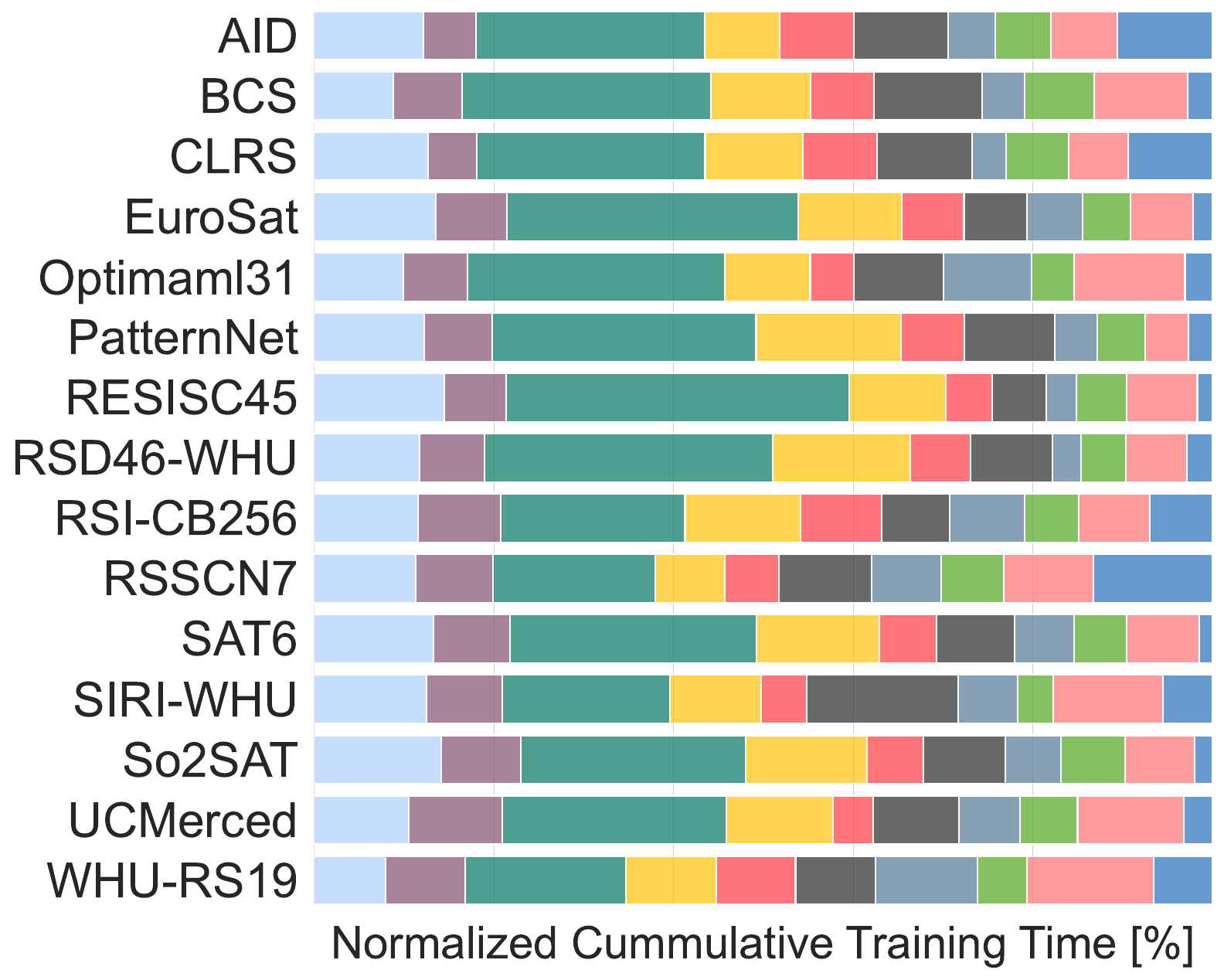}
    \subcaption{MCC: Trained from scratch}
    \end{subfigure}
    \hfill
    \begin{subfigure}[t]{0.25\linewidth}
    \includegraphics[width=\linewidth]{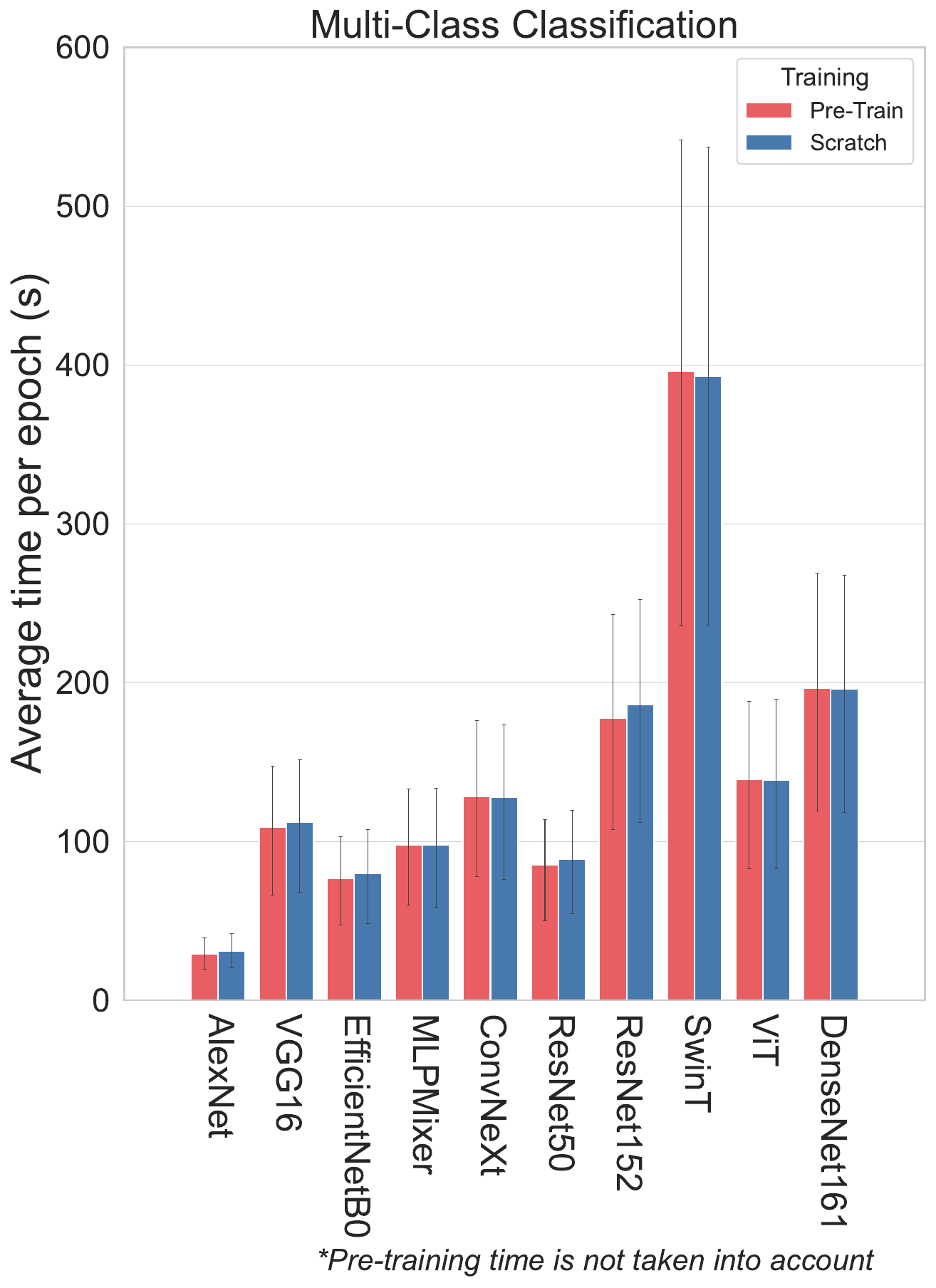}
    \subcaption{MCC: Average time per epoch.}
    \end{subfigure}
    \hfill
    \begin{subfigure}[t]{0.35\linewidth}
    \includegraphics[width=\linewidth]{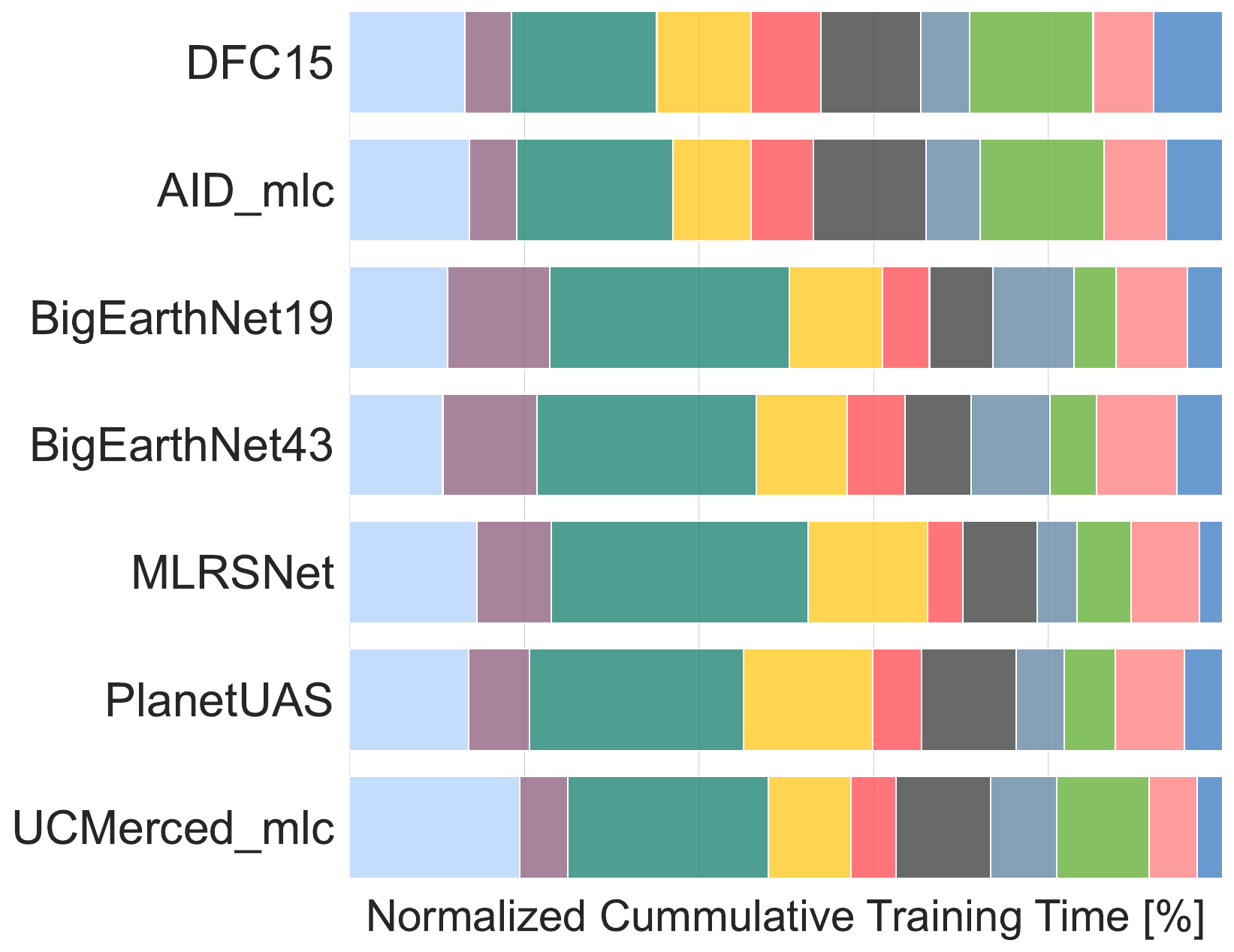}
    \subcaption{MLC: Pre-trained models}
    \end{subfigure}
    \begin{subfigure}[t]{0.35\linewidth}
    \includegraphics[width=\linewidth]{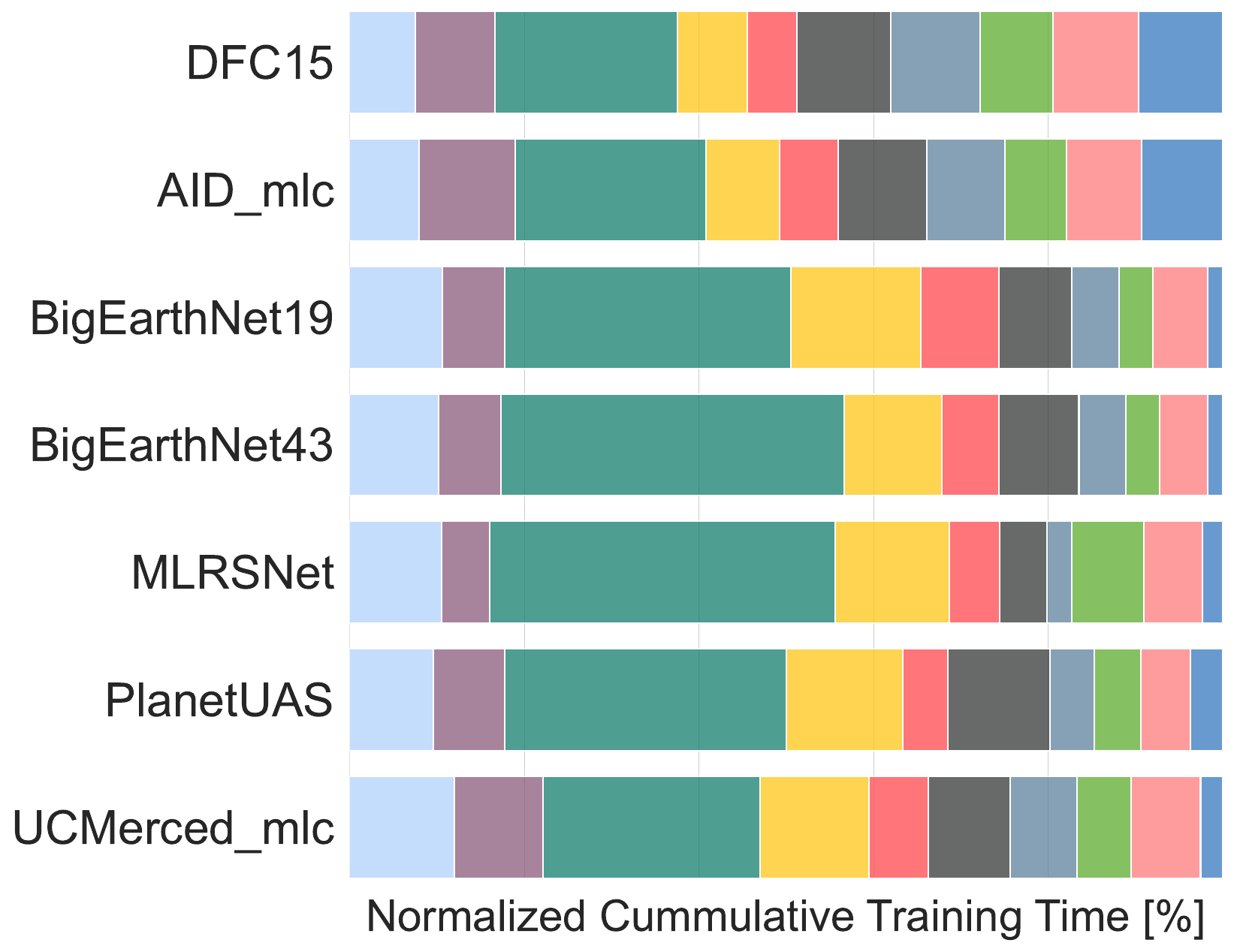}
    \subcaption{MLC: Trained from scratch}
    \end{subfigure}
    \hfill
    \begin{subfigure}[t]{0.25\linewidth}
    \includegraphics[width=\linewidth]{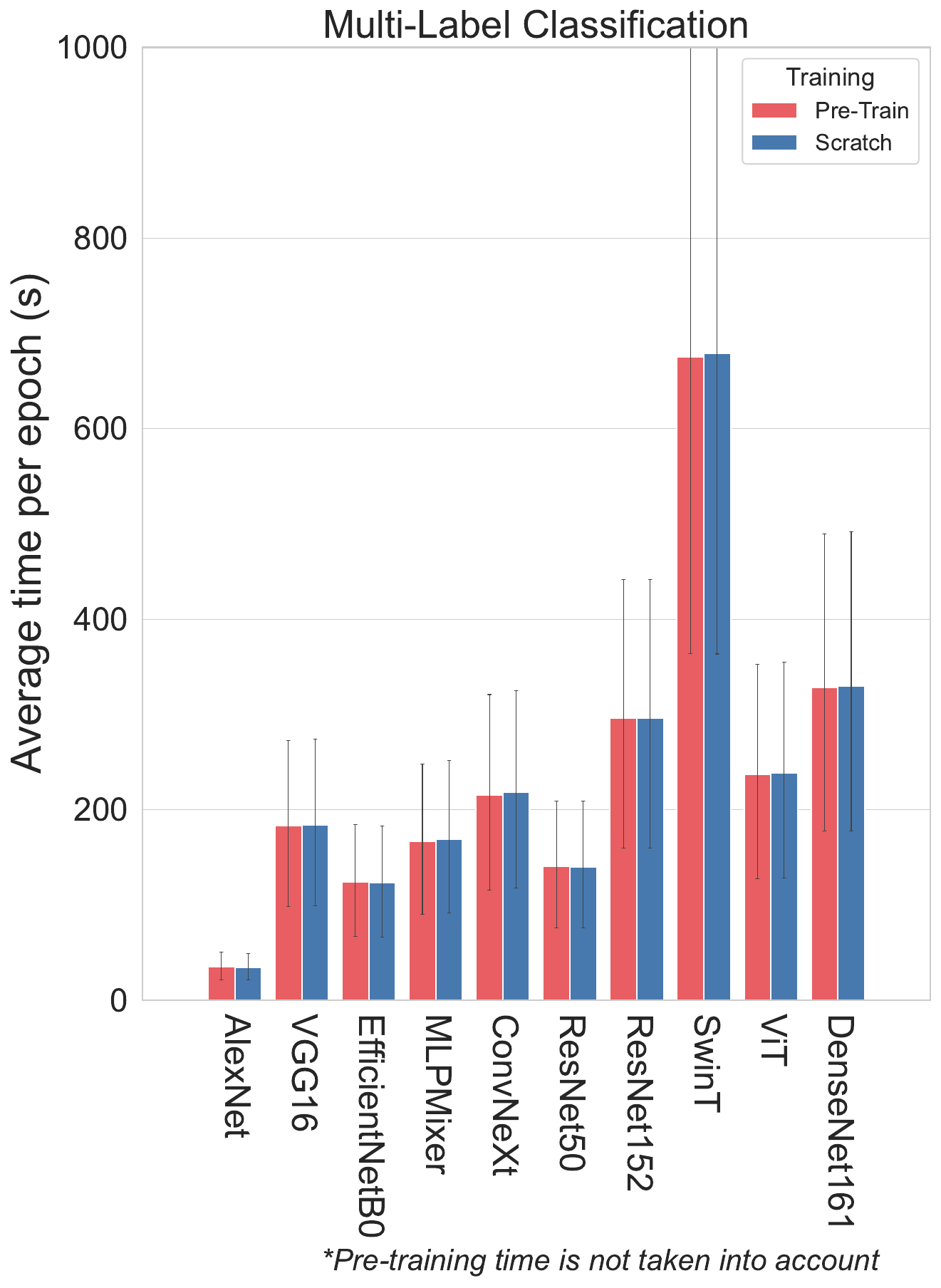}
    \subcaption{MLC: Average time per epoch.}
    \end{subfigure}
    \caption{\textbf{Training time} of models trained from scratch and pre-trained models for each of the \textbf{(a,b)} MCC and \textbf{(d,e)} MLC datasts. The training time of each model architecture (denoted with different colors) is depicted as a fraction (\%) of the cumulative training time for each dataset. Furthermore, (c) and (f) illustrate the average time per epoch of each model variant on \textbf{(c)} MCC  and \textbf{(f)} MLC tasks, comparing the \textbf{(red)} pre-trained model variants (from (a) and (b)) to their counterparts \textbf{(blue) }trained from scratch.}
    \label{fig:full_training_times}
\end{figure*}

\begin{table}[!ht]
  \centering
  \caption{Training-time details for multi-class classification tasks.}
  \begin{adjustbox}{width=1\linewidth,angle=90}
    \begin{tabular}{clrr|rr|rr|rr|rr|rr|rr|rr|rr|rr}\toprule
          &       & \multicolumn{2}{c}{AlexNet} & \multicolumn{2}{c}{VGG16} & \multicolumn{2}{c}{ResNet50} & \multicolumn{2}{c}{RestNet152} & \multicolumn{2}{c}{DenseNet161} & \multicolumn{2}{c}{EfficientNetB0} & \multicolumn{2}{c}{ViT} & \multicolumn{2}{c}{MLPMixer} & \multicolumn{2}{c}{ConvNeXt} & \multicolumn{2}{c}{SwinT}  \\
          &       & \multicolumn{1}{c}{Pt.} & \multicolumn{1}{c}{Sc.} & \multicolumn{1}{c}{Pt.} & \multicolumn{1}{c}{Sc.} & \multicolumn{1}{c}{Pt.} & \multicolumn{1}{c}{Sc.} & \multicolumn{1}{c}{Pt.} & \multicolumn{1}{c}{Sc.} & \multicolumn{1}{c}{Pt.} & \multicolumn{1}{c}{Sc.} & \multicolumn{1}{c}{Pt.} & \multicolumn{1}{c}{Sc.} & \multicolumn{1}{c}{Pt.} & \multicolumn{1}{c}{Sc.} & \multicolumn{1}{c}{Pt.} & \multicolumn{1}{c}{Sc.} & \multicolumn{1}{c}{Pt.} & \multicolumn{1}{c}{Sc.} \\\midrule
    \multirow{3}[0]{*}{Eurosat} & Avg time/epoch (sec.) & 8.88  & 8.02  & 33.69 & 33.62 & 26.56 & 26.45 & 56    & 56.21 & 61.12 & 62.5  & 23.47 & 24.19 & 43.19 & 44.22 & 30.41 & 31.45 & 40.38 & 40.03 & 124.17 & 122.44 \\
      & Total training time (sec.) & 426   & 802   & 977   & 2622  & 1912  & 2619  & 1904  & 4328  & 2078  & 5125  & 1056  & 2032  & 1123  & 2963  & 669   & 2327  & 1050  & 2642  & 2980  & 12244 \\
      & Best epoch & 38    & 95    & 19    & 63    & 62    & 84    & 24    & 62    & 24    & 67    & 35    & 69    & 16    & 52    & 12    & 59    & 16    & 51    & 14    & 90 \\\midrule
    \multirow{3}[0]{*}{ UCMerced} & Avg time/epoch (sec.) & 1.29  & 1.3   & 3.16  & 4.66  & 2.85  & 2.54  & 5.05  & 5.02  & 5.41  & 5.46  & 2.46  & 2.53  & 4     & 4.44  & 3.1   & 3.06  & 3.68  & 3.75  & 10.28 & 10.36 \\
      & Total training time (sec.) & 44    & 126   & 101   & 466   & 111   & 178   & 202   & 467   & 357   & 415   & 214   & 253   & 112   & 413   & 130   & 269   & 173   & 375   & 370   & 984 \\
      & Best epoch & 24    & 82    & 22    & 85    & 29    & 55    & 30    & 78    & 56    & 61    & 77    & 93    & 18    & 78    & 32    & 73    & 37    & 92    & 26    & 80  \\\midrule
    \multirow{3}[0]{*}{AID} & Avg time/epoch (sec.) & 21.32 & 19.46 & 21.35 & 19.65 & 20.29 & 19.66 & 22.2  & 22.25 & 24.36 & 24.48 & 20    & 19.33 & 20.45 & 19.63 & 19.78 & 19.06 & 23.06 & 19.15 & 46.65 & 46.94 \\
      & Total training time (sec.) & 725   & 1927  & 854   & 1356  & 1035  & 1514  & 1132  & 1513  & 1072  & 2228  & 800   & 1121  & 1145  & 1060  & 811   & 953   & 807   & 1915  & 1213  & 4647 \\
      & Best epoch & 24    & 84    & 30    & 54    & 41    & 62    & 41    & 53    & 34    & 76    & 30    & 43    & 46    & 39    & 31    & 35    & 25    & 96    & 16    & 84  \\\midrule
    \multirow{3}[0]{*}{RSSCN7} & Avg time/epoch (sec.) & 3.19  & 6.97  & 4.68  & 6.74  & 3.9   & 3.76  & 7.09  & 6.9   & 7.59  & 8.5   & 3.79  & 3.65  & 5.54  & 5.52  & 4.3   & 4.08  & 5.23  & 5.43  & 13.42 & 13.78 \\
      & Total training time (sec.) & 118   & 697   & 159   & 526   & 121   & 316   & 241   & 407   & 220   & 595   & 163   & 365   & 227   & 453   & 86    & 408   & 183   & 543   & 416   & 951 \\
      & Best epoch & 27    & 85    & 24    & 63    & 21    & 69    & 24    & 44    & 19    & 55    & 33    & 93    & 31    & 67    & 10    & 100   & 25    & 87    & 21    & 54 \\\midrule
    \multirow{3}[0]{*}{WHU-RS19} & Avg time/epoch (sec.) & 2.78  & 2.53  & 3     & 4.79  & 2.85  & 3.85  & 4.02  & 4.29  & 4.04  & 4.04  & 2.76  & 2.78  & 3.4   & 3.44  & 2.84  & 3.86  & 3.2   & 3.03  & 5.98  & 6.08 \\
      & Total training time (sec.) & 142   & 223   & 144   & 479   & 285   & 300   & 253   & 343   & 400   & 271   & 276   & 189   & 102   & 303   & 247   & 386   & 211   & 303   & 263   & 608 \\
      & Best epoch & 41    & 73    & 38    & 96    & 96    & 63    & 53    & 65    & 89    & 52    & 100   & 53    & 20    & 73    & 77    & 89    & 56    & 90    & 34    & 86 \\\midrule
    \multirow{3}[0]{*}{SIRI-WHU} & Avg time/epoch (sec.) & 4.28  & 3.54  & 4.98  & 7.32  & 4.66  & 3.81  & 6.65  & 6.54  & 7.3   & 7.49  & 4.57  & 3.61  & 5.37  & 5.08  & 4.55  & 3.92  & 5.64  & 11.99 & 11.87 & 12.1 \\
      & Total training time (sec.) & 197   & 326   & 214   & 732   & 191   & 305   & 226   & 608   & 365   & 749   & 329   & 238   & 322   & 503   & 150   & 392   & 203   & 1007  & 534   & 1113 \\
      & Best epoch & 36    & 77    & 33    & 93    & 31    & 65    & 24    & 78    & 40    & 94    & 62    & 51    & 50    & 84    & 23    & 98    & 26    & 69    & 35    & 77 \\\midrule
    \multirow{3}[0]{*}{RSI-CB256}& Avg time/epoch (sec.) & 34.84 & 34.99 & 34.04 & 34.9  & 33.69 & 36.39 & 51.9  & 51.86 & 56.6  & 56.75 & 33.5  & 26.5  & 41.18 & 41.08 & 35.29 & 29    & 40.35 & 36.93 & 113.14 & 113.6 \\
      & Total training time (sec.) & 1568  & 2414  & 885   & 2757  & 1078  & 3166  & 1609  & 4472  & 2717  & 4029  & 1340  & 2123  & 1400  & 3204  & 1235  & 2900  & 1977  & 2622  & 4752  & 7157 \\
      & Best epoch & 35    & 54    & 16    & 64    & 22    & 72    & 21    & 72    & 38    & 56    & 30    & 71    & 24    & 63    & 25    & 86    & 39    & 56    & 32    & 48 \\\midrule
    \multirow{3}[0]{*}{RESISC45} & Avg time/epoch (sec.) & 12.03 & 10.91 & 39.87 & 38.37 & 30.61 & 31.31 & 65.11 & 64.83 & 72.05 & 71.22 & 27.12 & 27.66 & 51.19 & 50.21 & 35.62 & 35.69 & 46.79 & 46.51 & 143.57 & 144.87 \\
      & Total training time (sec.) & 385   & 633   & 1196  & 2993  & 1163  & 1941  & 2409  & 4084  & 3098  & 5484  & 678   & 2102  & 2713  & 2611  & 1033  & 1285  & 1778  & 2279  & 4020  & 14487 \\
      & Best epoch & 22    & 43    & 20    & 63    & 28    & 47    & 27    & 48    & 33    & 62    & 15    & 61    & 43    & 37    & 19    & 21    & 28    & 34    & 18    & 85  \\\midrule
    \multirow{3}[0]{*}{PatternNet} & Avg time/epoch (sec.) & 15.17 & 13.75 & 37.74 & 37.47 & 29.1  & 35.65 & 62.94 & 69.05 & 68.87 & 71.08 & 25.86 & 27.54 & 48.5  & 49.05 & 33.8  & 34.54 & 45.93 & 45.06 & 138.65 & 138.59 \\
      & Total training time (sec.) & 637   & 1141  & 1321  & 2061  & 1193  & 3030  & 1070  & 6905  & 3168  & 5260  & 569   & 2286  & 1067  & 3237  & 1521  & 2038  & 1378  & 4326  & 2357  & 12612 \\
      & Best epoch & 32    & 68    & 25    & 40    & 31    & 70    & 7     & 88    & 36    & 59    & 12    & 68    & 12    & 51    & 35    & 44    & 20    & 81    & 7     & 76\\\midrule
    \multirow{3}[0]{*}{CLRS} & Avg time/epoch (sec.) & 20.48 & 20.35 & 20.23 & 19.33 & 18.6  & 19.43 & 31.96 & 32.05 & 35.46 & 35.81 & 19.73 & 20.71 & 25.32 & 24.96 & 19.75 & 17.98 & 23.62 & 23.09 & 68.93 & 69.19 \\
      & Total training time (sec.) & 635   & 2035  & 607   & 1450  & 279   & 1788  & 799   & 2373  & 993   & 2757  & 513   & 1512  & 785   & 1173  & 316   & 809   & 496   & 2309  & 1861  & 5535 \\
      & Best epoch & 21    & 92    & 20    & 60    & 15    & 77    & 15    & 60    & 18    & 62    & 16    & 58    & 21    & 32    & 6     & 30    & 11    & 96    & 17    & 65 \\\midrule
    \multirow{3}[0]{*}{RSD46-WHU} & Avg time/epoch (sec.) & 58.03 & 58.84 & 158.32 & 162.89 & 123.27 & 127.53 & 269.45 & 272.7 & 297.7 & 301.16 & 111.55 & 113.93 & 210.37 & 211.93 & 148.25 & 148.42 & 196.2 & 194.93 & 599.42 & 588.25 \\
      & Total training time (sec.) & 2031  & 3707  & 4433  & 8796  & 3205  & 8672  & 7814  & 19907 & 6847  & 15318 & 2231  & 6446  & 3997  & 9325  & 3558  & 4149  & 3924  & 11891 & 11389 & 41766 \\
      & Best epoch & 25    & 48    & 18    & 39    & 16    & 53    & 19    & 58    & 13    & 36    & 10    & 40    & 9     & 29    & 14    & 12    & 10    & 46    & 9     & 56 \\\midrule
    \multirow{3}[0]{*}{SAT6} & Avg time/epoch (sec.) & 92.48 & 107.26 & 550.04 & 579.1 & 410.33 & 457.04 & 872.87 & 987.21 & 970.39 & 956.03 & 363   & 420.37 & 692.5 & 687.12 & 476.34 & 479.37 & 630.78 & 627.69 & 2,003.62 & 1,973.44 \\
      & Total training time (sec.) & 5364  & 10726 & 29702 & 57910 & 37340 & 45704 & 61974 & 98721 & 55312 & 95603 & 8712  & 42037 & 42935 & 61841 & 15243 & 47937 & 42262 & 62769 & 106192 & 197344 \\
      & Best epoch & 48    & 98    & 44    & 98    & 81    & 99    & 61    & 94    & 47    & 85    & 14    & 95    & 52    & 75    & 22    & 95    & 57    & 97    & 43    & 99 \\\midrule
    \multirow{3}[0]{*}{Optimaml31} & Avg time/epoch (sec.) & 1.1   & 1.23  & 2.97  & 4.81  & 2.58  & 2.6   & 4.62  & 5.92  & 5.02  & 5.16  & 2.25  & 2.36  & 3.71  & 3.79  & 2.82  & 3.26  & 3.5   & 3.59  & 9.19  & 9.51 \\
      & Total training time (sec.) & 45    & 101   & 95    & 409   & 129   & 161   & 217   & 314   & 306   & 330   & 187   & 156   & 126   & 235   & 141   & 326   & 203   & 330   & 340   & 951 \\
      & Best epoch & 31    & 67    & 22    & 70    & 40    & 47    & 37    & 38    & 51    & 49    & 73    & 51    & 24    & 47    & 40    & 98    & 48    & 77    & 27    & 89 \\\midrule
    \multirow{3}[0]{*}{BCS} & Avg time/epoch (sec.) & 1.48  & 1.53  & 4.17  & 5.95  & 3.45  & 4.55  & 6.61  & 7.95  & 7.33  & 7.31  & 3.17  & 3.26  & 5.07  & 5.55  & 3.94  & 4.47  & 5.08  & 5.09  & 13.88 & 13.59 \\
      & Total training time (sec.) & 43    & 115   & 121   & 440   & 76    & 296   & 119   & 469   & 176   & 373   & 133   & 326   & 76    & 322   & 67    & 201   & 132   & 509   & 222   & 1169 \\
      & Best epoch & 19    & 60    & 19    & 59    & 12    & 50    & 8     & 44    & 14    & 36    & 32    & 98    & 5     & 43    & 7     & 30    & 16    & 95    & 6     & 71 \\\midrule
    \multirow{3}[0]{*}{So2Sat} & Avg time/epoch (sec.) & 158.09 & 174.74 & 716.09 & 723.72 & 565.55 & 558.79 & 1200.64 & 1198.37 & 1324.1 & 1325.67 & 510.45 & 499.21 & 925.09 & 926.5 & 643.91 & 651.31 & 853.91 & 851.06 & 2,636.45 & 2,631.44 \\
      & Total training time (sec.) & 1790  & 3320  & 7877  & 13027 & 6221  & 10617 & 13207 & 22769 & 14784 & 23862 & 5615  & 11981 & 10176 & 14824 & 7278  & 10421 & 9393  & 15319 & 29001 & 42103 \\
      & Best epoch & 1     & 4     & 1     & 3     & 1     & 4     & 1     & 4     & 1     & 3     & 1     & 9     & 1     & 1     & 1     & 1     & 1     & 3     & 1     & 1 \\\bottomrule
    \end{tabular}%
    \end{adjustbox}
  \label{tab:MCC_training_time_details}%
\end{table}%

\begin{table}[!ht]
  \centering
  \caption{Training-time details for multi-Label classification tasks.}
    \begin{adjustbox}{width=1.2\linewidth,angle=90}
    \begin{tabular}{clrr|rr|rr|rr|rr|rr|rr|rr|rr|rr}\toprule
          &       & \multicolumn{2}{c}{AlexNet} & \multicolumn{2}{c}{VGG16} & \multicolumn{2}{c}{ResNet50} & \multicolumn{2}{c}{ResNet152} & \multicolumn{2}{c}{DenseNet161} & \multicolumn{2}{c}{EfficientNetB0} & \multicolumn{2}{c}{ConvNeXt} & \multicolumn{2}{c}{ViT} & \multicolumn{2}{c}{MLPMixer} & \multicolumn{2}{c}{SwinT}\\
          &       & \multicolumn{1}{c}{Pt.} & \multicolumn{1}{c}{Sc.} & \multicolumn{1}{c}{Pt.} & \multicolumn{1}{c}{Sc.} & \multicolumn{1}{c}{Pt.} & \multicolumn{1}{c}{Sc.} & \multicolumn{1}{c}{Pt.} & \multicolumn{1}{c}{Sc.} & \multicolumn{1}{c}{Pt.} & \multicolumn{1}{c}{Sc.} & \multicolumn{1}{c}{Pt.} & \multicolumn{1}{c}{Sc.} & \multicolumn{1}{c}{Pt.} & \multicolumn{1}{c}{Sc.} & \multicolumn{1}{c}{Pt.} & \multicolumn{1}{c}{Sc.} & \multicolumn{1}{c}{Pt.} & \multicolumn{1}{c}{Sc.} & \multicolumn{1}{c}{Pt.} & \multicolumn{1}{c}{Sc.} \\\midrule
    \multirow{3}[0]{*}{AID} & Avg time/epoch (sec.) & 5.55  & 5.82  & 6.33  & 6.28  & 5.94  & 5.74  & 7.97  & 8.08  & 8.71  & 8.47  & 6.15  & 5.94  & 6.63  & 6.4   & 6.95  & 6.82  & 6.35  & 6.41 & 15.9 & 15.87 \\
          & Total training time (sec.) & 172   & 524   & 190   & 490   & 190   & 379   & 239   & 477   & 366   & 449   & 381   & 398   & 345   & 576   & 146   & 627   & 165   & 506 & 477 & 1238\\
          & Best epoch & 21    & 75    & 20    & 63    & 22    & 51    & 20    & 44    & 32    & 38    & 52    & 52    & 42    & 75    & 11    & 77    & 16    & 64 & 20 & 63 \\\midrule
    \multirow{3}[0]{*}{UCMerced } & Avg time/epoch (sec.) & 1.31  & 1.03  & 3.3   & 3.24  & 2.76  & 2.76  & 5.04  & 5.06  & 5.64  & 5.6   & 2.54  & 2.23  & 3.92  & 3.81  & 4.13  & 4.12  & 3.25  & 3.11 & 10.22 & 10.12 \\
          & Total training time (sec.) & 71    & 103   & 132   & 324   & 124   & 276   & 227   & 506   & 468   & 487   & 254   & 252   & 259   & 381   & 132   & 412   & 182   & 311 & 552 & 1012\\
          & Best epoch & 44    & 91    & 30    & 99    & 35    & 99    & 35    & 86    & 73    & 72    & 98    & 99    & 56    & 100   & 22    & 95    & 46    & 99 & 44 & 99 \\\midrule
    \multirow{3}[0]{*}{DFC15} & Avg time/epoch (sec.) & 7.74  & 7.83  & 8.94  & 8.5   & 8.49  & 8.92  & 9.45  & 9.66  & 9.54  & 9.89  & 8.33  & 8.47  & 8.72  & 8.8   & 8.76  & 8.85  & 8.18  & 8.31 & 17.59 & 17.09 \\
          & Total training time (sec.) & 325   & 783   & 286   & 799   & 331   & 464   & 444   & 647   & 544   & 613   & 583   & 686   & 471   & 880   & 219   & 743   & 229   & 831 & 686 & 1709\\
          & Best epoch & 32    & 99    & 22    & 79    & 29    & 37    & 37    & 52    & 47    & 47    & 60    & 66    & 44    & 91    & 15    & 69    & 18    & 100 & 29 & 95 \\\midrule
    \multirow{3}[0]{*}{MLRSNet} & Avg time/epoch (sec.) & 34.09 & 34.92 & 132.2 & 132.22 & 101.67 & 102.26 & 214.11 & 214.47 & 237.35 & 237.96 & 86.8  & 89.34 & 155.65 & 159.35 & 170.9 & 170.71 & 121.38 & 123.2 & 496.72 & 482.72 \\
          & Total training time (sec.) & 1125  & 2549  & 3306  & 7272  & 1726  & 6238  & 5781  & 14155 & 6171  & 11422 & 2604  & 8934  & 3580  & 5896  & 3589  & 5975  & 1942  & 3080 & 12418 & 42962\\
          & Best epoch & 23    & 58    & 15    & 40    & 16    & 46    & 17    & 51    & 16    & 33    & 20    & 87    & 13    & 22    & 11    & 20    & 6     & 10 &15 &74\\\midrule
    \multirow{3}[0]{*}{PlanetUAS} & Avg time/epoch (sec.) & 17.45 & 18.65 & 50.38 & 50.68 & 37    & 37.57 & 81.83 & 80.86 & 90.4  & 90.11 & 33.52 & 33.47 & 59.63 & 59.35 & 65.71 & 65.52 & 45.94 & 45.93 & 180.89 & 181.93 \\
          & Total training time (sec.) & 576   & 1865  & 1058  & 2889  & 740   & 2592  & 1964  & 6792  & 1808  & 4866  & 771   & 2711  & 1431  & 5935  & 920   & 4128  & 735   & 2572 & 3256 & 16365 \\
          & Best epoch & 23    & 87    & 11    & 42    & 10    & 54    & 14    & 69    & 10    & 39    & 13    & 66    & 14    & 90    & 4     & 48    & 6     & 41  & 8 & 75\\\midrule
    \multirow{3}[0]{*}{BigEarthNet 19} & Avg time/epoch (sec.) & 90.43 & 84.18 & 537.9 & 544.28 & 413.24 & 409.87 & 874.56 & 878   & 976.93 & 982.63 & 366.35 & 364.13 & 631.67 & 645.51 & 698.5 & 709.86 & 488.68 & 500.77 & 1990 & 2011.29 \\
          & Total training time (sec.) & 5245  & 5051  & 10758 & 15784 & 7025  & 18854 & 13993 & 32486 & 14654 & 29479 & 6228  & 11288 & 9475  & 26466 & 15367 & 20586 & 12217 & 15524 & 35820 & 96542  \\
          & Best epoch & 48    & 45    & 10    & 14    & 7     & 31    & 6     & 22    & 5     & 15    & 7     & 16    & 5     & 26    & 12    & 14    & 15    & 16 & 8 & 33 \\\midrule
    \multirow{3}[0]{*}{BigEarthNet 43} & Avg time/epoch (sec.) & 89.85 & 86.78 & 542.3 & 542.24 & 414.18 & 413.89 & 881.69 & 875.2 & 969.67 & 975.34 & 365.4 & 359.16 & 642.81 & 643.66 & 702   & 702.53 & 492.84 & 495.88 & 2016 & 2031.64 \\
          & Total training time (sec.) & 7188  & 5120  & 12473 & 18436 & 9112  & 26489 & 14107 & 43760 & 14545 & 31211 & 7308  & 11493 & 10285 & 24459 & 14742 & 21076 & 12321 & 15868 & 34272 & 113772  \\
          & Best epoch & 70    & 44    & 13    & 19    & 12    & 49    & 6     & 35    & 5     & 17    & 10    & 17    & 6     & 23    & 11    & 15    & 15    & 17 & 7 & 41\\\bottomrule
    \end{tabular}%
    \end{adjustbox}
  \label{tab:MLC_training_time_details}%
\end{table}%

\clearpage

\section{Extended results on model generalization performance}\label{appendix:model_generalization_results}
\setcounter{table}{0}
\setcounter{figure}{0}

\subsection{Evaluation on a same holdout set}
\begin{figure}[h]
    \centering
    \includegraphics[width=.7\linewidth]{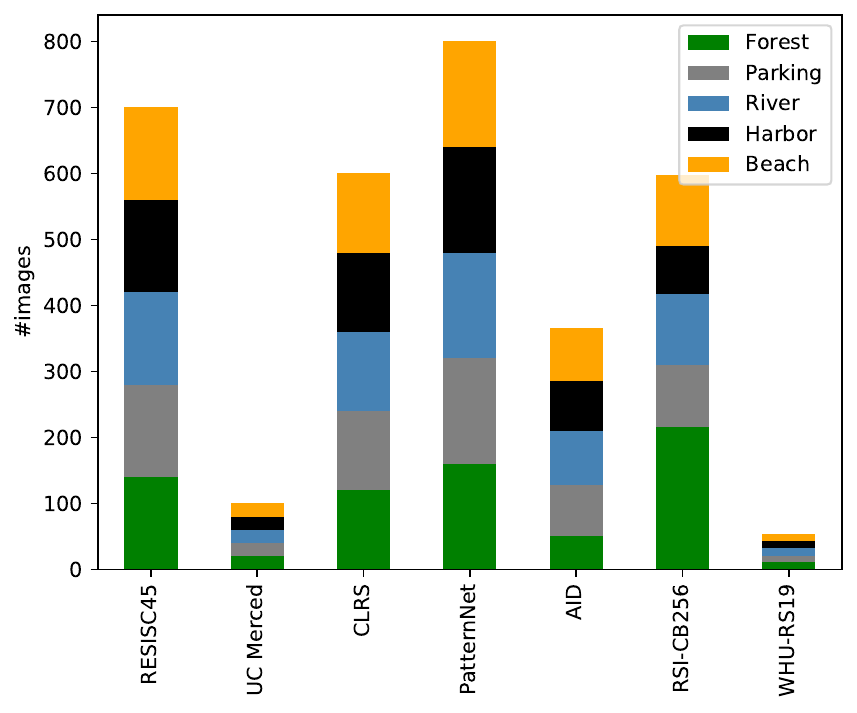}
     \caption{Distribution of images in the holdout set w.r.t. source datasets and labels.}
     \label{fig:multiclass_common_labels_aux}
\end{figure}

\begin{table}[h]\centering
\caption{Accuracy (\%) of models pre-trained on ImageNet-1K and fine-tuned on a specific source dataset and evaluated on the common test dataset with shared labels. Bold indicates best performing model for a given source dataset.}\label{tab:multiclass_common_labels_results_aux}
\scriptsize
\begin{tabular}{l|rrrrrrrrrr}\toprule
Dataset \textbackslash Model &AlexNet &VGG16 &ResNet50 &ResNet152 &DenseNet161 &EfficientNetB0 &ViT &MLPMixer &ConvNeXt &SwinT \\\midrule
RESISC45 & 66.853 & 78.514 & 81.063 & 84.08 & 84.111 & 77.985 & \textbf{86.007} & 82.121 & 84.422 & 83.706 \\
UC Merced & 63.371 & 67.04 & 76.057 & 73.01 & 74.254 & 74.44 & 75.995 & \textbf{79.478} & 75.902 & 72.326 \\
CLRS & 80.037 & 83.427 & 89.801 & 88.557 & 89.024 & 86.07 & \textbf{92.6} & 89.646 & 89.303 & 90.299 \\ 
PatternNet & 43.501 & 52.332 & 56.965 & 54.54 & 56.716 & 60.044 & 64.739 & 62.687 & 59.391 & \textbf{65.205} \\
AID & 71.393 & 69.714 & 79.384 & 80.1 & 66.169 & 77.892 & \textbf{83.862} & 77.954 & 79.851 & 79.789 \\
RSI-CB256 & 56.872 & 61.412 & 58.893 & 63.65 & 64.832 & 61.723 & 66.014 & \textbf{66.791} & 64.677 & 66.294 \\
WHU-RS19 & 61.101 & 62.624 & 71.953 & 73.321 & 72.388 & 68.284 & 72.917 & 74.036 & \textbf{74.876} & 71.144 \\
\midrule
\textit{Avg. Rank} & 9.71	& 8.71	& 5.43	& 5.29	& 5.86	& 6.86	& \textbf{2.14}	& 3.29	& 3.57	& 4.14
 \\
\bottomrule
\end{tabular}
\end{table}

\clearpage
\subsection{Results from pairwise comparisons}

\begin{figure}[h]
    \centering
    \begin{subfigure}[t]{0.6\textwidth}
    \includegraphics[width=1\linewidth]{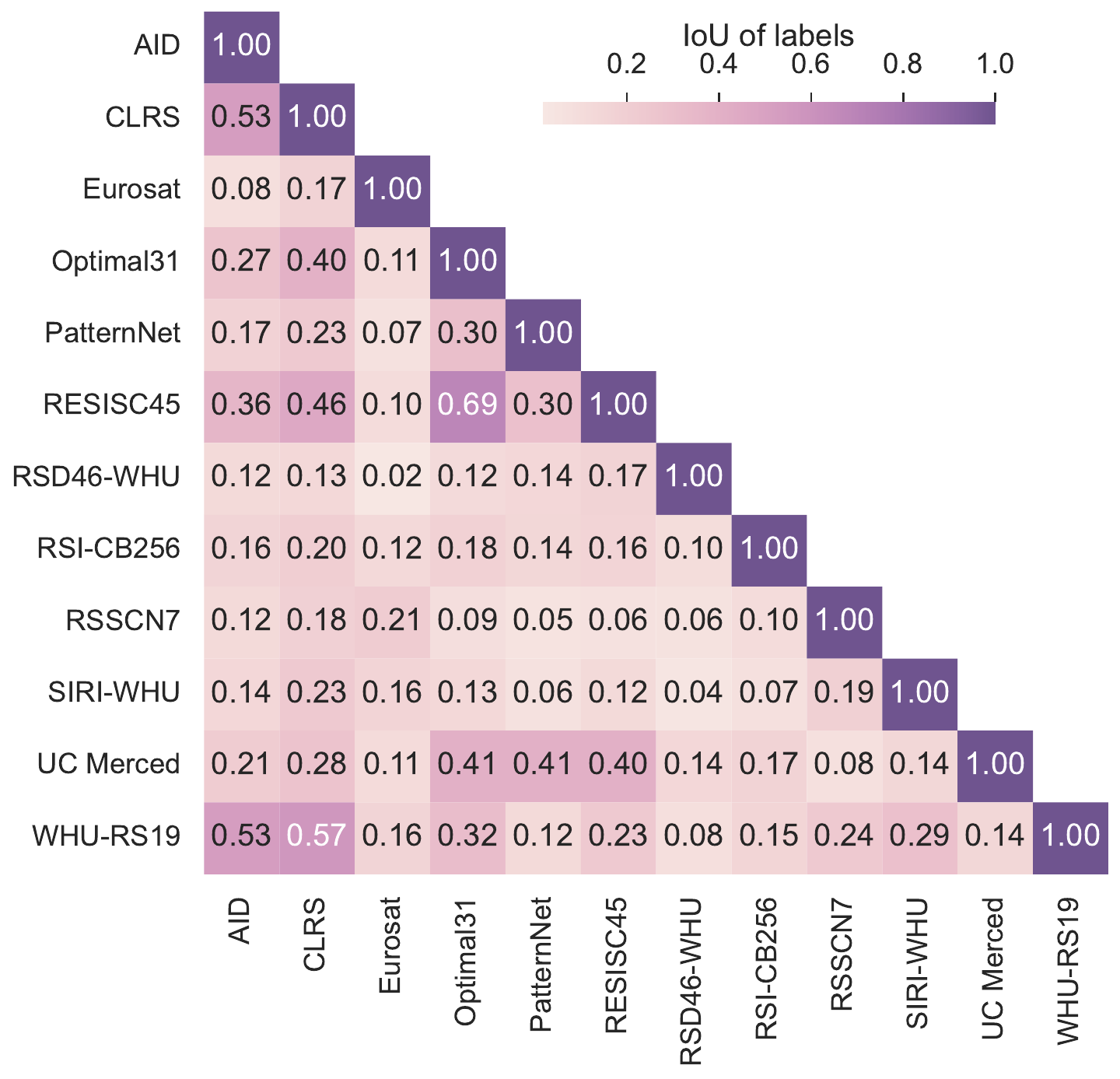}
    \subcaption{}
    \end{subfigure}
    \hfill
    \begin{subfigure}[t]{0.39\textwidth}
    \includegraphics[width=1\linewidth]{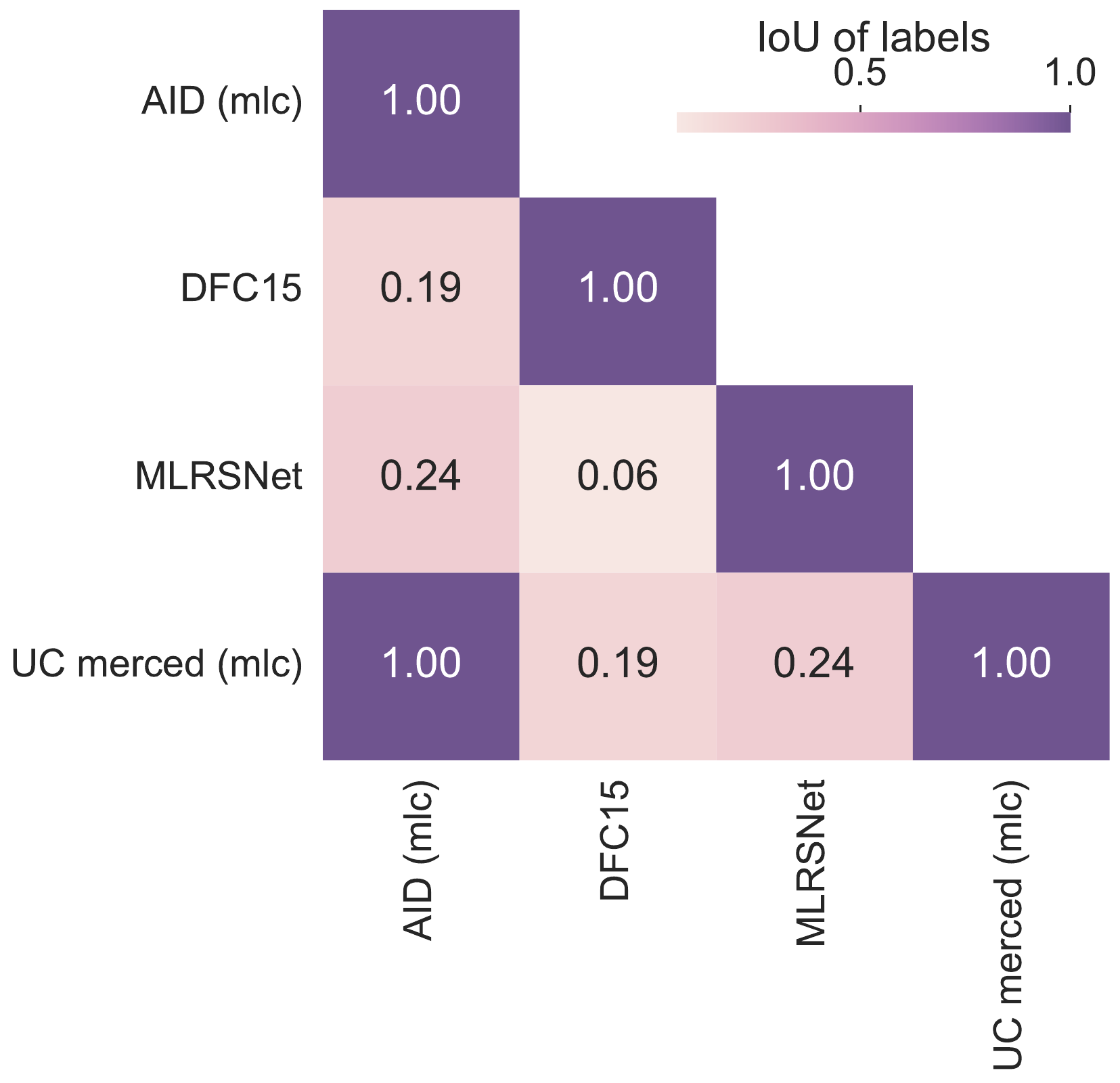}
    \subcaption{}
    \end{subfigure}
    \caption{Label overlap between each pair of datasets, in terms of Intersection over Union (IoU) for (a) MCC and (b) MLC datasets.}
    \label{fig:generalization_results_aux}
\end{figure}

\begin{table*}[!ht]
\centering
\caption{AlexNet on MCC tasks: Generalization performance in terms of accuracy (\%) of ImageNet-1K pre-trained models, fine-tuned on \textit{source} and evaluated on images with shared labels in \textit{target} dataset.}
\begin{adjustbox}{width=1.0\linewidth}
\begin{tabular}{l|ccccccccccccc}
\toprule
Source \textbackslash Target dataset & RESISC45 & UC Merced & CLRS & Optimal31 & PatternNet & AID & RSI-CB256 & WHU-RS19 & SIRI-WHU & RSD46-WHU & Eurosat & SAT6 & RSSCN7 \\\midrule
RESISC45 & 90.492 & 57.368 & 49.053 & 96.505 & 76.217 & 70.802 & 57.793 & 67.717 & 43.333 & 30.377 & 0.704 & 0.0 & 39.167 \\
UC Merced & 51.729 & 92.143 & 37.583 & 62.222 & 74.522 & 50.94 & 75.182 & 56.604 & 9.375 & 20.656 & 0.438 & 0.57 & 46.875\\
CLRS & 67.64 & 60.5 & 84.1 & 71.875 & 72.135 & 79.601 & 67.363 & 72.024 & 54.643 & 53.717 & 16.963 & 0.0 & 51.0 \\
Optimal31 & 74.654 & 65.333 & 44.635 & 80.914 & 77.656 & 65.704 & 59.596 & 66.142 & 35.0 & 27.216 & 1.045 & 0.0 & 33.333 \\
PatternNet & 39.248 & 47.059 & 20.278 & 57.292 & 99.161 & 31.322 & 33.952 & 23.81 & 11.667 & 16.095 & 30.062 & 0.0 & 11.875 \\
AID & 67.107 & 57.778 & 54.298 & 78.846 & 62.125 & 92.9 & 57.5 & 98.333 & 27.5 & 37.144 & 1.312 & 7.437 & 45.0 \\
RSI-CB256 & 45.13 & 70.0 & 32.833 & 50.833 & 71.319 & 50.469 & 99.354 & 61.644 & 29.167 & 36.944 & 29.821 & 10.597 & 16.25 \\
WHU-RS19 & 54.702 & 74.0 & 42.396 & 60.417 & 63.125 & 72.968 & 64.938 & 93.532 & 26.429 & 42.055 & 0.0 & 0.0 & 48.25 \\
SIRI-WHU & 59.167 & 26.25 & 59.167 & 63.333 & 58.542 & 57.735 & 29.446 & 51.316 & 92.292 & 63.991 & 28.875 & 8.481 & 26.25 \\
RSD46-WHU & 35.714 & 35.625 & 44.896 & 54.167 & 44.062 & 49.643 & 38.433 & 59.615 & 60.0 & 90.646 & 97.0 & 2.379 & 48.75 \\
Eurosat & 24.286 & 15.0 & 39.667 & 27.083 & 37.083 & 25.714 & 37.092 & 46.512 & 67.5 & 26.108 & 97.574 & 0.017 & 57.083 \\
SAT6 & 0.0 & 7.5 & 14.167 & 0.0 & 0.0 & 31.25 & 5.053 & 0.0 & 5.0 & 6.528 & 13.333 & 99.98 & 88.75 \\
RSSCN7 & 73.095 & 70.0 & 59.833 & 69.444 & 68.125 & 75.357 & 43.833 & 84.615 & 92.5 & 45.268 & 2.529 & 0.032 & 91.964 \\
\bottomrule
\end{tabular}
\end{adjustbox}
\label{table:pair_wise_alexnet_mcc}
\end{table*}

\begin{table*}[!ht]
\centering
\caption{VGG16 on MCC tasks: Generalization performance in terms of accuracy (\%) of ImageNet-1K pre-trained models, fine-tuned on \textit{source} and evaluated on images with shared labels in \textit{target} dataset.}
\begin{adjustbox}{width=1.0\linewidth}
\begin{tabular}{l|ccccccccccccc}
\toprule
Source \textbackslash Target dataset & RESISC45 & UC Merced & CLRS & Optimal31 & PatternNet & AID & RSI-CB256 & WHU-RS19 & SIRI-WHU & RSD46-WHU & Eurosat & SAT6 & RSSCN7 \\\midrule
RESISC45 & 93.905 & 76.579 & 65.227 & 98.656 & 81.941 & 78.752 & 67.052 & 76.378 & 48.333 & 43.589 & 0.407 & 0.0 & 52.083 \\
UC Merced & 55.94 & 95.476 & 40.417 & 67.222 & 79.449 & 58.464 & 83.577 & 67.925 & 3.75 & 26.019 & 0.125 & 0.223 & 41.25 \\
CLRS & 76.615 & 68.0 & 89.9 & 81.771 & 75.781 & 86.273 & 73.392 & 82.143 & 66.429 & 65.137 & 14.259 & 0.012 & 52.75 \\
Optimal31 & 79.378 & 71.333 & 53.229 & 88.71 & 80.977 & 67.09 & 59.091 & 63.78 & 50.5 & 36.299 & 6.0 & 0.0 & 37.917 \\
PatternNet & 45.263 & 55.588 & 29.306 & 61.458 & 99.424 & 41.379 & 52.293 & 31.746 & 25.0 & 23.98 & 29.0 & 0.0 & 26.25 \\
AID & 70.071 & 61.111 & 62.105 & 79.487 & 67.062 & 96.1 & 61.083 & 97.778 & 42.0 & 45.984 & 1.188 & 0.12 & 47.188 \\
RSI-CB256 & 47.922 & 74.375 & 35.083 & 57.5 & 74.931 & 58.75 & 99.051 & 69.863 & 24.167 & 39.985 & 23.143 & 1.621 & 35.0 \\
WHU-RS19 & 55.179 & 72.0 & 44.01 & 61.111 & 68.125 & 78.69 & 66.183 & 99.005 & 23.571 & 47.469 & 0.0 & 0.0 & 43.25 \\
SIRI-WHU & 53.452 & 11.25 & 61.429 & 68.333 & 36.667 & 56.906 & 27.988 & 61.842 & 93.958 & 74.734 & 30.375 & 18.881 & 25.833 \\
RSD46-WHU & 41.319 & 44.375 & 50.938 & 54.167 & 40.562 & 40.893 & 31.946 & 57.692 & 48.75 & 92.422 & 97.5 & 3.569 & 50.417 \\
Eurosat & 35.571 & 23.333 & 55.667 & 43.75 & 50.0 & 74.286 & 25.272 & 55.814 & 76.667 & 45.32 & 98.148 & 3.958 & 55.417 \\
SAT6 & 81.429 & 37.5 & 52.083 & 58.333 & 100.0 & 63.393 & 7.979 & 90.909 & 0.0 & 18.62 & 0.167 & 99.993 & 31.25 \\
RSSCN7 & 78.81 & 90.0 & 75.5 & 77.778 & 86.562 & 86.429 & 63.188 & 94.231 & 75.833 & 66.911 & 7.059 & 0.143 & 93.929 \\
\bottomrule
\end{tabular}
\end{adjustbox}
\label{table:pair_wise_vgg16_mcc}
\end{table*}

\begin{table*}[!ht]
\centering
\caption{ResNet50 on MCC tasks: Generalization performance in terms of accuracy (\%) of ImageNet-1K pre-trained models, fine-tuned on \textit{source} and evaluated on images with shared labels in \textit{target} dataset.}
\begin{adjustbox}{width=1.0\linewidth}
\begin{tabular}{l|ccccccccccccc}
\toprule
Source \textbackslash Target dataset & RESISC45 & UC Merced & CLRS & Optimal31 & PatternNet & AID & RSI-CB256 & WHU-RS19 & SIRI-WHU & RSD46-WHU & Eurosat & SAT6 & RSSCN7 \\\midrule
RESISC45 & 96.46 & 72.105 & 67.083 & 99.194 & 84.178 & 79.941 & 72.377 & 81.89 & 56.25 & 42.872 & 9.185 & 0.021 & 56.667 \\
UC Merced & 63.496 & 98.571 & 49.25 & 75.0 & 88.088 & 68.182 & 91.241 & 86.792 & 20.0 & 34.753 & 4.75 & 7.073 & 46.25\\
CLRS & 80.435 & 73.5 & 91.567 & 86.979 & 85.573 & 88.574 & 79.984 & 87.5 & 71.071 & 67.582 & 33.407 & 0.427 & 51.25\\
Optimal31 & 86.636 & 73.333 & 63.906 & 92.204 & 87.07 & 82.91 & 75.337 & 77.953 & 60.0 & 37.291 & 5.955 & 2.806 & 43.333\\
PatternNet & 46.128 & 54.412 & 27.847 & 65.104 & 99.737 & 39.224 & 64.847 & 44.444 & 17.5 & 23.73 & 29.938 & 0.0 & 26.25\\
AID & 75.321 & 64.444 & 66.886 & 80.769 & 73.75 & 96.55 & 72.167 & 99.444 & 42.0 & 47.474 & 17.938 & 3.662 & 50.0\\
RSI-CB256 & 49.545 & 68.75 & 36.583 & 56.667 & 74.931 & 58.125 & 99.677 & 60.274 & 29.167 & 39.985 & 37.036 & 0.743 & 17.812\\
WHU-RS19 & 71.905 & 89.0 & 56.198 & 78.472 & 69.375 & 86.816 & 66.805 & 99.502 & 44.643 & 48.221 & 0.364 & 30.899 & 53.25\\
SIRI-WHU & 55.0 & 20.0 & 60.476 & 61.667 & 42.083 & 52.762 & 27.697 & 55.263 & 95.0 & 71.901 & 47.562 & 41.35 & 22.917\\
RSD46-WHU & 41.374 & 45.0 & 45.833 & 56.25 & 42.688 & 35.714 & 36.475 & 44.231 & 42.5 & 94.153 & 98.833 & 10.93 & 44.167\\
Eurosat & 32.571 & 10.0 & 45.333 & 25.0 & 7.292 & 42.381 & 32.473 & 62.791 & 78.333 & 87.438 & 98.833 & 1.999 & 55.417\\
SAT6 & 0.714 & 7.5 & 17.083 & 8.333 & 0.0 & 37.5 & 5.053 & 0.0 & 0.0 & 7.196 & 0.167 & 100.0 & 68.75\\
RSSCN7 & 79.048 & 97.5 & 74.333 & 77.778 & 92.5 & 82.143 & 57.306 & 88.462 & 71.667 & 68.379 & 4.765 & 0.008 & 95.0\\
\bottomrule
\end{tabular}
\end{adjustbox}
\label{table:pair_wise_resnet50_mcc}
\end{table*}

\begin{table*}[!ht]
\centering
\caption{ResNet152 on MCC tasks: Generalization performance in terms of accuracy (\%) of ImageNet-1K pre-trained models, fine-tuned on \textit{source} and evaluated on images with shared labels in \textit{target} dataset.}
\begin{adjustbox}{width=1.0\linewidth}
\begin{tabular}{l|ccccccccccccc}
\toprule
Source \textbackslash Target dataset & RESISC45 & UC Merced & CLRS & Optimal31 & PatternNet & AID & RSI-CB256 & WHU-RS19 & SIRI-WHU & RSD46-WHU & Eurosat & SAT6 & RSSCN7 \\\midrule
RESISC45 & 96.54 & 72.895 & 68.485 & 98.656 & 85.066 & 80.684 & 73.38 & 83.465 & 59.583 & 47.194 & 18.63 & 0.331 & 52.5\\
UC Merced & 61.729 & 98.81 & 46.417 & 73.889 & 89.743 & 63.95 & 89.173 & 69.811 & 22.5 & 34.018 & 15.188 & 13.09 & 47.5\\
CLRS & 81.553 & 70.0 & 91.933 & 84.896 & 80.469 & 88.42 & 79.984 & 85.119 & 70.0 & 71.065 & 46.259 & 2.62 & 51.75\\
Optimal31 & 87.972 & 72.667 & 65.99 & 92.473 & 86.797 & 82.217 & 82.071 & 78.74 & 56.5 & 38.252 & 6.273 & 18.893 & 42.5\\
PatternNet & 51.053 & 49.412 & 33.681 & 63.542 & 99.49 & 35.776 & 61.463 & 28.571 & 30.0 & 30.013 & 24.875 & 0.0 & 15.0\\
AID & 79.143 & 61.667 & 69.298 & 80.769 & 73.125 & 97.2 & 73.083 & 98.889 & 36.5 & 45.045 & 11.438 & 17.338 & 44.688\\
RSI-CB256 & 53.312 & 73.125 & 37.75 & 62.5 & 78.472 & 58.438 & 99.859 & 64.384 & 27.5 & 44.325 & 27.643 & 19.238 & 20.625\\
WHU-RS19 & 72.798 & 87.0 & 57.708 & 79.167 & 66.25 & 86.567 & 73.755 & 98.01 & 40.0 & 51.83 & 10.409 & 0.141 & 49.5\\
SIRI-WHU & 66.19 & 50.0 & 67.262 & 73.333 & 74.375 & 66.298 & 40.233 & 57.895 & 96.25 & 84.061 & 55.062 & 18.092 & 27.083\\
RSD46-WHU & 44.396 & 43.75 & 49.062 & 56.25 & 45.438 & 43.929 & 39.045 & 51.923 & 46.25 & 94.404 & 95.5 & 1.968 & 40.0\\
Eurosat & 28.857 & 6.667 & 38.167 & 27.083 & 3.75 & 50.476 & 29.62 & 27.907 & 71.667 & 85.222 & 99.0 & 1.034 & 47.083\\
SAT6 & 0.0 & 10.0 & 26.25 & 0.0 & 0.0 & 46.429 & 19.681 & 0.0 & 15.0 & 6.825 & 2.833 & 100.0 & 68.75\\
RSSCN7 & 76.19 & 87.5 & 73.167 & 83.333 & 83.75 & 86.786 & 55.977 & 94.231 & 88.333 & 52.311 & 36.176 & 0.087 & 95.0\\
\bottomrule
\end{tabular}
\end{adjustbox}
\label{table:pair_wise_resnet152_mcc}
\end{table*}

\begin{table*}[!ht]
\centering
\caption{DenseNet161 on MCC tasks: Generalization performance in terms of accuracy (\%) of ImageNet-1K pre-trained models, fine-tuned on \textit{source} and evaluated on images with shared labels in \textit{target} dataset.}
\begin{adjustbox}{width=1.0\linewidth}
\begin{tabular}{l|ccccccccccccc}
\toprule
Source \textbackslash Target dataset & RESISC45 & UC Merced & CLRS & Optimal31 & PatternNet & AID & RSI-CB256 & WHU-RS19 & SIRI-WHU & RSD46-WHU & Eurosat & SAT6 & RSSCN7 \\\midrule
RESISC45 & 96.508 & 75.0 & 69.015 & 98.925 & 87.862 & 78.975 & 76.543 & 81.89 & 60.0 & 49.385 & 10.111 & 29.235 & 56.25\\
UC Merced & 63.271 & 98.333 & 50.917 & 73.889 & 87.059 & 68.025 & 92.214 & 75.472 & 16.875 & 31.995 & 12.25 & 5.067 & 55.0\\
CLRS & 80.994 & 72.5 & 92.2 & 88.021 & 81.146 & 88.344 & 79.502 & 86.31 & 72.857 & 71.199 & 38.519 & 1.229 & 49.75\\
Optimal31 & 88.641 & 77.0 & 65.469 & 94.355 & 87.461 & 83.603 & 78.283 & 78.74 & 55.5 & 40.67 & 6.409 & 30.969 & 44.167\\
PatternNet & 52.105 & 60.294 & 32.083 & 69.792 & 99.737 & 41.236 & 63.1 & 39.683 & 20.833 & 25.006 & 19.5 & 0.021 & 26.25\\
AID & 66.25 & 39.444 & 45.175 & 76.923 & 58.75 & 88.85 & 49.083 & 93.333 & 24.5 & 26.295 & 3.125 & 20.976 & 38.125\\
RSI-CB256 & 52.208 & 68.75 & 41.75 & 55.833 & 72.292 & 63.125 & 99.737 & 71.233 & 30.833 & 45.03 & 39.464 & 0.145 & 23.75\\
WHU-RS19 & 74.821 & 91.0 & 59.167 & 79.861 & 69.167 & 87.396 & 69.295 & 100.0 & 45.0 & 53.233 & 1.591 & 0.007 & 55.5\\
SIRI-WHU & 55.833 & 33.75 & 65.714 & 55.0 & 53.333 & 59.945 & 34.111 & 56.579 & 95.625 & 76.269 & 26.188 & 82.087 & 35.833\\
RSD46-WHU & 41.758 & 44.375 & 49.271 & 50.0 & 41.25 & 47.5 & 36.353 & 53.846 & 38.75 & 94.507 & 96.0 & 0.711 & 48.75\\
Eurosat & 5.857 & 1.667 & 20.5 & 0.0 & 5.417 & 2.381 & 39.402 & 25.581 & 57.5 & 76.601 & 98.889 & 0.2 & 30.417\\
SAT6 & 0.0 & 5.0 & 32.083 & 0.0 & 0.0 & 50.893 & 33.511 & 0.0 & 0.0 & 0.89 & 0.0 & 100.0 & 63.75\\
RSSCN7 & 85.0 & 97.5 & 80.5 & 80.556 & 93.125 & 91.429 & 59.013 & 92.308 & 81.667 & 63.316 & 11.941 & 0.921 & 94.821\\
\bottomrule
\end{tabular}
\end{adjustbox}
\label{table:pair_wise_densenet161_mcc}
\end{table*}

\begin{table*}[!ht]
\centering
\caption{EfficientNetB0 on MCC tasks: Generalization performance in terms of accuracy (\%) of ImageNet-1K pre-trained models, fine-tuned on \textit{source} and evaluated on images with shared labels in \textit{target} dataset.}
\begin{adjustbox}{width=1.0\linewidth}
\begin{tabular}{l|ccccccccccccc}
\toprule
Source \textbackslash Target dataset & RESISC45 & UC Merced & CLRS & Optimal31 & PatternNet & AID & RSI-CB256 & WHU-RS19 & SIRI-WHU & RSD46-WHU & Eurosat & SAT6 & RSSCN7 \\\midrule
RESISC45 & 94.873 & 71.053 & 65.303 & 98.656 & 81.316 & 76.003 & 68.364 & 74.016 & 53.333 & 45.576 & 3.185 & 0.0 & 45.417\\
UC Merced & 63.571 & 98.571 & 48.333 & 75.0 & 89.338 & 63.95 & 87.47 & 71.698 & 11.25 & 36.163 & 10.75 & 33.058 & 41.25\\
CLRS & 80.373 & 71.0 & 90.5 & 83.333 & 82.604 & 87.807 & 76.206 & 82.738 & 68.929 & 67.716 & 11.37 & 0.541 & 56.5\\
Optimal31 & 86.06 & 74.333 & 64.583 & 91.667 & 85.469 & 78.984 & 74.327 & 72.441 & 61.0 & 41.538 & 4.864 & 0.804 & 35.0\\
PatternNet & 50.338 & 58.824 & 35.833 & 69.271 & 99.539 & 42.529 & 65.611 & 49.206 & 17.5 & 25.732 & 22.75 & 1.17 & 34.375\\
AID & 75.286 & 65.556 & 63.377 & 80.769 & 73.25 & 96.25 & 73.75 & 98.333 & 44.5 & 43.102 & 4.875 & 1.346 & 44.062\\
RSI-CB256 & 49.481 & 75.625 & 38.75 & 55.833 & 76.458 & 58.281 & 99.717 & 71.233 & 27.5 & 43.731 & 31.893 & 2.586 & 18.75\\
WHU-RS19 & 70.774 & 89.0 & 51.771 & 74.306 & 67.188 & 85.489 & 79.357 & 99.502 & 47.5 & 47.268 & 0.0 & 0.0 & 48.0\\
SIRI-WHU & 61.548 & 40.0 & 60.833 & 65.0 & 65.625 & 52.21 & 36.735 & 55.263 & 95.0 & 74.262 & 11.188 & 4.447 & 25.833\\
RSD46-WHU & 42.418 & 43.75 & 47.812 & 46.875 & 44.562 & 41.786 & 38.433 & 51.923 & 31.25 & 93.399 & 95.0 & 4.666 & 41.25\\
Eurosat & 8.857 & 13.333 & 8.5 & 10.417 & 0.0 & 8.571 & 13.315 & 6.977 & 5.833 & 50.493 & 98.907 & 54.596 & 2.5\\
SAT6 & 6.429 & 10.0 & 5.417 & 8.333 & 1.875 & 12.5 & 13.298 & 0.0 & 0.0 & 7.196 & 0.0 & 99.988 & 30.0\\
RSSCN7 & 84.762 & 85.0 & 75.833 & 86.111 & 93.125 & 87.143 & 56.546 & 90.385 & 72.5 & 63.023 & 8.176 & 0.079 & 95.536\\
\bottomrule
\end{tabular}
\end{adjustbox}
\label{table:pair_wise_efficientnetb0_mcc}
\end{table*}

\begin{table*}[!ht]
\centering
\caption{ViT on MCC tasks: Generalization performance in terms of accuracy (\%) of ImageNet-1K pre-trained models, fine-tuned on \textit{source} and evaluated on images with shared labels in \textit{target} dataset.}
\begin{adjustbox}{width=1.0\linewidth}
\begin{tabular}{l|ccccccccccccc}
\toprule
Source \textbackslash Target dataset & RESISC45 & UC Merced & CLRS & Optimal31 & PatternNet & AID & RSI-CB256 & WHU-RS19 & SIRI-WHU & RSD46-WHU & Eurosat & SAT6 & RSSCN7 \\\midrule
RESISC45 & 97.079 & 87.368 & 71.553 & 98.118 & 90.461 & 83.804 & 81.096 & 88.189 & 62.5 & 54.015 & 15.519 & 5.992 & 51.667\\
UC Merced & 65.038 & 98.333 & 51.333 & 75.556 & 89.007 & 69.436 & 88.2 & 83.019 & 25.625 & 41.404 & 24.375 & 28.516 & 41.875\\
CLRS & 82.484 & 83.5 & 93.2 & 86.979 & 88.177 & 91.411 & 86.897 & 85.714 & 74.643 & 73.543 & 45.741 & 7.806 & 56.75\\
Optimal31 & 89.055 & 87.667 & 69.844 & 94.624 & 87.344 & 86.374 & 85.017 & 81.89 & 68.0 & 47.892 & 4.5 & 54.036 & 45.0\\
PatternNet & 55.038 & 70.882 & 42.639 & 74.479 & 99.655 & 51.724 & 70.197 & 63.492 & 49.167 & 36.496 & 30.875 & 0.0 & 38.75\\
AID & 81.393 & 81.111 & 74.342 & 87.821 & 78.5 & 97.75 & 70.083 & 98.333 & 55.5 & 54.145 & 37.875 & 33.276 & 46.562\\
RSI-CB256 & 57.857 & 81.25 & 48.25 & 67.5 & 84.236 & 68.75 & 99.758 & 83.562 & 27.5 & 54.117 & 33.821 & 11.852 & 27.812\\
WHU-RS19 & 66.369 & 81.0 & 60.729 & 77.083 & 77.396 & 87.313 & 71.473 & 99.502 & 43.214 & 56.391 & 5.455 & 23.313 & 52.0\\
SIRI-WHU & 70.119 & 60.0 & 72.262 & 76.667 & 81.667 & 62.155 & 34.111 & 67.105 & 95.625 & 88.548 & 31.062 & 7.107 & 24.167\\
RSD46-WHU & 47.637 & 47.5 & 52.708 & 60.417 & 50.875 & 57.857 & 43.819 & 73.077 & 81.25 & 94.238 & 99.0 & 10.62 & 51.25\\
Eurosat & 51.857 & 48.333 & 70.167 & 56.25 & 58.958 & 93.333 & 55.435 & 83.721 & 77.5 & 70.936 & 98.722 & 2.132 & 62.5\\
SAT6 & 90.714 & 30.0 & 55.0 & 91.667 & 80.625 & 57.143 & 7.181 & 90.909 & 0.0 & 4.599 & 1.833 & 99.998 & 86.25\\
RSSCN7 & 81.429 & 85.0 & 82.833 & 83.333 & 90.625 & 90.0 & 69.829 & 86.538 & 45.0 & 65.297 & 20.412 & 56.78 & 95.893\\
\bottomrule
\end{tabular}
\end{adjustbox}
\label{table:pair_wise_vit_mcc}
\end{table*}

\begin{table*}[!ht]
\centering
\caption{MLPMixer on MCC tasks: Generalization performance in terms of accuracy (\%) of ImageNet-1K pre-trained models, fine-tuned on \textit{source} and evaluated on images with shared labels in \textit{target} dataset.}
\begin{adjustbox}{width=1.0\linewidth}
\begin{tabular}{l|ccccccccccccc}
\toprule
Source \textbackslash Target dataset & RESISC45 & UC Merced & CLRS & Optimal31 & PatternNet & AID & RSI-CB256 & WHU-RS19 & SIRI-WHU & RSD46-WHU & Eurosat & SAT6 & RSSCN7 \\\midrule
RESISC45 & 95.952 & 81.842 & 65.947 & 98.925 & 86.25 & 78.455 & 78.164 & 83.465 & 52.5 & 47.255 & 11.926 & 1.255 & 47.5\\
UC Merced & 64.474 & 98.333 & 52.417 & 72.778 & 91.471 & 68.966 & 93.187 & 86.792 & 27.5 & 36.929 & 29.75 & 27.488 & 53.75\\
CLRS & 80.373 & 67.0 & 90.1 & 83.333 & 80.365 & 88.42 & 80.305 & 84.524 & 69.286 & 68.252 & 34.37 & 3.88 & 49.25\\
Optimal31 & 87.12 & 74.667 & 65.938 & 92.742 & 86.797 & 82.679 & 80.303 & 86.614 & 64.0 & 39.678 & 8.364 & 6.486 & 41.667\\
PatternNet & 52.481 & 68.529 & 39.792 & 67.708 & 99.704 & 51.868 & 67.686 & 49.206 & 48.333 & 29.662 & 26.312 & 0.233 & 24.375\\
AID & 77.929 & 68.333 & 67.544 & 78.205 & 71.688 & 96.7 & 68.417 & 99.444 & 49.5 & 46.179 & 34.438 & 21.946 & 45.312\\
RSI-CB256 & 56.169 & 82.5 & 45.5 & 67.5 & 80.139 & 65.312 & 99.657 & 80.822 & 32.5 & 45.438 & 36.179 & 13.455 & 20.938\\
WHU-RS19 & 68.452 & 84.0 & 59.167 & 77.778 & 69.896 & 86.816 & 84.44 & 98.507 & 47.143 & 58.095 & 2.5 & 6.472 & 55.25\\
SIRI-WHU & 67.262 & 38.75 & 72.024 & 80.0 & 77.292 & 66.575 & 34.694 & 65.789 & 95.208 & 82.999 & 40.25 & 0.978 & 33.333\\
RSD46-WHU & 41.264 & 41.25 & 49.375 & 50.0 & 49.25 & 54.821 & 43.696 & 65.385 & 50.0 & 93.667 & 96.833 & 7.126 & 53.75\\
Eurosat & 53.857 & 35.0 & 63.5 & 54.167 & 60.208 & 93.333 & 56.929 & 81.395 & 70.0 & 68.473 & 98.741 & 0.685 & 60.833\\
SAT6 & 97.143 & 50.0 & 55.833 & 100.0 & 100.0 & 61.607 & 6.117 & 90.909 & 0.0 & 9.125 & 0.0 & 99.995 & 77.5\\
RSSCN7 & 84.286 & 95.0 & 81.333 & 86.111 & 91.562 & 92.857 & 65.844 & 96.154 & 95.833 & 62.362 & 13.588 & 2.358 & 95.179\\
\bottomrule
\end{tabular}
\end{adjustbox}
\label{table:pair_wise_mlpmixer_mcc}
\end{table*}

\begin{table*}[!ht]
\centering
\caption{ConvNeXt on MCC tasks: Generalization performance in terms of accuracy (\%) of ImageNet-1K pre-trained models, fine-tuned on \textit{source} and evaluated on images with shared labels in \textit{target} dataset.}
\begin{adjustbox}{width=1.0\linewidth}
\begin{tabular}{l|ccccccccccccc}
\toprule
Source \textbackslash Target dataset & RESISC45 & UC Merced & CLRS & Optimal31 & PatternNet & AID & RSI-CB256 & WHU-RS19 & SIRI-WHU & RSD46-WHU & Eurosat & SAT6 & RSSCN7 \\\midrule
RESISC45 & 96.27 & 82.895 & 71.136 & 98.656 & 90.099 & 81.798 & 78.318 & 81.89 & 56.25 & 52.253 & 11.444 & 0.049 & 55.0\\
UC Merced & 64.436 & 97.857 & 52.667 & 72.778 & 91.066 & 66.928 & 91.971 & 73.585 & 18.125 & 38.737 & 5.438 & 2.732 & 50.625\\
CLRS & 81.056 & 71.5 & 91.1 & 86.458 & 86.875 & 92.101 & 79.26 & 88.095 & 71.071 & 72.907 & 33.111 & 0.184 & 52.75\\
Optimal31 & 87.581 & 82.667 & 65.521 & 93.011 & 89.18 & 84.527 & 81.145 & 86.614 & 63.5 & 47.861 & 6.773 & 3.807 & 50.417\\
PatternNet & 53.947 & 72.059 & 40.208 & 70.312 & 99.671 & 53.879 & 59.716 & 52.381 & 39.167 & 31.79 & 7.062 & 0.0 & 45.0\\
AID & 76.143 & 63.889 & 68.158 & 82.051 & 73.812 & 96.95 & 71.583 & 99.444 & 51.5 & 52.85 & 21.375 & 18.89 & 42.188\\
RSI-CB256 & 53.701 & 82.5 & 42.75 & 60.833 & 79.028 & 66.094 & 99.596 & 76.712 & 25.0 & 50.408 & 27.786 & 14.977 & 29.688\\
WHU-RS19 & 63.75 & 80.0 & 53.958 & 70.139 & 76.354 & 88.723 & 70.539 & 99.005 & 38.571 & 51.579 & 0.0 & 0.0 & 56.75\\
SIRI-WHU & 68.214 & 41.25 & 69.524 & 80.0 & 76.458 & 60.773 & 33.819 & 64.474 & 96.25 & 86.423 & 48.812 & 8.474 & 27.917\\
RSD46-WHU & 49.121 & 46.875 & 51.562 & 62.5 & 48.125 & 47.321 & 39.168 & 55.769 & 62.5 & 93.627 & 97.333 & 7.778 & 51.25\\
Eurosat & 56.286 & 55.0 & 69.167 & 54.167 & 64.792 & 94.762 & 48.641 & 93.023 & 80.833 & 39.655 & 98.778 & 0.479 & 75.833\\
SAT6 & 86.429 & 50.0 & 45.0 & 91.667 & 65.0 & 42.857 & 3.723 & 90.909 & 0.0 & 10.015 & 1.833 & 99.999 & 47.5\\
RSSCN7 & 82.857 & 100.0 & 80.833 & 83.333 & 89.375 & 88.929 & 63.188 & 94.231 & 68.333 & 59.501 & 8.118 & 1.953 & 94.643\\
\bottomrule
\end{tabular}
\end{adjustbox}
\label{table:pair_wise_convnext_mcc}
\end{table*}

\begin{table*}[!ht]
\centering
\caption{SwinT on MCC tasks: Generalization performance in terms of accuracy (\%) of ImageNet-1K pre-trained models, fine-tuned on \textit{source} and evaluated on images with shared labels in \textit{target} dataset.}
\begin{adjustbox}{width=1.0\linewidth}
\begin{tabular}{l|ccccccccccccc}
\toprule
Source \textbackslash Target dataset & RESISC45 & UC Merced & CLRS & Optimal31 & PatternNet & AID & RSI-CB256 & WHU-RS19 & SIRI-WHU & RSD46-WHU & Eurosat & SAT6 & RSSCN7 \\\midrule
RESISC45 & 96.587 & 86.316 & 71.742 & 99.462 & 88.355 & 82.987 & 79.398 & 88.976 & 62.5 & 50.942 & 11.704 & 4.357 & 59.167\\
UC Merced & 60.714 & 98.571 & 46.0 & 72.778 & 89.779 & 64.734 & 88.2 & 60.377 & 10.0 & 40.239 & 11.875 & 51.12 & 55.625\\
CLRS & 82.453 & 77.0 & 92.533 & 86.979 & 87.865 & 91.948 & 83.601 & 89.881 & 72.857 & 74.079 & 31.704 & 4.285 & 53.0\\
Optimal31 & 86.29 & 84.0 & 64.844 & 92.473 & 84.492 & 82.217 & 82.828 & 81.89 & 66.5 & 44.513 & 9.591 & 20.367 & 40.0\\
PatternNet & 49.925 & 71.765 & 37.847 & 67.188 & 99.688 & 51.006 & 74.017 & 53.968 & 40.833 & 31.364 & 22.375 & 0.987 & 36.25\\
AID & 79.286 & 74.444 & 71.579 & 83.974 & 76.312 & 97.4 & 72.0 & 100.0 & 50.5 & 51.133 & 27.875 & 20.957 & 47.188\\
RSI-CB256 & 55.519 & 83.75 & 44.333 & 61.667 & 78.472 & 67.812 & 99.677 & 83.562 & 30.0 & 46.254 & 42.286 & 13.953 & 27.812\\
WHU-RS19 & 66.667 & 75.0 & 52.5 & 75.694 & 77.083 & 84.577 & 58.506 & 99.502 & 47.857 & 50.727 & 11.182 & 0.0 & 56.25\\
SIRI-WHU & 64.405 & 61.25 & 68.095 & 85.0 & 77.917 & 60.497 & 33.819 & 67.105 & 95.625 & 84.298 & 42.812 & 18.724 & 26.25\\
RSD46-WHU & 45.604 & 45.625 & 53.333 & 52.083 & 45.938 & 51.25 & 44.553 & 59.615 & 43.75 & 93.536 & 94.833 & 6.06 & 63.75\\
Eurosat & 59.286 & 56.667 & 72.333 & 62.5 & 79.375 & 92.857 & 62.908 & 79.07 & 78.333 & 80.542 & 98.944 & 5.963 & 56.667\\
SAT6 & 97.143 & 37.5 & 52.083 & 91.667 & 95.0 & 50.893 & 3.723 & 36.364 & 0.0 & 12.982 & 0.0 & 99.999 & 50.0\\
RSSCN7 & 82.381 & 97.5 & 80.333 & 88.889 & 88.125 & 88.214 & 70.209 & 90.385 & 91.667 & 56.713 & 19.647 & 0.016 & 95.179\\
\bottomrule
\end{tabular}
\end{adjustbox}
\label{table:pair_wise_swint_mcc}
\end{table*}

\begin{table*}[!ht]
\centering
\caption{AlexNet on MLC tasks: Generalization performance in terms of mean average precision (\% mAP) of ImageNet-1K pre-trained models, fine-tuned on \textit{source} and evaluated on images with shared labels in \textit{target} dataset.}
\scriptsize
\begin{adjustbox}{width=0.5\linewidth}
\begin{tabular}{l|cccc}
\toprule
Source \textbackslash Target dataset & MLRSNet & AID (mlc) & UC merced (mlc) & DFC15 \\\midrule
MLRSNet & 93.399 & 74.226 & 74.63 & 39.901 \\
AID (mlc) & 53.989 & 75.908 & 43.061 & 35.914\\
UC merced (mlc) & 45.323 & 46.986 & 92.638 & 37.287\\
DFC15 & 52.787 & 69.946 & 46.843 & 94.058\\
\bottomrule
\end{tabular}
\end{adjustbox}
\label{table:pair_wise_alexnet_mlc}
\end{table*}

\begin{table*}[!ht]
\centering
\caption{VGG16 on MLC tasks: Generalization performance in terms of mean average precision (\% mAP) of ImageNet-1K pre-trained models, fine-tuned on \textit{source} and evaluated on images with shared labels in \textit{target} dataset.}
\begin{adjustbox}{width=0.5\linewidth}
\begin{tabular}{l|cccc}
\toprule
Source \textbackslash Target dataset & MLRSNet & AID (mlc) & UC merced (mlc) & DFC15 \\\midrule
MLRSNet & 94.633 & 76.237 & 76.619 & 45.813\\
AID (mlc) & 57.292 & 79.892 & 53.055 & 41.056\\
UC merced (mlc) & 46.068 & 48.009 & 92.848 & 49.281\\
DFC15 & 51.959 & 71.354 & 58.691 & 96.565\\
\bottomrule
\end{tabular}
\end{adjustbox}
\label{table:pair_wise_vgg16_mlc}
\end{table*}

\begin{table*}[!ht]
\centering
\caption{ResNet50 on MLC tasks: Generalization performance in terms of mean average precision (\% mAP) of ImageNet-1K pre-trained models, fine-tuned on \textit{source} and evaluated on images with shared labels in \textit{target} dataset.}
\begin{adjustbox}{width=0.5\linewidth}
\begin{tabular}{l|cccc}
\toprule
Source \textbackslash Target dataset & MLRSNet & AID (mlc) & UC merced (mlc) & DFC15 \\\midrule
MLRSNet & 96.271 & 77.441 & 77.751 & 50.581\\
AID (mlc) & 59.084 & 80.755 & 50.511 & 36.098\\
UC merced (mlc) & 52.307 & 52.015 & 95.665 & 51.553\\
DFC15 & 48.23 & 67.506 & 55.907 & 97.664\\
\bottomrule
\end{tabular}
\end{adjustbox}
\label{table:pair_wise_resnet50_mlc}
\end{table*}

\begin{table*}[!ht]
\centering
\caption{ResNet152 on MLC tasks: Generalization performance in terms of mean average precision (\% mAP) of ImageNet-1K pre-trained models, fine-tuned on \textit{source} and evaluated on images with shared labels in \textit{target} dataset.}
\begin{adjustbox}{width=0.5\linewidth}
\begin{tabular}{l|cccc}
\toprule
Source \textbackslash Target dataset & MLRSNet & AID (mlc) & UC merced (mlc) & DFC15 \\\midrule
MLRSNet & 96.432 & 76.818 & 81.111 & 48.062\\
AID (mlc) & 62.899 & 80.943 & 53.994 & 48.793\\
UC merced (mlc) & 51.062 & 51.454 & 96.007 & 48.286\\
DFC15 & 53.242 & 73.216 & 56.132 & 97.606\\
\bottomrule
\end{tabular}
\end{adjustbox}
\label{table:pair_wise_resnet152_mlc}
\end{table*}

\begin{table*}[!ht]
\centering
\caption{DenseNet161 on MLC tasks: Generalization performance in terms of mean average precision (\% mAP) of ImageNet-1K pre-trained models, fine-tuned on \textit{source} and evaluated on images with shared labels in \textit{target} dataset.}
\begin{adjustbox}{width=0.5\linewidth}
\begin{tabular}{l|cccc}
\toprule
Source \textbackslash Target dataset & MLRSNet & AID (mlc) & UC merced (mlc) & DFC15 \\\midrule
MLRSNet & 96.306 & 77.359 & 80.212 & 52.007\\
AID (mlc) & 62.522 & 81.709 & 55.913 & 43.452\\
UC merced (mlc) & 55.494 & 54.165 & 96.057 & 52.486\\
DFC15 & 52.947 & 73.871 & 53.681 & 97.532\\
\bottomrule
\end{tabular}
\end{adjustbox}
\label{table:pair_wise_densenet161_mlc}
\end{table*}

\begin{table*}[!ht]
\centering
\caption{EfficientNetB0 on MLC tasks: Generalization performance in terms of mean average precision (\% mAP) of ImageNet-1K pre-trained models, fine-tuned on \textit{source} and evaluated on images with shared labels in \textit{target} dataset.}
\begin{adjustbox}{width=0.5\linewidth}
\begin{tabular}{l|cccc}
\toprule
Source \textbackslash Target dataset & MLRSNet & AID (mlc) & UC merced (mlc) & DFC15 \\\midrule
MLRSNet & 95.391 & 76.929 & 77.914 & 44.973\\
AID (mlc) & 58.154 & 78.003 & 50.246 & 41.025\\
UC merced (mlc) & 49.899 & 51.137 & 95.383 & 49.322\\
DFC15 & 46.145 & 67.706 & 45.535 & 96.784\\
\bottomrule
\end{tabular}
\end{adjustbox}
\label{table:pair_wise_efficientnetb0_mlc}
\end{table*}

\begin{table*}[!ht]
\centering
\caption{ViT on MLC tasks: Generalization performance in terms of mean average precision (\% mAP) of ImageNet-1K pre-trained models, fine-tuned on \textit{source} and evaluated on images with shared labels in \textit{target} dataset.}
\begin{adjustbox}{width=0.5\linewidth}
\begin{tabular}{l|cccc}
\toprule
Source \textbackslash Target dataset & MLRSNet & AID (mlc) & UC merced (mlc) & DFC15 \\\midrule
MLRSNet & 96.408 & 77.141 & 82.344 & 63.666\\
AID (mlc) & 62.51 & 81.541 & 52.498 & 49.896\\
UC merced (mlc) & 57.166 & 52.044 & 96.699 & 63.075\\
DFC15 & 63.734 & 76.698 & 65.742 & 97.617\\
\bottomrule
\end{tabular}
\end{adjustbox}
\label{table:pair_wise_vit_mlc}
\end{table*}

\begin{table*}[!ht]
\centering
\caption{MLPMixer on MLC tasks: Generalization performance in terms of mean average precision (\% mAP) of ImageNet-1K pre-trained models, fine-tuned on \textit{source} and evaluated on images with shared labels in \textit{target} dataset.}
\begin{adjustbox}{width=0.5\linewidth}
\begin{tabular}{l|cccc}
\toprule
Source \textbackslash Target dataset & MLRSNet & AID (mlc) & UC merced (mlc) & DFC15 \\\midrule
MLRSNet & 95.048 & 77.694 & 78.951 & 54.953\\
AID (mlc) & 62.023 & 80.878 & 53.97 & 48.563\\
UC merced (mlc) & 53.876 & 52.737 & 96.34 & 52.036\\
DFC15 & 55.864 & 79.241 & 58.727 & 97.941\\
\bottomrule
\end{tabular}
\end{adjustbox}
\label{table:pair_wise_mlpmixer_mlc}
\end{table*}

\begin{table*}[!ht]
\centering
\caption{ConvNeXt on MLC tasks: Generalization performance in terms of mean average precision (\% mAP) of ImageNet-1K pre-trained models, fine-tuned on \textit{source} and evaluated on images with shared labels in \textit{target} dataset.}
\begin{adjustbox}{width=0.5\linewidth}
\begin{tabular}{l|cccc}
\toprule
Source \textbackslash Target dataset & MLRSNet & AID (mlc) & UC merced (mlc) & DFC15 \\\midrule
MLRSNet & 95.807 & 76.333 & 80.243 & 56.256\\
AID (mlc) & 62.766 & 82.298 & 57.133 & 42.201\\
UC merced (mlc) & 56.192 & 53.227 & 96.43 & 50.419\\
DFC15 & 55.426 & 74.261 & 65.665 & 97.994\\
\bottomrule
\end{tabular}
\end{adjustbox}
\label{table:pair_wise_convnext_mlc}
\end{table*}

\begin{table*}[!ht]
\centering
\caption{SwinT on MLC tasks: Generalization performance in terms of mean average precision (\% mAP) of ImageNet-1K pre-trained models, fine-tuned on \textit{source} and evaluated on images with shared labels in \textit{target} dataset.}
\begin{adjustbox}{width=0.5\linewidth}
\begin{tabular}{l|cccc}
\toprule
Source \textbackslash Target dataset & MLRSNet & AID (mlc) & UC merced (mlc) & DFC15 \\\midrule
MLRSNet & 96.62 & 78.452 & 81.812 & 59.951\\
AID (mlc) & 61.463 & 82.254 & 58.374 & 52.293\\
UC merced (mlc) & 58.263 & 55.748 & 96.831 & 63.277\\
DFC15 & 60.869 & 76.023 & 63.476 & 98.111\\
\bottomrule
\end{tabular}
\end{adjustbox}
\label{table:pair_wise_swint_mlc}
\end{table*}

\clearpage

\section{Detailed data descriptions \& extended results per task}\label{appendix:data}
\setcounter{table}{0}
\setcounter{figure}{0}

\subsection{UC Merced}

\begin{figure}[ht]
  \centering
  \includegraphics[width=0.7\linewidth]{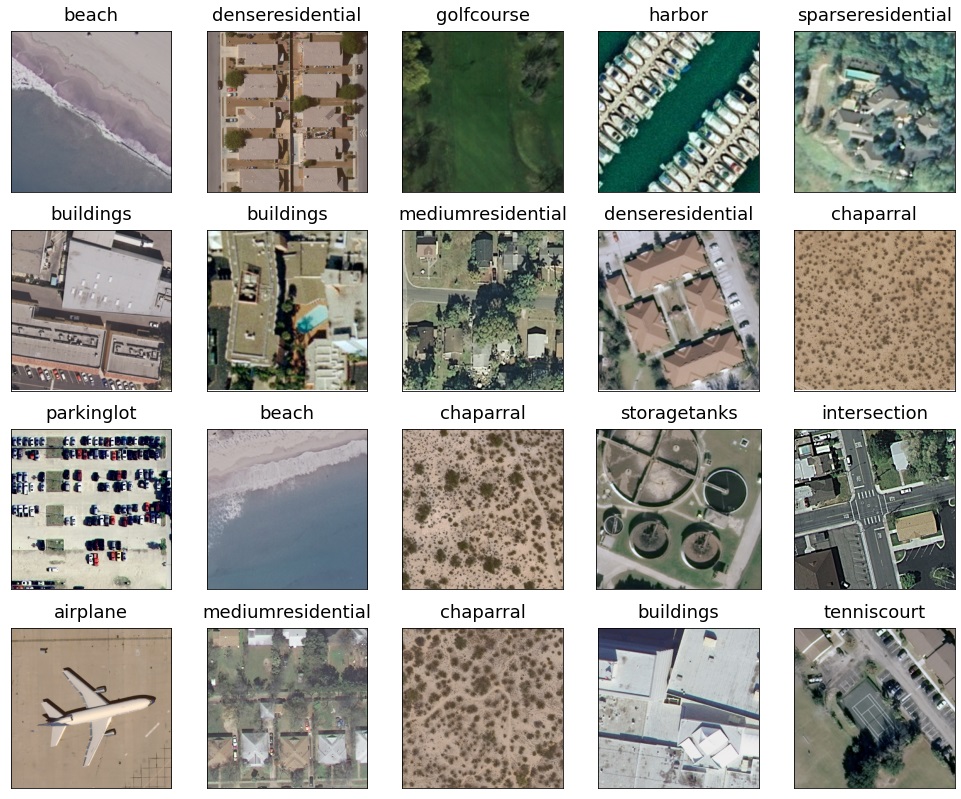}
  \caption{Example images with labels from the UC Merced dataset.}
  \label{fig:ucmerced_samples}
\end{figure}

The UC Merced dataset \citep{yang2010uc_merced} consists of 2100 images divided into 21 land-use scene classes. Each class has 100 RGB aerial image which are 256x256 pixels and have a spatial resolution of 0.3m per pixel. The images were manually extracted from large images from the United States Geological Survey (USGS) National Map of the following US regions: Birmingham, Boston, Buffalo, Columbus, Dallas, Harrisburg, Houston, Jacksonville, Las Vegas, Los Angeles, Miami, Napa, New York, Reno, San Diego, Santa Barbara, Seattle, Tampa, Tucson, and Ventura. Samples from the datasets can be seen on Figure~\ref{fig:ucmerced_samples}.

\begin{figure}[ht]
  \centering
  \includegraphics[width=0.5\linewidth]{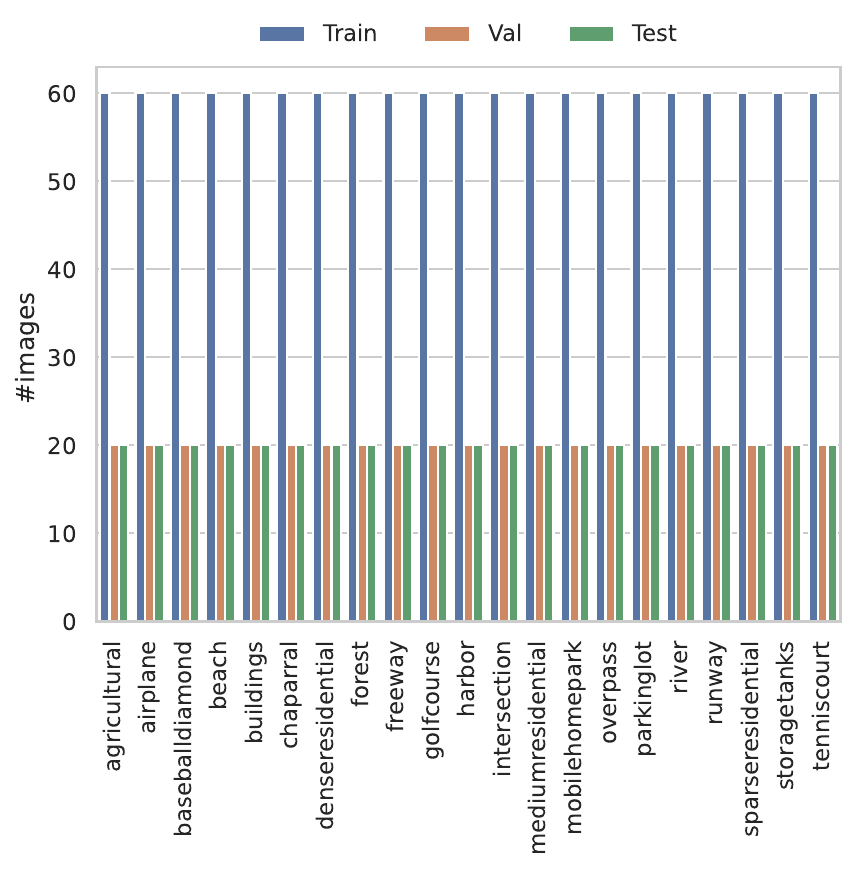}
  \caption{Class distribution for the UC Merced dataset.}
  \label{fig:ucmerced_distribution}
\end{figure}

The 21 classes are: agricultural, airplane, baseball diamond, beach, buildings, chaparral, dense residential, forest, freeway, golf course, harbor, intersection, medium density residential, mobile home park, overpass, parking lot, river, runway, sparse residential, storage tanks, and tennis courts. The authors have not set predefined train-test splits, so we have made such for our study (Figure~\ref{fig:ucmerced_distribution}).

\begin{figure}[ht]
  \centering
  \includegraphics[width=0.7\linewidth]{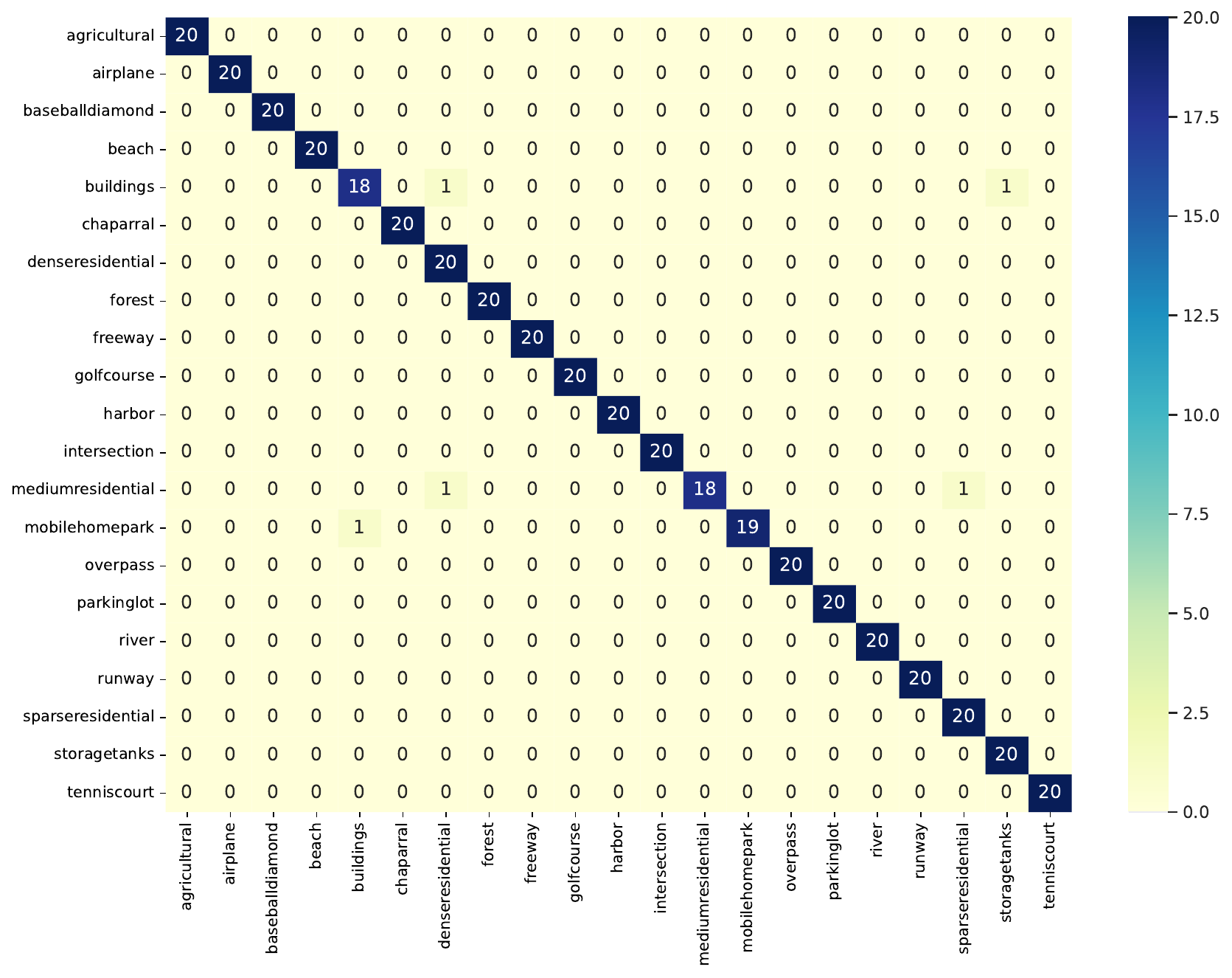}
  \caption{Confusion matrix for the pre-trained ResNet152 model on the UC Merced dataset.}
  \label{fig:ucmerced_confusionmatrix}
\end{figure}

The detailed results for all pre-trained models are shown on Table~\ref{tab:pre-trained_ucmerced} and for all the models learned from scratch are presented on Table~\ref{tab:scratch_ucmerced}. The best performing model is the pre-trained ResNet152. The results on a class level are show on Table~\ref{tab:perclass_ucmerced} along with a confusion matrix on Figure~\ref{fig:ucmerced_confusionmatrix}.

\begin{table}[ht]\centering
\caption{Detailed results for pre-trained models on UCMerced}\label{tab:pre-trained_ucmerced}
\scriptsize \begin{adjustbox}{width=0.75\linewidth}
\begin{tabular}{lrrrrrrrrrrr}\toprule
Model \textbackslash Metric &\rotatebox{90}{Accuracy} &\rotatebox{90}{Macro Precision} &\rotatebox{90}{Weighted Precision} &\rotatebox{90}{Macro Recall} &\rotatebox{90}{Weighted Recall} &\rotatebox{90}{Macro F1 score} &\rotatebox{90}{Weighted F1 score} &\rotatebox{90}{Avg. time / epoch (sec.)} &\rotatebox{90}{Total time (sec.)} &\rotatebox{90}{Best epoch}\\\midrule
AlexNet &92.14 &92.24 &92.24 &92.14 &92.14 &92.03 &92.03 &1.29 &44 &24 \\
VGG16 &95.48 &95.64 &95.64 &95.48 &95.48 &95.48 &95.48 &3.16 &101 &22 \\
ResNet50 &98.57 &98.64 &98.64 &98.57 &98.57 &98.59 &98.59 &2.85 &111 &29 \\
RestNet152 &\textbf{98.81} &98.86 &98.86 &98.81 &98.81 &98.80 &98.80 &5.05 &202 &30 \\
DenseNet161 &98.33 &98.40 &98.40 &98.33 &98.33 &98.34 &98.34 &5.41 &357 &56 \\
EfficientNetB0 &98.57 &98.61 &98.61 &98.57 &98.57 &98.57 &98.57 &2.46 &214 &77 \\
ConvNeXt &97.86 &97.99 &97.99 &97.86 &97.86 &97.87 &97.87 &3.68 &173 &37 \\
Vision Transformer &98.33 &98.44 &98.44 &98.33 &98.33 &98.36 &98.36 &4.00 &112 &18 \\
MLP Mixer &98.33 &98.40 &98.40 &98.33 &98.33 &98.34 &98.34 &3.10 &130 &32 \\
Swin Transformer &98.57 &98.62 &98.62 &98.57 &98.57 &98.58 &98.58 &10.28 &370 &26 \\
\bottomrule
\end{tabular} \end{adjustbox}
\end{table}

\begin{table}[ht]\centering
\caption{Detailed results for models trained from scratch on the UC Merced dataset.}\label{tab:scratch_ucmerced}
\scriptsize \begin{adjustbox}{width=0.75\linewidth}
\begin{tabular}{lrrrrrrrrrrr}\toprule
Model \textbackslash Metric &\rotatebox{90}{Accuracy} &\rotatebox{90}{Macro Precision} &\rotatebox{90}{Weighted Precision} &\rotatebox{90}{Macro Recall} &\rotatebox{90}{Weighted Recall} &\rotatebox{90}{Macro F1 score} &\rotatebox{90}{Weighted F1 score} &\rotatebox{90}{Avg. time / epoch (sec.)} &\rotatebox{90}{Total time (sec.)} &\rotatebox{90}{Best epoch}\\\midrule
AlexNet &81.19 &81.30 &81.30 &81.19 &81.19 &80.87 &80.87 &1.30 &126 &82\\
VGG16 &78.57 &78.96 &78.96 &78.57 &78.57 &78.30 &78.30 &4.66 &466 &85\\
ResNet50 &85.24 &85.20 &85.20 &85.24 &85.24 &84.75 &84.75 &2.54 &178 &55\\
RestNet152 &84.05 &84.02 &84.02 &84.05 &84.05 &83.68 &83.68 &5.02 &467 &78\\
DenseNet161 &\textbf{86.19} &86.42 &86.42 &86.19 &86.19 &85.75 &85.75 &5.46 &415 &61\\
EfficientNetB0 &84.29 &85.27 &85.27 &84.29 &84.29 &84.16 &84.16 &2.53 &253 &93\\
ConvNeXt &84.29 &84.51 &84.51 &84.29 &84.29 &84.14 &84.14 &3.75 &375 &92\\
Vision Transformer &83.10 &83.64 &83.64 &83.10 &83.10 &82.76 &82.76 &4.44 &413 &78\\
MLP Mixer &82.38 &82.12 &82.12 &82.38 &82.38 &82.01 &82.01 &3.06 &269 &73\\
Swin Transformer &81.43 &81.74 &81.74 &81.43 &81.43 &81.14 &81.14 &10.36 &984 &80 \\
\bottomrule
\end{tabular} \end{adjustbox}
\end{table}

\begin{table}[ht]\centering
\caption{Per class results for the pre-trained ResNet152 model on the UC Merced dataset.}\label{tab:perclass_ucmerced}
\scriptsize 
\begin{tabular}{lrrrr}\toprule
Label &Precision &Recall &F1 score \\\midrule
agricultural &100.00 &100.00 &100.00\\
airplane &100.00 &100.00 &100.00\\
baseballdiamond &100.00 &100.00 &100.00\\
beach &100.00 &100.00 &100.00\\
buildings &94.74 &90.00 &92.31\\
chaparral &100.00 &100.00 &100.00\\
denseresidential &90.91 &100.00 &95.24\\
forest &100.00 &100.00 &100.00\\
freeway &100.00 &100.00 &100.00\\
golfcourse &100.00 &100.00 &100.00\\
harbor &100.00 &100.00 &100.00\\
intersection &100.00 &100.00 &100.00\\
mediumresidential &100.00 &90.00 &94.74\\
mobilehomepark &100.00 &95.00 &97.44\\
overpass &100.00 &100.00 &100.00\\
parkinglot &100.00 &100.00 &100.00\\
river &100.00 &100.00 &100.00\\
runway &100.00 &100.00 &100.00\\
sparseresidential &95.24 &100.00 &97.56\\
storagetanks &95.24 &100.00 &97.56\\
tenniscourt &100.00 &100.00 &100.00\\
\bottomrule
\end{tabular} 
\end{table}

\clearpage

\subsection{WHU-RS19}
\begin{figure}[ht]
  \centering
  \includegraphics[width=0.5\linewidth]{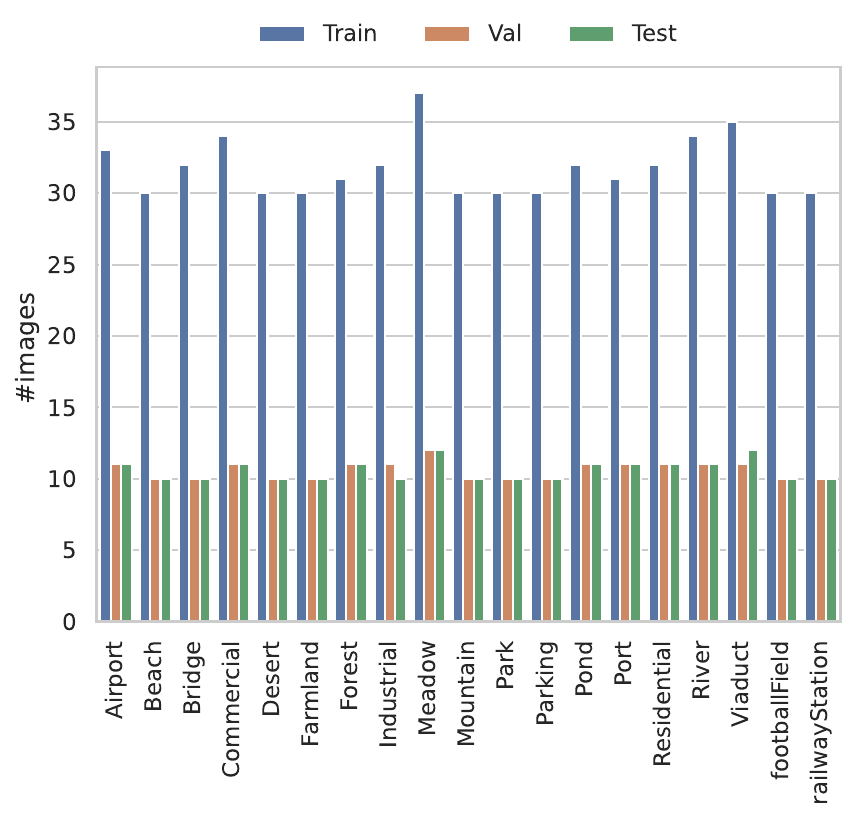}
  \caption{Class distribution for the WHU-RS19 dataset.}
  \label{fig:whurs19_distribution}
\end{figure}

WHU-RS19 is a set of satellite images exported from Google Earth, which provides high-resolution satellite images up to 0.5m and red, green and blue spectral bands \citep{xia2010whurs19}. It contains 19 classes of meaningful scenes in high-resolution satellite imagery, including: airport, beach, bridge, commercial area, desert, farmland, football field, forest, industrial area, meadow, mountain, park, parking lot, pond, port, railway station, residential area, river, and viaduct. For each class, there are about 50 samples with a total of 1005 images in the entire dataset. The data does not come with predefined train and test splits, so per standard we have made splits (Figure~\ref{fig:whurs19_distribution}).

\begin{figure}[ht]
  \centering
  \includegraphics[width=0.7\linewidth]{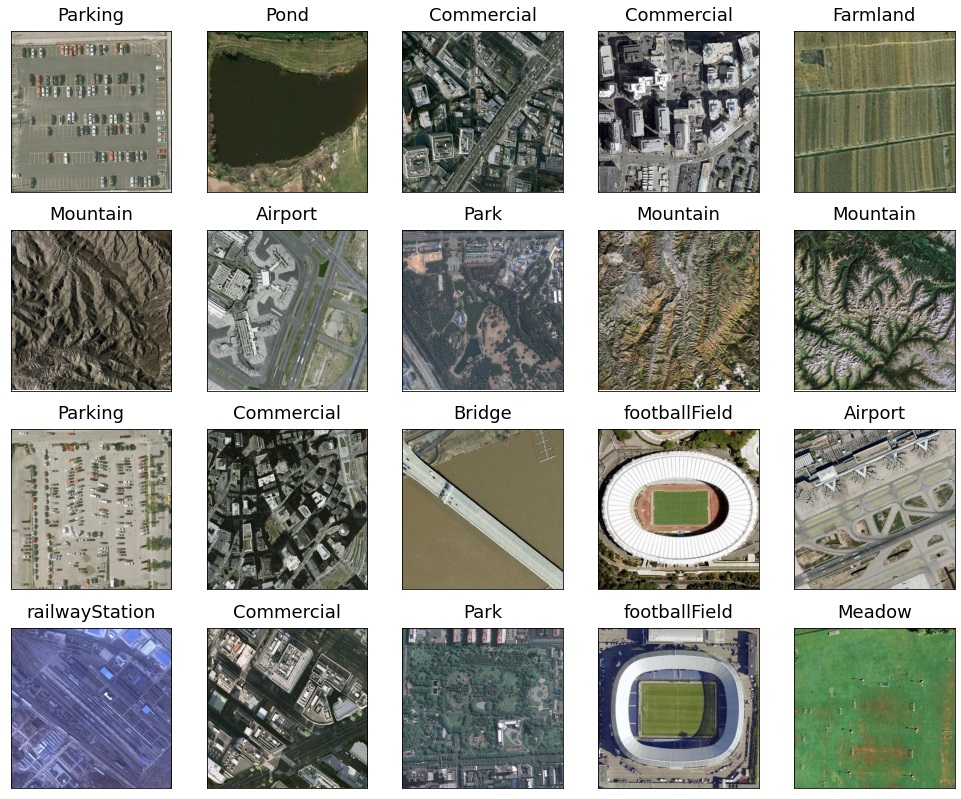}
  \caption{Example images with labels from the WHU-RS19 dataset.}
  \label{fig:whurs19_samples}
\end{figure}

The size of images is 600x600 pixel. The image samples of the same class are collected from different regions in satellite images of different resolutions and then might have different scales, orientations and illuminations. This makes the dataset challenging, however, the number of images is relatively small compared to the other datasets. Sample images from the dataset are shown in Figure~\ref{fig:whurs19_samples}.

\begin{figure}[ht]
  \centering
  \includegraphics[width=0.7\linewidth]{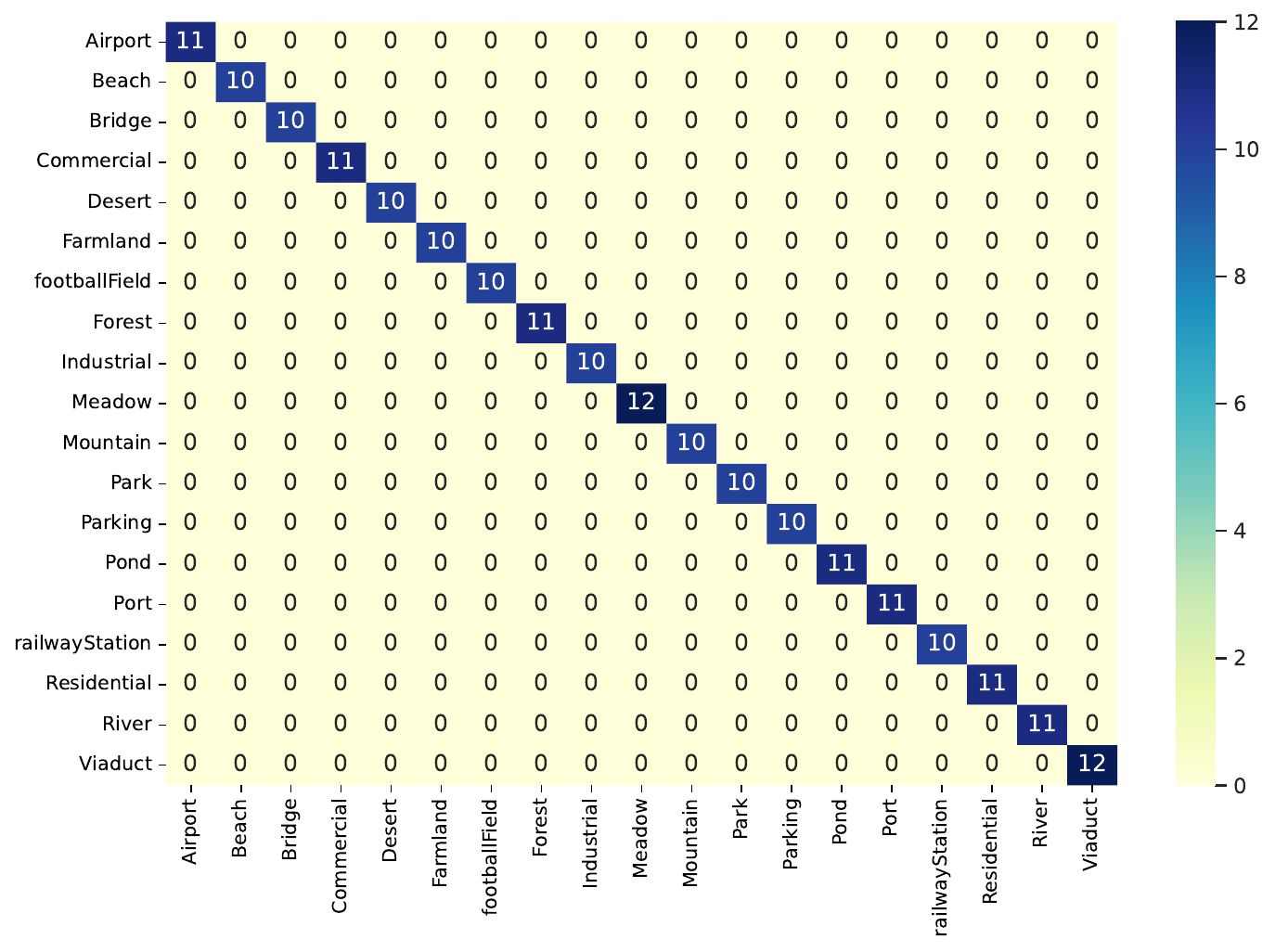}
  \caption{Confusion matrix for the pre-trained DenseNet161 model on the WHU-RS19 dataset.}
  \label{fig:whurs19_confusionmatrix}
\end{figure}

Detailed results for all pre-trained models are shown on Table~\ref{tab:pre-trained_whurs19} and for all the models learned from scratch are presented on Table~\ref{tab:scratch_whurs19}. The best performing model is the pre-trained DenseNet161. The results on a class level are show on Table~\ref{tab:perclass_whurs19} along with a confusion matrix on Figure~\ref{fig:whurs19_confusionmatrix}.

\begin{table}[ht]\centering
\caption{Detailed results for pre-trained models the WHU-RS19 dataset.}\label{tab:pre-trained_whurs19}
\scriptsize \begin{adjustbox}{width=0.75\linewidth}
\begin{tabular}{lrrrrrrrrrrr}\toprule
Model \textbackslash Metric &\rotatebox{90}{Accuracy} &\rotatebox{90}{Macro Precision} &\rotatebox{90}{Weighted Precision} &\rotatebox{90}{Macro Recall} &\rotatebox{90}{Weighted Recall} &\rotatebox{90}{Macro F1 score} &\rotatebox{90}{Weighted F1 score} &\rotatebox{90}{Avg. time / epoch (sec.)} &\rotatebox{90}{Total time (sec.)} &\rotatebox{90}{Best epoch}\\\midrule
AlexNet &93.53 &94.44 &94.30 &93.63 &93.53 &93.73 &93.59 &2.78 &142 &41\\
VGG16 &99.00 &99.08 &99.09 &99.04 &99.00 &99.01 &99.00 &3.00 &144 &38\\
ResNet50 &99.50 &99.56 &99.54 &99.52 &99.50 &99.52 &99.50 &2.85 &285 &96\\
RestNet152 &98.01 &98.21 &98.22 &97.99 &98.01 &98.01 &98.03 &4.02 &253 &53\\
DenseNet161 &\textbf{100.00} &100.00 &100.00 &100.00 &100.00 &100.00 &100.00 &4.04 &400 &89\\
EfficientNetB0 &99.50 &99.56 &99.54 &99.47 &99.50 &99.49 &99.50 &2.76 &276 &100\\
ConvNeXt &99.00 &99.04 &99.05 &99.00 &99.00 &98.99 &99.00 &3.20 &211 &56\\
Vision Transformer &99.50 &99.56 &99.54 &99.52 &99.50 &99.52 &99.50 &3.40 &102 &20\\
MLP Mixer &98.51 &98.64 &98.64 &98.47 &98.51 &98.49 &98.50 &2.84 &247 &77\\
Swin Transformer &99.50 &99.56 &99.54 &99.47 &99.50 &99.49 &99.50 &5.98 &263 &34 \\
\bottomrule
\end{tabular} \end{adjustbox}
\end{table}

\begin{table}[ht]\centering
\caption{Detailed results for models trained from scratch the WHU-RS19 dataset.}\label{tab:scratch_whurs19}
\scriptsize \begin{adjustbox}{width=0.75\linewidth}
\begin{tabular}{lrrrrrrrrrrr}\toprule
Model \textbackslash Metric &\rotatebox{90}{Accuracy} &\rotatebox{90}{Macro Precision} &\rotatebox{90}{Weighted Precision} &\rotatebox{90}{Macro Recall} &\rotatebox{90}{Weighted Recall} &\rotatebox{90}{Macro F1 score} &\rotatebox{90}{Weighted F1 score} &\rotatebox{90}{Avg. time / epoch (sec.)} &\rotatebox{90}{Total time (sec.)} &\rotatebox{90}{Best epoch}\\\midrule
AlexNet &66.17 &67.93 &67.68 &66.28 &66.17 &66.53 &66.36 &2.53 &223 &73\\
VGG16 &68.66 &70.53 &70.25 &68.69 &68.66 &69.02 &68.87 &4.79 &479 &96\\
ResNet50 &79.60 &82.28 &81.91 &79.75 &79.60 &79.88 &79.67 &3.85 &300 &63\\
RestNet152 &80.60 &82.62 &82.27 &80.63 &80.60 &81.08 &80.91 &4.29 &343 &65\\
DenseNet161 &\textbf{80.60} &82.75 &82.44 &80.59 &80.60 &80.75 &80.60 &4.04 &271 &52\\
EfficientNetB0 &75.62 &77.50 &77.00 &76.08 &75.62 &76.02 &75.54 &2.78 &189 &53\\
ConvNeXt &72.14 &73.09 &72.63 &72.41 &72.14 &72.36 &71.99 &3.03 &303 &90\\
Vision Transformer &74.63 &75.96 &75.69 &74.89 &74.63 &75.05 &74.78 &3.44 &303 &73\\
MLP Mixer &69.65 &70.70 &70.51 &69.91 &69.65 &69.10 &68.83 &3.86 &386 &89\\
Swin Transformer &78.61 &78.84 &78.66 &78.84 &78.61 &78.53 &78.33 &6.08 &608 &86 \\
\bottomrule
\end{tabular} \end{adjustbox}
\end{table}

\begin{table}[ht]\centering
\caption{Per class results for the pre-trained DenseNet161 model on the WHU-RS19 dataset.}\label{tab:perclass_whurs19}
\scriptsize 
\begin{tabular}{lrrrr}\toprule
Label &Precision &Recall &F1 score \\\midrule
Airport &100.00 &100.00 &100.00\\
Beach &100.00 &100.00 &100.00\\
Bridge &100.00 &100.00 &100.00\\
Commercial &100.00 &100.00 &100.00\\
Desert &100.00 &100.00 &100.00\\
Farmland &100.00 &100.00 &100.00\\
footballField &100.00 &100.00 &100.00\\
Forest &100.00 &100.00 &100.00\\
Industrial &100.00 &100.00 &100.00\\
Meadow &100.00 &100.00 &100.00\\
Mountain &100.00 &100.00 &100.00\\
Park &100.00 &100.00 &100.00\\
Parking &100.00 &100.00 &100.00\\
Pond &100.00 &100.00 &100.00\\
Port &100.00 &100.00 &100.00\\
railwayStation &100.00 &100.00 &100.00\\
Residential &100.00 &100.00 &100.00\\
River &100.00 &100.00 &100.00\\
Viaduct &100.00 &100.00 &100.00\\
\bottomrule
\end{tabular} 
\end{table}

\clearpage

\subsection{AID}
\label{app:aid}
\begin{figure}[ht]
 \centering
  \includegraphics[width=0.7\linewidth]{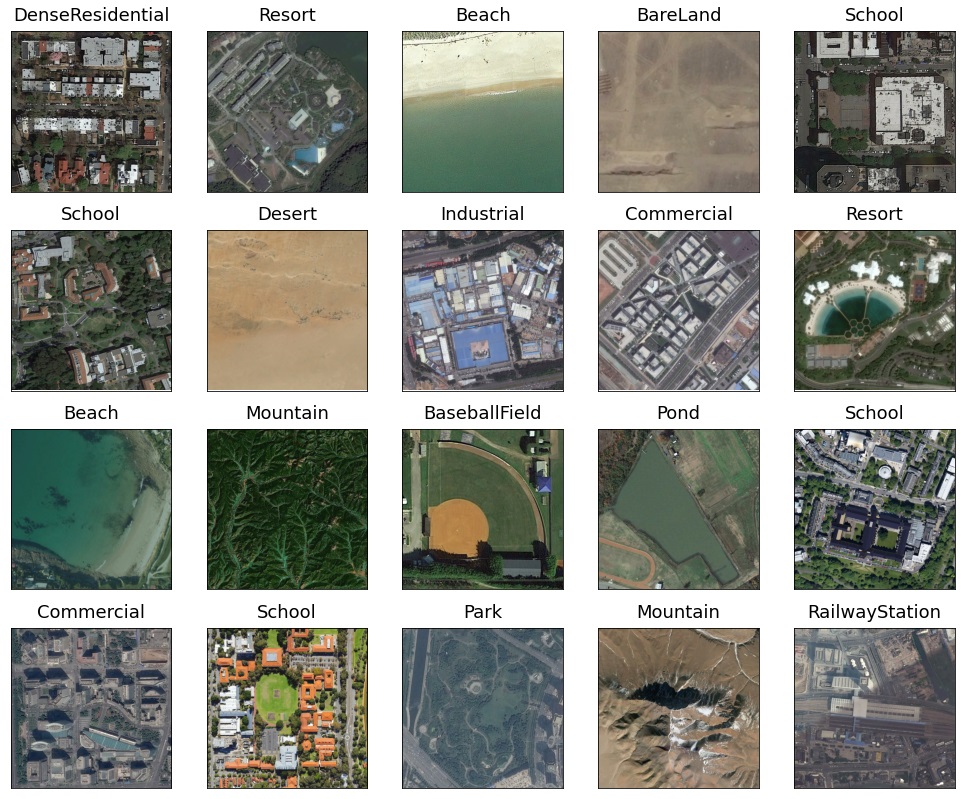}
  \caption{Example images with labels from the AID dataset.}
  \label{fig:aid_samples}
\end{figure}

Aerial Image Dataset (AID) is a large-scale aerial image dataset generated by collecting sample images from Google Earth imagery. The goal of AID is to advance the state-of-the-art in scene classification of remote sensing images. For creating AID, more than ten thousands aerial scene images have been collected and annotated. It consists of 10000 RGB images with 600x600 pixels resolution (Figure~\ref{fig:aid_samples}). The dataset is made up of the following 30 classes (aerial scene types): airport, bare land, baseball field, beach, bridge, center, church, commercial, dense residential, desert, farmland, forest, industrial, meadow, medium residential, mountain, park, parking, playground, pond, port, railway station, resort, river, school, sparse residential, square, stadium, storage tanks and viaduct. 

All the images were labeled by the specialists in the field of remote sensing image interpretation. All samples from each class are chosen from different countries and regions around the world, but mainly in China, USA, England, France, Italy, Japan, Germany etc. They are extracted at different time and seasons under different image conditions. Although, all images have a 600x600 pixels resolution, their spatial resolution varies from 8 to 0.5 meters.

\begin{figure}[ht]
  \centering
  \includegraphics[width=0.5\linewidth]{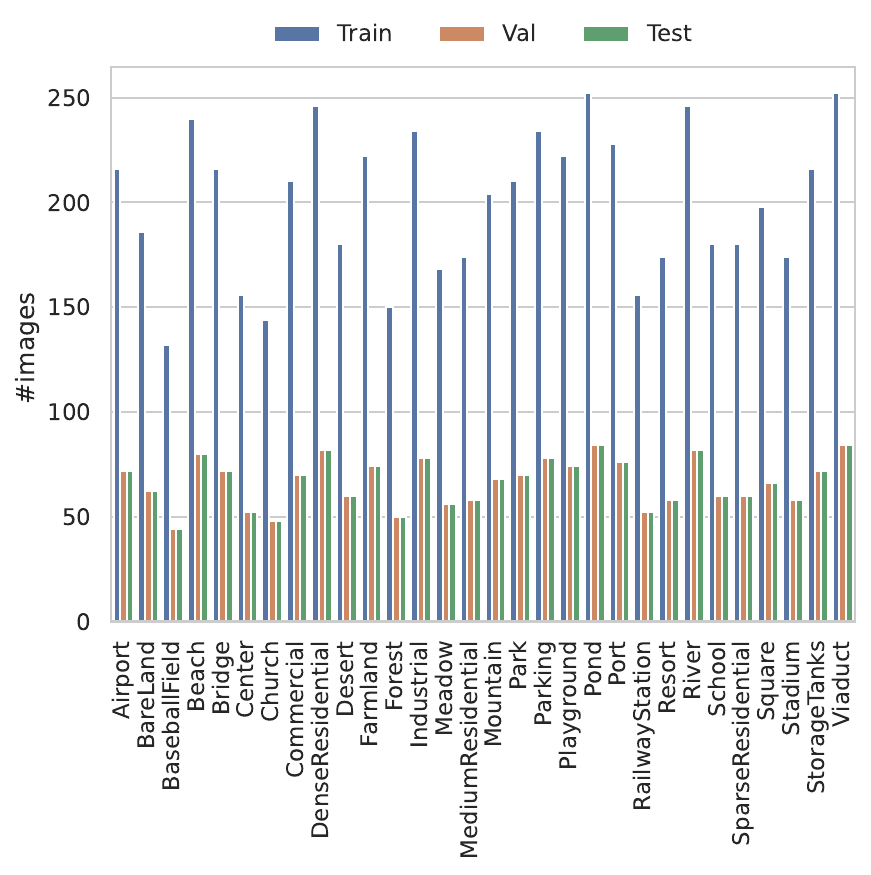}
  \caption{Class distribution for the AID dataset.}
  \label{fig:aid_distribution}
\end{figure}

\begin{figure}[ht]
  \centering
  \includegraphics[width=0.8\linewidth]{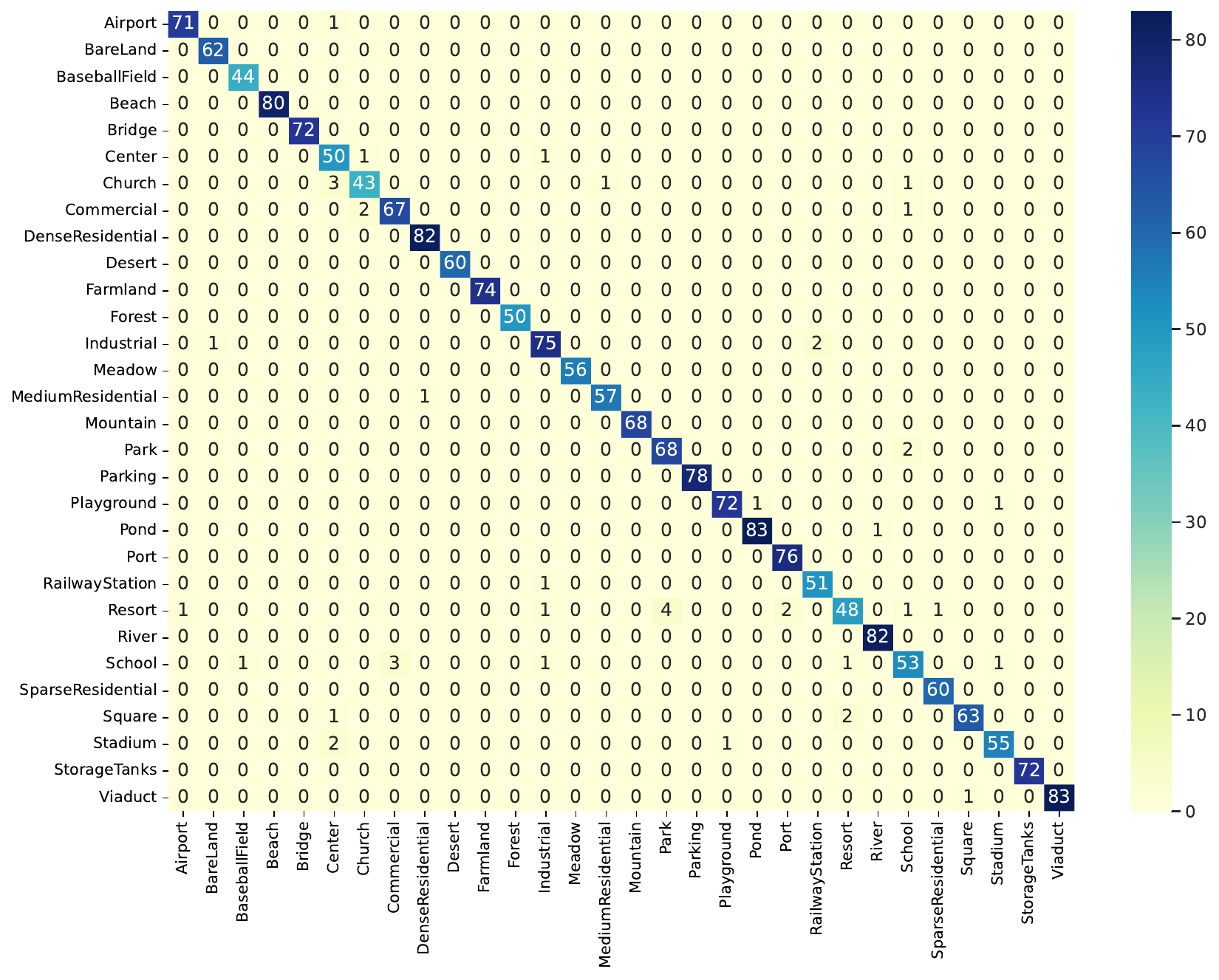}
  \caption{Confusion matrix for the pre-trained Vision Transformer model on the AID dataset.}
  \label{fig:aid_confusionmatrix}
\end{figure}

The dataset has no predefined train-test splits, so for properly conducting the study we have made train, test and validation splits. The distribution of the splits is presented on Figure~\ref{fig:aid_distribution}. Detailed results for all pre-trained models are shown on Table~\ref{tab:pre-trained_aid} and for all the models learned from scratch are presented on Table~\ref{tab:scratch_aid}. The best performing model is the pre-trained ViT model. The results on a class level are show on Table~\ref{tab:perclass_aid} along with a confusion matrix on Figure~\ref{fig:aid_confusionmatrix}.

\begin{table}[ht]\centering
\caption{Detailed results for pre-trained models on the AID dataset.}\label{tab:pre-trained_aid}
\scriptsize \begin{adjustbox}{width=0.75\linewidth}
\begin{tabular}{lrrrrrrrrrrr}\toprule
Model \textbackslash Metric &\rotatebox{90}{Accuracy} &\rotatebox{90}{Macro Precision} &\rotatebox{90}{Weighted Precision} &\rotatebox{90}{Macro Recall} &\rotatebox{90}{Weighted Recall} &\rotatebox{90}{Macro F1 score} &\rotatebox{90}{Weighted F1 score} &\rotatebox{90}{Avg. time / epoch (sec.)} &\rotatebox{90}{Total time (sec.)} &\rotatebox{90}{Best epoch}\\\midrule
AlexNet &92.90 &92.90 &92.94 &92.65 &92.90 &92.72 &92.87 &21.32 &725 &24\\
VGG16 &96.10 &95.95 &96.11 &95.91 &96.10 &95.90 &96.08 &21.35 &854 &30\\
ResNet50 &96.55 &96.48 &96.56 &96.26 &96.55 &96.30 &96.50 &20.29 &1035 &41\\
RestNet152 &97.20 &97.14 &97.24 &97.07 &97.20 &97.08 &97.19 &22.20 &1132 &41\\
DenseNet161 &97.25 &97.25 &97.30 &97.10 &97.25 &97.12 &97.23 &24.36 &1072 &34\\
EfficientNetB0 &96.25 &96.24 &96.26 &96.15 &96.25 &96.16 &96.23 &20.00 &800 &30\\
ConvNeXt &96.95 &96.95 &96.97 &96.81 &96.95 &96.85 &96.93 &23.06 &807 &25\\
Vision Transformer &\textbf{97.75} &97.56 &97.76 &97.53 &97.75 &97.52 &97.73 &20.45 &1145 &46\\
MLP Mixer &96.70 &96.58 &96.74 &96.52 &96.70 &96.51 &96.69 &19.78 &811 &31\\
Swin Transformer &97.40 &97.43 &97.41 &97.26 &97.40 &97.32 &97.38 &46.65 &1213 &16 \\
\bottomrule
\end{tabular} \end{adjustbox}
\end{table}

\begin{table}[ht]\centering
\caption{Detailed results for models trained from scratch on the AID dataset.}\label{tab:scratch_aid}
\scriptsize \begin{adjustbox}{width=0.75\linewidth}
\begin{tabular}{lrrrrrrrrrrr}\toprule
Model \textbackslash Metric &\rotatebox{90}{Accuracy} &\rotatebox{90}{Macro Precision} &\rotatebox{90}{Weighted Precision} &\rotatebox{90}{Macro Recall} &\rotatebox{90}{Weighted Recall} &\rotatebox{90}{Macro F1 score} &\rotatebox{90}{Weighted F1 score} &\rotatebox{90}{Avg. time / epoch (sec.)} &\rotatebox{90}{Total time (sec.)} &\rotatebox{90}{Best epoch}\\\midrule
AlexNet &81.35 &81.23 &81.32 &81.14 &81.35 &81.07 &81.23 &19.46 &1927 &84\\
VGG16 &81.95 &81.80 &82.04 &81.52 &81.95 &81.50 &81.84 &19.65 &1356 &54\\
ResNet50 &89.05 &89.09 &89.23 &88.82 &89.05 &88.85 &89.04 &19.66 &1514 &62\\
RestNet152 &89.90 &90.08 &90.09 &89.60 &89.90 &89.73 &89.88 &22.25 &1513 &53\\
DenseNet161 &\textbf{93.30} &93.32 &93.42 &93.13 &93.30 &93.17 &93.30 &24.48 &2228 &76\\
EfficientNetB0 &90.05 &90.19 &90.32 &89.88 &90.05 &89.92 &90.08 &19.33 &1121 &43\\
ConvNeXt &81.10 &81.51 &81.18 &80.87 &81.10 &81.03 &80.98 &19.15 &1915 &96\\
Vision Transformer &79.35 &79.27 &79.27 &79.51 &79.35 &79.30 &79.21 &19.63 &1060 &39\\
MLP Mixer &71.75 &72.02 &71.87 &72.01 &71.75 &71.73 &71.52 &19.06 &953 &35\\
Swin Transformer &87.70 &87.96 &87.85 &87.62 &87.70 &87.66 &87.66 &46.94 &4647 &84 \\
\bottomrule
\end{tabular} \end{adjustbox}
\end{table}

\begin{table}[ht]\centering
\caption{Per class results for the pre-trained Vision Transformer on the AID dataset.}\label{tab:perclass_aid}
\scriptsize 
\begin{tabular}{lrrrr}\toprule
Label &Precision &Recall &F1 score \\\midrule
Airport &98.61 &98.61 &98.61\\
BareLand &98.41 &100.00 &99.20\\
BaseballField &97.78 &100.00 &98.88\\
Beach &100.00 &100.00 &100.00\\
Bridge &100.00 &100.00 &100.00\\
Center &87.72 &96.15 &91.74\\
Church &93.48 &89.58 &91.49\\
Commercial &95.71 &95.71 &95.71\\
DenseResidential &98.80 &100.00 &99.39\\
Desert &100.00 &100.00 &100.00\\
Farmland &100.00 &100.00 &100.00\\
Forest &100.00 &100.00 &100.00\\
Industrial &94.94 &96.15 &95.54\\
Meadow &100.00 &100.00 &100.00\\
MediumResidential &98.28 &98.28 &98.28\\
Mountain &100.00 &100.00 &100.00\\
Park &94.44 &97.14 &95.77\\
Parking &100.00 &100.00 &100.00\\
Playground &98.63 &97.30 &97.96\\
Pond &98.81 &98.81 &98.81\\
Port &97.44 &100.00 &98.70\\
RailwayStation &96.23 &98.08 &97.14\\
Resort &94.12 &82.76 &88.07\\
River &98.80 &100.00 &99.39\\
School &91.38 &88.33 &89.83\\
SparseResidential &98.36 &100.00 &99.17\\
Square &98.44 &95.45 &96.92\\
Stadium &96.49 &94.83 &95.65\\
StorageTanks &100.00 &100.00 &100.00\\
Viaduct &100.00 &98.81 &99.40\\
\bottomrule
\end{tabular} 
\end{table}

\clearpage

\subsection{Eurosat}
\begin{figure}[ht]
 \centering
  \includegraphics[width=0.7\linewidth]{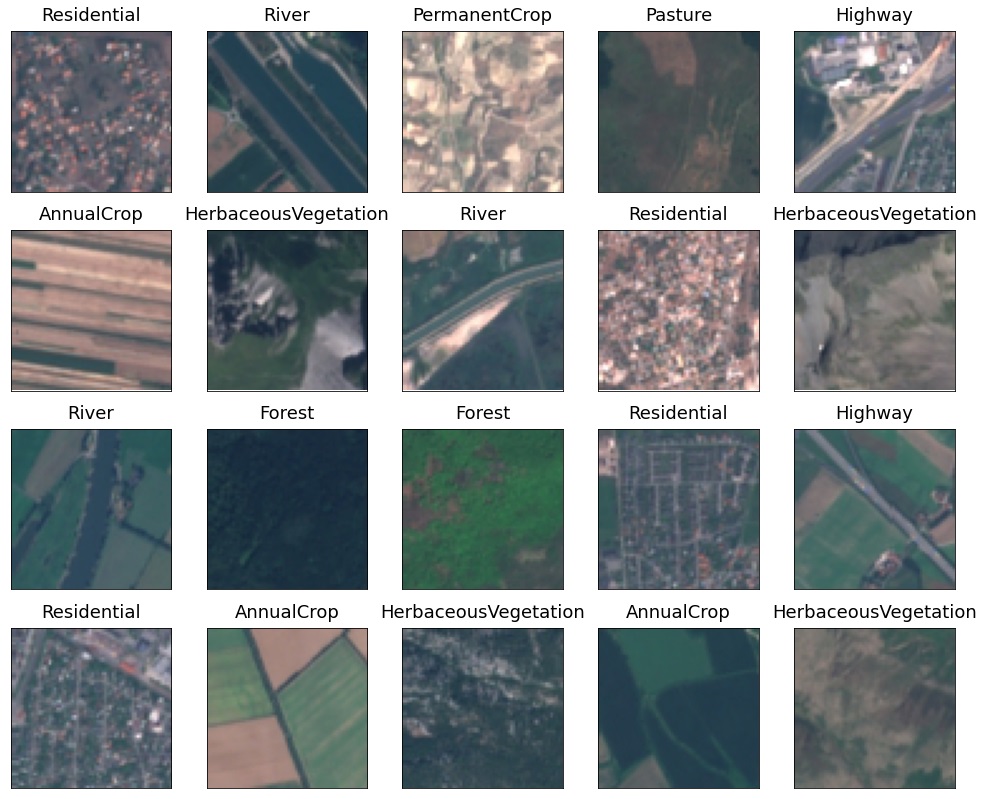}
  \caption{Example images with labels from the Eurosat dataset.}
  \label{fig:eurosat_samples}
\end{figure}

\begin{figure}[ht]
  \centering
  \includegraphics[width=0.5\linewidth]{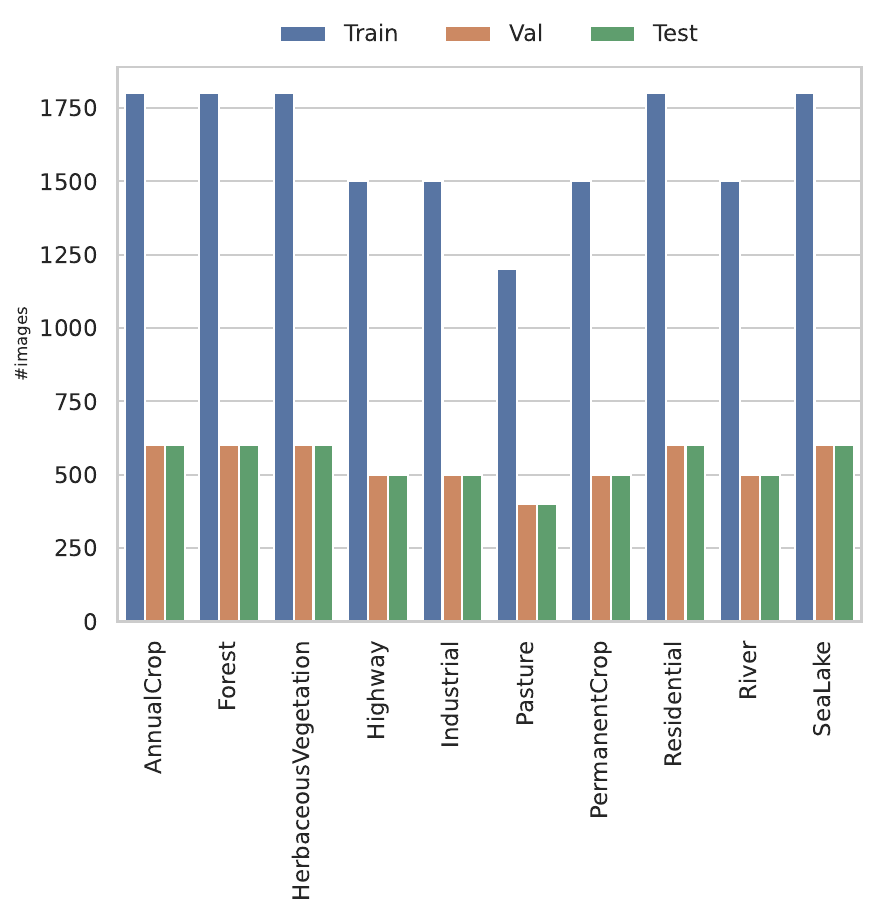}
  \caption{Class distribution for the Eurosat dataset.}
  \label{fig:eurosat_distribution}
\end{figure}

EuroSAT \citep{helber2019eurosat} is a land use and land cover classification dataset based on Sentinel-2 satellite images covering 13 spectral bands and consisting out of 10 classes with in total 27000 labeled and geo-referenced images. The dataset provides RGB and multi-spectral (MS) version of the data. The spectral bands and their respective spatial resolutions are presented on Table~\ref{table:eurosat_overview}. The 10 image classes are the following: Annual Crop, Forest, Herbaceous Vegetation, Highway, Industrial, Pasture, Permanent Crop, Residential, River, Sea/Lake. Some samples from the dataset are presented on Figure~\ref{fig:eurosat_samples}.The class distrubtion of our train, test and validation splits are provided on Figure~\ref{fig:eurosat_distribution}.

\begin{figure}[ht]
  \centering
  \includegraphics[width=0.7\linewidth]{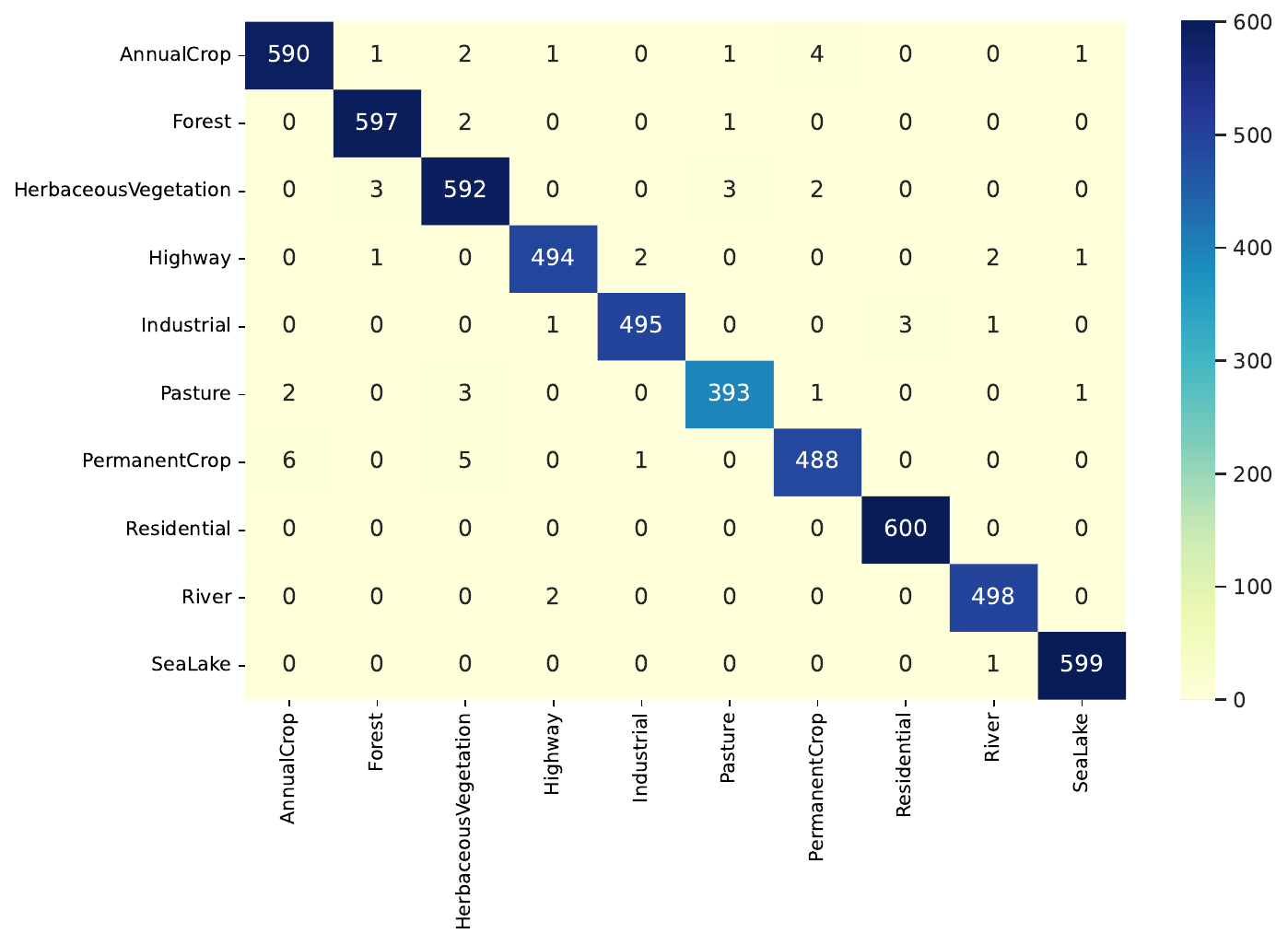}
  \caption{Confusion matrix for the pre-trained ResNet152 model on the Eurosat dataset.}
  \label{fig:eurosat_confusionmatrix}
\end{figure}

Detailed results for all pre-trained models are shown on Table~\ref{tab:pre-trained_eurosat} and for all the models learned from scratch are presented on Table~\ref{tab:scratch_eurosat}. The best performing model is the pre-trained ResNet152 model. The results on a class level are show on Table~\ref{tab:perclass_eurosat} along with a confusion matrix on Figure~\ref{fig:eurosat_confusionmatrix}.

\begin{table}[ht]
\centering
\caption{Eurosat bands and spatial resolutions.}
\begin{tabular}{ccc}
\hline
\textbf{Band} & \textbf{\rotatebox[origin=c]{0}{Spatial resolution \textit{m}}} \\
\hline
B01 - Aerosols & 60  \\
B02 - Blue & 10  \\
B03 - Green & 10  \\
B04 - Red & 10  \\
B05 - Red edge 1 & 20 \\
B06 - Red edge 2 & 20 \\
B07 - Red edge 3 & 20 \\
B08 - NIR & 10  \\
B08A - Red edge 4 & 20 \\
B09 - Water vapor & 60 \\
B10 - Cirrus & 60 \\
B11 - SWIR 1 & 20 \\
B12 - SWIR 2 & 20  \\
\hline
\end{tabular} 
\label{table:eurosat_overview}
\end{table}

\begin{table}[!htp]\centering
\caption{Detailed results for pre-trained models on the Eurosat dataset.}\label{tab:pre-trained_eurosat}
\scriptsize 
\begin{tabular}{lrrrrrrrrrrr}\toprule
Model \textbackslash Metric &\rotatebox{90}{Accuracy} &\rotatebox{90}{Macro Precision} &\rotatebox{90}{Weighted Precision} &\rotatebox{90}{Macro Recall} &\rotatebox{90}{Weighted Recall} &\rotatebox{90}{Macro F1 score} &\rotatebox{90}{Weighted F1 score} &\rotatebox{90}{Avg. time / epoch (sec.)} &\rotatebox{90}{Total time (sec.)} &\rotatebox{90}{Best epoch}\\\midrule
AlexNet &97.57 &97.48 &97.58 &97.48 &97.57 &97.48 &97.57 &8.88 &426 &38\\
VGG16 &98.15 &98.14 &98.15 &98.06 &98.15 &98.09 &98.15 &33.69 &977 &19\\
ResNet50 &98.83 &98.82 &98.83 &98.77 &98.83 &98.79 &98.83 &26.56 &1912 &62\\
RestNet152 &\textbf{99.00} &99.00 &99.00 &98.96 &99.00 &98.98 &99.00 &56.00 &1904 &24\\
DenseNet161 &98.89 &98.88 &98.89 &98.82 &98.89 &98.85 &98.89 &61.12 &2078 &24\\
EfficientNetB0 &98.91 &98.91 &98.91 &98.86 &98.91 &98.88 &98.91 &23.47 &1056 &35\\
ConvNeXt &98.78 &98.76 &98.78 &98.75 &98.78 &98.75 &98.78 &40.38 &1050 &16\\
Vision Transformer &98.72 &98.71 &98.73 &98.64 &98.72 &98.68 &98.72 &43.19 &1123 &16\\
MLP Mixer &98.74 &98.73 &98.74 &98.65 &98.74 &98.68 &98.74 &30.41 &669 &12\\
Swin Transformer &98.94 &98.94 &98.95 &98.89 &98.94 &98.91 &98.94 &124.17 &2980 &14 \\
\bottomrule
\end{tabular} 
\end{table}

\begin{table}[!htp]\centering
\caption{Detailed results for models trained from scratch on the Eurosat dataset.}\label{tab:scratch_eurosat}
\scriptsize 
\begin{tabular}{lrrrrrrrrrrr}\toprule
Model \textbackslash Metric &\rotatebox{90}{Accuracy} &\rotatebox{90}{Macro Precision} &\rotatebox{90}{Weighted Precision} &\rotatebox{90}{Macro Recall} &\rotatebox{90}{Weighted Recall} &\rotatebox{90}{Macro F1 score} &\rotatebox{90}{Weighted F1 score} &\rotatebox{90}{Avg. time / epoch (sec.)} &\rotatebox{90}{Total time (sec.)} &\rotatebox{90}{Best epoch}\\\midrule
AlexNet &96.17 &96.02 &96.18 &96.10 &96.17 &96.06 &96.17 &8.02 &802 &95\\
VGG16 &97.19 &97.17 &97.19 &97.04 &97.19 &97.10 &97.18 &33.62 &2622 &63\\
ResNet50 &97.00 &96.93 &97.01 &96.85 &97.00 &96.88 &97.00 &26.45 &2619 &84\\
RestNet152 &97.41 &97.36 &97.41 &97.27 &97.41 &97.31 &97.40 &56.21 &4328 &62\\
DenseNet161 &97.63 &97.57 &97.64 &97.51 &97.63 &97.54 &97.63 &62.50 &5125 &67\\
EfficientNetB0 &\textbf{97.80} &97.76 &97.80 &97.72 &97.80 &97.74 &97.79 &24.19 &2032 &69\\
ConvNeXt &95.43 &95.25 &95.44 &95.29 &95.43 &95.27 &95.43 &40.03 &2642 &51\\
Vision Transformer &95.04 &94.86 &95.02 &94.80 &95.04 &94.82 &95.02 &44.22 &2963 &52\\
MLP Mixer &95.50 &95.29 &95.50 &95.35 &95.50 &95.31 &95.49 &31.45 &2327 &59\\
Swin Transformer &95.72 &95.78 &95.78 &95.45 &95.72 &95.58 &95.72 &122.44 &12244 &90 \\
\bottomrule
\end{tabular} 
\end{table}

\begin{table}[ht]\centering
\caption{Per class results for the pre-trained ResNet152 model on the Eurosat dataset.}\label{tab:perclass_eurosat}
\scriptsize 
\begin{tabular}{lrrrr}\toprule
Label &Precision &Recall &F1 score \\\midrule
Annual Crop &98.66 &98.33 &98.50 \\
Forest &99.17 &99.50 &99.33 \\
Herbaceous Vegetation &98.01 &98.67 &98.34 \\
Highway &99.20 &98.80 &99.00 \\
Industrial &99.40 &99.00 &99.20 \\
Pasture &98.74 &98.25 &98.50 \\
Permanent Crop &98.59 &97.60 &98.09 \\
Residential &99.50 &100.00 &99.75 \\
River &99.20 &99.60 &99.40 \\
Sea Lake &99.50 &99.83 &99.67 \\
\bottomrule
\end{tabular} 
\end{table}

\clearpage

\subsection{PatternNet}
\begin{figure}[ht]
  \centering
  \includegraphics[width=0.8\linewidth]{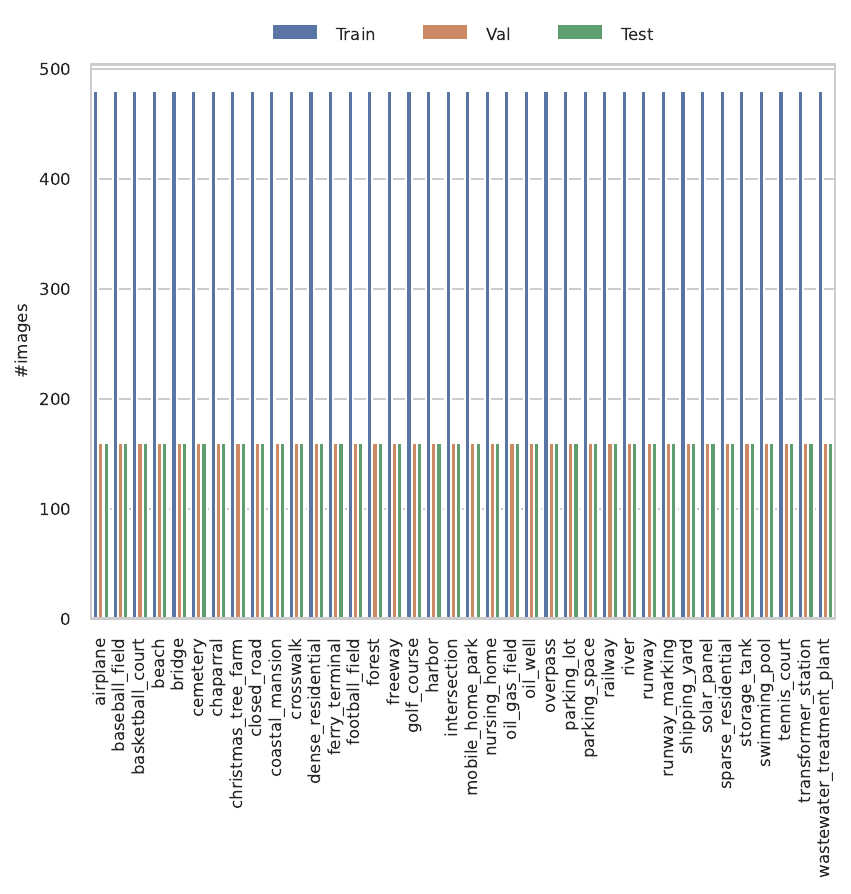}
  \caption{Class distribution for the PatternNet dataset.}
  \label{fig:patternnet_distribution}
\end{figure}

PatternNet is a large-scale remote sensing dataset that was collected specifically for Remote sensing image retrieval. It contains 38 classes: airplane, baseball field, basketball court, beach, bridge, cemetery, chaparral, christmas tree farm, closed road, coastal mansion, crosswalk, dense residential, ferry terminal, football field, forest, freeway, golf course, harbor, intersection, mobile home park, nursing home, oil gas field, oil well, overpass, parking lot, parking space, railway, river, runway, runway marking, shipping yard, solar panel, sparse residential, storage tank, swimming pool, tennis court, transformer station and wastewater treatment plant. There are a total of 38 classes with 800 images of size 256×256 pixels for each class. The class distribution of the train, test and validation splits we generated is presented on Figure~\ref{fig:patternnet_distribution}, since the dataset does not have predefined ones.

\begin{figure}[ht]
 \centering
  \includegraphics[width=0.7\linewidth]{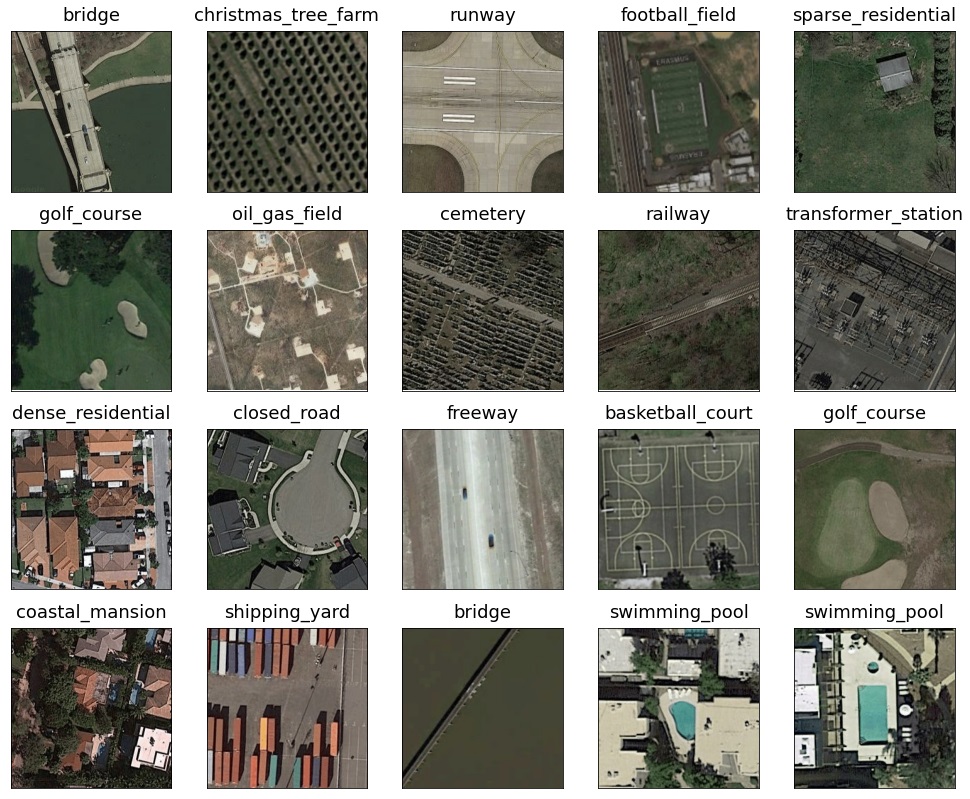}
  \caption{Example images with labels from the PatternNet dataset.}
  \label{fig:patternnet_samples}
\end{figure}

PatternNet dataset has the following main characteristics: It's the largest publicly available dataset specifically designed for remote sensing image retrieval. It has a higher spatial resolution, so that the classes of interest constitute a larger portion of the image. It has high inter-class similarity and high intra-class diversity. Some sample images are shown on Figure~\ref{fig:patternnet_samples}. 

\begin{figure}[ht]
  \centering
  \includegraphics[width=0.9\linewidth]{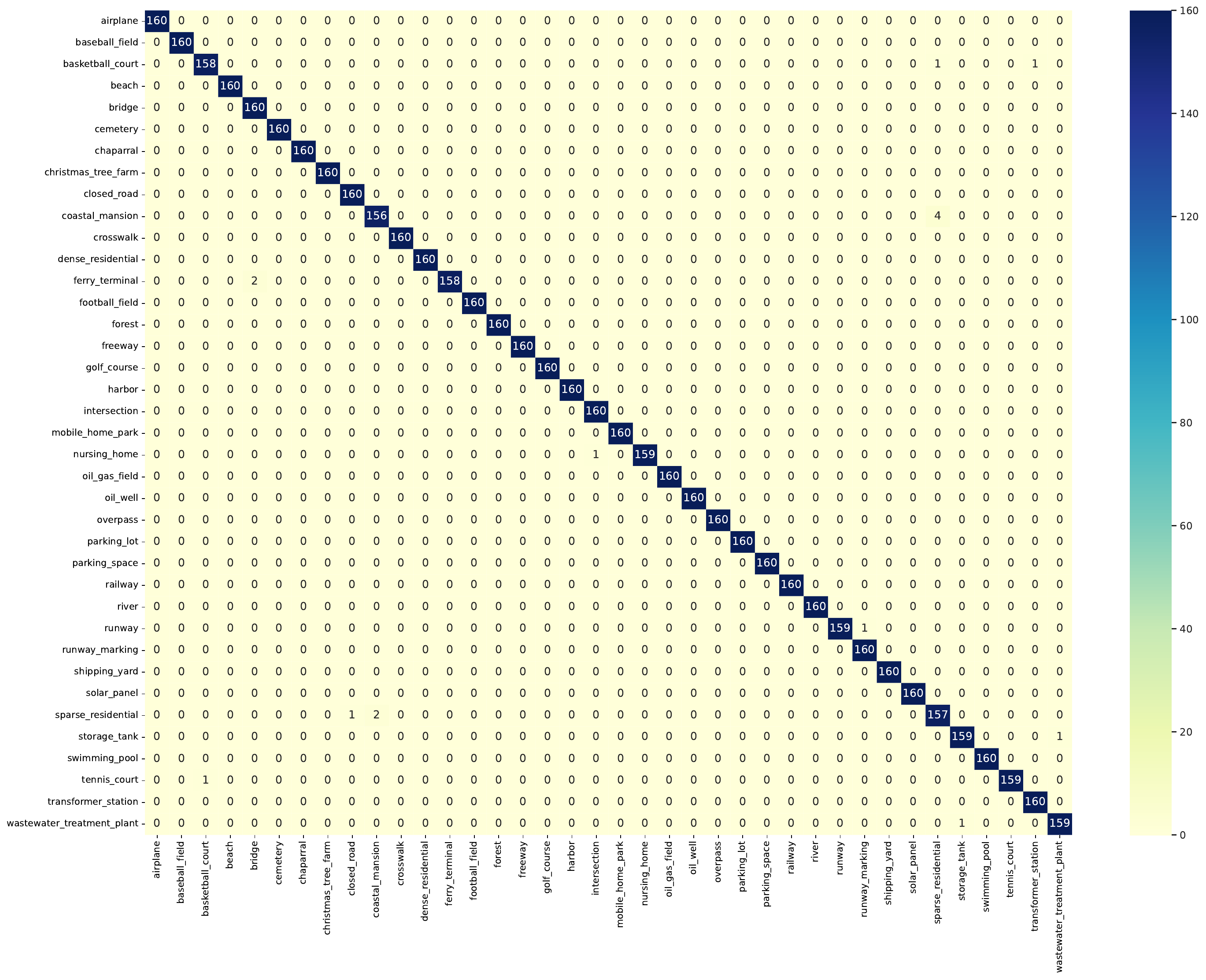}
  \caption{Confusion matrix for the pre-trained DenseNet161 model on the PatternNet dataset.}
  \label{fig:patternnet_confusionmatrix}
\end{figure}

Detailed results for all pre-trained models are shown on Table~\ref{tab:pre-trained_patternnet} and for all the models learned from scratch are presented on Table~\ref{tab:scratch_patternnet}. The best performing models are the pre-trained DenseNet161 and ResNet50 models. The results on a class level are show on Table~\ref{tab:perclass_patternnet} along with a confusion matrix on Figure~\ref{fig:patternnet_confusionmatrix}.

\begin{table}[ht]\centering
\caption{Detailed results for pre-trained models on the PatternNet dataset.}\label{tab:pre-trained_patternnet}
\scriptsize \begin{adjustbox}{width=0.75\linewidth}
\begin{tabular}{lrrrrrrrrrrr}\toprule
Model \textbackslash Metric &\rotatebox{90}{Accuracy} &\rotatebox{90}{Macro Precision} &\rotatebox{90}{Weighted Precision} &\rotatebox{90}{Macro Recall} &\rotatebox{90}{Weighted Recall} &\rotatebox{90}{Macro F1 score} &\rotatebox{90}{Weighted F1 score} &\rotatebox{90}{Avg. time / epoch (sec.)} &\rotatebox{90}{Total time (sec.)} &\rotatebox{90}{Best epoch}\\\midrule
AlexNet &99.16 &99.17 &99.17 &99.16 &99.16 &99.16 &99.16 &15.17 &637 &32\\
VGG16 &99.42 &99.43 &99.43 &99.42 &99.42 &99.42 &99.42 &37.74 &1321 &25\\
ResNet50 &\textbf{99.74} &99.74 &99.74 &99.74 &99.74 &99.74 &99.74 &29.10 &1193 &31\\
RestNet152 &99.49 &99.49 &99.49 &99.49 &99.49 &99.49 &99.49 &62.94 &1070 &7\\
DenseNet161 &\textbf{99.74} &99.74 &99.74 &99.74 &99.74 &99.74 &99.74 &68.87 &3168 &36\\
EfficientNetB0 &99.54 &99.54 &99.54 &99.54 &99.54 &99.54 &99.54 &25.86 &569 &12\\
ConvNeXt &99.67 &99.67 &99.67 &99.67 &99.67 &99.67 &99.67 &45.93 &1378 &20\\
Vision Transformer &99.65 &99.66 &99.66 &99.65 &99.65 &99.65 &99.65 &48.50 &1067 &12\\
MLP Mixer &99.70 &99.71 &99.71 &99.70 &99.70 &99.70 &99.70 &33.80 &1521 &35\\
Swin Transformer &99.69 &99.69 &99.69 &99.69 &99.69 &99.69 &99.69 &138.65 &2357 &7 \\
\bottomrule
\end{tabular} \end{adjustbox}
\end{table}

\begin{table}[ht]\centering
\caption{Detailed results for models trained from scratch on the PatternNet dataset.}\label{tab:scratch_patternnet}
\scriptsize 
\begin{tabular}{lrrrrrrrrrrr}\toprule
Model \textbackslash Metric &\rotatebox{90}{Accuracy} &\rotatebox{90}{Macro Precision} &\rotatebox{90}{Weighted Precision} &\rotatebox{90}{Macro Recall} &\rotatebox{90}{Weighted Recall} &\rotatebox{90}{Macro F1 score} &\rotatebox{90}{Weighted F1 score} &\rotatebox{90}{Avg. time / epoch (sec.)} &\rotatebox{90}{Total time (sec.)} &\rotatebox{90}{Best epoch}\\\midrule
AlexNet &97.83 &97.83 &97.83 &97.83 &97.83 &97.82 &97.82 &13.75 &1141 &68\\
VGG16 &97.91 &97.93 &97.93 &97.91 &97.91 &97.91 &97.91 &37.47 &2061 &40\\
ResNet50 &99.06 &99.07 &99.07 &99.06 &99.06 &99.06 &99.06 &35.65 &3030 &70\\
RestNet152 &98.88 &98.89 &98.89 &98.88 &98.88 &98.88 &98.88 &69.05 &6905 &88\\
DenseNet161 &\textbf{99.24} &99.25 &99.25 &99.24 &99.24 &99.24 &99.24 &71.08 &5260 &59\\
EfficientNetB0 &98.83 &98.84 &98.84 &98.83 &98.83 &98.83 &98.83 &27.54 &2286 &68\\
ConvNeXt &97.83 &97.83 &97.83 &97.83 &97.83 &97.82 &97.82 &45.06 &4326 &81\\
Vision Transformer &96.69 &96.69 &96.69 &96.69 &96.69 &96.68 &96.68 &49.05 &3237 &51\\
MLP Mixer &98.83 &98.84 &98.84 &98.83 &98.83 &98.83 &98.83 &34.54 &2038 &44\\
Swin Transformer &98.52 &98.53 &98.53 &98.52 &98.52 &98.52 &98.52 &138.59 &12612 &76 \\
\bottomrule
\end{tabular} 
\end{table}

\begin{table}[ht]\centering
\caption{Per class results for the pre-trained DenseNet161 model on the PatternNet dataset.}\label{tab:perclass_patternnet}
\scriptsize 
\begin{tabular}{lrrrr}\toprule
Label &Precision &Recall &F1 score \\\midrule
airplane &100.00 &100.00 &100.00\\
baseball field &100.00 &100.00 &100.00\\
basketball court &99.37 &98.75 &99.06\\
beach &100.00 &100.00 &100.00\\
bridge &98.77 &100.00 &99.38\\
cemetery &100.00 &100.00 &100.00\\
chaparral &100.00 &100.00 &100.00\\
christmas tree farm &100.00 &100.00 &100.00\\
closed\_road &99.38 &100.00 &99.69\\
coastal\_mansion &98.73 &97.50 &98.11\\
crosswalk &100.00 &100.00 &100.00\\
dense\_residential &100.00 &100.00 &100.00\\
ferry terminal &100.00 &98.75 &99.37\\
football field &100.00 &100.00 &100.00\\
forest &100.00 &100.00 &100.00\\
freeway &100.00 &100.00 &100.00\\
golf course &100.00 &100.00 &100.00\\
harbor &100.00 &100.00 &100.00\\
intersection &99.38 &100.00 &99.69\\
mobile home park &100.00 &100.00 &100.00\\
nursing home &100.00 &99.38 &99.69\\
oil gas field &100.00 &100.00 &100.00\\
oil well &100.00 &100.00 &100.00\\
overpass &100.00 &100.00 &100.00\\
parking lot &100.00 &100.00 &100.00\\
parking space &100.00 &100.00 &100.00\\
railway &100.00 &100.00 &100.00\\
river &100.00 &100.00 &100.00\\
runway &100.00 &99.38 &99.69\\
runway marking &99.38 &100.00 &99.69\\
shipping yard &100.00 &100.00 &100.00\\
solar panel &100.00 &100.00 &100.00\\
sparse residential &96.91 &98.13 &97.52\\
storage tank &99.38 &99.38 &99.38\\
swimming pool &100.00 &100.00 &100.00\\
tennis court &100.00 &99.38 &99.69\\
transformer station &99.38 &100.00 &99.69\\
wastewater treatment plant &99.38 &99.38 &99.38\\
\bottomrule
\end{tabular} 
\end{table}

\clearpage

\subsection{Resisc45}
RESISC45 \citep{Cheng2017_RSSurvey} dataset is a publicly available benchmark for Remote Sensing Image Scene Classification (RESISC), created by Northwestern Polytechnical University (NWPU). This dataset contains 31500 images, covering 45 scene classes with 700 images in each class. The 45 scene classes are as follows: airplane, airport, baseball diamond, basketball court, beach, bridge, chaparral, church, circular farmland, cloud, commercial area, dense residential, desert, forest, freeway, golf course, ground track field, harbor, industrial area, intersection, island, lake, meadow, medium residential, mobile home park, mountain, overpass, palace, parking lot, railway, railway station, rectangular farmland, river, roundabout, runway, sea ice, ship, snowberg, sparse residential, stadium, storage tank, tennis court, terrace, thermal power station, and wetland. Accordingly, these classes contain a variety of spatial patterns, some homogeneous with respect to texture, some homogeneous with respect to color, others not homogeneous at all.

\begin{figure}[ht]
\centering
  \includegraphics[width=0.7\linewidth]{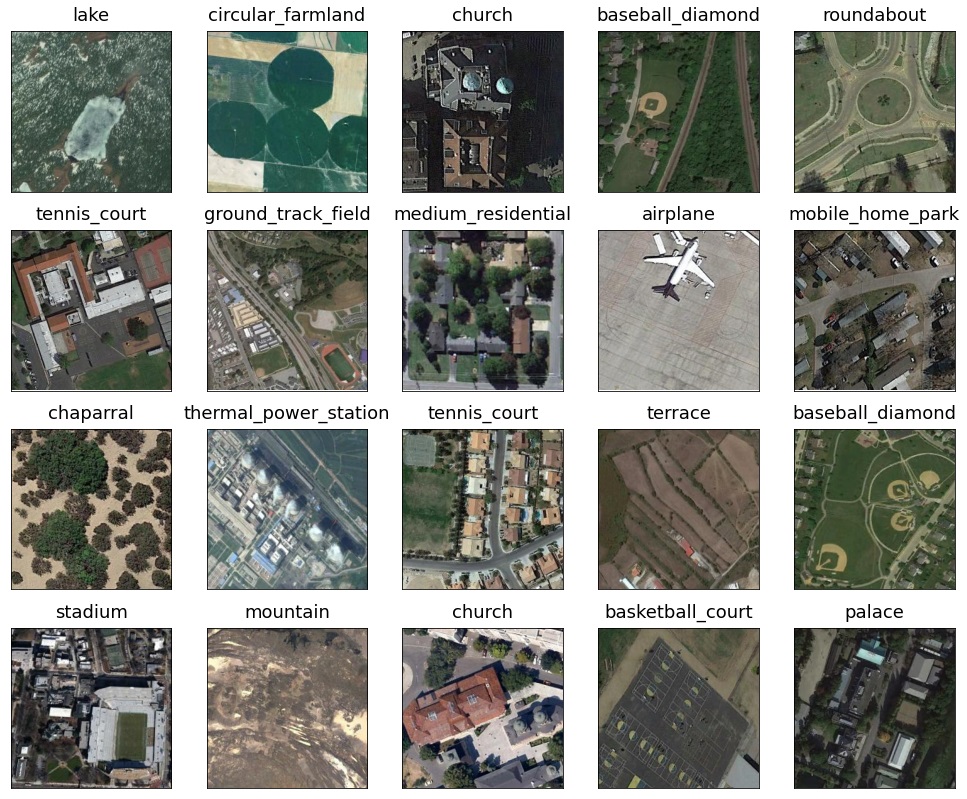}
  \caption{Example images with labels from the Resisc45 dataset.}
  \label{fig:resisc45_samples}
\end{figure}

\begin{figure}[ht]
  \centering
  \includegraphics[width=0.7\linewidth]{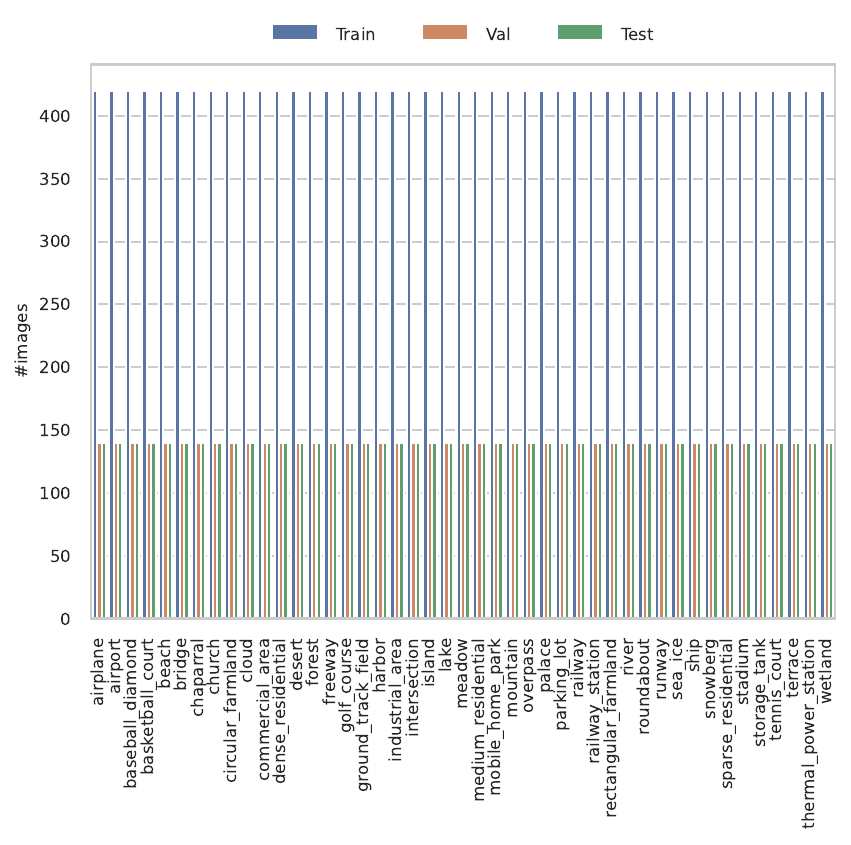}
  \caption{Class distribution for the Resisc45 dataset.}
  \label{fig:resisc45_distribution}
\end{figure}

The images are with a size of 256x256 pixels in the RGB color space. The spatial resolution varies from about 30m to 0.2m per pixel for most of the scene classes except for the classes of island, lake, mountain, and snowberg that have lower spatial resolutions. The 31500 images cover more than 100 countries and regions all over the world, including developing, transition, and highly developed economies (Figure~\ref{fig:resisc45_samples}). Our generated train, test and validation splits distribution is show on Figure~\ref{fig:resisc45_distribution}.

\begin{figure}[ht]
  \centering
  \includegraphics[width=0.9\linewidth]{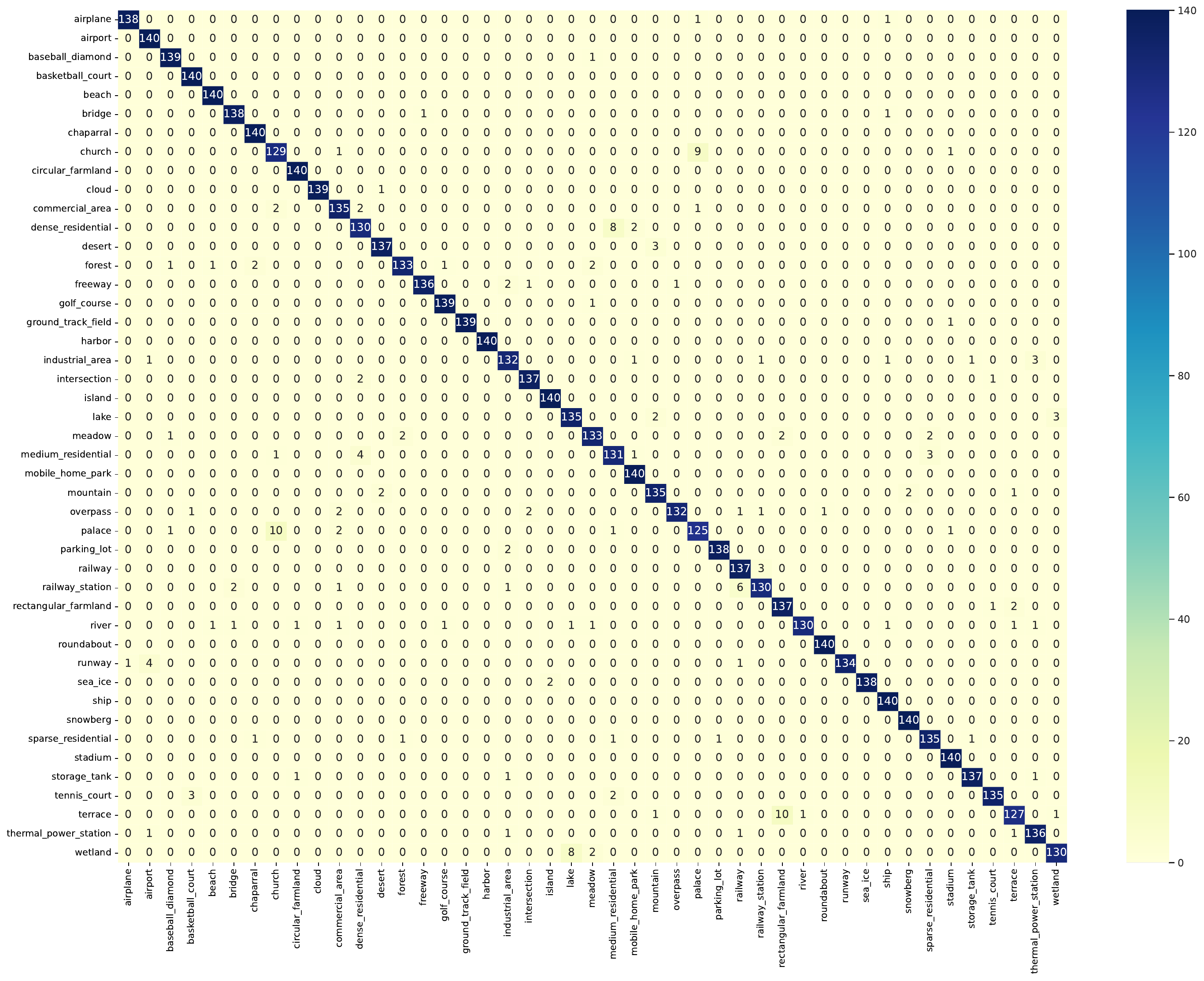}
  \caption{Confusion matrix for the pre-trained Vision Transformer model on the Resisc45 dataset.}
  \label{fig:resisc45_confusionmatrix}
\end{figure}

Detailed results for all pre-trained models are shown on Table~\ref{tab:pre-trained_resisc45} and for all the models learned from scratch are presented on Table~\ref{tab:scratch_resisc45}. The best performing model is the pre-trained Vision Transformer model. The results on a class level are show on Table~\ref{tab:perclass_resisc45} along with a confusion matrix on Figure~\ref{fig:resisc45_confusionmatrix}.

\begin{table}[ht]\centering
\caption{Detailed results for pre-trained models on the Resisc45 dataset.}\label{tab:pre-trained_resisc45}
\scriptsize \begin{adjustbox}{width=0.75\linewidth}
\begin{tabular}{lrrrrrrrrrrr}\toprule
Model \textbackslash Metric &\rotatebox{90}{Accuracy} &\rotatebox{90}{Macro Precision} &\rotatebox{90}{Weighted Precision} &\rotatebox{90}{Macro Recall} &\rotatebox{90}{Weighted Recall} &\rotatebox{90}{Macro F1 score} &\rotatebox{90}{Weighted F1 score} &\rotatebox{90}{Avg. time / epoch (sec.)} &\rotatebox{90}{Total time (sec.)} &\rotatebox{90}{Best epoch}\\\midrule
AlexNet &90.49 &90.56 &90.56 &90.49 &90.49 &90.49 &90.49 &12.03 &385 &22\\
VGG16 &93.90 &93.91 &93.91 &93.90 &93.90 &93.89 &93.89 &39.87 &1196 &20\\
ResNet50 &96.46 &96.50 &96.50 &96.46 &96.46 &96.46 &96.46 &30.61 &1163 &28\\
RestNet152 &96.54 &96.57 &96.57 &96.54 &96.54 &96.54 &96.54 &65.11 &2409 &27\\
DenseNet161 &96.51 &96.53 &96.53 &96.51 &96.51 &96.51 &96.51 &72.05 &3098 &33\\
EfficientNetB0 &94.87 &94.93 &94.93 &94.87 &94.87 &94.88 &94.88 &27.12 &678 &15\\
ConvNeXt &96.27 &96.28 &96.28 &96.27 &96.27 &96.26 &96.26 &46.79 &1778 &28\\
Vision Transformer &\textbf{97.08} &97.10 &97.10 &97.08 &97.08 &97.07 &97.07 &51.19 &2713 &43\\
MLP Mixer &95.95 &95.99 &95.99 &95.95 &95.95 &95.96 &95.96 &35.62 &1033 &19\\
Swin Transformer &96.59 &96.60 &96.60 &96.59 &96.59 &96.58 &96.58 &143.57 &4020 &18 \\
\bottomrule
\end{tabular} \end{adjustbox}
\end{table}

\begin{table}[ht]\centering
\caption{Detailed results for models trained from scratch on the Resisc45 dataset.}\label{tab:scratch_resisc45}
\scriptsize 
\begin{tabular}{lrrrrrrrrrrr}\toprule
Model \textbackslash Metric &\rotatebox{90}{Accuracy} &\rotatebox{90}{Macro Precision} &\rotatebox{90}{Weighted Precision} &\rotatebox{90}{Macro Recall} &\rotatebox{90}{Weighted Recall} &\rotatebox{90}{Macro F1 score} &\rotatebox{90}{Weighted F1 score} &\rotatebox{90}{Avg. time / epoch (sec.)} &\rotatebox{90}{Total time (sec.)} &\rotatebox{90}{Best epoch}\\\midrule
AlexNet &82.16 &82.29 &82.29 &82.16 &82.16 &82.10 &82.10 &10.91 &633 &43\\
VGG16 &83.89 &84.00 &84.00 &83.89 &83.89 &83.84 &83.84 &38.37 &2993 &63\\
ResNet50 &92.33 &92.40 &92.40 &92.33 &92.33 &92.33 &92.33 &31.31 &1941 &47\\
RestNet152 &90.68 &90.79 &90.79 &90.68 &90.68 &90.69 &90.69 &64.83 &4084 &48\\
DenseNet161 &\textbf{93.46} &93.50 &93.50 &93.46 &93.46 &93.46 &93.46 &71.22 &5484 &62\\
EfficientNetB0 &91.37 &91.47 &91.47 &91.37 &91.37 &91.38 &91.38 &27.66 &2102 &61\\
ConvNeXt &85.94 &86.30 &86.30 &85.94 &85.94 &86.05 &86.05 &46.51 &2279 &34\\
Vision Transformer &81.02 &81.18 &81.18 &81.02 &81.02 &80.98 &80.98 &50.21 &2611 &37\\
MLP Mixer &69.41 &69.67 &69.67 &69.41 &69.41 &69.22 &69.22 &35.69 &1285 &21\\
Swin Transformer &88.73 &88.82 &88.82 &88.73 &88.73 &88.71 &88.71 &144.87 &14487 &85 \\
\bottomrule
\end{tabular} 
\end{table}

\begin{table}[ht]\centering
\caption{Per class results for the pre-trained Vision Transformer model on the Resisc45 dataset,}\label{tab:perclass_resisc45}
\scriptsize 
\begin{tabular}{lrrrr}\toprule
Label &Precision &Recall &F1 score \\\midrule
airplane &99.28 &98.57 &98.92\\
airport &95.89 &100.00 &97.90\\
baseball\_diamond &97.89 &99.29 &98.58\\
basketball\_court &97.22 &100.00 &98.59\\
beach &98.59 &100.00 &99.29\\
bridge &97.87 &98.57 &98.22\\
chaparral &97.90 &100.00 &98.94\\
church &90.85 &92.14 &91.49\\
circular\_farmland &98.59 &100.00 &99.29\\
cloud &100.00 &99.29 &99.64\\
commercial\_area &95.07 &96.43 &95.74\\
dense\_residential &94.20 &92.86 &93.53\\
desert &97.86 &97.86 &97.86\\
forest &97.79 &95.00 &96.38\\
freeway &99.27 &97.14 &98.19\\
golf\_course &98.58 &99.29 &98.93\\
ground\_track\_field &100.00 &99.29 &99.64\\
harbor &100.00 &100.00 &100.00\\
industrial\_area &94.96 &94.29 &94.62\\
intersection &97.86 &97.86 &97.86\\
island &98.59 &100.00 &99.29\\
lake &93.75 &96.43 &95.07\\
meadow &95.00 &95.00 &95.00\\
medium\_residential &91.61 &93.57 &92.58\\
mobile\_home\_park &97.22 &100.00 &98.59\\
mountain &95.74 &96.43 &96.09\\
overpass &99.25 &94.29 &96.70\\
palace &91.91 &89.29 &90.58\\
parking\_lot &99.28 &98.57 &98.92\\
railway &93.84 &97.86 &95.80\\
railway\_station &96.30 &92.86 &94.55\\
rectangular\_farmland &91.95 &97.86 &94.81\\
river &99.24 &92.86 &95.94\\
roundabout &99.29 &100.00 &99.64\\
runway &100.00 &95.71 &97.81\\
sea\_ice &100.00 &98.57 &99.28\\
ship &97.22 &100.00 &98.59\\
snowberg &98.59 &100.00 &99.29\\
sparse\_residential &96.43 &96.43 &96.43\\
stadium &97.90 &100.00 &98.94\\
storage\_tank &98.56 &97.86 &98.21\\
tennis\_court &98.54 &96.43 &97.47\\
terrace &96.21 &90.71 &93.38\\
thermal\_power\_station &96.45 &97.14 &96.80\\
wetland &97.01 &92.86 &94.89\\
\bottomrule
\end{tabular} 
\end{table}

\clearpage

\subsection{RSI-CB256}
\begin{figure}[ht]
\centering
  \includegraphics[width=0.5\linewidth]{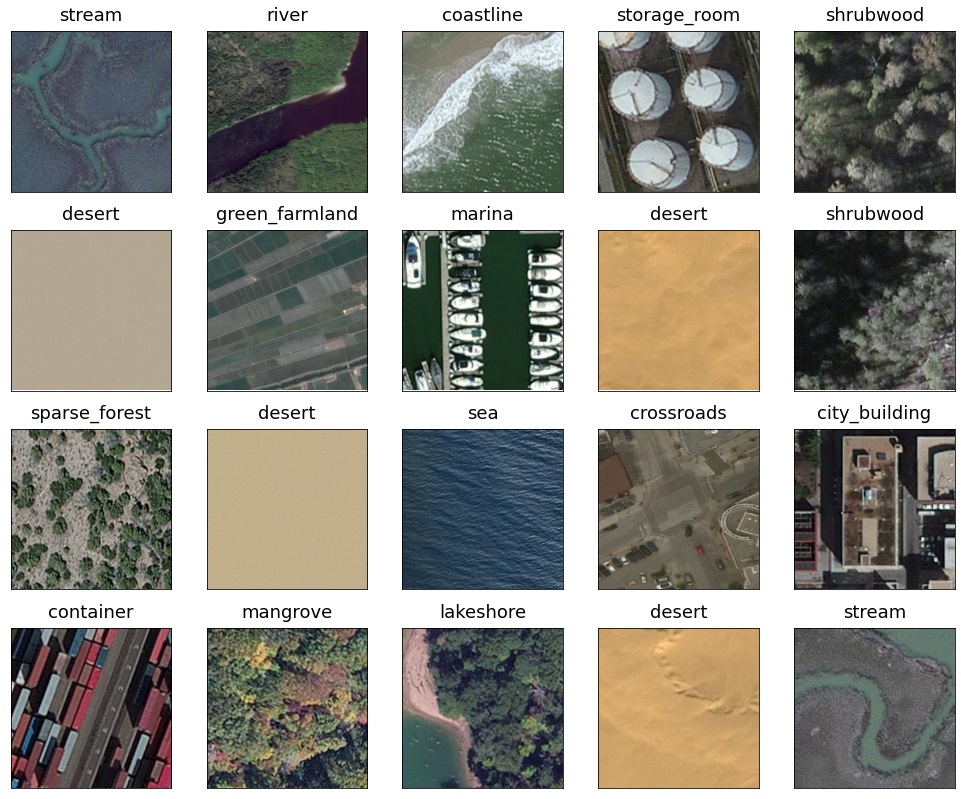}
  \caption{Example images with labels from the RSI-CB256 dataset.}
  \label{fig:rsicb256_samples}
\end{figure}

\begin{figure}[ht]
  \centering
  \includegraphics[width=0.7\linewidth]{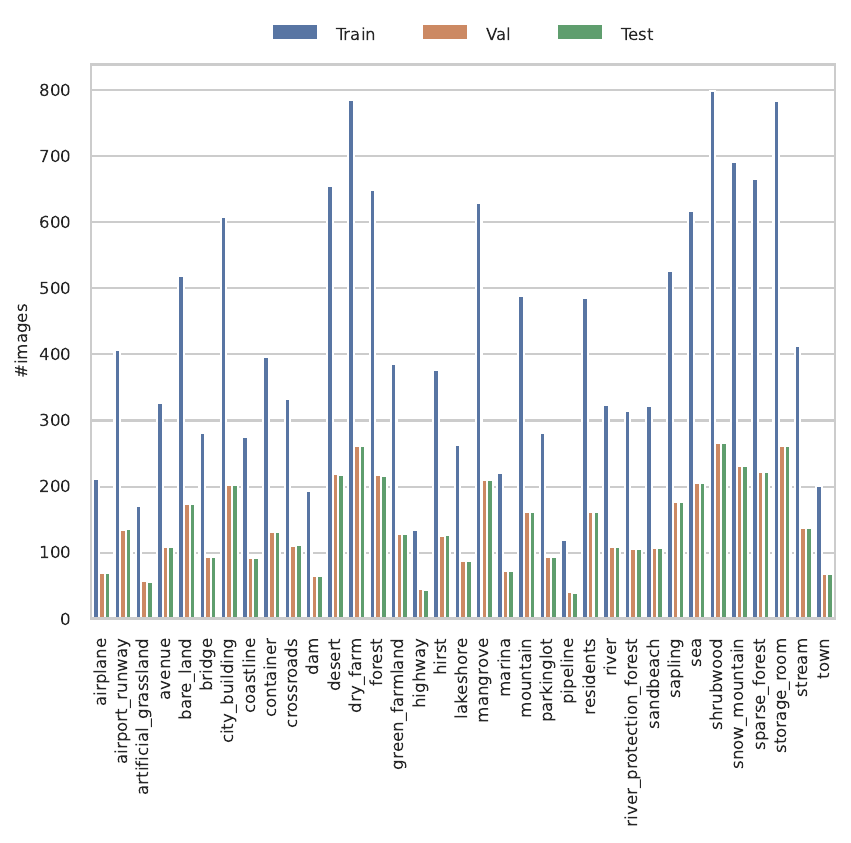}
  \caption{Class distribution for the RSI-CB526 dataset.}
  \label{fig:rsicb256_distribution}
\end{figure}

RSI-CB256 \citep{haifeng2020rsicb256} is a large scale remote sensing image classification benchmark via crowdsource data such as Open Street Map (OSM) data, ground objects in remote sensing images etc. It contains 35 categories and more than 24000 images with a size of 256x256 pixels (Figure~\ref{fig:rsicb256_samples}). A strict object category system according to the national standard of land-use classification in China and the hierarchical grading mechanism of ImageNet-1K has been established. Using crowd-source data as a supervisor facilitates machine self-learning through the Internet. The class distribution of the train, test and validation splits is presented in Figure~\ref{fig:rsicb256_distribution}.

\begin{figure}[ht]
  \centering
  \includegraphics[width=0.9\linewidth]{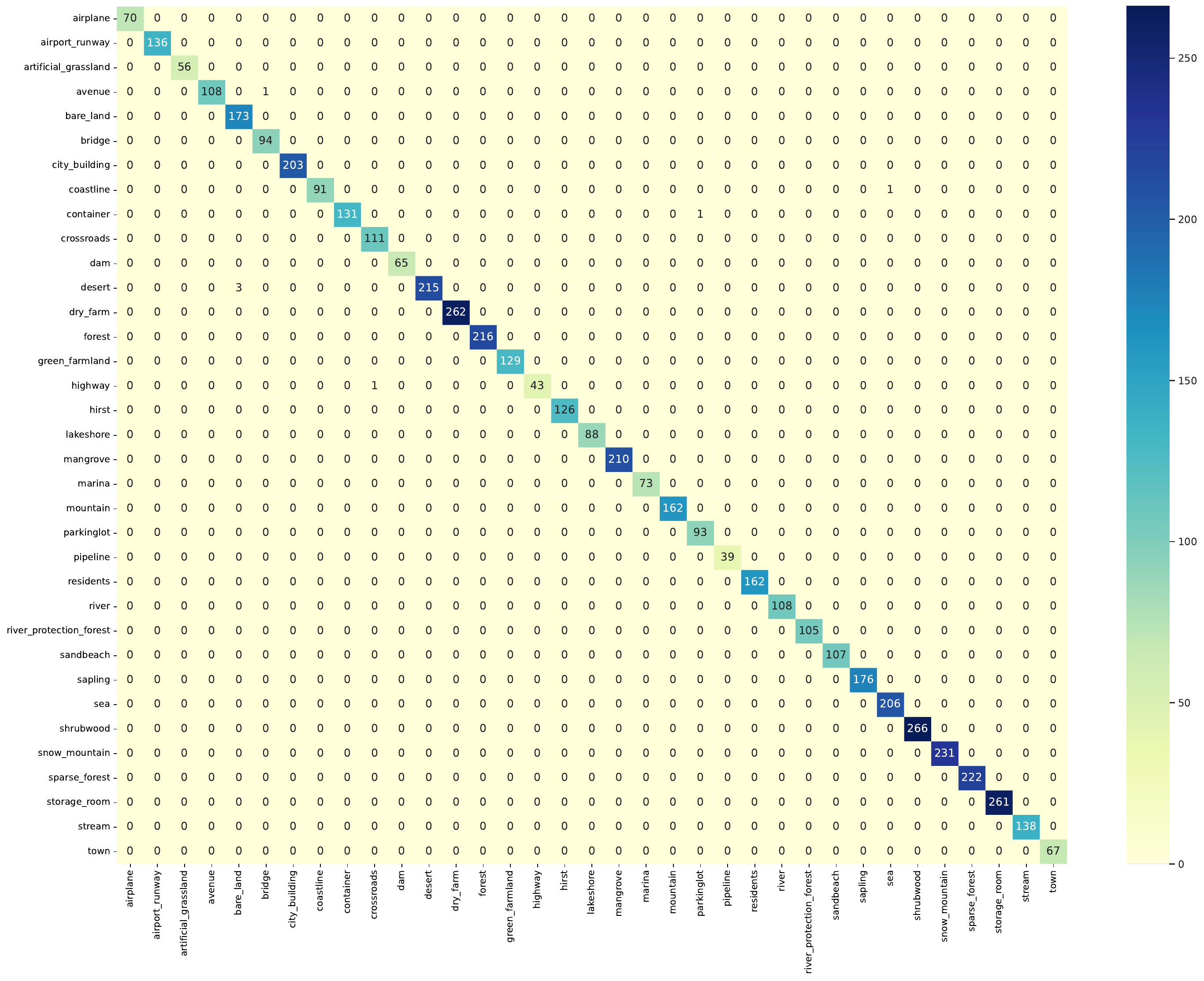}
  \caption{Confusion matrix for the pre-trained ResNet152 model on the RSI-CB256 dataset.}
  \label{fig:rsicb256_confusionmatrix}
\end{figure}

Detailed results for all pre-trained models are shown on Table~\ref{tab:pre-trained_rsicb256} and for all the models learned from scratch are presented on Table~\ref{tab:scratch_rsicb256}. The best performing model is the pre-trained ResNet152 model. The results on a class level are show on Table~\ref{tab:perclass_rsicb256} along with a confusion matrix on Figure~\ref{fig:rsicb256_confusionmatrix}.

\begin{table}[ht]\centering
\caption{Detailed results for pre-trained models on the RSI-CB256 dataset.}\label{tab:pre-trained_rsicb256}
\scriptsize \begin{adjustbox}{width=0.75\linewidth}
\begin{tabular}{lrrrrrrrrrrr}\toprule
Model \textbackslash Metric &\rotatebox{90}{Accuracy} &\rotatebox{90}{Macro Precision} &\rotatebox{90}{Weighted Precision} &\rotatebox{90}{Macro Recall} &\rotatebox{90}{Weighted Recall} &\rotatebox{90}{Macro F1 score} &\rotatebox{90}{Weighted F1 score} &\rotatebox{90}{Avg. time / epoch (sec.)} &\rotatebox{90}{Total time (sec.)} &\rotatebox{90}{Best epoch}\\\midrule
AlexNet &99.35 &99.13 &99.36 &99.06 &99.35 &99.09 &99.35 &34.84 &1568 &35\\
VGG16 &99.05 &98.93 &99.07 &98.75 &99.05 &98.83 &99.05 &34.04 &885 &16\\
ResNet50 &99.68 &99.53 &99.68 &99.54 &99.68 &99.53 &99.68 &33.69 &1078 &22\\
RestNet152 &\textbf{99.86} &99.85 &99.86 &99.82 &99.86 &99.83 &99.86 &51.90 &1609 &21\\
DenseNet161 &99.74 &99.68 &99.74 &99.64 &99.74 &99.66 &99.74 &56.60 &2717 &38\\
EfficientNetB0 &99.72 &99.63 &99.72 &99.65 &99.72 &99.64 &99.72 &33.50 &1340 &30\\
ConvNeXt &99.60 &99.50 &99.60 &99.55 &99.60 &99.52 &99.60 &40.35 &1977 &39\\
Vision Transformer &99.76 &99.75 &99.76 &99.71 &99.76 &99.73 &99.76 &41.18 &1400 &24\\
MLP Mixer &99.66 &99.54 &99.66 &99.61 &99.66 &99.57 &99.66 &35.29 &1235 &25\\
Swin Transformer &99.68 &99.54 &99.68 &99.62 &99.68 &99.57 &99.68 &113.14 &4752 &32 \\
\bottomrule
\end{tabular} \end{adjustbox}
\end{table}

\begin{table}[ht]\centering
\caption{Detailed results for models trained from scratch on the RSI-CB256 dataset.}\label{tab:scratch_rsicb256}
\scriptsize \begin{adjustbox}{width=0.75\linewidth}
\begin{tabular}{lrrrrrrrrrrr}\toprule
Model \textbackslash Metric &\rotatebox{90}{Accuracy} &\rotatebox{90}{Macro Precision} &\rotatebox{90}{Weighted Precision} &\rotatebox{90}{Macro Recall} &\rotatebox{90}{Weighted Recall} &\rotatebox{90}{Macro F1 score} &\rotatebox{90}{Weighted F1 score} &\rotatebox{90}{Avg. time / epoch (sec.)} &\rotatebox{90}{Total time (sec.)} &\rotatebox{90}{Best epoch}\\\midrule
AlexNet &97.35 &96.55 &97.39 &96.54 &97.35 &96.51 &97.35 &34.99 &2414 &54\\
VGG16 &98.83 &98.51 &98.84 &98.36 &98.83 &98.43 &98.83 &34.90 &2757 &64\\
ResNet50 &98.83 &98.51 &98.84 &98.36 &98.83 &98.43 &98.83 &36.39 &3166 &72\\
RestNet152 &\textbf{99.15} &98.98 &99.15 &98.81 &99.15 &98.89 &99.15 &51.86 &4472 &72\\
DenseNet161 &99.13 &98.80 &99.13 &98.71 &99.13 &98.75 &99.13 &56.75 &4029 &56\\
EfficientNetB0 &99.11 &98.85 &99.12 &98.91 &99.11 &98.87 &99.11 &26.50 &2123 &71\\
ConvNeXt &98.44 &97.75 &98.45 &97.74 &98.44 &97.73 &98.44 &36.93 &2622 &56\\
Vision Transformer &98.12 &97.52 &98.13 &97.12 &98.12 &97.31 &98.12 &41.08 &3204 &63\\
MLP Mixer &98.42 &97.81 &98.43 &97.80 &98.42 &97.79 &98.42 &29.00 &2900 &86\\
Swin Transformer &99.09 &98.83 &99.09 &98.70 &99.09 &98.76 &99.09 &113.60 &7157 &48 \\
\bottomrule
\end{tabular} \end{adjustbox}
\end{table}

\begin{table}[ht]\centering
\caption{Per class results for the pre-trained ResNet152 model on the RSI-CB256 dataset.}\label{tab:perclass_rsicb256}
\scriptsize 
\begin{tabular}{lrrrr}\toprule
Label &Precision &Recall &F1 score \\\midrule
airplane &100.00 &100.00 &100.00\\
airport\_runway &100.00 &100.00 &100.00\\
artificial\_grassland &100.00 &100.00 &100.00\\
avenue &100.00 &99.08 &99.54\\
bare\_land &98.30 &100.00 &99.14\\
bridge &98.95 &100.00 &99.47\\
city\_building &100.00 &100.00 &100.00\\
coastline &100.00 &98.91 &99.45\\
container &100.00 &99.24 &99.62\\
crossroads &99.11 &100.00 &99.55\\
dam &100.00 &100.00 &100.00\\
desert &100.00 &98.62 &99.31\\
dry\_farm &100.00 &100.00 &100.00\\
forest &100.00 &100.00 &100.00\\
green\_farmland &100.00 &100.00 &100.00\\
highway &100.00 &97.73 &98.85\\
hirst &100.00 &100.00 &100.00\\
lakeshore &100.00 &100.00 &100.00\\
mangrove &100.00 &100.00 &100.00\\
marina &100.00 &100.00 &100.00\\
mountain &100.00 &100.00 &100.00\\
parkinglot &98.94 &100.00 &99.47\\
pipeline &100.00 &100.00 &100.00\\
residents &100.00 &100.00 &100.00\\
river &100.00 &100.00 &100.00\\
river\_protection\_forest &100.00 &100.00 &100.00\\
sandbeach &100.00 &100.00 &100.00\\
sapling &100.00 &100.00 &100.00\\
sea &99.52 &100.00 &99.76\\
shrubwood &100.00 &100.00 &100.00\\
snow\_mountain &100.00 &100.00 &100.00\\
sparse\_forest &100.00 &100.00 &100.00\\
storage\_room &100.00 &100.00 &100.00\\
stream &100.00 &100.00 &100.00\\
town &100.00 &100.00 &100.00\\
\bottomrule
\end{tabular} 
\end{table}

\clearpage

\subsection{RSSCN7}
\begin{figure}[ht]
\centering
  \includegraphics[width=0.7\linewidth]{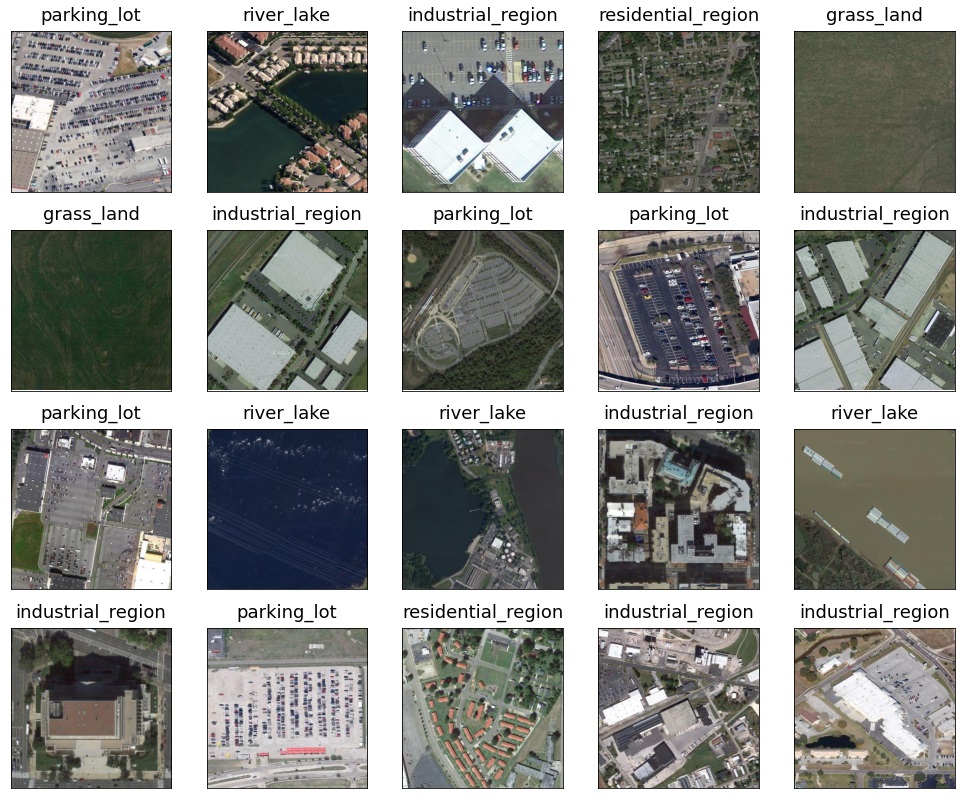}
  \caption{Example images with labels from the RSSCN7 dataset.}
  \label{fig:rsscn7_samples}
\end{figure}

\begin{figure}[ht]
  \centering
  \includegraphics[width=0.4\linewidth]{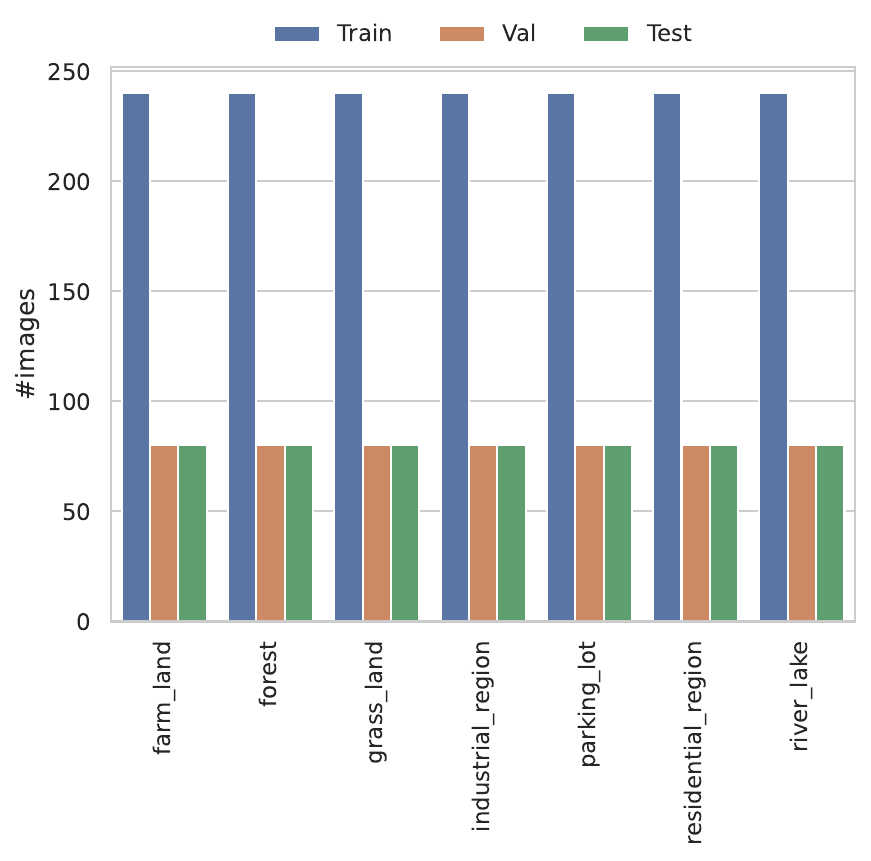}
  \caption{Class distribution for the RSSCN7 dataset.}
  \label{fig:rsscn7_distribution}
\end{figure}

RSSCN7 \citep{Zou2015RSSCN7} is a scene classification dataset. The images are obtained from Google Earth. This dataset was collected for academic research. It contains a total of 2800 remote sensing images, which are organized into 7 scene classes: grass land, forest, farm land, parking lot, residential region, industrial region, and river/lake (Figure~\ref{fig:rsscn7_samples}). For each, class there are 400 RGB images that are cropped on four different scales with 100 images per scale. Each image has a 400x400 pixels size. The main challenge of this dataset is the scale variations of the images. The class distribution over the train, test and validation splits is presented on Figure~\ref{fig:rsscn7_distribution}.

\begin{figure}[ht]
  \centering
  \includegraphics[width=0.5\linewidth]{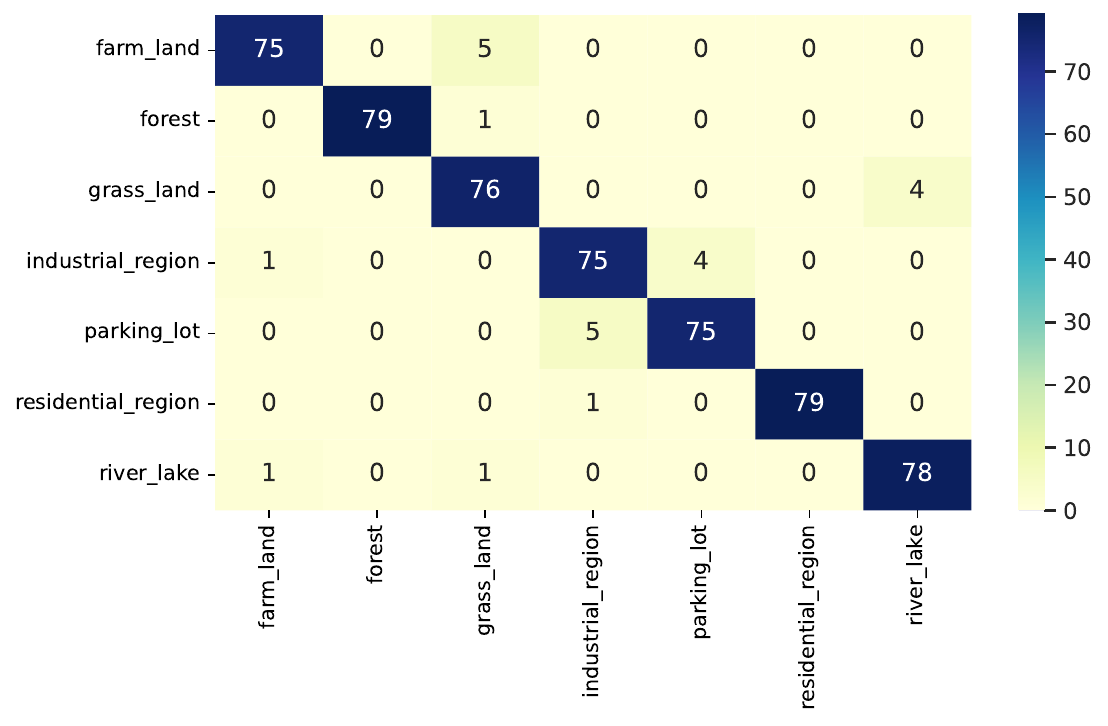}
  \caption{Confusion matrix for the pre-trained Vision Transformer model on the RSSCN7 dataset.}
  \label{fig:rsscn7_confusionmatrix}
\end{figure}

Detailed results for all pre-trained models are shown on Table~\ref{tab:pre-trained_rsscn7} and for all the models learned from scratch are presented on Table~\ref{tab:scratch_rsscn7}. The best performing model is the pre-trained Vision Transformer model. The results on a class level are show on Table~\ref{tab:perclass_rsscn7} along with a confusion matrix on Figure~\ref{fig:rsscn7_confusionmatrix}.

\begin{table}[ht]\centering
\caption{Detailed results for pre-trained models on the RSSCN7 dataset.}\label{tab:pre-trained_rsscn7}
\scriptsize \begin{adjustbox}{width=0.75\linewidth}
\begin{tabular}{lrrrrrrrrrrr}\toprule
Model \textbackslash Metric &\rotatebox{90}{Accuracy} &\rotatebox{90}{Macro Precision} &\rotatebox{90}{Weighted Precision} &\rotatebox{90}{Macro Recall} &\rotatebox{90}{Weighted Recall} &\rotatebox{90}{Macro F1 score} &\rotatebox{90}{Weighted F1 score} &\rotatebox{90}{Avg. time / epoch (sec.)} &\rotatebox{90}{Total time (sec.)} &\rotatebox{90}{Best epoch}\\\midrule
AlexNet &91.96 &92.05 &92.05 &91.96 &91.96 &91.92 &91.92 &3.19 &118 &27\\
VGG16 &93.93 &93.95 &93.95 &93.93 &93.93 &93.90 &93.90 &4.68 &159 &24\\
ResNet50 &95.00 &95.08 &95.08 &95.00 &95.00 &94.99 &94.99 &3.90 &121 &21\\
RestNet152 &95.00 &95.07 &95.07 &95.00 &95.00 &95.01 &95.01 &7.09 &241 &24\\
DenseNet161 &94.82 &94.83 &94.83 &94.82 &94.82 &94.82 &94.82 &7.59 &220 &19\\
EfficientNetB0 &95.54 &95.56 &95.56 &95.54 &95.54 &95.54 &95.54 &3.79 &163 &33\\
ConvNeXt &94.64 &94.76 &94.76 &94.64 &94.64 &94.61 &94.61 &5.23 &183 &25\\
Vision Transformer &\textbf{95.89} &95.95 &95.95 &95.89 &95.89 &95.91 &95.91 &5.54 &227 &31\\
MLP Mixer &95.18 &95.23 &95.23 &95.18 &95.18 &95.17 &95.17 &4.30 &86 &10\\
Swin Transformer &95.18 &95.23 &95.23 &95.18 &95.18 &95.18 &95.18 &13.42 &416 &21 \\
\bottomrule
\end{tabular} \end{adjustbox}
\end{table}

\begin{table}[ht]\centering
\caption{Detailed results for models trained from scratch on the RSSCN7 dataset.}\label{tab:scratch_rsscn7}
\scriptsize \begin{adjustbox}{width=0.75\linewidth}
\begin{tabular}{lrrrrrrrrrrr}\toprule
Model \textbackslash Metric &\rotatebox{90}{Accuracy} &\rotatebox{90}{Macro Precision} &\rotatebox{90}{Weighted Precision} &\rotatebox{90}{Macro Recall} &\rotatebox{90}{Weighted Recall} &\rotatebox{90}{Macro F1 score} &\rotatebox{90}{Weighted F1 score} &\rotatebox{90}{Avg. time / epoch (sec.)} &\rotatebox{90}{Total time (sec.)} &\rotatebox{90}{Best epoch}\\\midrule
AlexNet &80.54 &80.64 &80.64 &80.54 &80.54 &80.45 &80.45 &6.97 &697 &85\\
VGG16 &81.61 &81.50 &81.50 &81.61 &81.61 &81.41 &81.41 &6.74 &526 &63\\
ResNet50 &82.68 &82.65 &82.65 &82.68 &82.68 &82.41 &82.41 &3.76 &316 &69\\
RestNet152 &82.68 &82.65 &82.65 &82.68 &82.68 &82.41 &82.41 &6.90 &407 &44\\
DenseNet161 &\textbf{87.32} &87.55 &87.55 &87.32 &87.32 &87.38 &87.38 &8.50 &595 &55\\
EfficientNetB0 &83.93 &84.03 &84.03 &83.93 &83.93 &83.87 &83.87 &3.65 &365 &93\\
ConvNeXt &83.04 &82.84 &82.84 &83.04 &83.04 &82.90 &82.90 &5.43 &543 &87\\
Vision Transformer &86.07 &86.17 &86.17 &86.07 &86.07 &86.00 &86.00 &5.52 &453 &67\\
MLP Mixer &83.21 &83.29 &83.29 &83.21 &83.21 &83.17 &83.17 &4.08 &408 &100\\
Swin Transformer &82.50 &82.59 &82.59 &82.50 &82.50 &82.50 &82.50 &13.78 &951 &54 \\
\bottomrule
\end{tabular} \end{adjustbox}
\end{table}

\begin{table}[ht]\centering
\caption{Per class results for the pre-trained Vision Transformer model on the RSSCN7 dataset.}\label{tab:perclass_rsscn7}
\scriptsize 
\begin{tabular}{lrrrr}\toprule
Label &Precision &Recall &F1 score \\\midrule
farm\_land &97.40 &93.75 &95.54\\
forest &100.00 &98.75 &99.37\\
grass\_land &91.57 &95.00 &93.25\\
industrial\_region &92.59 &93.75 &93.17\\
parking\_lot &94.94 &93.75 &94.34\\
residential\_region &100.00 &98.75 &99.37\\
river\_lake &95.12 &97.50 &96.30\\
\bottomrule
\end{tabular} 
\end{table}

\clearpage

\subsection{SAT6}
\begin{figure}[ht]
\centering
  \includegraphics[width=0.5\linewidth]{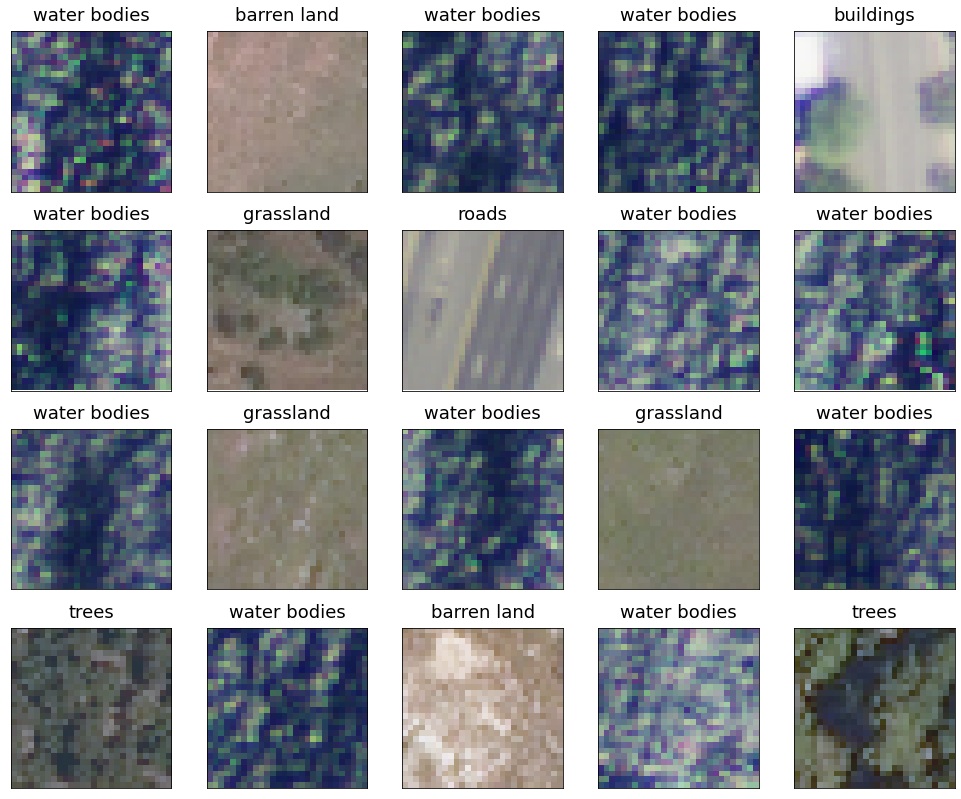}
  \caption{Example images with labels from the SAT6 dataset.}
  \label{fig:sat6_samples}
\end{figure}

SAT-6 \citep{Basu2015sat6} consists of a total of 405000 image patches each of size 28x28 and covering 6 land cover classes - barren land, trees, grassland, roads, buildings and water bodies (Figure~\ref{fig:sat6_samples}). The authors of the dataset selected 324000 images for the training dataset and 81000 were selected as testing dataset. Additionally we have selected 20\% of the images from the train dataset to create the validation split. The training and test datasets were selected from disjoint  National Agriculture Imagery Program (NAIP) tiles. The specifications for the various land cover classes of SAT-6 were adopted from those used in the National Land Cover Data (NLCD) algorithm. The class distribution of the train, test and validation splits is presented on Figure~\ref{fig:sat6_samples}.

\begin{figure}[ht]
  \centering
  \includegraphics[width=0.5\linewidth]{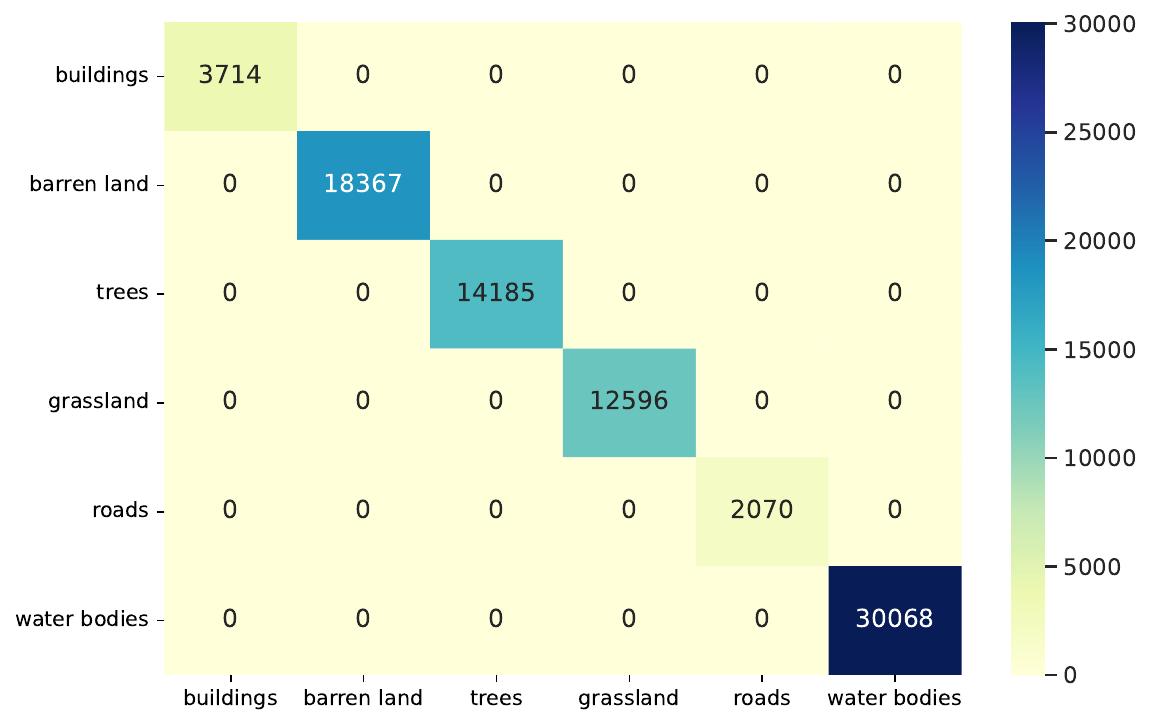}
  \caption{Confusion matrix for the pre-trained DenseNet161 model on the SAT6 dataset.}
  \label{fig:sat6_confusionmatrix}
\end{figure}

Detailed results for all pre-trained models are shown on Table~\ref{tab:pre-trained_sat6} and for all the models learned from scratch are presented on Table~\ref{tab:scratch_sat6}. All pre-trained model obtained excellent result on the dataset with ResNet50, ResNet152, DenseNet161, ConvNeXt, Vision Transformer, MLPMixer and Swin Transformer achieving 100 \% accuracy. The results on a class level are show on Table~\ref{tab:perclass_sat6} along with a confusion matrix on Figure~\ref{fig:sat6_confusionmatrix} for the DenseNet161 model.

\begin{figure}[ht]
  \centering
  \includegraphics[width=0.5\linewidth]{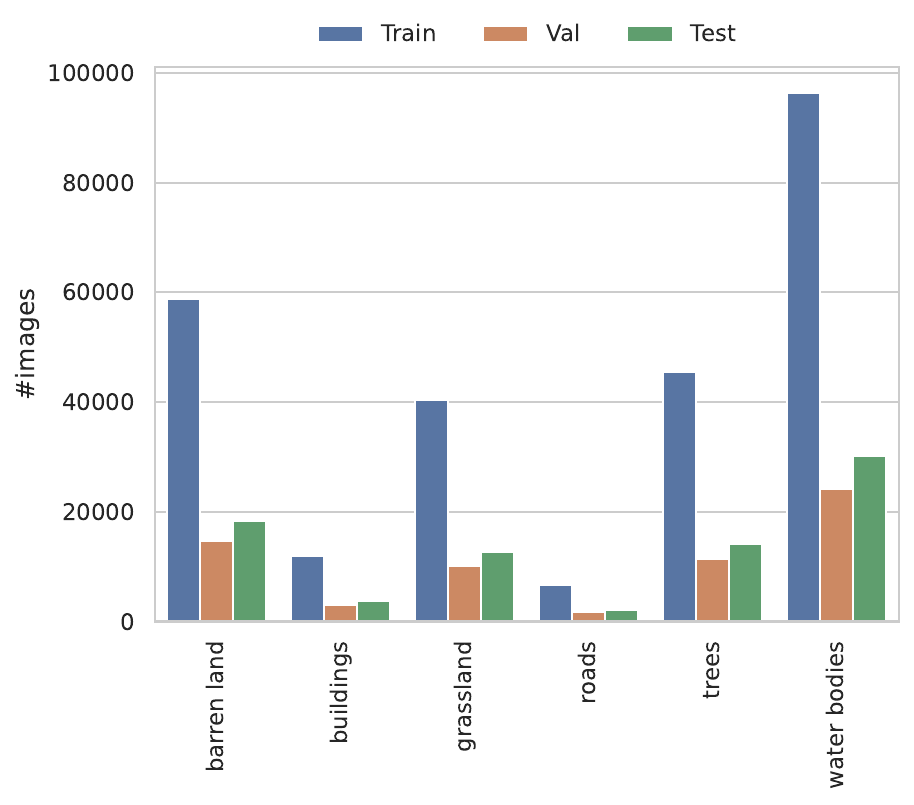}
  \caption{Class distribution for the SAT6 dataset.}
  \label{fig:sat6_distribution}
\end{figure}

\begin{table}[ht]\centering
\caption{Detailed results for pre-trained models on the SAT6 dataset.}\label{tab:pre-trained_sat6}
\scriptsize \begin{adjustbox}{width=0.75\linewidth}
\begin{tabular}{lrrrrrrrrrrr}\toprule
Model \textbackslash Metric &\rotatebox{90}{Accuracy} &\rotatebox{90}{Macro Precision} &\rotatebox{90}{Weighted Precision} &\rotatebox{90}{Macro Recall} &\rotatebox{90}{Weighted Recall} &\rotatebox{90}{Macro F1 score} &\rotatebox{90}{Weighted F1 score} &\rotatebox{90}{Avg. time / epoch (sec.)} &\rotatebox{90}{Total time (sec.)} &\rotatebox{90}{Best epoch}\\\midrule
AlexNet &99.98 &99.98 &99.98 &99.97 &99.98 &99.97 &99.98 &92.48 &5364 &48\\
VGG16 &99.99 &99.99 &99.99 &99.99 &99.99 &99.99 &99.99 &550.04 &29702 &44\\
ResNet50 &100.00 &100.00 &100.00 &100.00 &100.00 &100.00 &100.00 &410.33 &37340 &81\\
RestNet152 &100.00 &100.00 &100.00 &100.00 &100.00 &100.00 &100.00 &872.87 &61974 &61\\
DenseNet161 &100.00 &100.00 &100.00 &100.00 &100.00 &100.00 &100.00 &970.39 &55312 &47\\
EfficientNetB0 &99.99 &99.99 &99.99 &99.99 &99.99 &99.99 &99.99 &363.00 &8712 &14\\
ConvNeXt &100.00 &100.00 &100.00 &99.99 &100.00 &100.00 &100.00 &630.78 &42262 &57\\
Vision Transformer &100.00 &100.00 &100.00 &100.00 &100.00 &100.00 &100.00 &692.50 &42935 &52\\
MLP Mixer &100.00 &100.00 &100.00 &100.00 &100.00 &100.00 &100.00 &476.34 &15243 &22\\
Swin Transformer &100.00 &100.00 &100.00 &100.00 &100.00 &100.00 &100.00 &2,003.62 &106192 &43 \\
\bottomrule
\end{tabular} \end{adjustbox}
\end{table}

\begin{table}[ht]\centering
\caption{Detailed results for models trained from scratch on the SAT6 dataset.}\label{tab:scratch_sat6}
\scriptsize \begin{adjustbox}{width=0.75\linewidth}
\begin{tabular}{lrrrrrrrrrrr}\toprule
Model \textbackslash Metric &\rotatebox{90}{Accuracy} &\rotatebox{90}{Macro Precision} &\rotatebox{90}{Weighted Precision} &\rotatebox{90}{Macro Recall} &\rotatebox{90}{Weighted Recall} &\rotatebox{90}{Macro F1 score} &\rotatebox{90}{Weighted F1 score} &\rotatebox{90}{Avg. time / epoch (sec.)} &\rotatebox{90}{Total time (sec.)} &\rotatebox{90}{Best epoch}\\\midrule
AlexNet &99.27 &98.67 &99.27 &98.65 &99.27 &98.66 &99.27 &107.26 &10726 &98\\
VGG16 &99.56 &99.42 &99.56 &99.42 &99.56 &99.42 &99.56 &579.10 &57910 &98\\
ResNet50 &100.00 &100.00 &100.00 &100.00 &100.00 &100.00 &100.00 &457.04 &45704 &99\\
RestNet152 &100.00 &100.00 &100.00 &100.00 &100.00 &100.00 &100.00 &987.21 &98721 &94\\
DenseNet161 &100.00 &100.00 &100.00 &100.00 &100.00 &100.00 &100.00 &956.03 &95603 &85\\
EfficientNetB0 &100.00 &100.00 &100.00 &100.00 &100.00 &100.00 &100.00 &420.37 &42037 &95\\
ConvNeXt &100.00 &100.00 &100.00 &100.00 &100.00 &100.00 &100.00 &627.69 &62769 &97\\
Vision Transformer &99.99 &99.98 &99.99 &99.98 &99.99 &99.98 &99.99 &687.12 &61841 &75\\
MLP Mixer &99.98 &99.98 &99.98 &99.96 &99.98 &99.97 &99.98 &479.37 &47937 &95\\
Swin Transformer &99.98 &99.96 &99.98 &99.97 &99.98 &99.97 &99.98 &1,973.44 &197344 &99 \\
\bottomrule
\end{tabular} \end{adjustbox}
\end{table}

\begin{table}[ht]\centering
\caption{Per class results for the pre-trained DenseNet model on the SAT6 dataset.}\label{tab:perclass_sat6}
\scriptsize 
\begin{tabular}{lrrrr}\toprule
Label &Precision &Recall &F1 score \\\midrule
buildings &100.00 &100.00 &100.00\\
barren land &100.00 &100.00 &100.00\\
trees &100.00 &100.00 &100.00\\
grassland &100.00 &100.00 &100.00\\
roads &100.00 &100.00 &100.00\\
water bodies &100.00 &100.00 &100.00\\
\bottomrule
\end{tabular} 
\end{table}

\clearpage

\subsection{Siri-Whu}
\begin{figure}[ht]
\centering
  \includegraphics[width=0.7\linewidth]{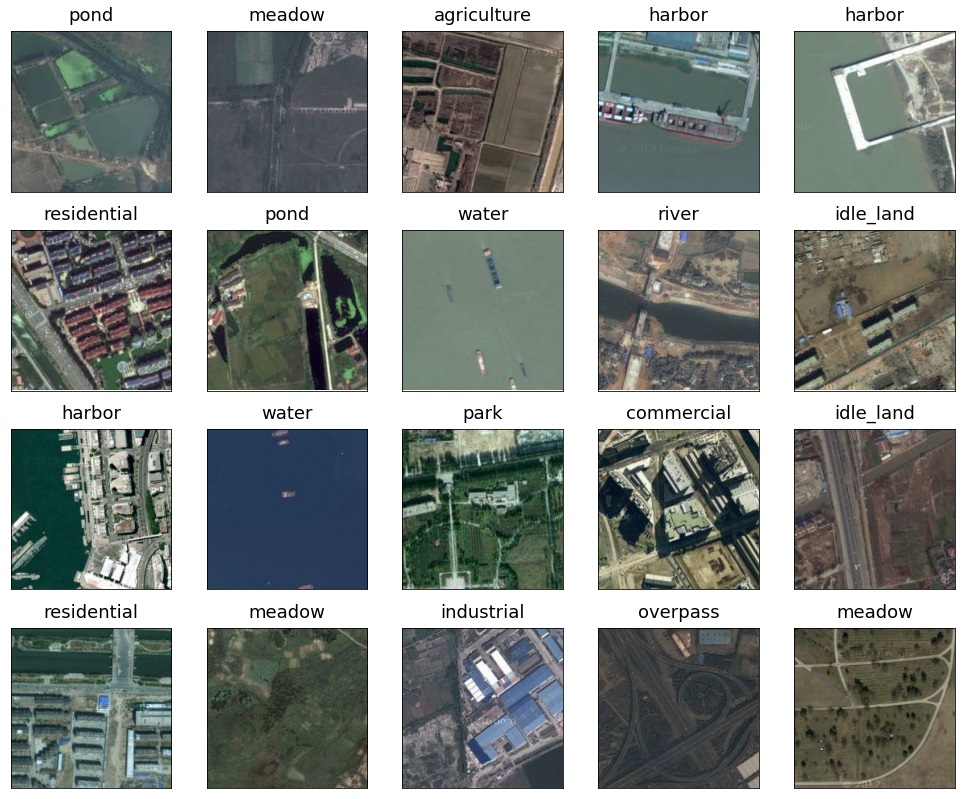}
  \caption{Example images with labels from the SIRI-WHU dataset.}
  \label{fig:siriwhu_samples}
\end{figure}

\begin{figure}[ht]
  \centering
  \includegraphics[width=0.5\linewidth]{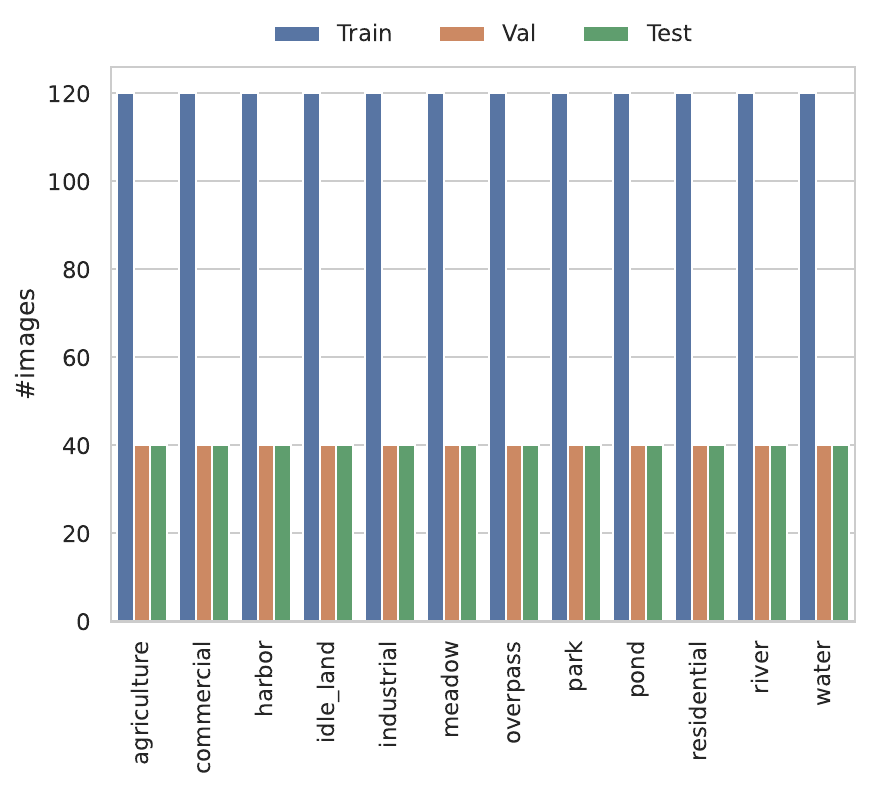}
  \caption{Class distribution for the SIRI-WHU dataset.}
  \label{fig:siriwhu_distribution}
\end{figure}

The SIRI-WHU \citep{zhu2016siriwhu} is a scene classification dataset comprised of 2400 images organized into 12 classes. Each class contains 200 images with a 2m spatial resolution and a size of 200×200 pixels (Figure~\ref{fig:siriwhu_samples}). It was collected from Google Earth (Google Inc.) by the Intelligent Data Extraction and Analysis of Remote Sensing (RS\_IDEA) Group in Wuhan University. The 12 land-use classes contain agriculture, commercial, harbor, idle land, industrial, meadow, overpass, park, pond, residential, river, and water. This dataset mainly covers urban areas in China, which means it lack diversity and is less challenging. The class distribution is presented on Figure~\ref{fig:siriwhu_distribution}.

\begin{figure}[ht]
  \centering
  \includegraphics[width=0.7\linewidth]{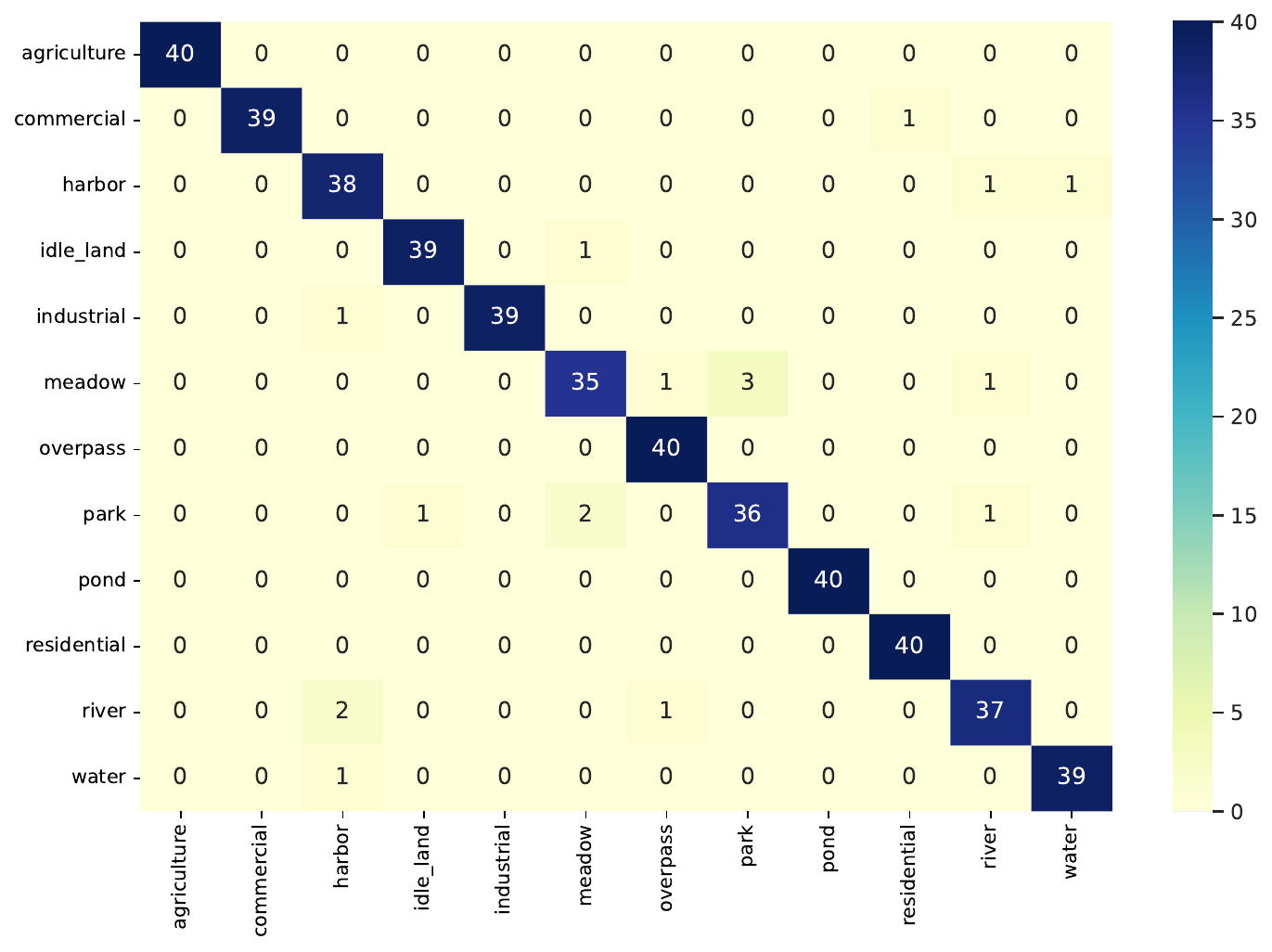}
  \caption{Confusion matrix for the pre-trained ResNet152 model on the SIRI-WHU dataset.}
  \label{fig:siriwhu_confusionmatrix}
\end{figure}

Detailed results for all pre-trained models are shown on Table~\ref{tab:pre-trained_siriwhu} and for all the models learned from scratch are presented on Table~\ref{tab:scratch_siriwhu}. The best performing model is the pre-trained ResNet152 model. The results on a class level are show on Table~\ref{tab:perclass_siriwhu} along with a confusion matrix on Figure~\ref{fig:siriwhu_confusionmatrix}.

\begin{table}[ht]\centering
\caption{Detailed results for pre-trained models on the SIRI-WHU dataset.}\label{tab:pre-trained_siriwhu}
\scriptsize \begin{adjustbox}{width=0.75\linewidth}
\begin{tabular}{lrrrrrrrrrrr}\toprule
Model \textbackslash Metric &\rotatebox{90}{Accuracy} &\rotatebox{90}{Macro Precision} &\rotatebox{90}{Weighted Precision} &\rotatebox{90}{Macro Recall} &\rotatebox{90}{Weighted Recall} &\rotatebox{90}{Macro F1 score} &\rotatebox{90}{Weighted F1 score} &\rotatebox{90}{Avg. time / epoch (sec.)} &\rotatebox{90}{Total time (sec.)} &\rotatebox{90}{Best epoch}\\\midrule
AlexNet &92.29 &92.64 &92.64 &92.29 &92.29 &92.31 &92.31 &4.28 &197 &36\\
VGG16 &93.96 &94.08 &94.08 &93.96 &93.96 &93.96 &93.96 &4.98 &214 &33\\
ResNet50 &95.00 &95.12 &95.12 &95.00 &95.00 &95.01 &95.01 &4.66 &191 &31\\
RestNet152 &\textbf{96.25} &96.27 &96.27 &96.25 &96.25 &96.24 &96.24 &6.65 &226 &24\\
DenseNet161 &95.63 &95.64 &95.64 &95.63 &95.63 &95.61 &95.61 &7.30 &365 &40\\
EfficientNetB0 &95.00 &95.09 &95.09 &95.00 &95.00 &95.01 &95.01 &4.57 &329 &62\\
ConvNeXt &96.25 &96.34 &96.34 &96.25 &96.25 &96.24 &96.24 &5.64 &203 &26\\
Vision Transformer &95.63 &95.73 &95.73 &95.63 &95.62 &95.63 &95.63 &5.37 &322 &50\\
MLP Mixer &95.21 &95.36 &95.36 &95.21 &95.21 &95.23 &95.23 &4.55 &150 &23\\
Swin Transformer &95.63 &95.60 &95.60 &95.63 &95.62 &95.57 &95.57 &11.87 &534 &35 \\
\bottomrule
\end{tabular} \end{adjustbox}
\end{table}

\begin{table}[ht]\centering
\caption{Detailed results for models trained from scratch on the SIRI-WHU dataset.}\label{tab:scratch_siriwhu}
\scriptsize \begin{adjustbox}{width=0.75\linewidth}
\begin{tabular}{lrrrrrrrrrrr}\toprule
Model \textbackslash Metric &\rotatebox{90}{Accuracy} &\rotatebox{90}{Macro Precision} &\rotatebox{90}{Weighted Precision} &\rotatebox{90}{Macro Recall} &\rotatebox{90}{Weighted Recall} &\rotatebox{90}{Macro F1 score} &\rotatebox{90}{Weighted F1 score} &\rotatebox{90}{Avg. time / epoch (sec.)} &\rotatebox{90}{Total time (sec.)} &\rotatebox{90}{Best epoch}\\\midrule
AlexNet &83.75 &83.83 &83.83 &83.75 &83.75 &83.66 &83.66 &3.54 &326 &77\\
VGG16 &84.79 &85.05 &85.05 &84.79 &84.79 &84.70 &84.70 &7.32 &732 &93\\
ResNet50 &\textbf{88.96} &89.14 &89.14 &88.96 &88.96 &88.94 &88.94 &3.81 &305 &65\\
RestNet152 &88.75 &88.67 &88.67 &88.75 &88.75 &88.62 &88.62 &6.54 &608 &78\\
DenseNet161 &86.67 &87.38 &87.38 &86.67 &86.67 &86.56 &86.56 &7.49 &749 &94\\
EfficientNetB0 &86.04 &86.23 &86.23 &86.04 &86.04 &85.94 &85.94 &3.61 &238 &51\\
ConvNeXt &84.17 &84.32 &84.32 &84.17 &84.17 &84.09 &84.09 &11.99 &1007 &69\\
Vision Transformer &86.25 &86.31 &86.31 &86.25 &86.25 &86.14 &86.14 &5.08 &503 &84\\
MLP Mixer &82.50 &82.40 &82.40 &82.50 &82.50 &82.34 &82.34 &3.92 &392 &98\\
Swin Transformer &85.83 &86.02 &86.02 &85.83 &85.83 &85.62 &85.62 &12.10 &1113 &77 \\
\bottomrule
\end{tabular} \end{adjustbox}
\end{table}

\begin{table}[ht]\centering
\caption{Per class results for the pre-trained ResNet152 model on the SIRI-WHU dataset.}\label{tab:perclass_siriwhu}
\scriptsize 
\begin{tabular}{lrrrr}\toprule
Label &Precision &Recall &F1 score \\\midrule
agriculture &100.00 &100.00 &100.00\\
commercial &100.00 &97.50 &98.73\\
harbor &90.48 &95.00 &92.68\\
idle\_land &97.50 &97.50 &97.50\\
industrial &100.00 &97.50 &98.73\\
meadow &92.11 &87.50 &89.74\\
overpass &95.24 &100.00 &97.56\\
park &92.31 &90.00 &91.14\\
pond &100.00 &100.00 &100.00\\
residential &97.56 &100.00 &98.77\\
river &92.50 &92.50 &92.50\\
water &97.50 &97.50 &97.50\\
\bottomrule
\end{tabular} 
\end{table}

\clearpage

\subsection{CLRS}
\begin{figure}[ht]
\centering
  \includegraphics[width=0.7\linewidth]{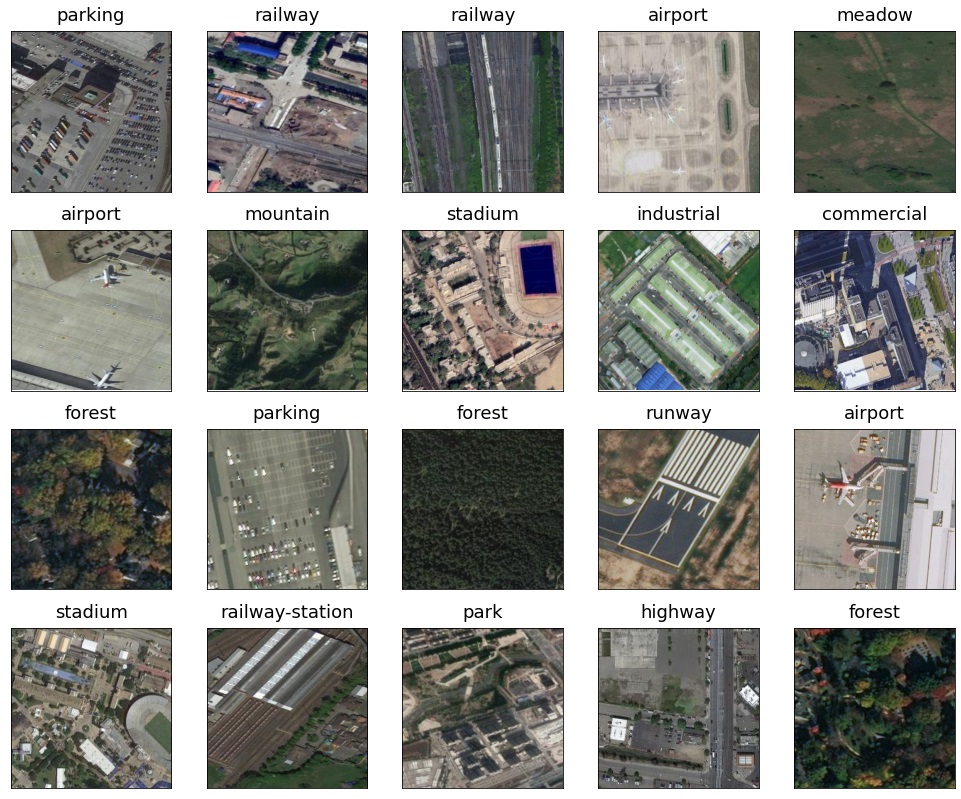}
  \caption{Example images with labels from the CLRS dataset.}
  \label{fig:clrs_samples}
\end{figure}

\begin{figure}[ht]
  \centering
  \includegraphics[width=0.6\linewidth]{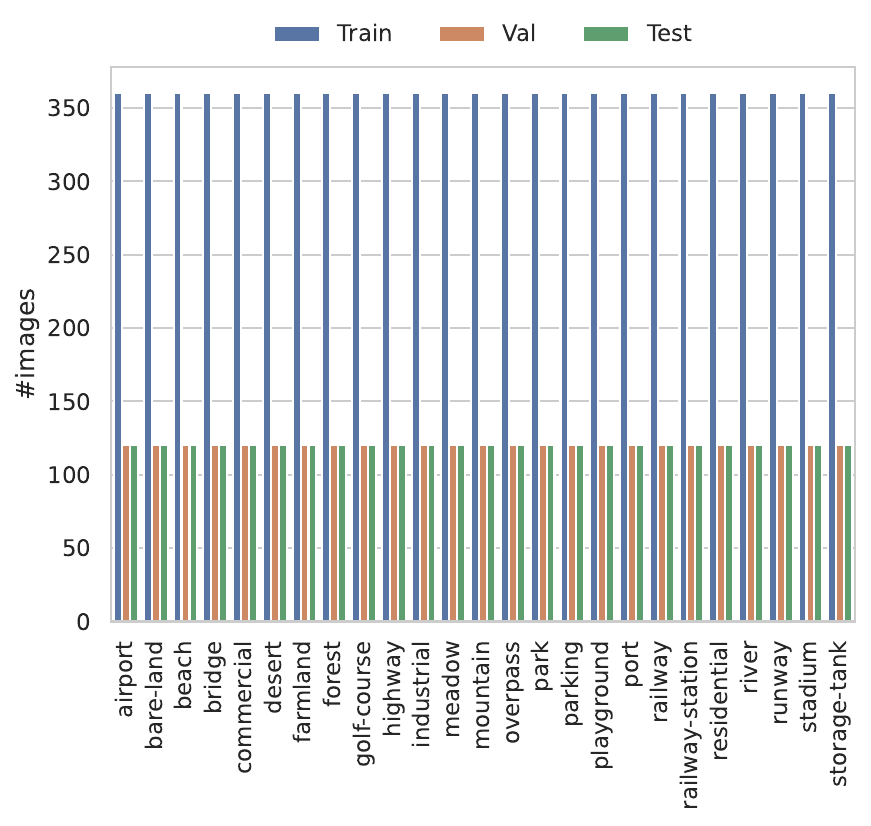}
  \caption{Class distribution for the CLRS dataset.}
  \label{fig:clrs_distribution}
\end{figure}

This dataset \citep{haifeng2020clrs} is a database designed for the task named Continual/Lifelong learning for remote sensing image scene classification. The proposed CLRS dataset consists of 15000 remote sensing images divided into 25 scene classes covering over 100 countries (Figure~\ref{fig:clrs_samples}). The images have a spatial resolution between 0.26~8.85 meters. The data is acquired from multiple sources such as: Google Earth, Bing Map, Google Map, and Tianditu. The class distribution of the train, test and validation splits is presented on Figure~\ref{fig:clrs_distribution}.

\begin{figure}[ht]
  \centering
  \includegraphics[width=0.9\linewidth]{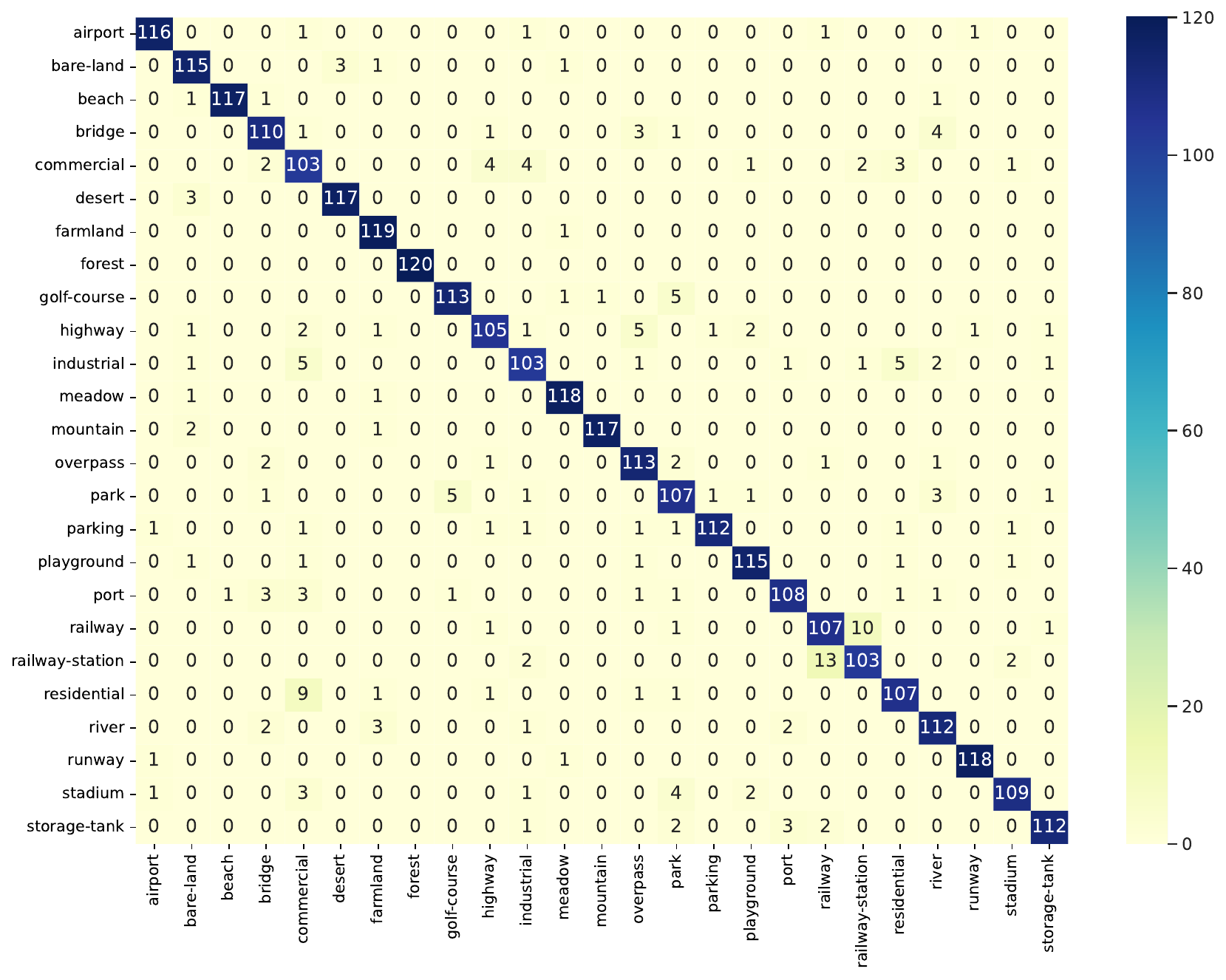}
  \caption{Confusion matrix for the pre-trained Vision Transformer model on the CLRS dataset.}
  \label{fig:clrs_confusionmatrix}
\end{figure}

Detailed results for all pre-trained models are shown on Table~\ref{tab:pre-trained_clrs} and for all the models learned from scratch are presented on Table~\ref{tab:scratch_clrs}. The best performing model is the pre-trained Vision Transformer model. The results on a class level are show on Table~\ref{tab:perclass_clrs} along with a confusion matrix on Figure~\ref{fig:clrs_confusionmatrix}.

\begin{table}[ht]\centering
\caption{Detailed results for pre-trained models on the CLRS dataset.}\label{tab:pre-trained_clrs}
\scriptsize \begin{adjustbox}{width=0.75\linewidth}
\begin{tabular}{lrrrrrrrrrrr}\toprule
Model \textbackslash Metric &\rotatebox{90}{Accuracy} &\rotatebox{90}{Macro Precision} &\rotatebox{90}{Weighted Precision} &\rotatebox{90}{Macro Recall} &\rotatebox{90}{Weighted Recall} &\rotatebox{90}{Macro F1 score} &\rotatebox{90}{Weighted F1 score} &\rotatebox{90}{Avg. time / epoch (sec.)} &\rotatebox{90}{Total time (sec.)} &\rotatebox{90}{Best epoch}\\\midrule
AlexNet &84.10 &84.19 &84.19 &84.10 &84.10 &84.03 &84.03 &20.48 &635 &21\\
VGG16 &89.90 &89.97 &89.97 &89.90 &89.90 &89.90 &89.90 &20.23 &607 &20\\
ResNet50 &91.57 &91.67 &91.67 &91.57 &91.57 &91.58 &91.58 &18.60 &279 &15\\
RestNet152 &91.90 &91.99 &91.99 &91.90 &91.90 &91.91 &91.91 &31.96 &799 &15\\
DenseNet161 &92.20 &92.29 &92.29 &92.20 &92.20 &92.20 &92.20 &35.46 &993 &18\\
EfficientNetB0 &90.50 &90.61 &90.61 &90.50 &90.50 &90.49 &90.49 &19.73 &513 &16\\
ConvNeXt &91.10 &91.29 &91.29 &91.10 &91.10 &91.12 &91.12 &23.62 &496 &11\\
Vision Transformer &\textbf{93.20} &93.29 &93.29 &93.20 &93.20 &93.22 &93.22 &25.32 &785 &21\\
MLP Mixer &90.10 &90.21 &90.21 &90.10 &90.10 &90.05 &90.05 &19.75 &316 &6\\
Swin Transformer &92.53 &92.53 &92.53 &92.53 &92.53 &92.51 &92.51 &68.93 &1861 &17 \\
\bottomrule
\end{tabular} \end{adjustbox}
\end{table}

\begin{table}[ht]\centering
\caption{Detailed results for models trained from scratch on the CLRS dataset.}\label{tab:scratch_clrs}
\scriptsize \begin{adjustbox}{width=0.75\linewidth}
\begin{tabular}{lrrrrrrrrrrr}\toprule
Model \textbackslash Metric &\rotatebox{90}{Accuracy} &\rotatebox{90}{Macro Precision} &\rotatebox{90}{Weighted Precision} &\rotatebox{90}{Macro Recall} &\rotatebox{90}{Weighted Recall} &\rotatebox{90}{Macro F1 score} &\rotatebox{90}{Weighted F1 score} &\rotatebox{90}{Avg. time / epoch (sec.)} &\rotatebox{90}{Total time (sec.)} &\rotatebox{90}{Best epoch}\\\midrule
AlexNet &71.40 &71.59 &71.59 &71.40 &71.40 &71.33 &71.33 &20.35 &2035 &92\\
VGG16 &76.07 &76.20 &76.20 &76.07 &76.07 &76.00 &76.00 &19.33 &1450 &60\\
ResNet50 &85.57 &85.72 &85.72 &85.57 &85.57 &85.57 &85.57 &19.43 &1788 &77\\
RestNet152 &82.30 &82.47 &82.47 &82.30 &82.30 &82.19 &82.19 &32.05 &2373 &60\\
DenseNet161 &\textbf{86.17} &86.29 &86.29 &86.17 &86.17 &86.18 &86.18 &35.81 &2757 &62\\
EfficientNetB0 &82.27 &82.55 &82.55 &82.27 &82.27 &82.31 &82.31 &20.71 &1512 &58\\
ConvNeXt &69.17 &69.02 &69.02 &69.17 &69.17 &69.01 &69.01 &23.09 &2309 &96\\
Vision Transformer &65.47 &66.41 &66.41 &65.47 &65.47 &65.49 &65.49 &24.96 &1173 &32\\
MLP Mixer &61.13 &62.18 &62.18 &61.13 &61.13 &60.87 &60.87 &17.98 &809 &30\\
Swin Transformer &80.00 &80.10 &80.10 &80.00 &80.00 &79.91 &79.91 &69.19 &5535 &65 \\
\bottomrule
\end{tabular} \end{adjustbox}
\end{table}

\begin{table}[ht]\centering
\caption{Per class results for the pre-trained Vision Transformer model on the CLRS dataset.}\label{tab:perclass_clrs}
\scriptsize 
\begin{tabular}{lrrrr}\toprule
Label &Precision &Recall &F1 score \\\midrule
airport &97.48 &96.67 &97.07\\
bare-land &92.00 &95.83 &93.88\\
beach &99.15 &97.50 &98.32\\
bridge &90.91 &91.67 &91.29\\
commercial &79.84 &85.83 &82.73\\
desert &97.50 &97.50 &97.50\\
farmland &93.70 &99.17 &96.36\\
forest &100.00 &100.00 &100.00\\
golf-course &94.96 &94.17 &94.56\\
highway &92.11 &87.50 &89.74\\
industrial &88.79 &85.83 &87.29\\
meadow &96.72 &98.33 &97.52\\
mountain &99.15 &97.50 &98.32\\
overpass &89.68 &94.17 &91.87\\
park &85.60 &89.17 &87.35\\
parking &98.25 &93.33 &95.73\\
playground &95.04 &95.83 &95.44\\
port &94.74 &90.00 &92.31\\
railway &86.29 &89.17 &87.70\\
railway-station &88.79 &85.83 &87.29\\
residential &90.68 &89.17 &89.92\\
river &90.32 &93.33 &91.80\\
runway &98.33 &98.33 &98.33\\
stadium &95.61 &90.83 &93.16\\
storage-tank &96.55 &93.33 &94.92\\
\bottomrule
\end{tabular} 
\end{table}

\clearpage

\subsection{RSD46-WHU}
\begin{figure}[ht]
\centering
  \includegraphics[width=0.7\linewidth]{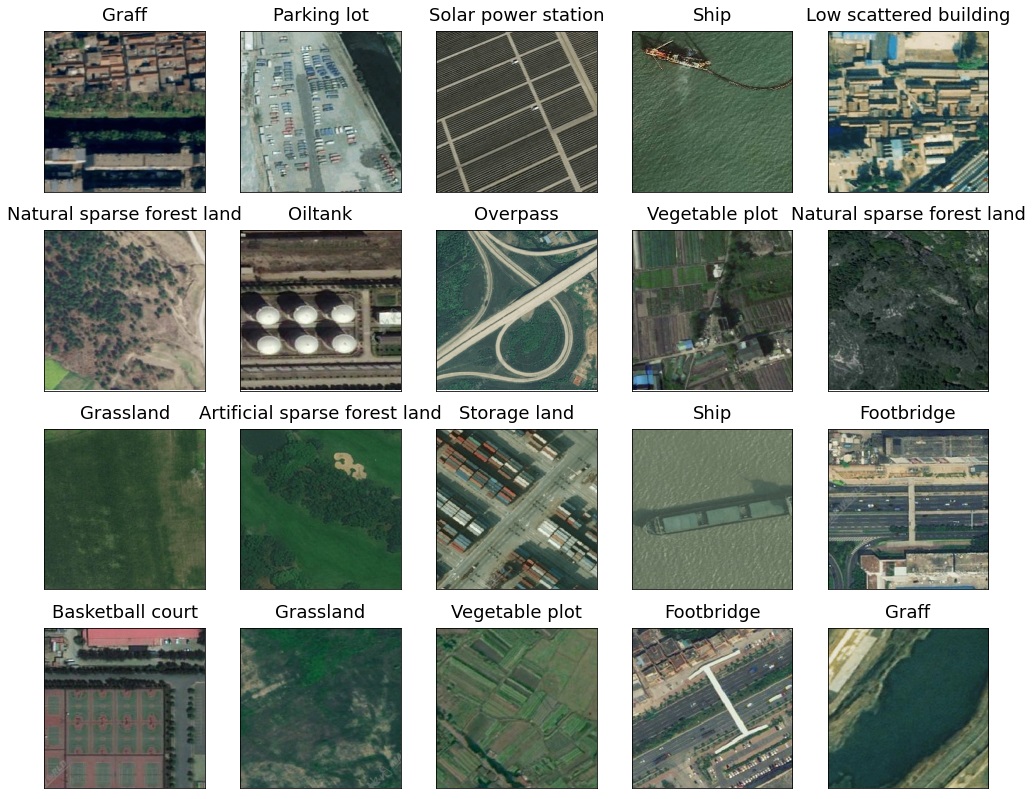}
  \caption{Example images with labels from the RSD46-WHU dataset.}
  \label{fig:rsd46whu_samples}
\end{figure}

\begin{figure}[ht]
  \centering
  \includegraphics[width=0.7\linewidth]{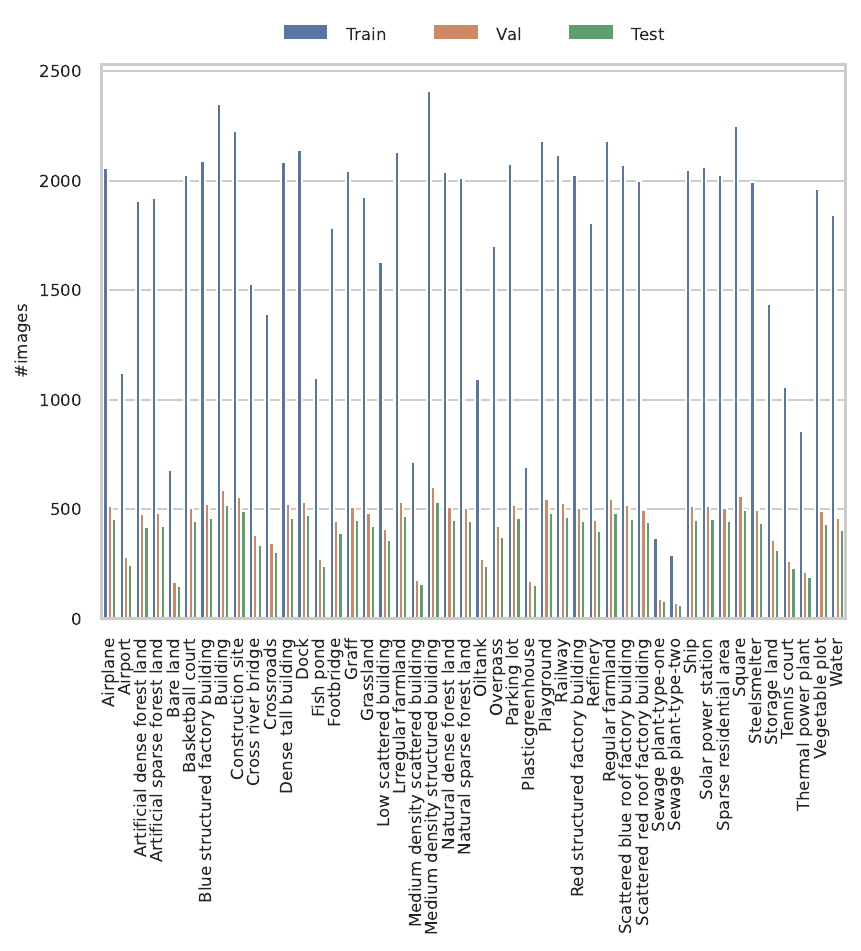}
  \caption{Class distribution for the RSD46-WHU dataset.}
  \label{fig:rsd46whu_distribution}
\end{figure}

RSD46-WHU is a large-scale open dataset for scene classification in remote sensing images. The dataset is manually collected from Google Earth and Tianditu. The ground resolution of most classes is 0.5m, and the others are about 2m. There are 500-3000 images in each class. The RSD46-WHU dataset contains around 117000 images with 46 classes (Figure~\ref{fig:rsd46whu_samples}). The image are not evenly distributed between classes and each class contains between 428 to 3000 images. The dataset comes with predefined train and test splits. For creating the validation split we used 20\% of the images from the train split. The class distribution of the different splits is presented on Figure~\ref{fig:rsd46whu_distribution}.

\begin{figure}[ht]
  \centering
  \includegraphics[width=0.9\linewidth]{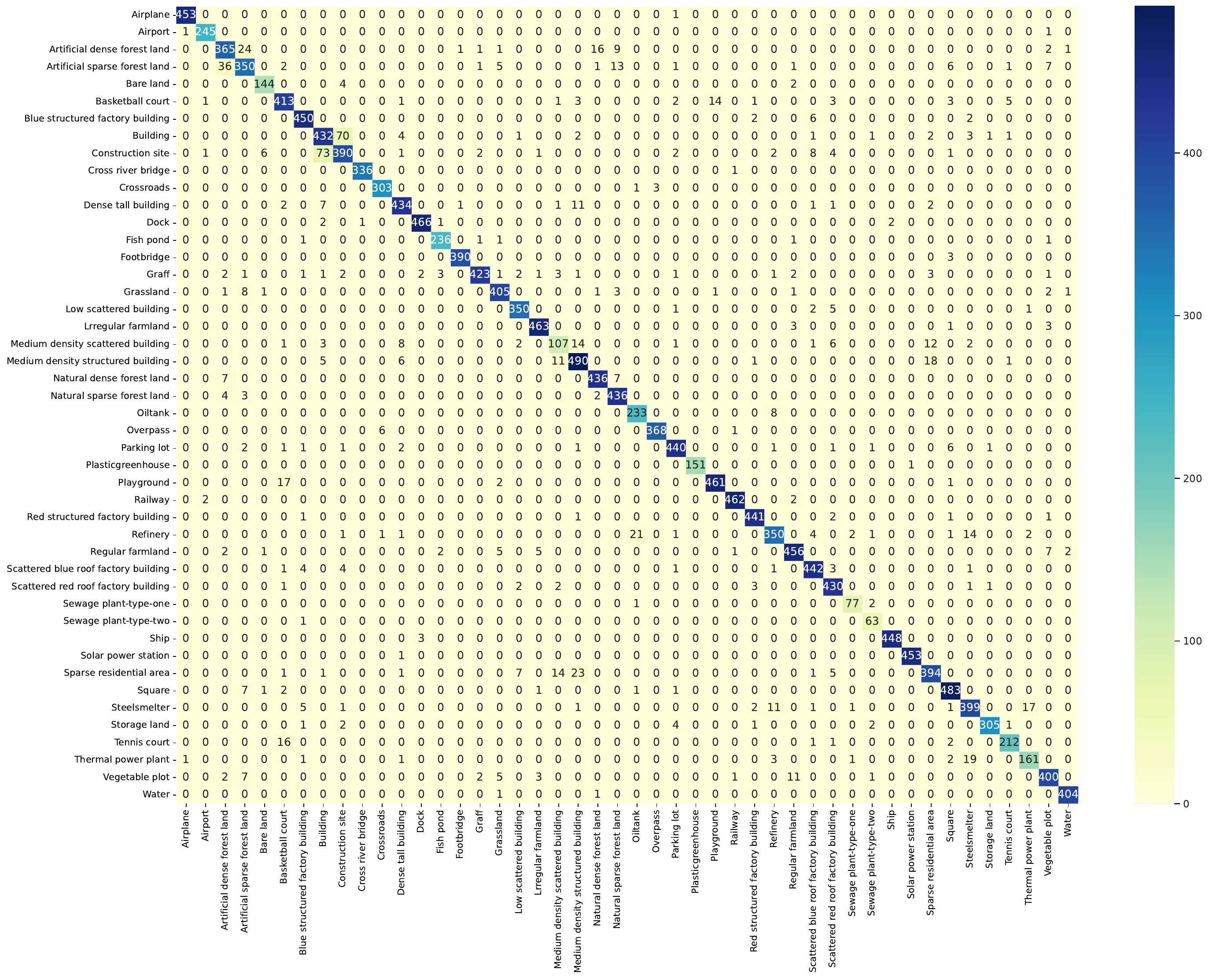}
  \caption{Confusion matrix for the pre-trained DenseNet161 model on the RSD46-WHU dataset.}
  \label{fig:rsd46whu_confusionmatrix}
\end{figure}

Detailed results for all pre-trained models are shown on Table~\ref{tab:pre-trained_rsd46whu} and for all the models learned from scratch are presented on Table~\ref{tab:scratch_rsd46whu}. The best performing model is the pre-trained DenseNet161 model. The results on a class level are show on Table~\ref{tab:perclass_rsd46whu} along with a confusion matrix on Figure~\ref{fig:rsd46whu_confusionmatrix}.

\begin{table}[ht]\centering
\caption{Detailed results for pre-trained models on the RSD46-WHU dataset.}\label{tab:pre-trained_rsd46whu}
\scriptsize \begin{adjustbox}{width=0.75\linewidth}
\begin{tabular}{lrrrrrrrrrrr}\toprule
Model \textbackslash Metric &\rotatebox{90}{Accuracy} &\rotatebox{90}{Macro Precision} &\rotatebox{90}{Weighted Precision} &\rotatebox{90}{Macro Recall} &\rotatebox{90}{Weighted Recall} &\rotatebox{90}{Macro F1 score} &\rotatebox{90}{Weighted F1 score} &\rotatebox{90}{Avg. time / epoch (sec.)} &\rotatebox{90}{Total time (sec.)} &\rotatebox{90}{Best epoch}\\\midrule
AlexNet &90.65 &90.43 &90.61 &90.35 &90.65 &90.36 &90.61 &58.03 &2031 &25\\
VGG16 &92.42 &92.30 &92.38 &92.25 &92.42 &92.22 &92.37 &158.32 &4433 &18\\
ResNet50 &94.16 &94.07 &94.15 &94.18 &94.16 &94.11 &94.14 &123.27 &3205 &16\\
RestNet152 &94.40 &94.33 &94.40 &94.41 &94.40 &94.36 &94.39 &269.45 &7814 &19\\
DenseNet161 &\textbf{94.51} &94.36 &94.49 &94.41 &94.51 &94.36 &94.48 &297.70 &6847 &13\\
EfficientNetB0 &93.39 &93.20 &93.38 &93.39 &93.39 &93.26 &93.35 &111.55 &2231 &10\\
ConvNeXt &93.63 &93.61 &93.67 &93.47 &93.63 &93.48 &93.60 &196.20 &3924 &10\\
Vision Transformer &94.24 &94.38 &94.23 &94.08 &94.24 &94.16 &94.20 &210.37 &3997 &9\\
MLP Mixer &93.67 &93.77 &93.69 &93.47 &93.67 &93.55 &93.65 &148.25 &3558 &14\\
Swin Transformer &93.54 &93.48 &93.62 &93.62 &93.54 &93.50 &93.52 &599.42 &11389 &9 \\
\bottomrule
\end{tabular} \end{adjustbox}
\end{table}

\begin{table}[ht]\centering
\caption{Detailed results for models trained from scratch on the RSD46-WHU dataset.}\label{tab:scratch_rsd46whu}
\scriptsize \begin{adjustbox}{width=0.75\linewidth}
\begin{tabular}{lrrrrrrrrrrr}\toprule
Model \textbackslash Metric &\rotatebox{90}{Accuracy} &\rotatebox{90}{Macro Precision} &\rotatebox{90}{Weighted Precision} &\rotatebox{90}{Macro Recall} &\rotatebox{90}{Weighted Recall} &\rotatebox{90}{Macro F1 score} &\rotatebox{90}{Weighted F1 score} &\rotatebox{90}{Avg. time / epoch (sec.)} &\rotatebox{90}{Total time (sec.)} &\rotatebox{90}{Best epoch}\\\midrule
AlexNet &86.03 &85.83 &86.03 &85.67 &86.03 &85.71 &85.99 &58.84 &3707 &48\\
VGG16 &88.62 &88.37 &88.56 &88.37 &88.62 &88.32 &88.55 &162.89 &8796 &39\\
ResNet50 &90.55 &90.40 &90.53 &90.26 &90.55 &90.30 &90.52 &127.53 &8672 &53\\
RestNet152 &89.94 &89.84 &89.99 &89.77 &89.94 &89.78 &89.95 &272.70 &19907 &58\\
DenseNet161 &\textbf{92.21} &92.11 &92.23 &92.03 &92.21 &92.06 &92.21 &301.16 &15318 &36\\
EfficientNetB0 &90.61 &90.57 &90.61 &90.25 &90.61 &90.37 &90.58 &113.93 &6446 &40\\
ConvNeXt &88.69 &88.66 &88.67 &88.33 &88.69 &88.46 &88.66 &194.93 &11891 &46\\
Vision Transformer &86.47 &86.22 &86.45 &85.94 &86.47 &86.02 &86.42 &211.93 &9325 &29\\
MLP Mixer &81.25 &81.56 &81.59 &80.11 &81.25 &80.51 &81.19 &148.42 &4149 &12\\
Swin Transformer &91.81 &91.50 &91.79 &91.48 &91.81 &91.47 &91.79 &588.25 &41766 &56 \\
\bottomrule
\end{tabular} \end{adjustbox}
\end{table}

\begin{table}[ht]\centering
\caption{Per class results for the pre-trained DenseNet161 model on the RSD46-WHU dataset.}\label{tab:perclass_rsd46whu}
\scriptsize 
\begin{tabular}{lrrrr}\toprule
Label &Precision &Recall &F1 score \\\midrule
Airplane &99.56 &99.78 &99.67\\
Airport &98.39 &99.19 &98.79\\
Artificial dense forest land &87.11 &86.90 &87.01\\
Artificial sparse forest land &87.06 &82.55 &84.75\\
Bare land &94.12 &96.00 &95.05\\
Basketball court &90.37 &92.39 &91.37\\
Blue structured factory building &96.57 &97.83 &97.19\\
Building &82.44 &83.40 &82.92\\
Construction site &82.11 &79.43 &80.75\\
Cross river bridge &99.70 &99.70 &99.70\\
Crossroads &97.74 &98.70 &98.22\\
Dense tall building &94.35 &94.35 &94.35\\
Dock &98.94 &98.73 &98.83\\
Fish pond &97.52 &97.93 &97.72\\
Footbridge &99.49 &99.24 &99.36\\
Graff &98.37 &93.79 &96.03\\
Grassland &95.07 &95.52 &95.29\\
Low scattered building &96.15 &97.49 &96.82\\
Lrregular farmland &97.68 &98.51 &98.09\\
Medium density scattered building &76.98 &68.15 &72.30\\
Medium density structured building &89.58 &92.11 &90.82\\
Natural dense forest land &95.40 &96.89 &96.14\\
Natural sparse forest land &93.16 &97.98 &95.51\\
Oiltank &90.66 &96.68 &93.57\\
Overpass &99.19 &98.13 &98.66\\
Parking lot &96.49 &96.07 &96.28\\
Plasticgreenhouse &100.00 &99.34 &99.67\\
Playground &96.85 &95.84 &96.34\\
Railway &99.14 &99.14 &99.14\\
Red structured factory building &97.78 &98.66 &98.22\\
Refinery &92.84 &87.72 &90.21\\
Regular farmland &95.20 &94.80 &95.00\\
Scattered blue roof factory building &94.44 &96.72 &95.57\\
Scattered red roof factory building &93.28 &97.73 &95.45\\
Sewage plant-type-one &95.06 &96.25 &95.65\\
Sewage plant-type-two &88.73 &98.44 &93.33\\
Ship &99.56 &99.33 &99.45\\
Solar power station &99.78 &99.78 &99.78\\
Sparse residential area &91.42 &88.14 &89.75\\
Square &94.52 &97.38 &95.93\\
Steelsmelter &90.48 &90.89 &90.68\\
Storage land &99.03 &96.52 &97.76\\
Tennis court &95.93 &91.38 &93.60\\
Thermal power plant &88.95 &85.19 &87.03\\
Vegetable plot &94.12 &92.59 &93.35\\
Water &99.02 &99.51 &99.26\\
\bottomrule
\end{tabular} 
\end{table}

\clearpage

\subsection{Brazilian Coffee Scenes}
\begin{figure}[ht]
\centering
  \includegraphics[width=0.5\linewidth]{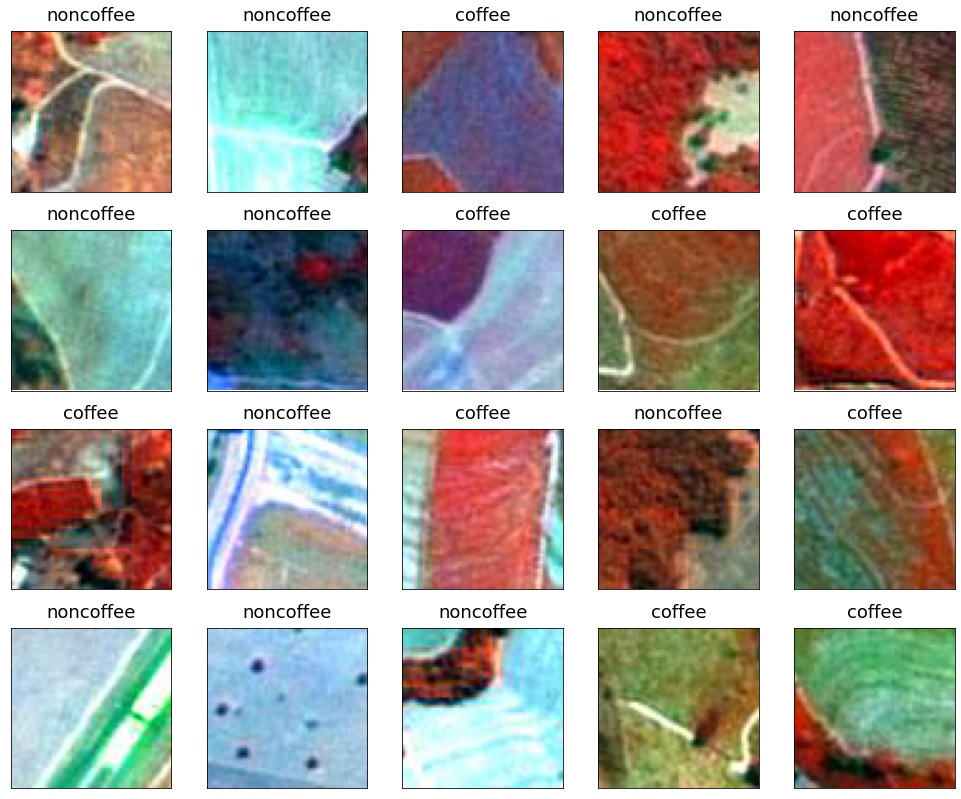}
  \caption{Example images with labels from the Brazilian Coffee Scenes dataset.}
  \label{fig:bcs_samples}
\end{figure}

\begin{figure}[ht]
  \centering
  \includegraphics[width=0.35\linewidth]{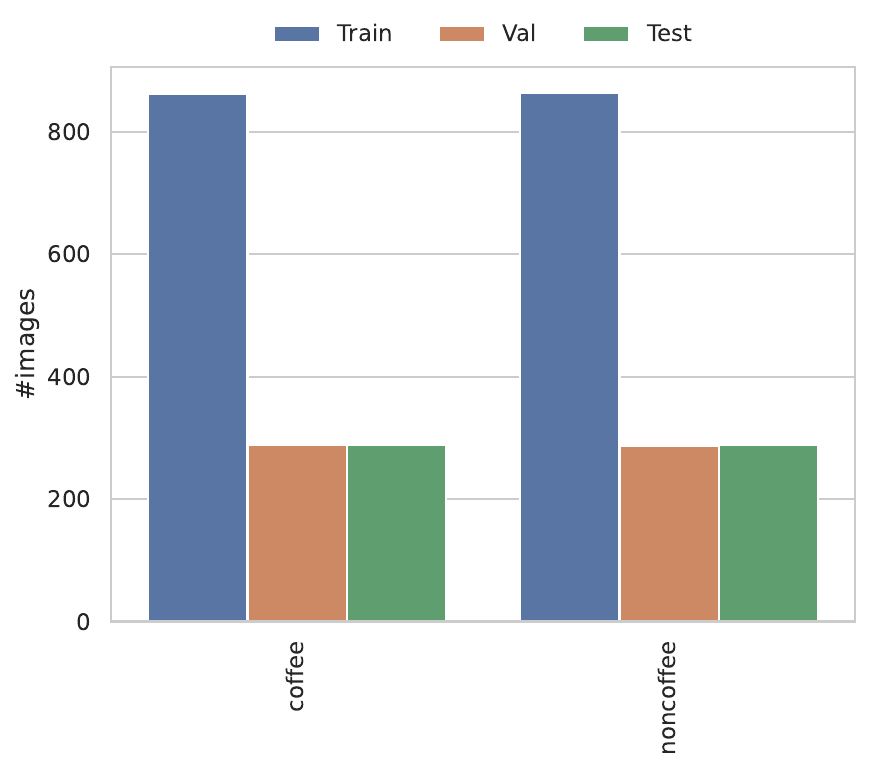}
  \caption{Class distribution for the Brazilian Coffee Scenes dataset.}
  \label{fig:bcs_distribution}
\end{figure}

The Brazilian Coffee Scenes dataset \citep{penatti2015deep} consists of only two classes: coffee and non-coffee class. Each class has 1438 images with 64x64 pixels cropped from SPOT satellite images over four counties in the state of Minas Gerais, Brazil: Arceburgo, Guaranesia, Guaxupe, and Monte Santo (Figure~\ref{fig:bcs_samples}). The images in the dataset are in green, red and near-infrared spectral bands, since these are most useful and representative for distinguishing vegetation areas. The dataset is manually annotated by agricultural researchers. Images which contain coffee pixels in at least 85\% of the image were assigned to the coffee class. Image with less than 10\% of coffee pixels are assigned to the non-coffee class. The number of classes and the degree to which the data is tailored, should make this less challenging dataset. The class distribution is presented on Figure~\ref{fig:bcs_distribution}.

\begin{figure}[ht]
  \centering
  \includegraphics[width=.5\linewidth]{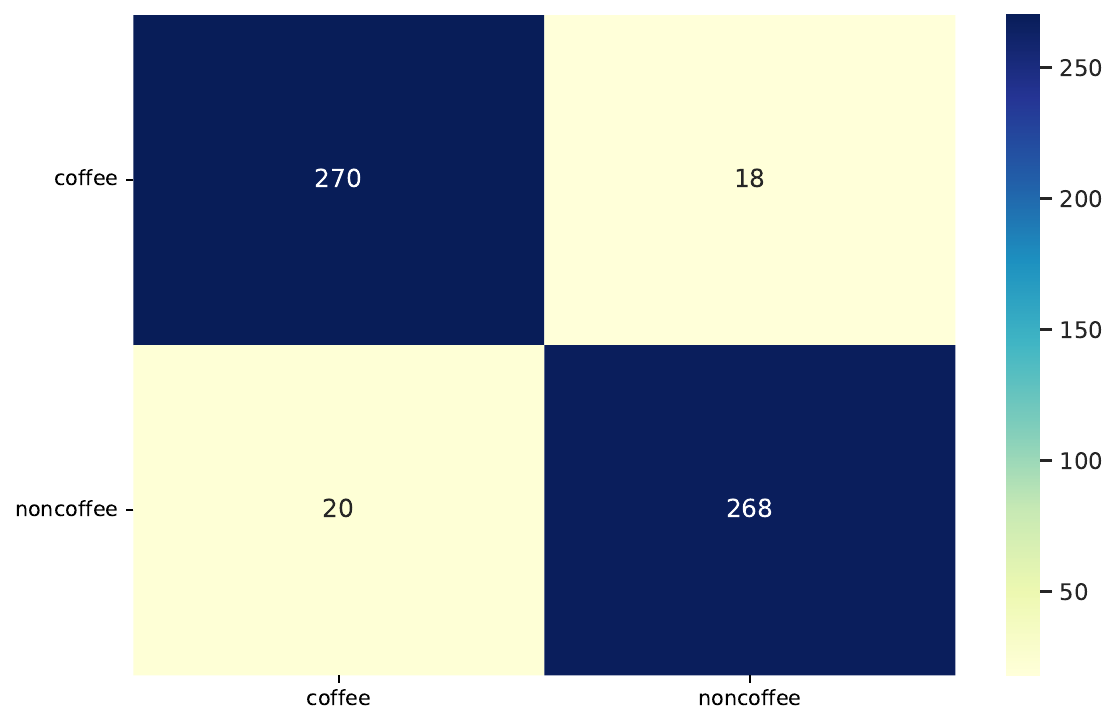}
  \caption{Confusion matrix for Swin Transformer on the Brazilian Coffee Scenes dataset.}
  \label{fig:bcs_confusionmatrix}
\end{figure}

Detailed results for all pre-trained models are shown on Table~\ref{tab:pre-trained_bcs} and for all the models learned from scratch are presented on Table~\ref{tab:scratch_bcs}. The best performing model is the pre-trained Swin Transformer model. The results on a class level are show on Table~\ref{tab:perclass_bcs} along with a confusion matrix on Figure~\ref{fig:bcs_confusionmatrix}.

\begin{table}[ht]\centering
\caption{Detailed results for pre-trained models on the Brazilian Coffee Scenes dataset.}\label{tab:pre-trained_bcs}
\scriptsize \begin{adjustbox}{width=0.75\linewidth}
\begin{tabular}{lrrrrrrrrrrr}\toprule
Model \textbackslash Metric &\rotatebox{90}{Accuracy} &\rotatebox{90}{Macro Precision} &\rotatebox{90}{Weighted Precision} &\rotatebox{90}{Macro Recall} &\rotatebox{90}{Weighted Recall} &\rotatebox{90}{Macro F1 score} &\rotatebox{90}{Weighted F1 score} &\rotatebox{90}{Avg. time / epoch (sec.)} &\rotatebox{90}{Total time (sec.)} &\rotatebox{90}{Best epoch}\\\midrule
AlexNet &89.58 &89.59 &89.59 &89.58 &89.58 &89.58 &89.58 &1.48 &43 &19\\
VGG16 &90.97 &91.00 &91.00 &90.97 &90.97 &90.97 &90.97 &4.17 &121 &19\\
ResNet50 &92.01 &92.06 &92.06 &92.01 &92.01 &92.01 &92.01 &3.45 &76 &12\\
RestNet152 &92.36 &92.37 &92.37 &92.36 &92.36 &92.36 &92.36 &6.61 &119 &8\\
DenseNet161 &92.71 &92.81 &92.81 &92.71 &92.71 &92.70 &92.70 &7.33 &176 &14\\
EfficientNetB0 &91.32 &91.32 &91.32 &91.32 &91.32 &91.32 &91.32 &3.17 &133 &32\\
ConvNeXt &91.49 &91.58 &91.58 &91.49 &91.49 &91.49 &91.49 &5.08 &132 &16\\
Vision Transformer &92.01 &92.03 &92.03 &92.01 &92.01 &92.01 &92.01 &5.07 &76 &5\\
MLP Mixer &93.06 &93.07 &93.07 &93.06 &93.06 &93.05 &93.05 &3.94 &67 &7\\
Swin Transformer &\textbf{93.40} &93.40 &93.40 &93.40 &93.40 &93.40 &93.40 &13.88 &222 &6 \\
\bottomrule
\end{tabular} \end{adjustbox}
\end{table}

\begin{table}[ht]\centering
\caption{Detailed results for models trained from scratch on the Brazilian Coffee Scenes dataset.}\label{tab:scratch_bcs}
\scriptsize \begin{adjustbox}{width=0.75\linewidth}
\begin{tabular}{lrrrrrrrrrrr}\toprule
Model \textbackslash Metric &\rotatebox{90}{Accuracy} &\rotatebox{90}{Macro Precision} &\rotatebox{90}{Weighted Precision} &\rotatebox{90}{Macro Recall} &\rotatebox{90}{Weighted Recall} &\rotatebox{90}{Macro F1 score} &\rotatebox{90}{Weighted F1 score} &\rotatebox{90}{Avg. time / epoch (sec.)} &\rotatebox{90}{Total time (sec.)} &\rotatebox{90}{Best epoch}\\\midrule
AlexNet &89.41 &89.62 &89.62 &89.41 &89.41 &89.40 &89.40 &1.53 &115 &60\\
VGG16 &89.41 &89.45 &89.45 &89.41 &89.41 &89.41 &89.41 &5.95 &440 &59\\
ResNet50 &89.24 &89.39 &89.39 &89.24 &89.24 &89.23 &89.23 &4.55 &296 &50\\
RestNet152 &88.54 &88.56 &88.56 &88.54 &88.54 &88.54 &88.54 &7.95 &469 &44\\
DenseNet161 &\textbf{90.80} &90.80 &90.80 &90.80 &90.80 &90.80 &90.80 &7.31 &373 &36\\
EfficientNetB0 &85.42 &85.71 &85.71 &85.42 &85.42 &85.39 &85.39 &3.26 &326 &98\\
ConvNeXt &84.38 &84.39 &84.39 &84.38 &84.38 &84.37 &84.37 &5.09 &509 &95\\
Vision Transformer &87.85 &87.89 &87.89 &87.85 &87.85 &87.84 &87.84 &5.55 &322 &43\\
MLP Mixer &86.28 &86.29 &86.29 &86.28 &86.28 &86.28 &86.28 &4.47 &201 &30\\
Swin Transformer &89.24 &89.33 &89.33 &89.24 &89.24 &89.23 &89.23 &13.59 &1169 &71 \\
\bottomrule
\end{tabular} \end{adjustbox}
\end{table}

\begin{table}[ht]\centering
\caption{Per class results for Swin Transformer on the Brazilian Coffee Scenes dataset.}\label{tab:perclass_bcs}
\scriptsize 
\begin{tabular}{lrrrr}\toprule
Label &Precision &Recall &F1 score \\\midrule
coffee &93.10 &93.75 &93.43 \\
noncoffee &93.71 &93.06 &93.38 \\
\bottomrule
\end{tabular} 
\end{table}

\clearpage

\subsection{Optimal 31}
\begin{figure}[ht]
\centering
  \includegraphics[width=0.7\linewidth]{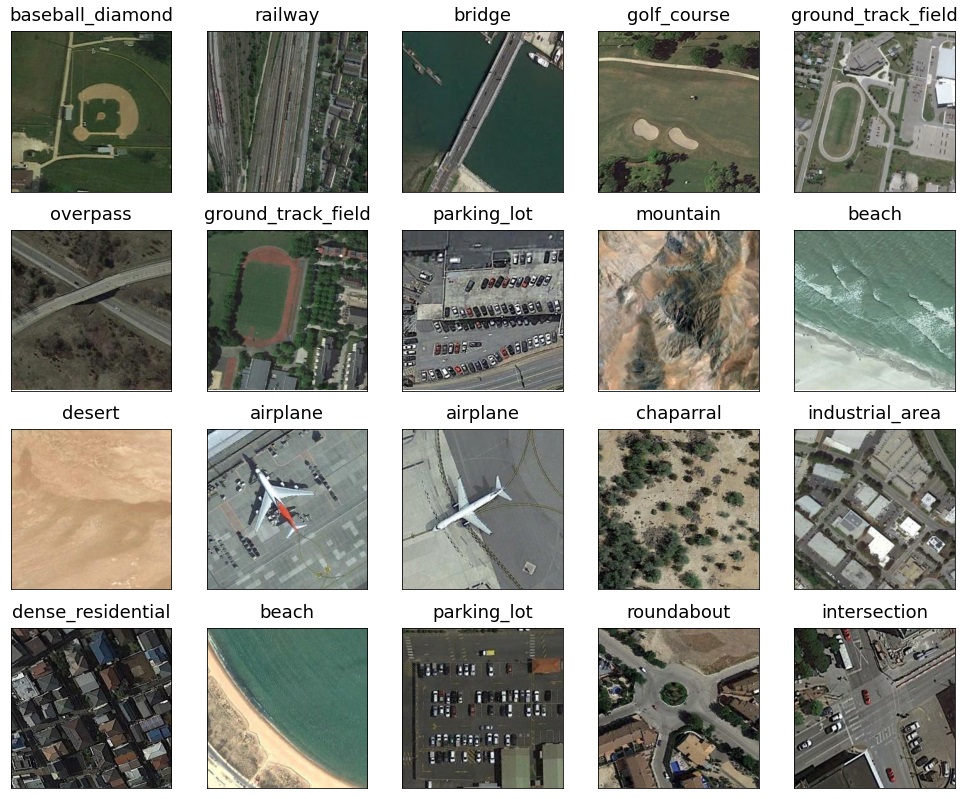}
  \caption{Example images with labels from the Optimal 31 dataset.}
  \label{fig:optimal31_samples}
\end{figure}

\begin{figure}[ht]
  \centering
  \includegraphics[width=0.6\linewidth]{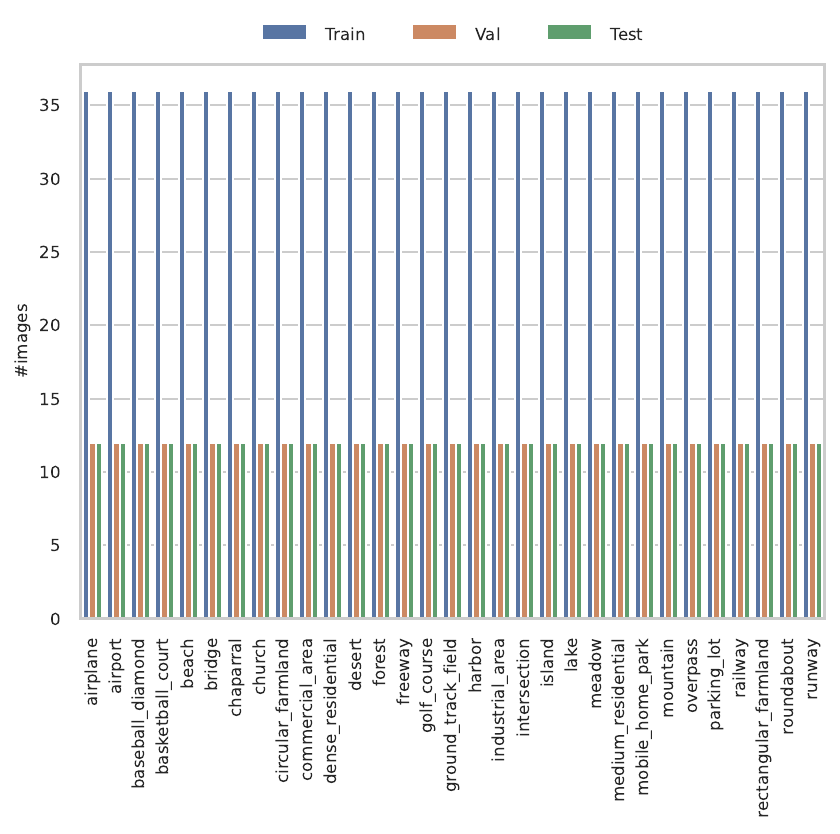}
  \caption{Class distribution for the Optimal 31 dataset.}
  \label{fig:optimal31_distribution}
\end{figure}

The Optimal 31 dataset \citep{qi2019optimal31} is for remote sensing image scene classification. The dataset contains 31 classes, each class contains 60 images with a size of 256×256 pixels. Totaling 1860 aerial RGB images (Figure~\ref{fig:optimal31_samples}). These classes include: airplane, airport, basketball court, baseball field, bridge, beach, bushes, crossroads, church, round farmland, business district, desert, harbor, dense houses, factory, forest, freeway, golf field, island, lake, meadow, medium houses, mountain, mobile house area, overpass, playground, parking lot, roundabout, runway, railway, and square farmland. It is considered challenging due to small number of images dispersed across many classes. We have generated train, test and validation spits for our study and their class distribution is presented on Figure~\ref{fig:optimal31_distribution}.

\begin{figure}[ht]
  \centering
  \includegraphics[width=0.9\linewidth]{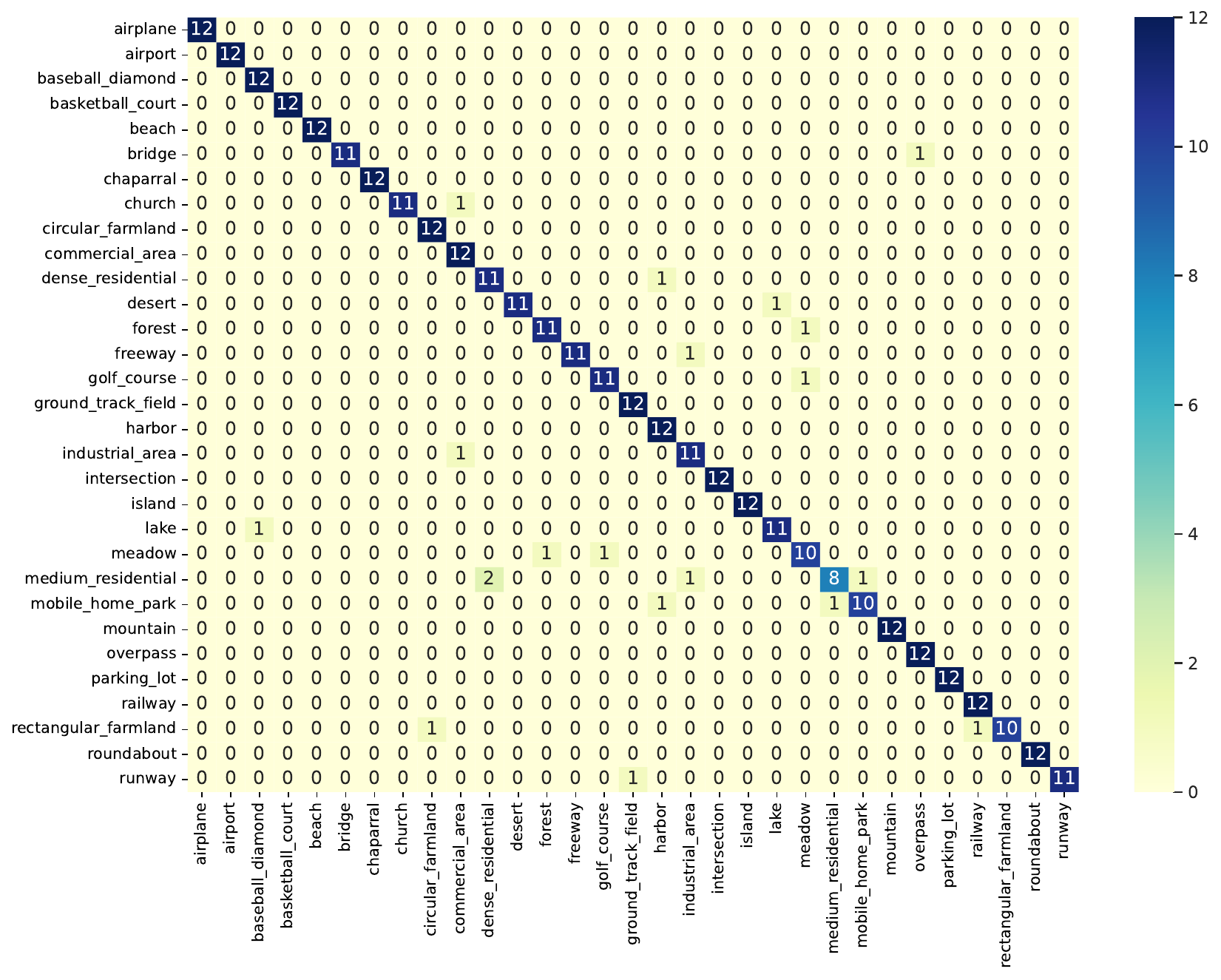}
  \caption{Confusion matrix for the pre-trained Vision Transformer model on the Optimal 31 dataset.}
  \label{fig:optimal31_confusionmatrix}
\end{figure}

Detailed results for all pre-trained models are shown on Table~\ref{tab:pre-trained_optimal31} and for all the models learned from scratch are presented on Table~\ref{tab:scratch_optimal31}. The best performing model is the pre-trained Vision Transformer model. The results on a class level are show on Table~\ref{tab:perclass_optimal31} along with a confusion matrix on Figure~\ref{fig:optimal31_confusionmatrix}.

\begin{table}[ht]\centering
\caption{Detailed results for pre-trained models on the Optimal 31 dataset.}\label{tab:pre-trained_optimal31}
\scriptsize \begin{adjustbox}{width=0.75\linewidth}
\begin{tabular}{lrrrrrrrrrrr}\toprule
Model \textbackslash Metric &\rotatebox{90}{Accuracy} &\rotatebox{90}{Macro Precision} &\rotatebox{90}{Weighted Precision} &\rotatebox{90}{Macro Recall} &\rotatebox{90}{Weighted Recall} &\rotatebox{90}{Macro F1 score} &\rotatebox{90}{Weighted F1 score} &\rotatebox{90}{Avg. time / epoch (sec.)} &\rotatebox{90}{Total time (sec.)} &\rotatebox{90}{Best epoch}\\\midrule
AlexNet &80.91 &81.90 &81.90 &80.91 &80.91 &80.74 &80.74 &1.10 &45 &31\\
VGG16 &88.71 &89.58 &89.58 &88.71 &88.71 &88.79 &88.79 &2.97 &95 &22\\
ResNet50 &92.20 &92.85 &92.85 &92.20 &92.20 &92.25 &92.25 &2.58 &129 &40\\
RestNet152 &92.47 &92.99 &92.99 &92.47 &92.47 &92.47 &92.47 &4.62 &217 &37\\
DenseNet161 &94.35 &94.92 &94.92 &94.35 &94.35 &94.43 &94.43 &5.02 &306 &51\\
EfficientNetB0 &91.67 &92.04 &92.04 &91.67 &91.67 &91.60 &91.60 &2.25 &187 &73\\
ConvNeXt &93.01 &93.33 &93.33 &93.01 &93.01 &92.99 &92.99 &3.50 &203 &48\\
Vision Transformer &\textbf{94.62} &94.85 &94.85 &94.62 &94.62 &94.56 &94.56 &3.71 &126 &24\\
MLP Mixer &92.74 &93.17 &93.17 &92.74 &92.74 &92.74 &92.74 &2.82 &141 &40\\
Swin Transformer &92.47 &92.92 &92.92 &92.47 &92.47 &92.51 &92.51 &9.19 &340 &27 \\
\bottomrule
\end{tabular} \end{adjustbox}
\end{table}

\begin{table}[ht]\centering
\caption{Detailed results for models trained from scratch on the Optimal 31 dataset.}\label{tab:scratch_optimal31}
\scriptsize \begin{adjustbox}{width=0.75\linewidth}
\begin{tabular}{lrrrrrrrrrrr}\toprule
Model \textbackslash Metric &\rotatebox{90}{Accuracy} &\rotatebox{90}{Macro Precision} &\rotatebox{90}{Weighted Precision} &\rotatebox{90}{Macro Recall} &\rotatebox{90}{Weighted Recall} &\rotatebox{90}{Macro F1 score} &\rotatebox{90}{Weighted F1 score} &\rotatebox{90}{Avg. time / epoch (sec.)} &\rotatebox{90}{Total time (sec.)} &\rotatebox{90}{Best epoch}\\\midrule
AlexNet &55.11 &55.61 &55.61 &55.11 &55.11 &54.24 &54.24 &1.23 &101 &67\\
VGG16 &56.72 &58.89 &58.89 &56.72 &56.72 &56.58 &56.58 &4.81 &409 &70\\
ResNet50 &67.20 &69.56 &69.56 &67.20 &67.20 &67.17 &67.17 &2.60 &161 &47\\
RestNet152 &62.90 &64.95 &64.95 &62.90 &62.90 &62.78 &62.78 &5.92 &314 &38\\
DenseNet161 &\textbf{71.24} &72.01 &72.01 &71.24 &71.24 &70.65 &70.65 &5.16 &330 &49\\
EfficientNetB0 &68.55 &70.59 &70.59 &68.55 &68.55 &68.70 &68.70 &2.36 &156 &51\\
ConvNeXt &58.87 &60.69 &60.69 &58.87 &58.87 &58.92 &58.92 &3.59 &330 &77\\
Vision Transformer &62.63 &63.89 &63.89 &62.63 &62.63 &62.32 &62.32 &3.79 &235 &47\\
MLP Mixer &59.14 &60.36 &60.36 &59.14 &59.14 &58.47 &58.47 &3.26 &326 &98\\
Swin Transformer &66.13 &67.47 &67.47 &66.13 &66.13 &65.62 &65.62 &9.51 &951 &89 \\
\bottomrule
\end{tabular} \end{adjustbox}
\end{table}

\begin{table}[ht]\centering
\caption{Per class results for the pre-trained Vision Transformer model on the Optimal 31 dataset.}\label{tab:perclass_optimal31}
\scriptsize 
\begin{tabular}{lrrrr}\toprule
Label &Precision &Recall &F1 score \\\midrule
airplane &100.00 &100.00 &100.00\\
airport &100.00 &100.00 &100.00\\
baseball\_diamond &92.31 &100.00 &96.00\\
basketball\_court &100.00 &100.00 &100.00\\
beach &100.00 &100.00 &100.00\\
bridge &100.00 &91.67 &95.65\\
chaparral &100.00 &100.00 &100.00\\
church &100.00 &91.67 &95.65\\
circular\_farmland &92.31 &100.00 &96.00\\
commercial\_area &85.71 &100.00 &92.31\\
dense\_residential &84.62 &91.67 &88.00\\
desert &100.00 &91.67 &95.65\\
forest &91.67 &91.67 &91.67\\
freeway &100.00 &91.67 &95.65\\
golf\_course &91.67 &91.67 &91.67\\
ground\_track\_field &92.31 &100.00 &96.00\\
harbor &85.71 &100.00 &92.31\\
industrial\_area &84.62 &91.67 &88.00\\
intersection &100.00 &100.00 &100.00\\
island &100.00 &100.00 &100.00\\
lake &91.67 &91.67 &91.67\\
meadow &83.33 &83.33 &83.33\\
medium\_residential &88.89 &66.67 &76.19\\
mobile\_home\_park &90.91 &83.33 &86.96\\
mountain &100.00 &100.00 &100.00\\
overpass &92.31 &100.00 &96.00\\
parking\_lot &100.00 &100.00 &100.00\\
railway &92.31 &100.00 &96.00\\
rectangular\_farmland &100.00 &83.33 &90.91\\
roundabout &100.00 &100.00 &100.00\\
runway &100.00 &91.67 &95.65\\
\bottomrule
\end{tabular} 
\end{table}

\clearpage

\subsection{So2Sat}
\label{app:so2Sat}
This dataset \citep{Zhu2020So2Sat} consists of co-registered synthetic aperture radar and multispectral optical image patches acquired by the Sentinel-1 and Sentinel-2 remote sensing satellites, and the corresponding local climate zones (LCZ) label. So2Sat has a total of 400673 images of size 32x32 pixels organized into 17 classes. Sample images are shown on Figure~\ref{fig:optimal31_samples}.  

The dataset is distributed over 42 cities across different continents and cultural regions of the world. The classes include: compact high rise, compact middle rise, compact low rise, open high rise, open middle rise, open low rise, lightweight low rise, large low rise, sparsely built, heavy industry, dense trees, scattered trees, bush scrub, low plants, bare rock or paved, bare soil or sand, and water. 

\begin{figure}[ht]
  \centering
  \includegraphics[width=0.6\linewidth]{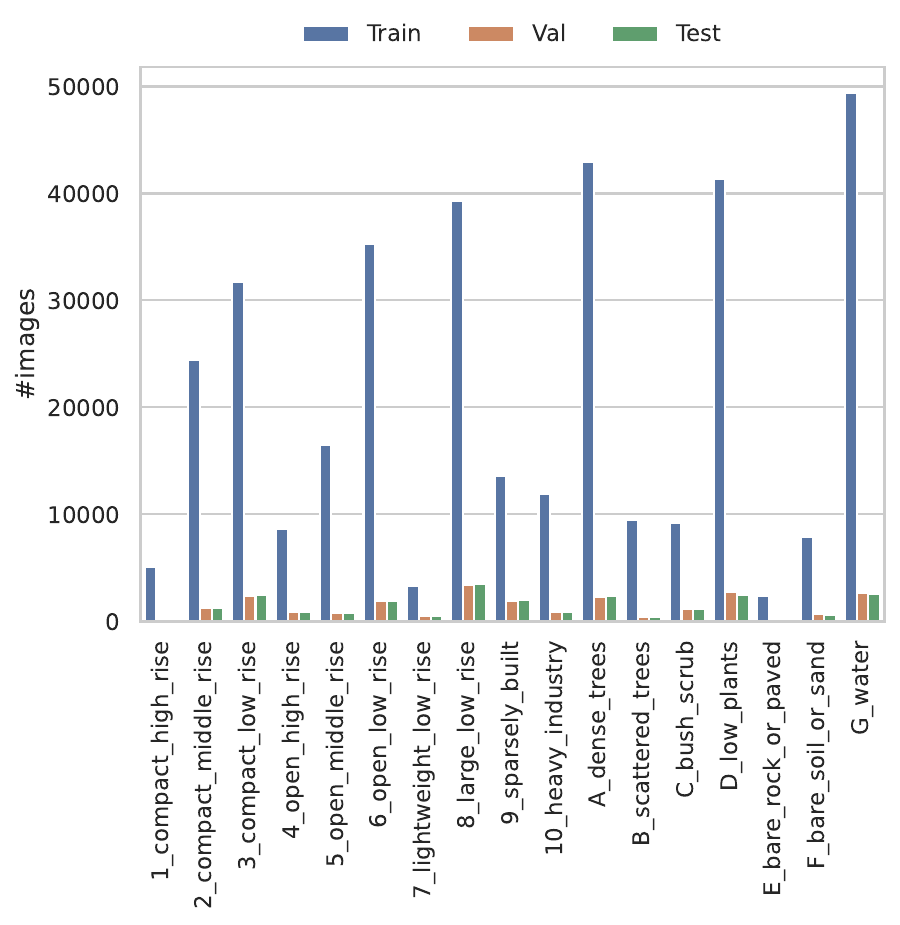}
  \caption{Class distribution for the So2Sat dataset.}
  \label{fig:so2sat_distribution}
\end{figure}

The creators of So2Sat have provided different versions for train, test and validation splits for the dataset. The class distribution of the splits is depicted on Figure~\ref{fig:so2sat_distribution}. We are using Version~2 \footnote{available at So2Sat-LCZ42 repo \url{https://github.com/zhu-xlab/So2Sat-LCZ42}.} with only Sentinel 2 data. Version 2 provides a training set covering 42 cities around the world, a validation set covering western half of 10 other cities covering 10 cultural zones and a test set containing the eastern half of the 10 other cities.

\begin{figure}[t]
  \centering
  \includegraphics[width=0.9\linewidth]{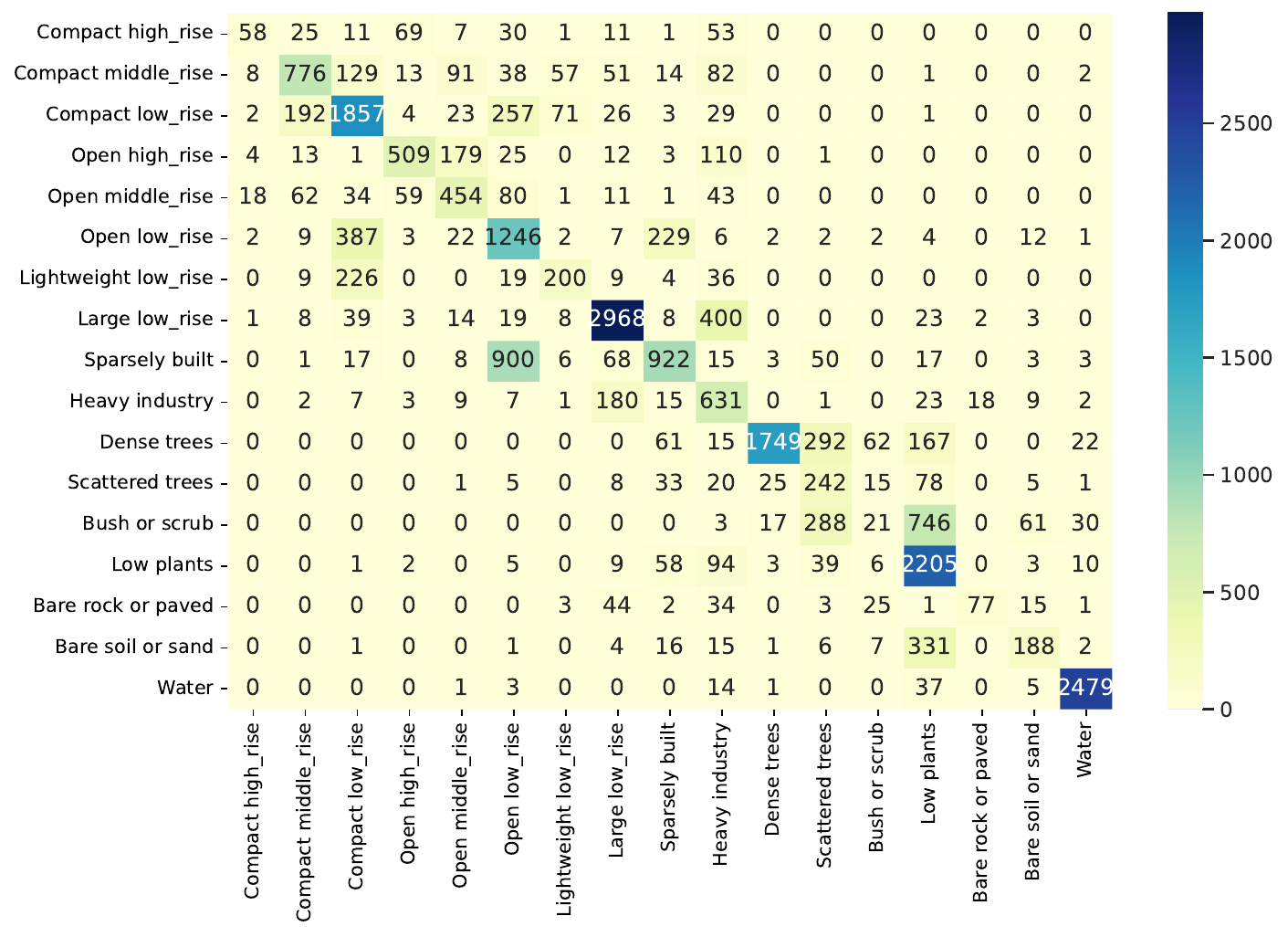}
  \caption{Confusion matrix for the pre-trained Vision Transformer model on the So2Sat dataset.}
  \label{fig:so2sat_confusionmatrix}
\end{figure}

Detailed results for all pre-trained models are shown on Table~\ref{tab:pre-trained_so2sat} and for all the models learned from scratch are presented on Table~\ref{tab:scratch_so2sat}. The best performing model is the pre-trained Vision Transformer model. The results on a class level are show on Table~\ref{tab:perclass_so2sat} along with a confusion matrix on Figure~\ref{fig:so2sat_confusionmatrix}.

\begin{figure}[ht]
\centering
  \includegraphics[width=0.7\linewidth]{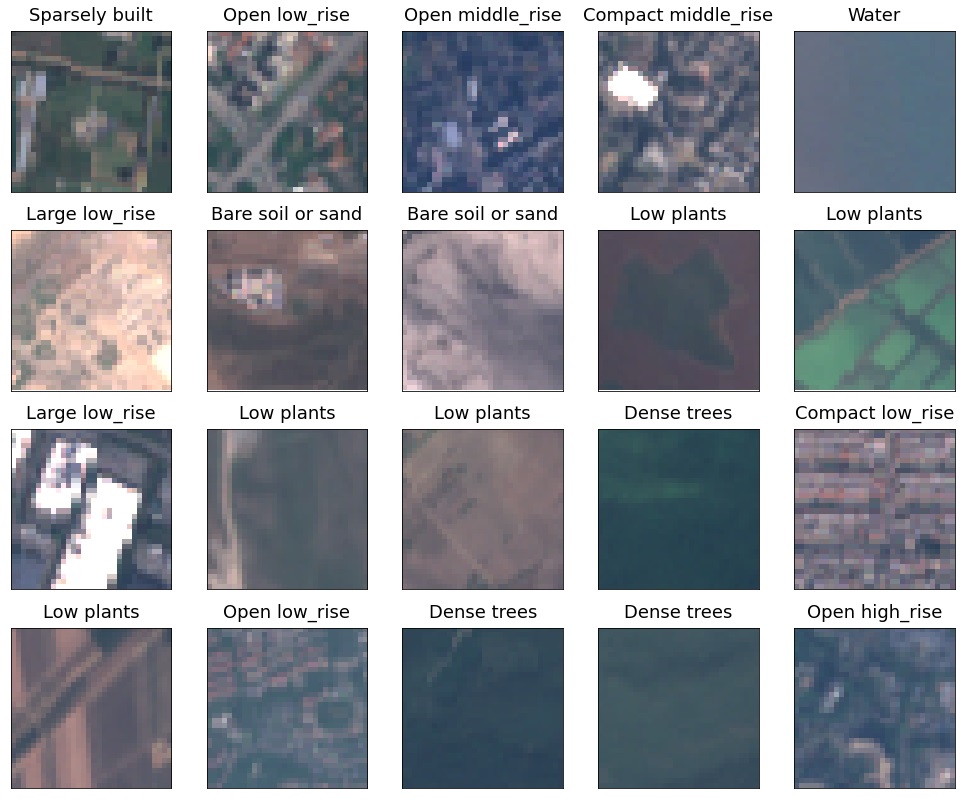}
  \caption{Example images with labels from the So2Sat dataset.}
  \label{fig:so2sat_samples}
\end{figure}

\begin{table}[ht]\centering
\caption{Detailed results for pre-trained models on the So2Sat dataset.}\label{tab:pre-trained_so2sat}
\scriptsize \begin{adjustbox}{width=0.75\linewidth}
\begin{tabular}{lrrrrrrrrrrr}\toprule
Model \textbackslash Metric &\rotatebox{90}{Accuracy} &\rotatebox{90}{Macro Precision} &\rotatebox{90}{Weighted Precision} &\rotatebox{90}{Macro Recall} &\rotatebox{90}{Weighted Recall} &\rotatebox{90}{Macro F1 score} &\rotatebox{90}{Weighted F1 score} &\rotatebox{90}{Avg. time / epoch (sec.)} &\rotatebox{90}{Total time (sec.)} &\rotatebox{90}{Best epoch}\\\midrule
AlexNet &59.20 &46.01 &59.31 &42.70 &59.20 &41.57 &57.59 &158.09 &1790 &1\\
VGG16 &65.38 &57.30 &64.34 &50.00 &65.38 &49.64 &63.00 &716.09 &7877 &1\\
ResNet50 &61.90 &51.01 &60.88 &48.45 &61.90 &48.35 &60.41 &565.55 &6221 &1\\
ResNet152 &65.17 &56.66 &64.48 &53.42 &65.17 &52.93 &63.75 &1,200.64 &13207 &1\\
DenseNet161 &65.76 &55.47 &64.58 &48.59 &65.76 &48.67 &63.81 &1,324.09 &14784 &1\\
EfficientNetB0 &65.80 &56.30 &65.64 &53.37 &65.80 &53.65 &64.77 &510.45 &5615 &1\\
ConvNeXt &66.17 &59.11 &66.87 &54.87 &66.17 &54.71 &65.56 &853.91 &9393 &1\\
Vision Transformer &\textbf{68.55} &62.95 &69.64 &57.17 &68.55 &57.26 &67.48 &925.09 &10176 &1\\
MLP Mixer &67.07 &63.74 &68.25 &51.34 &67.07 &51.94 &65.66 &643.91 &7278 &1\\
Swin Transformer &65.95 &59.11 &66.82 &53.20 &65.95 &52.89 &64.60 &2,636.45 &29001 &1 \\
\bottomrule
\end{tabular} \end{adjustbox}
\end{table}

\begin{table}[ht]\centering
\caption{Detailed results for models trained from scratch on the So2Sat dataset.}\label{tab:scratch_so2sat}
\scriptsize \begin{adjustbox}{width=0.75\linewidth}
\begin{tabular}{lrrrrrrrrrrr}\toprule
Model \textbackslash Metric &\rotatebox{90}{Accuracy} &\rotatebox{90}{Macro Precision} &\rotatebox{90}{Weighted Precision} &\rotatebox{90}{Macro Recall} &\rotatebox{90}{Weighted Recall} &\rotatebox{90}{Macro F1 score} &\rotatebox{90}{Weighted F1 score} &\rotatebox{90}{Avg. time / epoch (sec.)} &\rotatebox{90}{Total time (sec.)} &\rotatebox{90}{Best epoch}\\\midrule
AlexNet &56.51 &41.86 &54.97 &40.70 &56.51 &39.72 &54.65 &174.74 &3320 &4\\
VGG16 &62.27 &51.36 &61.08 &45.40 &62.27 &45.54 &59.78 &723.72 &13027 &3\\
ResNet50 &59.59 &46.54 &59.35 &43.94 &59.59 &43.37 &58.18 &558.79 &10617 &4\\
ResNet152 &61.48 &49.43 &62.30 &48.71 &61.48 &46.98 &60.22 &1,198.37 &22769 &4\\
DenseNet161 &55.43 &48.87 &60.98 &42.53 &55.43 &40.76 &54.11 &1,325.67 &23862 &3\\
EfficientNetB0 &\textbf{65.17} &53.75 &64.00 &50.34 &65.17 &50.36 &63.88 &499.21 &11981 &9\\
ConvNeXt &60.15 &50.97 &61.52 &48.03 &60.15 &47.17 &59.73 &851.06 &15319 &3\\
Vision Transformer &55.33 &43.56 &55.31 &37.42 &55.33 &37.01 &52.20 &926.50 &14824 &1\\
MLP Mixer &53.58 &42.31 &53.80 &36.73 &53.58 &36.61 &51.19 &651.31 &10421 &1\\
Swin Transformer &57.13 &47.93 &56.48 &36.29 &57.13 &35.29 &52.28 &2,631.44 &42103 &1 \\
\bottomrule
\end{tabular} \end{adjustbox}
\end{table}

\begin{table}[ht]\centering

\begin{minipage}[]{0.49\linewidth}
\caption{Per class results for the pre-trained ViT model\\on the So2Sat dataset.}\label{tab:perclass_so2sat}
\scriptsize 
\begin{tabular}{lrrrr}\toprule
Label &Precision &Recall &F1 score \\\midrule
Compact high\_rise &62.37 &21.80 &32.31\\
Compact middle\_rise &70.74 &61.49 &65.79\\
Compact low\_rise &68.52 &75.33 &71.77\\
Open high\_rise &76.54 &59.39 &66.89\\
Open middle\_rise &56.12 &59.50 &57.76\\
Open low\_rise &47.29 &64.36 &54.52\\
Lightweight low\_rise &57.14 &39.76 &46.89\\
Large low\_rise &87.11 &84.87 &85.98\\
Sparsely built &67.30 &45.80 &54.51\\
Heavy industry &39.39 &69.49 &50.28\\
Dense trees &97.11 &73.86 &83.91\\
Scattered trees &26.16 &55.89 &35.64\\
Bush or scrub &15.22 &1.80 &3.22\\
Low plants &60.68 &90.55 &72.66\\
Bare rock or paved &79.38 &37.56 &50.99\\
Bare soil or sand &62.05 &32.87 &42.97\\
Water &97.10 &97.60 &97.35\\
\bottomrule
\end{tabular} 
\end{minipage}
\begin{minipage}[]{0.49\linewidth}
\centering
  \includegraphics[width=1\linewidth]{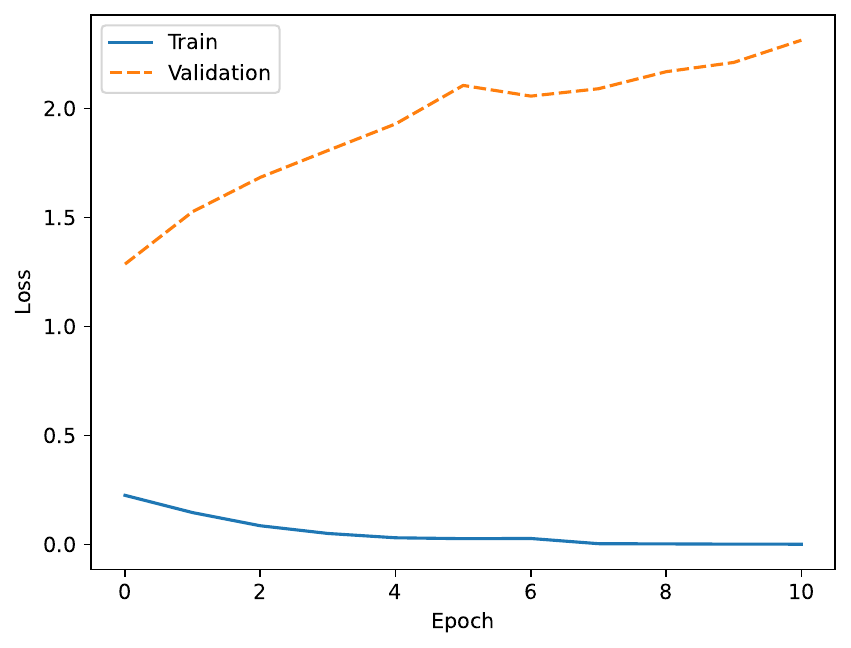}
  \captionof{figure}{Train and validation learning curves showing an over-fit of a ViT model on the So2Sat dataset.}
  \label{fig:so2sat_learning_curve}
\end{minipage}

\end{table}

\begin{figure}[ht]

\end{figure}

\clearpage

\subsection{UC Merced multi-label}
\begin{figure}[ht] \centering  \includegraphics[width=0.7\linewidth]{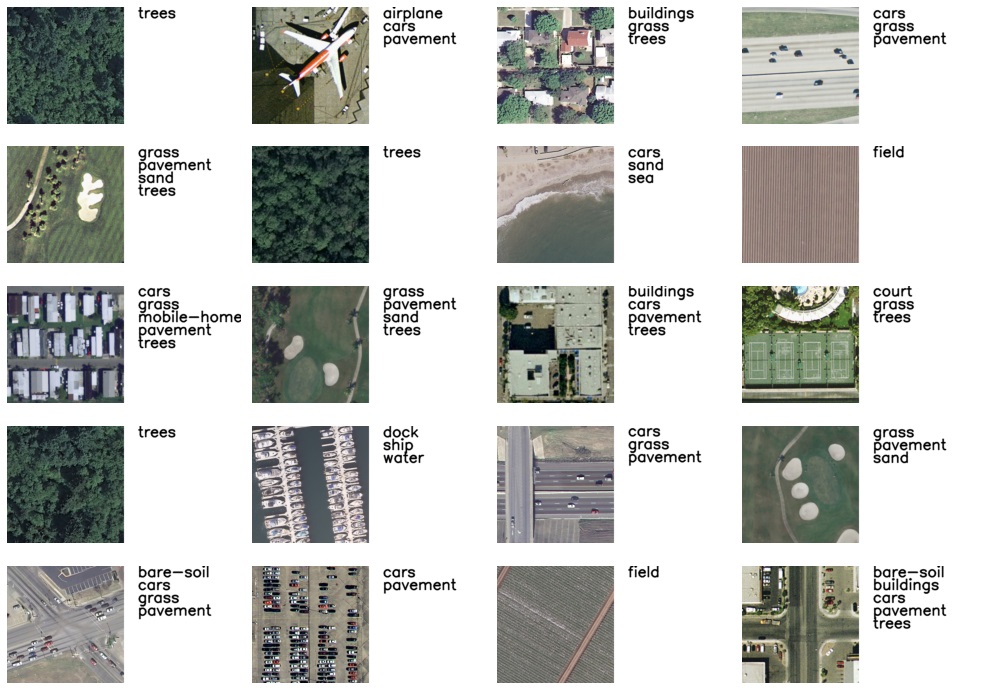}
  \caption{Example images with labels from the UC Merced multi-label dataset.}
  \label{fig:ucmerced_multilabel_samples}
\end{figure}

\begin{figure}[ht]
  \centering
  \includegraphics[width=0.5\linewidth]{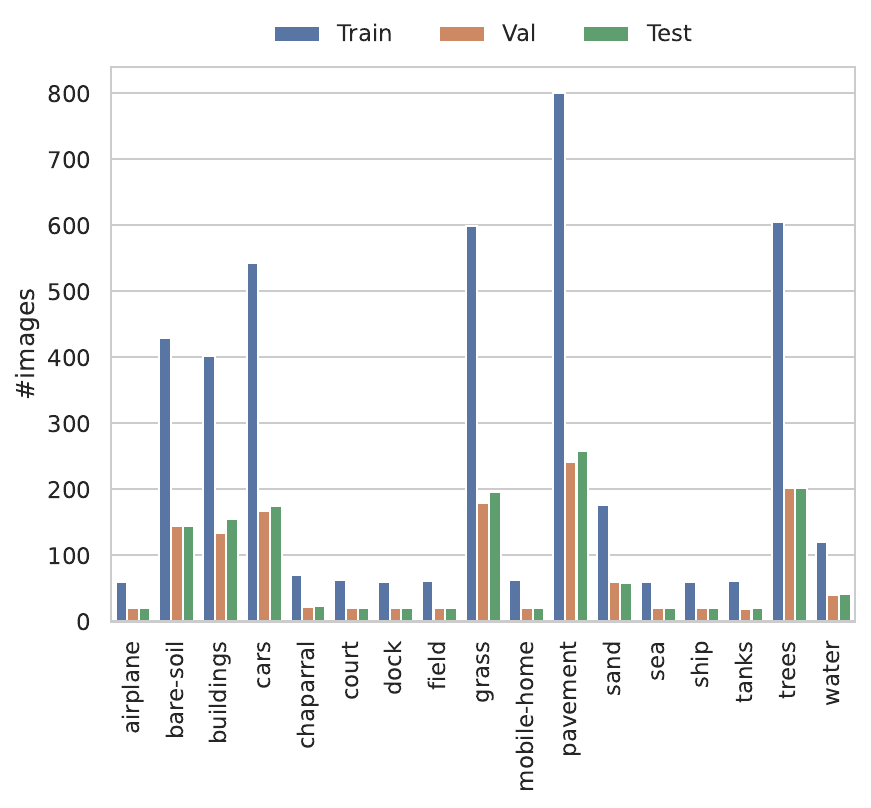}
  \caption{Label distribution for the UC Merced multi-label dataset.}
  \label{fig:ucmerced_multilabel_distribution}
\end{figure}

The UC Merced dataset was extended in \citep{chaudhuri2018ucmerced} for multi-label classification. The dataset still has the same number of 2100 images of 256x256 pixels size (Figure~\ref{fig:ucmerced_multilabel_samples}). The difference is in the number of classes (labels) and the number of annotations (classes) an image belongs to. Each image in the dataset has been manually labeled with one or more (maximum seven) labels based on visual inspection in order to create the ground truth data (the multilabels are available at http://bigearth.eu/datasets). The total number of distinct class labels in the dataset is 17. The labels are: airplane, bare-soil, buildings, cars, chaparral, court, dock, field, grass, mobile-home, pavement, sand, sea, ship, tanks, trees, water. The average number of labels per image is 3.3. This dataset has no predefined train-test splits by the authors. For our study, we made appropriate splits and their distribution is presented on Figure~\ref{fig:ucmerced_multilabel_distribution}.

\begin{figure}[ht]
  \centering
  \includegraphics[width=0.7\linewidth]{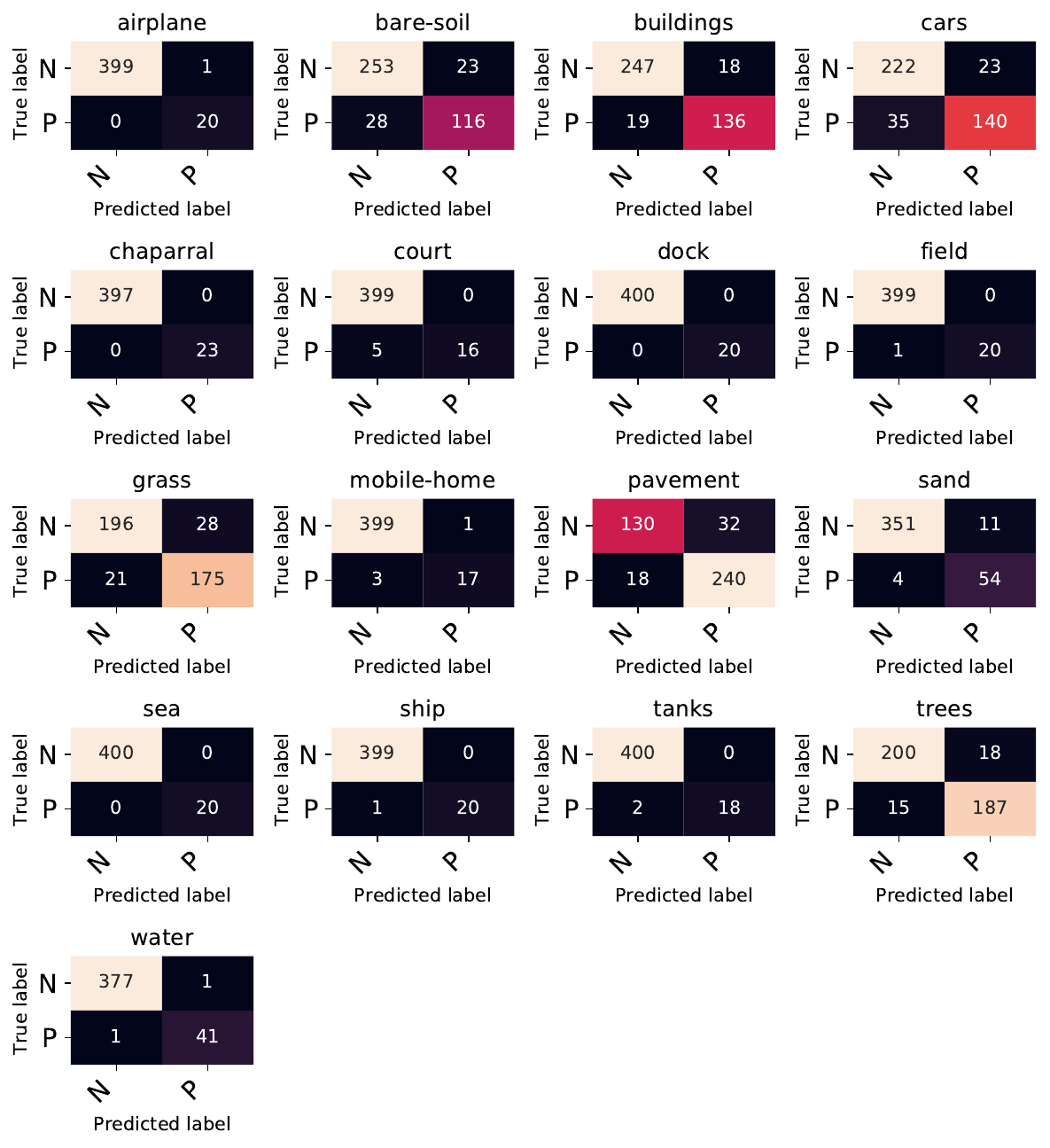}
  \caption{Confusion matrix for the pre-trained Swin Transformer model on the UC Merced multi-label dataset.}
  \label{fig:ucmerced_multilabel_confusionmatrix}
\end{figure}

Detailed results for all pre-trained models are shown on Table~\ref{tab:pre-trained_ucmerced_multilabel} and for all the models learned from scratch are presented on Table~\ref{tab:scratch_ucmerced_multilabel}. The best performing model is the pre-trained Swin Transformer model. The results on a class level are show on Table~\ref{tab:perclass_ucmerced_multilabel} along with a confusion matrix on Figure~\ref{fig:ucmerced_multilabel_confusionmatrix}.

\begin{table}[ht]\centering
\caption{Detailed results for pre-trained models on the UC Merced multi-label dataset.}\label{tab:pre-trained_ucmerced_multilabel}
\scriptsize \begin{adjustbox}{width=0.75\linewidth}
\begin{tabular}{lrrrrrrrrrrrrrr}\toprule
&\rotatebox{90}{mAP} &\rotatebox{90}{Micro Precision} &\rotatebox{90}{Macro Precision} &\rotatebox{90}{Weighted Precision} &\rotatebox{90}{Micro Recall} &\rotatebox{90}{Macro Recall} &\rotatebox{90}{Weighted Recall} &\rotatebox{90}{Micro F1 score} &\rotatebox{90}{Macro F1 score} &\rotatebox{90}{Weighted F1 score} &\rotatebox{90}{Avg. time / epoch (sec.)} &\rotatebox{90}{Total time (sec.)} &\rotatebox{90}{Best epoch}\\\midrule
AlexNet &92.64 &82.78 &88.47 &83.14 &86.23 &86.07 &86.23 &84.47 &86.91 &84.52 &1.31 &71 &44\\
VGG16 &92.85 &86.43 &91.38 &86.61 &86.37 &87.84 &86.37 &86.40 &89.33 &86.39 &3.30 &132 &30\\
ResNet50 &95.66 &86.19 &92.37 &86.53 &87.71 &88.84 &87.71 &86.94 &90.23 &86.95 &2.76 &124 &35\\
ResNet152 &96.01 &88.10 &93.19 &88.33 &86.23 &89.45 &86.23 &87.15 &91.07 &87.13 &5.04 &227 &35\\
DenseNet161 &96.06 &88.82 &93.99 &88.90 &87.01 &89.69 &87.01 &87.91 &91.51 &87.76 &5.64 &468 &73\\
EfficientNetB0 &95.38 &87.98 &93.22 &88.23 &87.36 &89.19 &87.36 &87.67 &90.92 &87.65 &2.54 &254 &98\\
ConvNeXt &96.43 &88.80 &94.30 &88.91 &87.92 &89.92 &87.92 &88.36 &91.84 &88.32 &3.92 &259 &56\\
Vision Transformer &96.70 &88.87 &94.16 &89.09 &89.62 &90.55 &89.62 &89.24 &92.14 &89.16 &4.13 &132 &22\\
MLP Mixer &96.34 &88.62 &94.38 &88.75 &87.99 &88.16 &87.99 &88.31 &90.77 &88.21 &3.25 &182 &46\\
Swin Transformer &\textbf{96.83} &89.01 &93.75 &89.08 &89.19 &91.50 &89.19 &89.10 &92.46 &89.06 &10.22 &552 &44 \\
\bottomrule
\end{tabular} \end{adjustbox}
\end{table}

\begin{table}[ht]\centering
\caption{Detailed results for models trained from scratch on the UC Merced multi-label dataset.}\label{tab:scratch_ucmerced_multilabel}
\scriptsize \begin{adjustbox}{width=0.75\linewidth}
\begin{tabular}{lrrrrrrrrrrrrrr}\toprule
&\rotatebox{90}{mAP} &\rotatebox{90}{Micro Precision} &\rotatebox{90}{Macro Precision} &\rotatebox{90}{Weighted Precision} &\rotatebox{90}{Micro Recall} &\rotatebox{90}{Macro Recall} &\rotatebox{90}{Weighted Recall} &\rotatebox{90}{Micro F1 score} &\rotatebox{90}{Macro F1 score} &\rotatebox{90}{Weighted F1 score} &\rotatebox{90}{Avg. time / epoch (sec.)} &\rotatebox{90}{Total time (sec.)} &\rotatebox{90}{Best epoch}\\\midrule
AlexNet &75.52 &72.54 &67.64 &70.50 &73.87 &63.95 &73.87 &73.20 &64.95 &71.73 &1.03 &103 &91\\
VGG16 &76.80 &74.33 &72.59 &73.65 &78.53 &70.75 &78.53 &76.37 &71.14 &75.77 &3.24 &324 &99\\
ResNet50 &79.87 &76.72 &77.52 &76.42 &78.67 &71.21 &78.67 &77.68 &72.73 &76.99 &2.76 &276 &99\\
ResNet152 &73.66 &76.89 &69.85 &74.78 &73.80 &65.05 &73.80 &75.32 &66.81 &73.92 &5.06 &506 &86\\
DenseNet161 &85.41 &81.30 &84.62 &81.61 &79.52 &76.19 &79.52 &80.40 &79.63 &80.26 &5.60 &487 &72\\
EfficientNetB0 &79.87 &78.45 &74.10 &76.91 &75.85 &72.13 &75.85 &77.13 &72.89 &76.25 &2.23 &252 &99\\
ConvNeXt &72.27 &72.40 &69.27 &71.19 &74.65 &62.31 &74.65 &73.50 &63.50 &71.89 &3.81 &381 &100\\
Vision Transformer &\textbf{87.14} &81.02 &85.66 &81.10 &79.31 &75.95 &79.31 &80.16 &79.29 &79.69 &4.12 &412 &95\\
MLP Mixer &75.68 &75.29 &73.64 &74.60 &73.38 &64.54 &73.38 &74.32 &67.44 &73.43 &3.11 &311 &99\\
Swin Transformer &81.07 &76.88 &75.54 &76.02 &79.38 &72.02 &79.38 &78.11 &72.50 &77.27 &10.12 &1012 &99 \\
\bottomrule
\end{tabular} \end{adjustbox}
\end{table}

\begin{table}[ht]\centering
\caption{Per label results for the pre-trained Swin Transformer model on the UC Merced multi-label dataset.}\label{tab:perclass_ucmerced_multilabel}
\scriptsize 
\begin{tabular}{lrrrr}\toprule
Label &Precision &Recall &F1 score \\\midrule
airplane &95.24 &100.00 &97.56 \\
bare-soil &83.45 &80.56 &81.98 \\
buildings &88.31 &87.74 &88.03 \\
cars &85.89 &80.00 &82.84 \\
chaparral &100.00 &100.00 &100.00 \\
court &100.00 &76.19 &86.49 \\
dock &100.00 &100.00 &100.00 \\
field &100.00 &95.24 &97.56 \\
grass &86.21 &89.29 &87.72 \\
mobile-home &94.44 &85.00 &89.47 \\
pavement &88.24 &93.02 &90.57 \\
sand &83.08 &93.10 &87.80 \\
sea &100.00 &100.00 &100.00 \\
ship &100.00 &95.24 &97.56 \\
tanks &100.00 &90.00 &94.74 \\
trees &91.22 &92.57 &91.89 \\
water &97.62 &97.62 &97.62 \\
\bottomrule
\end{tabular}
\end{table}

\clearpage

\subsection{BigEarthNet}
\label{app:bigearth}
BigEarthNet is a new large-scale multi-label Sentinel-2 benchmark archive \citep{Sumbul2019BigearthnetAL} \citep{Sumbul2021BigearthnetAL}. The BigEarthNet consists of 590326 Sentinel-2 image patches, each of which is a section of: 120x120 pixels for 10m bands; 60x60 pixels for 20m bands; and 20x20 pixels for 60m bands. Each image patch is annotated by multiple land-cover classes (i.e., multi-labels) that are provided from the CORINE Land Cover database. It was constructed by selecting 125 Sentinel-2 tiles acquired between June 2017 and May 2018. Covering different countries and seasonal period. More precisely, the number of images acquired in autumn, winter, spring and summer seasons are 154943, 117156, 189276 and 128951 respectively. The image patches are geographically distributed across 10 countries (Austria, Belgium, Finland, Ireland, Kosovo, Lithuania, Luxembourg, Portugal, Serbia, Switzerland) of Europe. The images are stored in tiff format and accompanied with additional metadata in JSON format.

The authors provide a predefined set of train-validation-test splits. Additionally, they proposed 2 versions of the labels in the dataset. 

\begin{figure}[ht]
  \centering
  \includegraphics[width=0.8\linewidth]{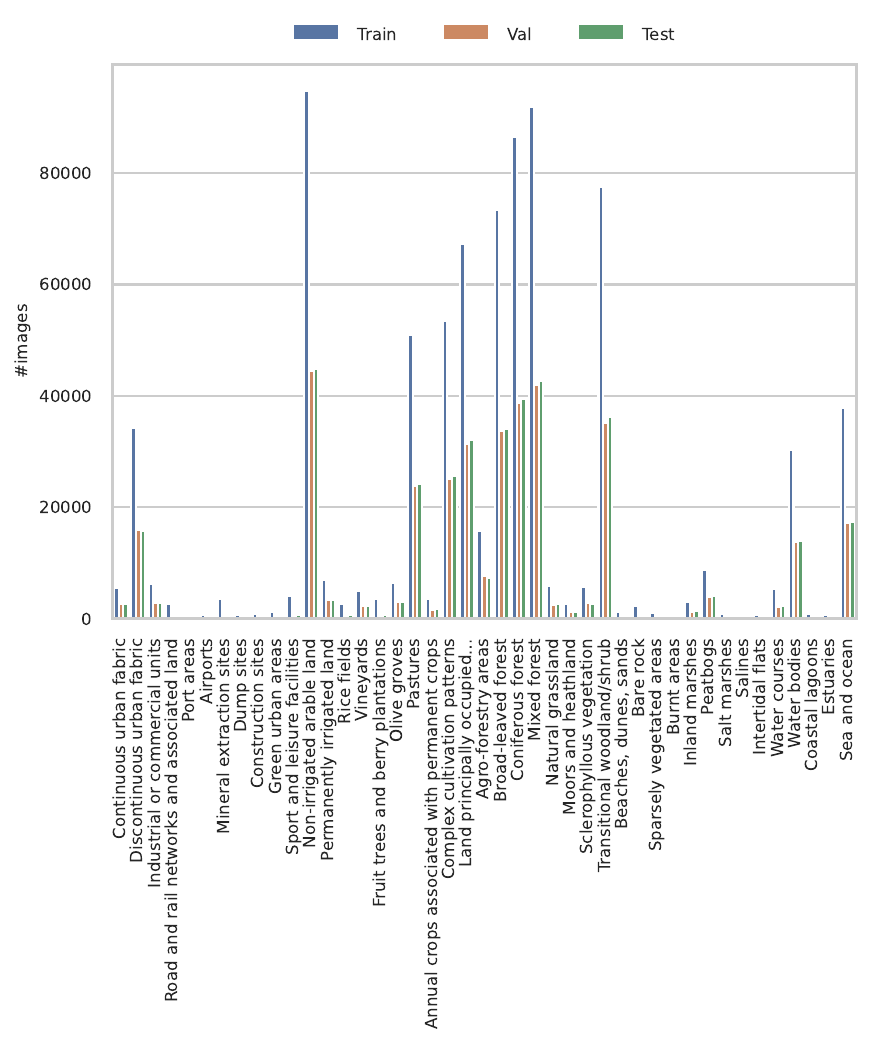}
  \caption{Label distribution for the BigEarthNet 43 dataset.}
  \label{fig:bigearthnet43_distribution}
\end{figure}

The first version of the dataset contains 43 labels with an 3.0 labels per image (Figure~\ref{fig:bigearthnet43_distribution}). The labels in this version are: Continuous urban fabric, Discontinuous urban fabric, Industrial or commercial units, Road and rail networks and associated land, Port areas, Airports, Mineral extraction sites, Dump sites, Construction sites, Green urban areas, Sport and leisure facilities, Non-irrigated arable land, Permanently irrigated land, Rice fields, Vineyards, Fruit trees and berry plantations, Olive groves, Pastures, Annual crops associated with permanent crops, Complex cultivation patterns, Land principally occupied by agriculture, with significant areas of natural vegetation, Agro-forestry areas, Broad-leaved forest, Coniferous forest, Mixed forest, Natural grassland, Moors and heathland, Sclerophyllous vegetation, Transitional woodland/shrub, Beaches, dunes, sands, Bare rock, Sparsely vegetated areas, Burnt areas, Inland marshes, Peatbogs, Salt marshes, Salines, Intertidal flats, Water courses, Water bodies, Coastal lagoons, Estuaries, Sea and ocean. The largest class (label), Mixed forest, appeared in 217119 image, whereas the label with fewest appearances, Burnt areas, appeared in 328 images. This high imbalance should make the dataset more challenging. 

\begin{figure}[ht]
  \centering
  \includegraphics[width=0.9\linewidth]{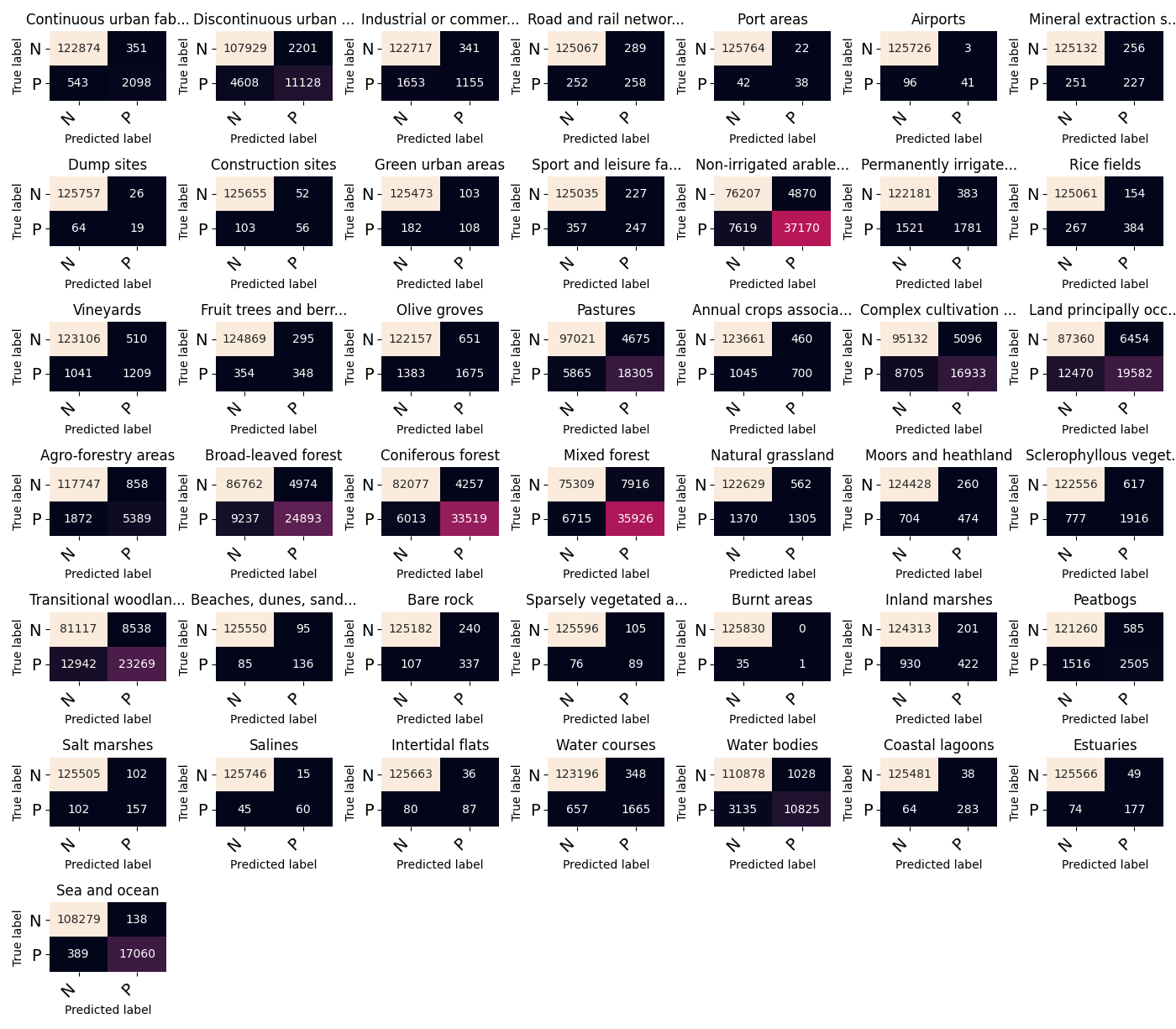}
  \caption{Confusion matrix for the pre-trained Swin Transformer model on the BigEarthNet 43 dataset.}
  \label{fig:bigearthnet43_confusionmatrix}
\end{figure}

Detailed results for all pre-trained models are shown on Table~\ref{tab:pre-trained_bigearthnet43} and for all the models learned from scratch are presented on Table~\ref{tab:scratch_bigearthnet43}. The best performing model is the pre-trained Swin Transformer model. The results on a class level are show on Table~\ref{tab:perclass_bigearthnet43} along with a confusion matrix on Figure~\ref{fig:bigearthnet43_confusionmatrix}.

\begin{figure}[ht] \centering  \includegraphics[width=0.7\linewidth]{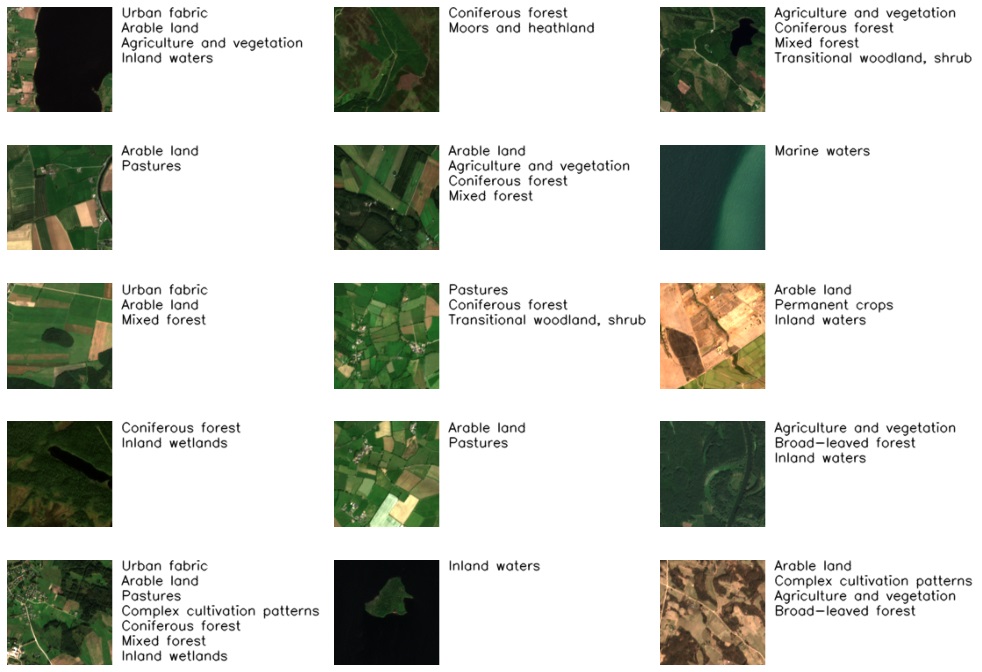}
  \caption{Example images with labels from the BigEarthNet dataset.}
  \label{fig:bigearthnet19_samples}
\end{figure}

\begin{figure}[ht]
  \centering
  \includegraphics[width=0.6\linewidth]{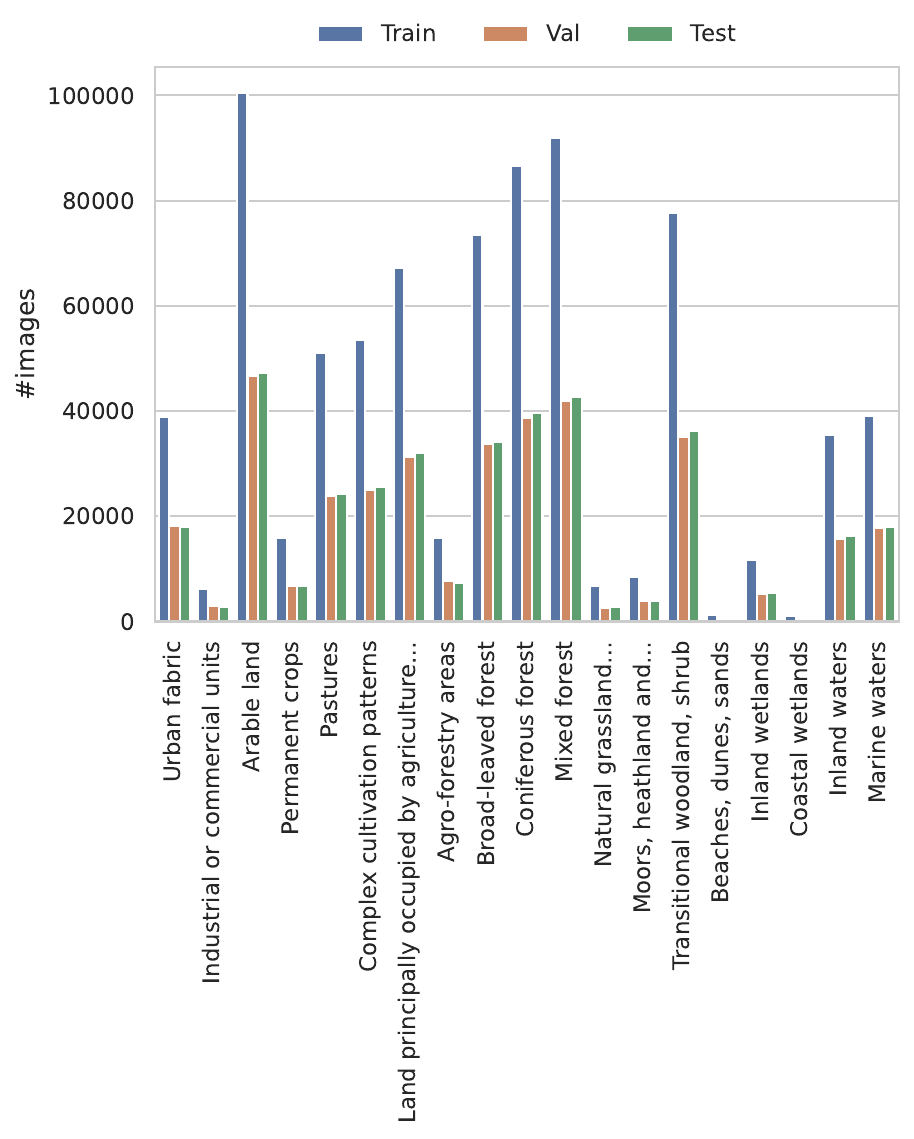}
  \caption{Label distribution for the BigEarthNet 19 dataset.}
  \label{fig:bigearthnet19_distribution}
\end{figure}

The second version of the dataset contains 19 labels with 2.9 labels per image on average (Figure~\ref{fig:bigearthnet19_distribution}). The labels contained here are: Urban fabric, Industrial or commercial units,  Arable land,  Permanent crops,  Pastures,  Complex cultivation patterns,  Land principally occupied by agriculture, with significant areas of natural vegetation,  Agro-forestry areas,  Broad-leaved forest,  Coniferous forest,  Mixed forest, Natural grassland and sparsely vegetated areas, Moors, heath-land and sclerophyllous vegetation, Transitional woodland, shrub, Beaches, dunes, sands, Inland wetlands, Coastal wetlands, Inland waters, Marine waters. The label Mixed forest is most commonly found and is present in 176546 images, whereas Beaches, dunes, sands appears in 1536 images and is the least frequently used label. Sample images are shown on Figure~\ref{fig:bigearthnet19_samples}. 

\begin{figure}[ht]
  \centering
  \includegraphics[width=0.7\linewidth]{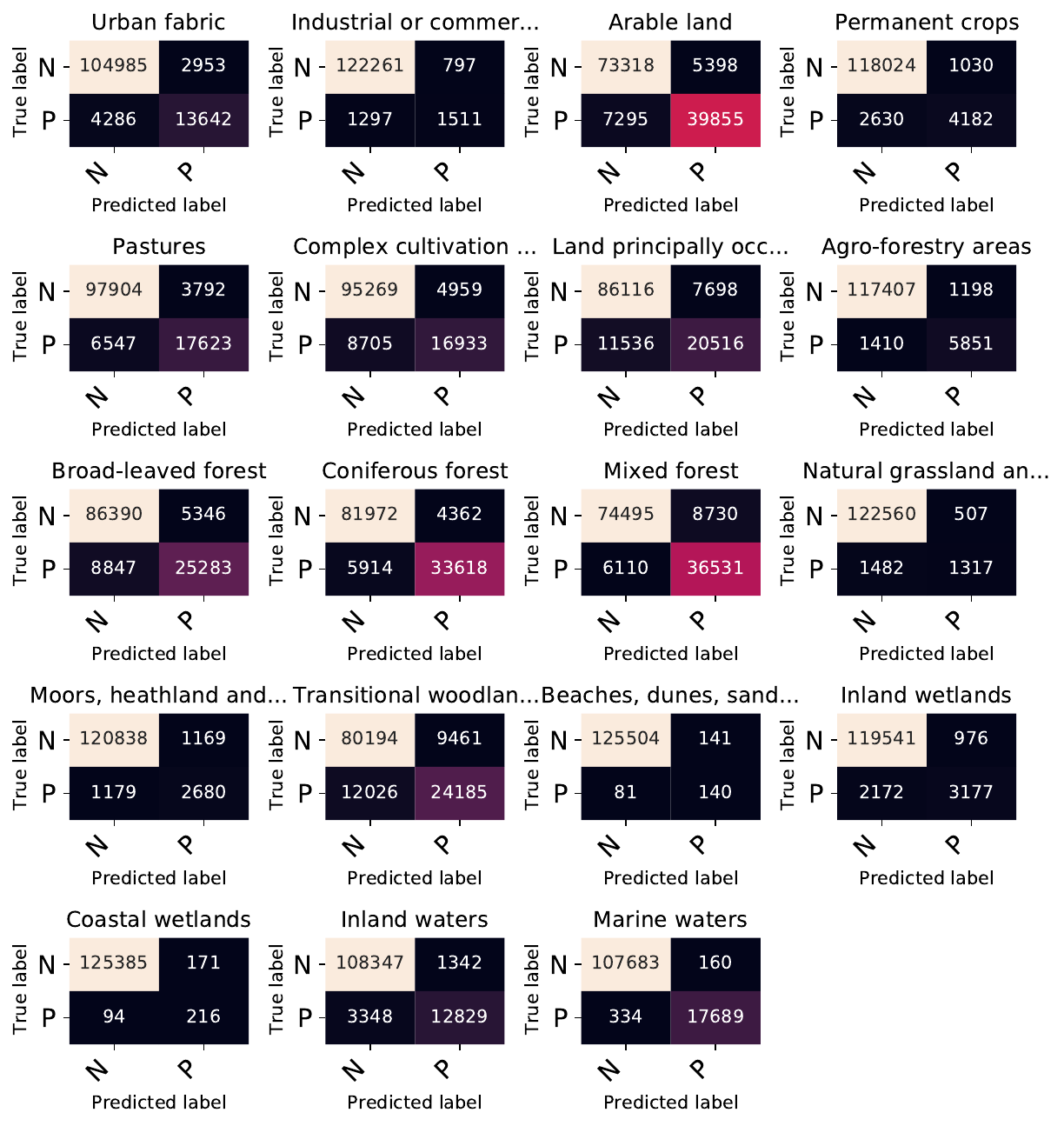}
  \caption{Confusion matrix for the pre-trained Swin Transformer model on the BigEarthNet 19 dataset.}
  \label{fig:bigearthnet19_confusionmatrix}
\end{figure}

Detailed results for all pre-trained models are shown on Table~\ref{tab:pre-trained_bigearthnet19} and for all the models learned from scratch are presented on Table~\ref{tab:scratch_bigearthnet19}. The best performing model is the pre-trained Swin Transformer model. The results on a class level are show on Table~\ref{tab:perclass_bigearthnet19} along with a confusion matrix on Figure~\ref{fig:bigearthnet19_confusionmatrix}.

\clearpage

\subsubsection{BigEarthNet 43}

\begin{table}[ht]\centering
\caption{Detailed results for pre-trained models on the BigEarthNet 43 dataset.}\label{tab:pre-trained_bigearthnet43}
\scriptsize \begin{adjustbox}{width=0.75\linewidth}
\begin{tabular}{lrrrrrrrrrrrrrr}\toprule
&\rotatebox{90}{mAP} &\rotatebox{90}{Micro Precision} &\rotatebox{90}{Macro Precision} &\rotatebox{90}{Weighted Precision} &\rotatebox{90}{Micro Recall} &\rotatebox{90}{Macro Recall} &\rotatebox{90}{Weighted Recall} &\rotatebox{90}{Micro F1 score} &\rotatebox{90}{Macro F1 score} &\rotatebox{90}{Weighted F1 score} &\rotatebox{90}{Avg. time / epoch (sec.)} &\rotatebox{90}{Total time (sec.)} &\rotatebox{90}{Best epoch}\\\midrule
AlexNet &58.55 &80.15 &61.88 &79.67 &72.14 &51.99 &72.14 &75.93 &55.62 &75.48 &89.85 &7188 &70\\	
VGG16 &61.21 &80.71 &64.71 &80.29 &72.74 &53.97 &72.74 &76.52 &57.74 &76.08 &542.30 &12473 &13\\	
ResNet50 &66.26 &81.99 &67.47 &81.64 &74.14 &58.15 &74.14 &77.87 &61.87 &77.54 &414.18 &9112 &12\\	
ResNet152 &64.07 &82.17 &70.42 &81.73 &72.08 &52.11 &72.08 &76.80 &58.27 &76.17 &881.69 &14107 &6\\	
DenseNet161 &64.23 &81.87 &68.31 &81.39 &72.63 &53.58 &72.63 &76.97 &58.80 &76.48 &969.67 &14545 &5\\	
EfficientNetB0 &64.59 &82.14 &70.17 &81.75 &73.37 &53.93 &73.37 &77.51 &59.71 &77.08 &365.40 &7308 &10\\	
ConvNeXt &66.17 &81.67 &69.24 &81.31 &73.93 &56.11 &73.93 &77.61 &61.12 &77.23 &642.81 &10285 &6\\	
Vision Transformer &59.00 &79.77 &65.42 &79.39 &71.39 &48.98 &71.39 &75.35 &54.65 &74.81 &702.00 &14742 &11\\	
MLP Mixer &59.65 &81.18 &67.47 &80.55 &71.30 &48.85 &71.30 &75.92 &54.95 &75.28 &492.84 &12321 &15\\
Swin Transformer &\textbf{67.73} &82.43 &72.36 &82.12 &74.18 &58.08 &74.18 &78.09 &62.78 &77.75 &2,016.00 &34272 &7 \\
\bottomrule
\end{tabular} \end{adjustbox}
\end{table}

\begin{table}[ht]\centering
\caption{Detailed results for models trained from scratch on the BigEarthNet 43 dataset.}\label{tab:scratch_bigearthnet43}
\scriptsize \begin{adjustbox}{width=0.75\linewidth}
\begin{tabular}{lrrrrrrrrrrrrrr}\toprule
&\rotatebox{90}{mAP} &\rotatebox{90}{Micro Precision} &\rotatebox{90}{Macro Precision} &\rotatebox{90}{Weighted Precision} &\rotatebox{90}{Micro Recall} &\rotatebox{90}{Macro Recall} &\rotatebox{90}{Weighted Recall} &\rotatebox{90}{Micro F1 score} &\rotatebox{90}{Macro F1 score} &\rotatebox{90}{Weighted F1 score} &\rotatebox{90}{Avg. time / epoch (sec.)} &\rotatebox{90}{Total time (sec.)} &\rotatebox{90}{Best epoch}\\\midrule
AlexNet &56.08 &79.15 &58.19 &78.68 &71.41 &50.79 &71.41 &75.08 &53.54 &74.65 &84.18 &5051 &45\\
VGG16 &58.97 &80.56 &64.94 &80.13 &71.99 &48.02 &71.99 &76.03 &53.38 &75.49 &544.28 &15784 &14\\
ResNet50 &64.34 &82.07 &67.06 &81.65 &73.47 &55.64 &73.47 &77.53 &60.14 &77.12 &409.87 &18854 &31\\
ResNet152 &62.74 &80.72 &66.55 &80.30 &72.96 &53.88 &72.96 &76.64 &58.59 &76.12 &878.00 &32486 &22\\
DenseNet161 &63.39 &82.20 &66.27 &81.74 &71.83 &53.84 &71.83 &76.67 &58.40 &76.00 &982.63 &29479 &15\\
EfficientNetB0 &62.17 &81.25 &66.61 &80.90 &73.01 &52.02 &73.01 &76.91 &56.94 &76.48 &364.13 &11288 &16\\
ConvNeXt &60.47 &80.71 &67.02 &80.19 &72.40 &51.09 &72.40 &76.33 &56.51 &75.81 &645.51 &26466 &26\\
Vision Transformer &57.41 &79.12 &63.50 &78.74 &71.20 &47.96 &71.20 &74.95 &52.94 &74.31 &709.86 &20586 &14\\
MLP Mixer &58.77 &80.82 &65.97 &80.10 &71.12 &48.10 &71.12 &75.66 &53.38 &74.90 &500.77 &15524 &16\\
Swin Transformer &\textbf{67.49} &81.91 &68.28 &81.62 &75.50 &59.70 &75.50 &78.58 &63.02 &78.29 &2,031.64 &113772 &41 \\
\bottomrule
\end{tabular} \end{adjustbox}
\end{table}

\begin{table}[ht]\centering
\caption{Per label results for the pre-trained  Swin Transformer model on the BigEarthNet 43 dataset.}\label{tab:perclass_bigearthnet43}
\scriptsize 
\begin{tabular}{lrrrr}\toprule
Label &Precision &Recall &F1 score \\\midrule
Continuous urban fabric &85.67 &79.44 &82.44 \\
Discontinuous urban fabric &83.49 &70.72 &76.57 \\
Industrial or commercial units &77.21 &41.13 &53.67 \\
Road and rail networks and associated land &47.17 &50.59 &48.82 \\
Port areas &63.33 &47.50 &54.29 \\
Airports &93.18 &29.93 &45.30 \\
Mineral extraction sites &47.00 &47.49 &47.24 \\
Dump sites &42.22 &22.89 &29.69 \\
Construction sites &51.85 &35.22 &41.95 \\
Green urban areas &51.18 &37.24 &43.11 \\
Sport and leisure facilities &52.11 &40.89 &45.83 \\
Non-irrigated arable land &88.42 &82.99 &85.62 \\
Permanently irrigated land &82.30 &53.94 &65.17 \\
Rice fields &71.38 &58.99 &64.59 \\
Vineyards &70.33 &53.73 &60.92 \\
Fruit trees and berry plantations &54.12 &49.57 &51.75 \\
Olive groves &72.01 &54.77 &62.22 \\
Pastures &79.66 &75.73 &77.65 \\
Annual crops associated with permanent crops &60.34 &40.11 &48.19 \\
Complex cultivation patterns &76.87 &66.05 &71.05 \\
Land principally occupied by agriculture, with significant areas of natural vegetation &75.21 &61.09 &67.42 \\
Agro-forestry areas &86.27 &74.22 &79.79 \\
Broad-leaved forest &83.35 &72.94 &77.79 \\
Coniferous forest &88.73 &84.79 &86.72 \\
Mixed forest &81.94 &84.25 &83.08 \\
Natural grassland &69.90 &48.79 &57.46 \\
Moors and heathland &64.58 &40.24 &49.58 \\
Sclerophyllous vegetation &75.64 &71.15 &73.33 \\
Transitional woodland/shrub &73.16 &64.26 &68.42 \\
Beaches, dunes, sands &58.87 &61.54 &60.18 \\
Bare rock &58.41 &75.90 &66.01 \\
Sparsely vegetated areas &45.88 &53.94 &49.58 \\
Burnt areas &100.00 &2.78 &5.41 \\
Inland marshes &67.74 &31.21 &42.73 \\
Peatbogs &81.07 &62.30 &70.45 \\
Salt marshes &60.62 &60.62 &60.62 \\
Salines &80.00 &57.14 &66.67 \\
Intertidal flats &70.73 &52.10 &60.00 \\
Water courses &82.71 &71.71 &76.82 \\
Water bodies &91.33 &77.54 &83.87 \\
Coastal lagoons &88.16 &81.56 &84.73 \\
Estuaries &78.32 &70.52 &74.21 \\
Sea and ocean &99.20 &97.77 &98.48 \\
\bottomrule
\end{tabular} 
\end{table}

\clearpage

\subsubsection{BigEarthNet 19}

\begin{table}[ht]\centering
\caption{Detailed results for pre-trained models on the BigEarthNet 19 dataset.}\label{tab:pre-trained_bigearthnet19}
\scriptsize \begin{adjustbox}{width=0.75\linewidth}
\begin{tabular}{lrrrrrrrrrrrrrr}\toprule
&\rotatebox{90}{mAP} &\rotatebox{90}{Micro Precision} &\rotatebox{90}{Macro Precision} &\rotatebox{90}{Weighted Precision} &\rotatebox{90}{Micro Recall} &\rotatebox{90}{Macro Recall} &\rotatebox{90}{Weighted Recall} &\rotatebox{90}{Micro F1 score} &\rotatebox{90}{Macro F1 score} &\rotatebox{90}{Weighted F1 score} &\rotatebox{90}{Avg. time / epoch (sec.)} &\rotatebox{90}{Total time (sec.)} &\rotatebox{90}{Best epoch}\\\midrule
AlexNet &77.15 &80.90 &75.60 &80.58 &73.59 &66.06 &73.59 &77.07 &70.04 &76.77 &90.43 &5245 &48\\
VGG16 &78.42 &81.33 &77.61 &81.02 &73.92 &66.27 &73.92 &77.45 &70.92 &77.11 &537.90 &10758 &10\\
ResNet50 &79.98 &82.65 &78.57 &82.37 &73.62 &67.74 &73.62 &77.88 &72.12 &77.51 &413.24 &7025 &7\\
ResNet152 &79.78 &82.58 &80.36 &82.43 &73.95 &66.57 &73.95 &78.03 &71.79 &77.57 &874.56 &13993 &6\\
DenseNet161 &79.69 &81.92 &78.55 &81.83 &74.42 &66.99 &74.42 &77.99 &71.61 &77.72 &976.93 &14654 &5\\
EfficientNetB0 &80.22 &82.87 &80.56 &82.61 &74.36 &66.32 &74.36 &78.38 &72.14 &78.09 &366.35 &6228 &7\\
ConvNeXt &80.28	&80.95	&78.78	&80.99	&76.72	&68.71	&76.72	&78.78	&72.66	&78.62 &631.67 &9475 &5\\
Vision Transformer &77.31 &82.31 &76.93 &81.85 &70.99 &64.08 &70.99 &76.23 &69.18 &75.70 &698.50 &15367 &12\\
MLP Mixer &77.29 &81.41 &78.12 &80.97 &73.20 &64.33 &73.20 &77.09 &69.68 &76.62 &488.68 &12217 &15\\
Swin Transformer &\textbf{81.38} &82.19 &77.29 &82.03 &76.51 &71.44 &76.51 &79.25 &73.89 &79.06 &1,990.00 &35820 &8 \\
\bottomrule
\end{tabular} \end{adjustbox}
\end{table}

\begin{table}[ht]\centering
\caption{Detailed results for models trained from scratch on the BigEarthNet 19 dataset.}\label{tab:scratch_bigearthnet19}
\scriptsize \begin{adjustbox}{width=0.75\linewidth}
\begin{tabular}{lrrrrrrrrrrrrrr}\toprule
&\rotatebox{90}{mAP} &\rotatebox{90}{Micro Precision} &\rotatebox{90}{Macro Precision} &\rotatebox{90}{Weighted Precision} &\rotatebox{90}{Micro Recall} &\rotatebox{90}{Macro Recall} &\rotatebox{90}{Weighted Recall} &\rotatebox{90}{Micro F1 score} &\rotatebox{90}{Macro F1 score} &\rotatebox{90}{Weighted F1 score} &\rotatebox{90}{Avg. time / epoch (sec.)} &\rotatebox{90}{Total time (sec.)} &\rotatebox{90}{Best epoch}\\\midrule
AlexNet &75.71 &80.27 &74.63 &79.88 &72.73 &64.83 &72.73 &76.31 &68.89 &75.96 &86.78 &5120 &44\\
VGG16 &77.99 &80.45 &75.61 &80.21 &74.91 &67.63 &74.91 &77.58 &70.90 &77.28 &542.24 &18436 &19\\
ResNet50 &78.73 &82.94 &78.20 &82.44 &72.61 &66.15 &72.61 &77.44 &71.28 &76.99 &413.89 &26489 &49\\
ResNet152 &78.52 &81.06 &75.86 &81.02 &74.69 &68.18 &74.69 &77.74 &71.34 &77.55 &875.20 &43760 &35\\
DenseNet161 &79.73 &82.24 &77.81 &82.05 &74.77 &67.99 &74.77 &78.33 &71.98 &78.08 &975.34 &31211 &17\\
EfficientNetB0 &79.21 &82.25 &78.89 &82.02 &74.68 &66.53 &74.68 &78.28 &71.65 &78.01 &359.16 &11493 &17\\
ConvNeXt &77.91 &81.39 &78.16 &81.18 &73.57 &64.64 &73.57 &77.29 &70.08 &76.95 &643.66 &24459 &23\\
Vision Transformer &75.87 &80.48 &75.45 &80.14 &71.36 &63.85 &71.36 &75.65 &68.59 &75.23 &702.53 &21076 &15\\
MLP Mixer &77.01 &81.39 &77.37 &81.12 &72.59 &64.34 &72.59 &76.74 &69.74 &76.42 &495.88 &15868 &17\\
Swin Transformer &\textbf{80.59} &83.12 &80.38 &82.76 &74.60 &66.96 &74.60 &78.63 &72.35 &78.22 &2,011.29 &96542 &33 \\
\bottomrule
\end{tabular} \end{adjustbox}
\end{table}

\begin{table}[ht]\centering
\caption{Per label results for the pre-trained Swin Transformer model on the BigEarthNet 19 dataset.}\label{tab:perclass_bigearthnet19}
\scriptsize 
\begin{tabular}{lrrrr}\toprule
Label &Precision &Recall &F1 score \\\midrule
Urban fabric &82.21 &76.09 &79.03 \\
Industrial or commercial units &65.47 &53.81 &59.07 \\
Arable land &88.07 &84.53 &86.26 \\
Permanent crops &80.24 &61.39 &69.56 \\
Pastures &82.29 &72.91 &77.32 \\
Complex cultivation patterns &77.35 &66.05 &71.25 \\
Land principally occupied by agriculture, with significant areas of natural vegetation &72.72 &64.01 &68.08 \\
Agro-forestry areas &83.00 &80.58 &81.77 \\
Broad-leaved forest &82.55 &74.08 &78.08 \\
Coniferous forest &88.52 &85.04 &86.74 \\
Mixed forest &80.71 &85.67 &83.12 \\
Natural grassland and sparsely vegetated areas &72.20 &47.05 &56.98 \\
Moors, heathland and sclerophyllous vegetation &69.63 &69.45 &69.54 \\
Transitional woodland, shrub &71.88 &66.79 &69.24 \\
Beaches, dunes, sands &49.82 &63.35 &55.78 \\
Inland wetlands &76.50 &59.39 &66.87 \\
Coastal wetlands &55.81 &69.68 &61.98 \\
Inland waters &90.53 &79.30 &84.55 \\
Marine waters &99.10 &98.15 &98.62 \\
\bottomrule
\end{tabular} 
\end{table}

\clearpage

\subsection{MLRSNet}

\begin{figure}[ht] \centering  \includegraphics[width=0.7\linewidth]{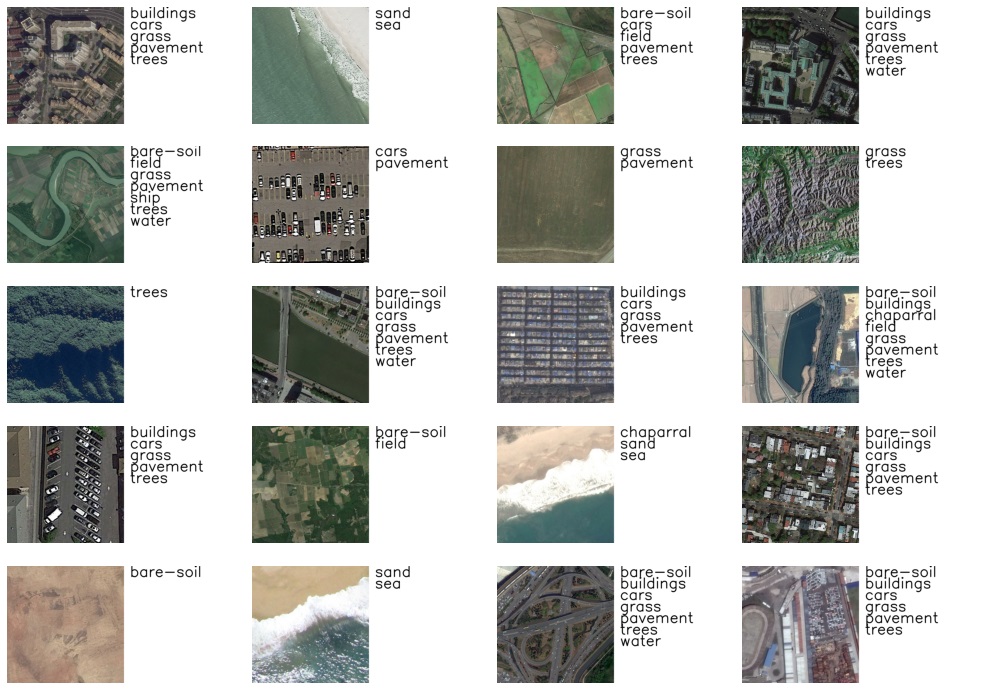}
  \caption{Example images with labels from the MLRSNet dataset.}
  \label{fig:mlrsnet_samples}
\end{figure}

\begin{figure}[ht]
  \centering
  \includegraphics[width=0.9\linewidth]{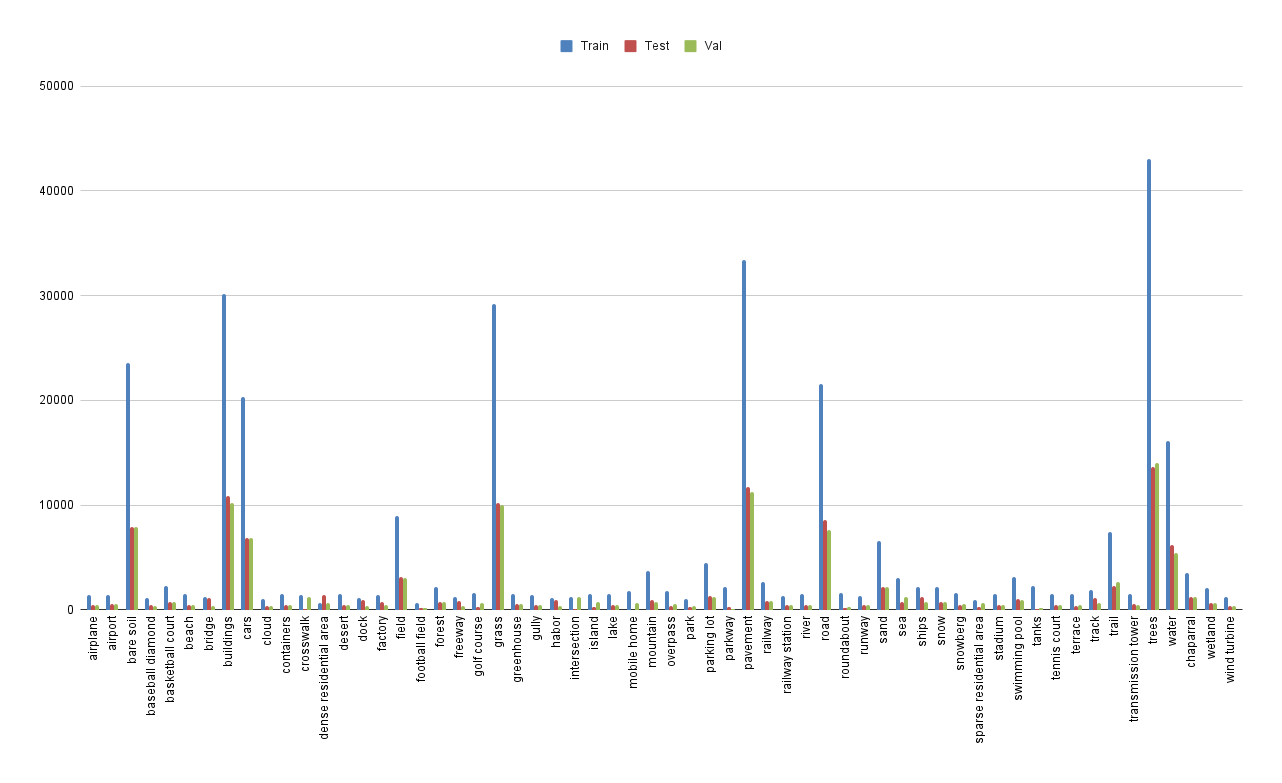}
  \caption{Label distribution for the MLRSNet dataset.}
  \label{fig:mlrsnet_distribution}
\end{figure}

MLRSNet \citep{qi2020mlrsnet} is a multi-label high spatial resolution remote sensing dataset for semantic scene understanding. It is composed of high-resolution optical satellite or aerial RGB images. MLRSNet contains a total of 109161 images (Figure~\ref{fig:mlrsnet_samples}) within 46 scene categories, and each image has at least one of 60 predefined labels. The number of labels associated with each image varies between 1 and 13, but averages at 5.0 labels per image (Figure~\ref{fig:mlrsnet_distribution}). The labels annotating the images are: airplane, airport, bare soil, baseball diamond, basketball court, beach, bridge, buildings, cars, cloud, containers, crosswalk, dense residential area, desert, dock, factory, field, football field, forest, freeway, golf course, grass, greenhouse, gully, habor, intersection, island, lake, mobile home, mountain, overpass, park, parking lot, parkway, pavement, railway, railway station, river, road, roundabout, runway, sand, sea, ships, snow, snowberg, sparse residential area, stadium, swimming pool, tanks, tennis court, terrace, track, trail, transmission tower, trees, water, chaparral, wetland, wind turbine. The dataset does not have predefined train-tests splits. 

\begin{figure}[ht]
  \centering
  \includegraphics[width=0.9\linewidth]{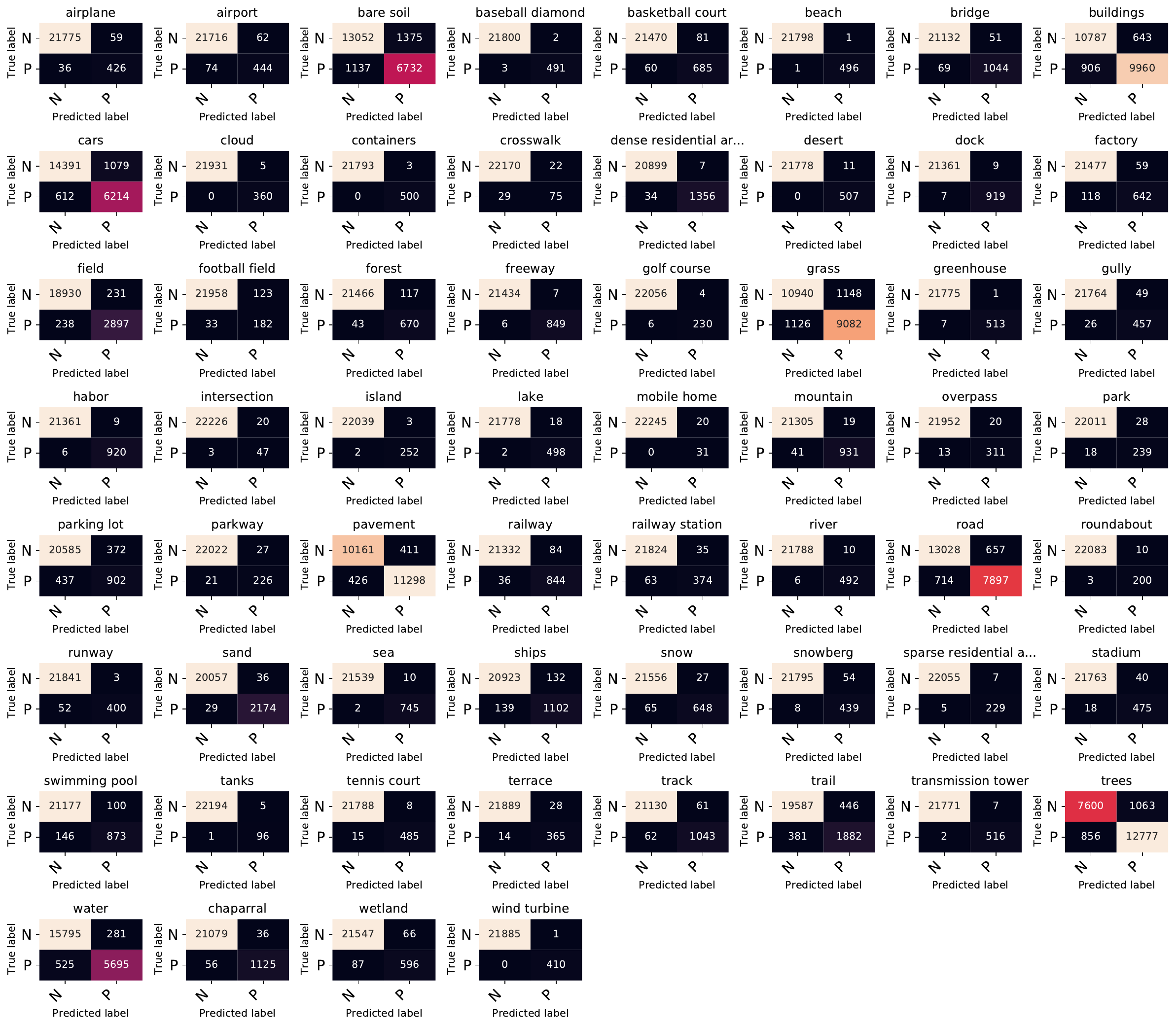}
  \caption{Confusion matrix for the pre-trained Swin Transformer model on the MLRSNet dataset.}
  \label{fig:mlrsnet_confusionmatrix}
\end{figure}

Detailed results for all pre-trained models are shown on Table~\ref{tab:pre-trained_mlrsnet} and for all the models learned from scratch are presented on Table~\ref{tab:scratch_mlrsnet}. The best performing model is the pre-trained Swin Transformer model. The results on a class level are show on Table~\ref{tab:perclass_mlrsnet} along with a confusion matrix on Figure~\ref{fig:mlrsnet_confusionmatrix}.

\begin{table}[ht]\centering
\caption{Detailed results for pre-trained models on the MLRSNet dataset.}\label{tab:pre-trained_mlrsnet}
\scriptsize \begin{adjustbox}{width=0.75\linewidth}
\begin{tabular}{lrrrrrrrrrrrrrr}\toprule
&\rotatebox{90}{mAP} &\rotatebox{90}{Micro Precision} &\rotatebox{90}{Macro Precision} &\rotatebox{90}{Weighted Precision} &\rotatebox{90}{Micro Recall} &\rotatebox{90}{Macro Recall} &\rotatebox{90}{Weighted Recall} &\rotatebox{90}{Micro F1 score} &\rotatebox{90}{Macro F1 score} &\rotatebox{90}{Weighted F1 score} &\rotatebox{90}{Avg. time / epoch (sec.)} &\rotatebox{90}{Total time (sec.)} &\rotatebox{90}{Best epoch}\\\midrule
AlexNet &93.40 &87.93 &87.37 &88.15 &88.54 &88.95 &88.54 &88.24 &87.73 &88.25 &34.09 &1125 &23\\
VGG16 &94.63 &89.56 &89.05 &89.73 &89.39 &90.06 &89.39 &89.48 &89.18 &89.48 &132.24 &3306 &15\\
ResNet50 &96.27 &91.33 &92.54 &91.38 &90.72 &91.79 &90.72 &91.03 &92.00 &91.00 &101.67 &1726 &16\\
ResNet152 &96.43 &91.83 &92.51 &91.84 &90.74 &92.27 &90.74 &91.28 &92.26 &91.25 &214.11 &5781 &17\\
DenseNet161 &96.31 &91.61 &92.35 &91.63 &90.85 &92.18 &90.85 &91.23 &92.07 &91.21 &237.35 &6171 &16\\
EfficientNetB0 &95.39 &91.35 &91.63 &91.37 &90.09 &90.52 &90.09 &90.71 &90.84 &90.67 &86.80 &2604 &20\\
ConvNeXt &95.81 &91.04 &90.71 &91.12 &90.60 &91.90 &90.60 &90.82 &91.10 &90.81 &155.65 &3580 &13\\
Vision Transformer &96.41 &91.81 &91.89 &91.84 &91.75 &93.16 &91.75 &91.78 &92.33 &91.77 &170.90 &3589 &11\\
MLP Mixer &95.05 &90.77 &91.21 &90.83 &89.14 &89.23 &89.14 &89.95 &89.86 &89.88 &121.38 &1942 &6\\
Swin Transformer &\textbf{96.62} &91.74 &91.91 &91.83 &92.13 &93.65 &92.13 &91.93 &92.60 &91.95 &496.72 &12418 &15 \\
\bottomrule
\end{tabular} \end{adjustbox}
\end{table}

\begin{table}[ht]\centering
\caption{Detailed results for models trained from scratch on the MLRSNet dataset.}\label{tab:scratch_mlrsnet}
\scriptsize \begin{adjustbox}{width=0.75\linewidth}
\begin{tabular}{lrrrrrrrrrrrrrr}\toprule
&\rotatebox{90}{mAP} &\rotatebox{90}{Micro Precision} &\rotatebox{90}{Macro Precision} &\rotatebox{90}{Weighted Precision} &\rotatebox{90}{Micro Recall} &\rotatebox{90}{Macro Recall} &\rotatebox{90}{Weighted Recall} &\rotatebox{90}{Micro F1 score} &\rotatebox{90}{Macro F1 score} &\rotatebox{90}{Weighted F1 score} &\rotatebox{90}{Avg. time / epoch (sec.)} &\rotatebox{90}{Total time (sec.)} &\rotatebox{90}{Best epoch}\\\midrule
AlexNet &90.85 &86.53 &83.69 &86.69 &86.58 &86.54 &86.58 &86.56 &84.70 &86.58 &34.92 &2549 &58\\
VGG16 &91.52 &86.63 &83.24 &87.00 &87.98 &88.23 &87.98 &87.30 &85.33 &87.41 &132.22 &7272 &40\\
ResNet50 &\textbf{95.26} &90.65 &90.76 &90.68 &89.42 &90.33 &89.42 &90.03 &90.37 &90.00 &102.26 &6238 &46\\
ResNet152 &93.98 &89.47 &88.92 &89.54 &88.45 &88.55 &88.45 &88.96 &88.51 &88.92 &214.47 &14155 &51\\
DenseNet161 &94.74 &90.23 &89.59 &90.23 &88.13 &88.86 &88.13 &89.17 &88.87 &89.08 &237.96 &11422 &33\\
EfficientNetB0 &94.40 &89.90 &89.09 &89.99 &89.22 &90.19 &89.22 &89.56 &89.40 &89.54 &89.34 &8934 &87\\
ConvNeXt &90.71 &87.86 &84.80 &88.00 &85.38 &84.73 &85.38 &86.60 &84.36 &86.60 &159.35 &5896 &22\\
Vision Transformer &87.25 &85.78 &82.28 &85.81 &84.64 &80.90 &84.64 &85.20 &81.06 &85.03 &170.71 &5975 &20\\
MLP Mixer &85.28 &84.45 &82.59 &84.45 &82.19 &75.60 &82.19 &83.31 &78.11 &83.01 &123.20 &3080 &10\\
Swin Transformer &94.10 &89.56 &88.56 &89.65 &89.93 &90.17 &89.93 &89.74 &89.11 &89.73 &482.72 &42962 &74 \\
\bottomrule
\end{tabular} \end{adjustbox}
\end{table}

\begin{table}[ht]\centering
\caption{Per label results for the pre-trained Swin Transformer model on the MLRSNet dataset.}\label{tab:perclass_mlrsnet}
\scriptsize
\begin{tabular}{lrrrr}\toprule
Label &Precision &Recall &F1 score \\\midrule
airplane &87.84 &92.21 &89.97 \\
airport &87.75 &85.71 &86.72 \\
bare soil &83.04 &85.55 &84.28 \\
baseball diamond &99.59 &99.39 &99.49 \\
basketball court &89.43 &91.95 &90.67 \\
beach &99.80 &99.80 &99.80 \\
bridge &95.34 &93.80 &94.57 \\
buildings &93.94 &91.66 &92.78 \\
cars &85.20 &91.03 &88.02 \\
cloud &98.63 &100.00 &99.31 \\
containers &99.40 &100.00 &99.70 \\
crosswalk &77.32 &72.12 &74.63 \\
dense residential area &99.49 &97.55 &98.51 \\
desert &97.88 &100.00 &98.93 \\
dock &99.03 &99.24 &99.14 \\
factory &91.58 &84.47 &87.89 \\
field &92.62 &92.41 &92.51 \\
football field &59.67 &84.65 &70.00 \\
forest &85.13 &93.97 &89.33 \\
freeway &99.18 &99.30 &99.24 \\
golf course &98.29 &97.46 &97.87 \\
grass &88.78 &88.97 &88.87 \\
greenhouse &99.81 &98.65 &99.23 \\
gully &90.32 &94.62 &92.42 \\
habor &99.03 &99.35 &99.19 \\
intersection &70.15 &94.00 &80.34 \\
island &98.82 &99.21 &99.02 \\
lake &96.51 &99.60 &98.03 \\
mobile home &60.78 &100.00 &75.61 \\
mountain &98.00 &95.78 &96.88 \\
\bottomrule
\end{tabular}
\begin{tabular}{lrrrr}\toprule
Label &Precision &Recall &F1 score \\\midrule
overpass &93.96 &95.99 &94.96 \\
park &89.51 &93.00 &91.22 \\
parking lot &70.80 &67.36 &69.04 \\
parkway &89.33 &91.50 &90.40 \\
pavement &96.49 &96.37 &96.43 \\
railway &90.95 &95.91 &93.36 \\
railway station &91.44 &85.58 &88.42 \\
river &98.01 &98.80 &98.40 \\
road &92.32 &91.71 &92.01 \\
roundabout &95.24 &98.52 &96.85 \\
runway &99.26 &88.50 &93.57 \\
sand &98.37 &98.68 &98.53 \\
sea &98.68 &99.73 &99.20 \\
ships &89.30 &88.80 &89.05 \\
snow &96.00 &90.88 &93.37 \\
snowberg &89.05 &98.21 &93.40 \\
sparse residential area &97.03 &97.86 &97.45 \\
stadium &92.23 &96.35 &94.25 \\
swimming pool &89.72 &85.67 &87.65 \\
tanks &95.05 &98.97 &96.97 \\
tennis court &98.38 &97.00 &97.68 \\
terrace &92.88 &96.31 &94.56 \\
track &94.47 &94.39 &94.43 \\
trail &80.84 &83.16 &81.99 \\
transmission tower &98.66 &99.61 &99.14 \\
trees &92.32 &93.72 &93.01 \\
water &95.30 &91.56 &93.39 \\
chaparral &96.90 &95.26 &96.07 \\
wetland &90.03 &87.26 &88.62 \\
wind turbine &99.76 &100.00 &99.88 \\
\bottomrule
\end{tabular}
\end{table}

\clearpage

\subsection{DFC15}

\begin{figure}[ht] \centering  \includegraphics[width=0.7\linewidth]{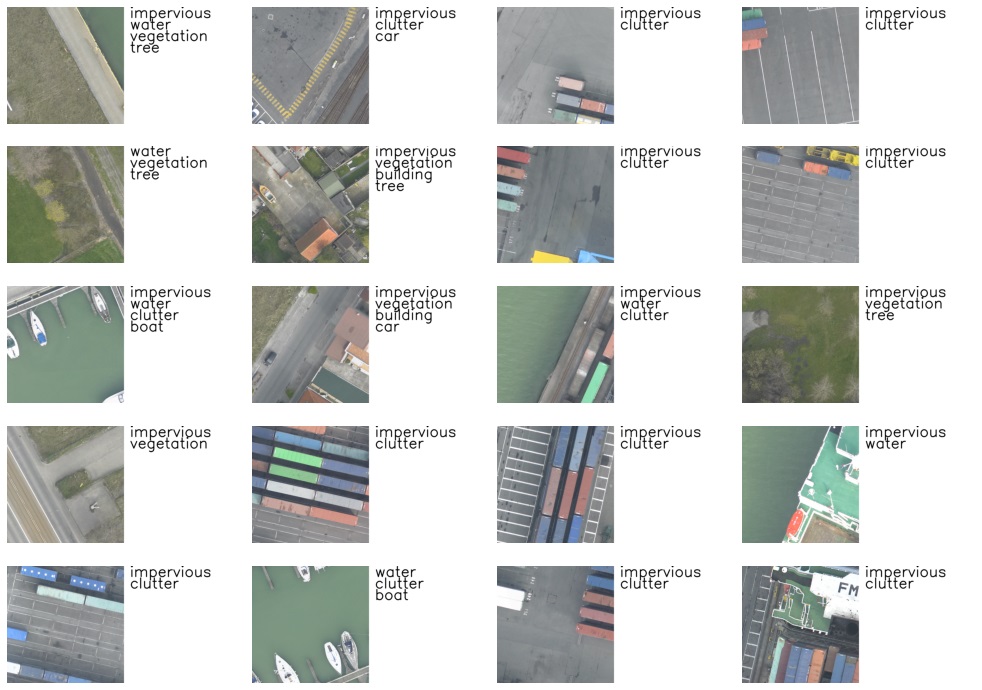}
  \caption{Example images with labels from the DFC15 dataset.}
  \label{fig:dfc15_multilabel_samples}
\end{figure}

\begin{figure}[ht]
  \centering
  \includegraphics[width=0.5\linewidth]{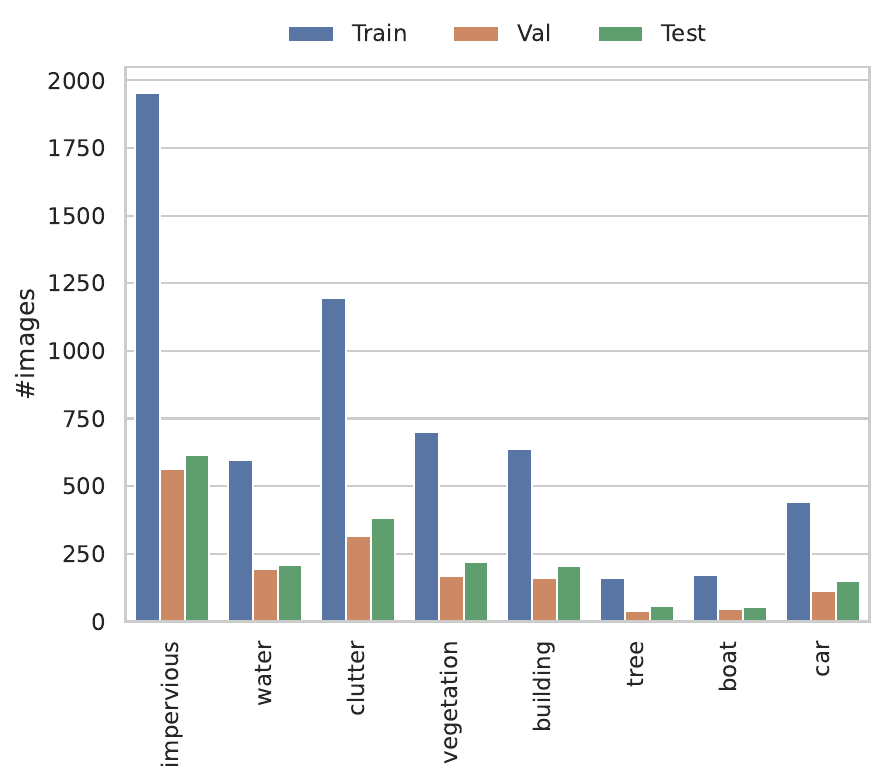}
  \caption{Label distribution for the DFC15 dataset.}
  \label{fig:dfc15_multilabel_distribution}
\end{figure}

DFC15 \citep{hua2019dfc15} is a multi-label dataset created from the semantic segmentation dataset, DFC15 (IEEE GRSS data fusion contest, 2015), which was published and first used in 2015 IEEE GRSS Data Fusion Contest. The dataset is acquired over Zeebrugge with an airborne sensor, which is 300m off the ground. In total, 7 tiles are collected in DFC dataset, and each of them is pixels with a spatial resolution of 5cm. All tiles in DFC15 dataset are labeled in pixel-level, and each pixel is categorized into 8 distinct object classes: impervious, water, clutter, vegetation, building, tree, boat, and car. As a result of this process, the dataset contains 3342 images with a size of 600x600 pixels (Figure~\ref{fig:dfc15_multilabel_samples}). The images are annotated with one or more of the 8 labels in the dataset, with an average of 2.8 labels per image (Figure~\ref{fig:dfc15_multilabel_distribution}). The most frequent labels is \textit{impervious} and it appears in 3133 image. The label \textit{tree} is least frequent and it appears in 258 images. 

\begin{figure}[ht]
  \centering
  \includegraphics[width=0.4\linewidth]{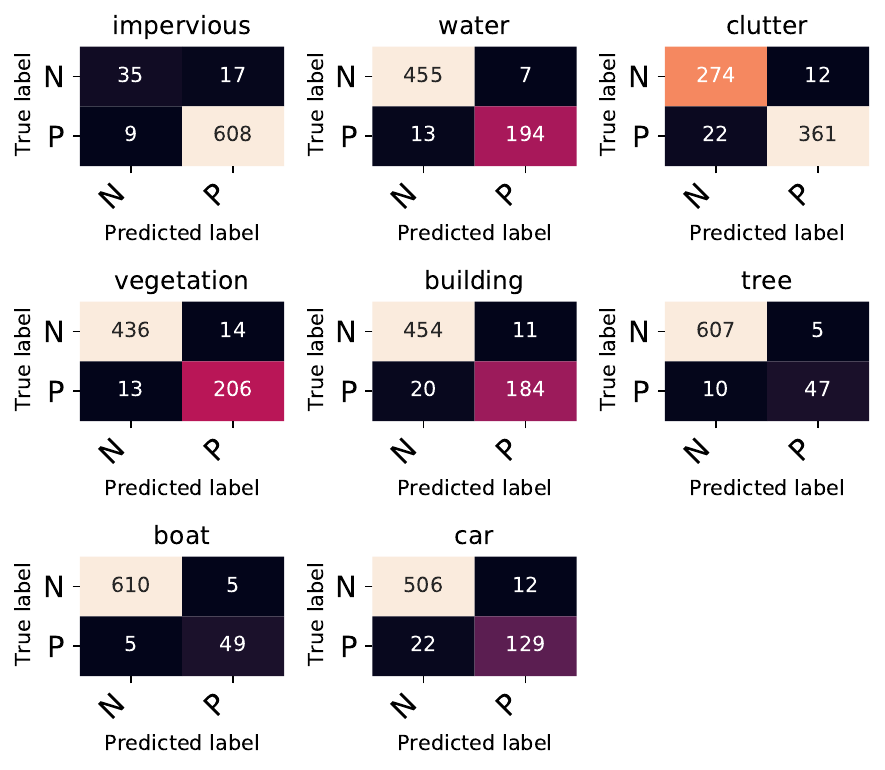}
  \caption{Confusion matrix for the pre-trained Swin Transformer model on the DFC15 dataset.}
  \label{fig:dfc15_confusionmatrix}
\end{figure}

Detailed results for all pre-trained models are shown on Table~\ref{tab:pre-trained_dfc15_multilabel} and for all the models learned from scratch are presented on Table~\ref{tab:scratch_dfc15_multilabel}. The best performing model is the pre-trained Swin Transformer model. The results on a class level are show on Table~\ref{tab:perclass_dfc15_multilabel} along with a confusion matrix on Figure~\ref{fig:dfc15_confusionmatrix}.

\begin{table}[ht]\centering
\caption{Detailed results for pre-trained models on the DFC15 dataset.}\label{tab:pre-trained_dfc15_multilabel}
\scriptsize \begin{adjustbox}{width=0.75\linewidth}
\begin{tabular}{lrrrrrrrrrrrrrr}\toprule
&\rotatebox{90}{mAP} &\rotatebox{90}{Micro Precision} &\rotatebox{90}{Macro Precision} &\rotatebox{90}{Weighted Precision} &\rotatebox{90}{Micro Recall} &\rotatebox{90}{Macro Recall} &\rotatebox{90}{Weighted Recall} &\rotatebox{90}{Micro F1 score} &\rotatebox{90}{Macro F1 score} &\rotatebox{90}{Weighted F1 score} &\rotatebox{90}{Avg. time / epoch (sec.)} &\rotatebox{90}{Total time (sec.)} &\rotatebox{90}{Best epoch}\\\midrule
AlexNet &94.06 &92.33 &89.03 &92.32 &91.01 &86.21 &91.01 &91.67 &87.52 &91.60 &7.74 &325 &32\\
VGG16 &96.57 &94.09 &91.75 &94.30 &92.60 &88.57 &92.60 &93.34 &89.79 &93.30 &8.94 &286 &22\\
ResNet50 &97.66 &95.21 &94.19 &95.19 &93.50 &91.54 &93.50 &94.35 &92.81 &94.31 &8.49 &331 &29\\
ResNet152 &97.60 &95.08 &93.78 &95.04 &93.97 &90.88 &93.97 &94.52 &92.25 &94.46 &9.45 &444 &37\\
DenseNet161 &97.53 &95.07 &93.52 &95.03 &94.71 &91.43 &94.71 &94.89 &92.43 &94.85 &9.54 &544 &47\\
EfficientNetB0 &96.79 &95.54 &94.09 &95.51 &94.08 &90.97 &94.08 &94.81 &92.48 &94.77 &8.33 &583 &60\\
ConvNeXt &97.99 &94.99 &93.84 &94.98 &94.24 &91.39 &94.24 &94.61 &92.55 &94.58 &8.72 &471 &44\\
Vision Transformer &97.62 &96.40 &94.75 &96.33 &93.34 &89.45 &93.34 &94.84 &91.96 &94.77 &8.76 &219 &15\\
MLP Mixer &97.94 &95.23 &94.29 &95.20 &93.92 &90.82 &93.92 &94.57 &92.48 &94.53 &8.18 &229 &18\\
Swin Transformer &\textbf{98.11} &95.54 &93.90 &95.50 &93.97 &91.18 &93.97 &94.75 &92.49 &94.71 &17.59 &686 &29 \\
\bottomrule
\end{tabular} \end{adjustbox}
\end{table}

\begin{table}[ht]\centering
\caption{Detailed results for models trained from scratch on the DFC15 dataset.}\label{tab:scratch_dfc15_multilabel}
\scriptsize \begin{adjustbox}{width=0.75\linewidth}
\begin{tabular}{lrrrrrrrrrrrrrr}\toprule
&\rotatebox{90}{mAP} &\rotatebox{90}{Micro Precision} &\rotatebox{90}{Macro Precision} &\rotatebox{90}{Weighted Precision} &\rotatebox{90}{Micro Recall} &\rotatebox{90}{Macro Recall} &\rotatebox{90}{Weighted Recall} &\rotatebox{90}{Micro F1 score} &\rotatebox{90}{Macro F1 score} &\rotatebox{90}{Weighted F1 score} &\rotatebox{90}{Avg. time / epoch (sec.)} &\rotatebox{90}{Total time (sec.)} &\rotatebox{90}{Best epoch}\\\midrule
AlexNet &88.10 &90.40 &84.01 &90.16 &85.57 &76.29 &85.57 &87.92 &79.75 &87.69 &7.83 &783 &99\\
VGG16 &89.87 &91.07 &86.30 &91.03 &87.37 &79.82 &87.37 &89.18 &82.38 &88.98 &8.50 &799 &79\\
ResNet50 &94.67 &92.88 &89.33 &92.84 &91.75 &87.01 &91.75 &92.32 &88.11 &92.26 &8.92 &464 &37\\
ResNet152 &94.19 &92.05 &89.36 &91.91 &89.96 &83.80 &89.96 &90.99 &86.36 &90.82 &9.66 &647 &52\\
DenseNet161 &\textbf{95.85} &94.23 &92.10 &94.19 &92.28 &87.62 &92.28 &93.24 &89.65 &93.15 &9.89 &613 &47\\
EfficientNetB0 &93.97 &93.90 &91.67 &93.77 &91.91 &85.64 &91.91 &92.90 &88.40 &92.75 &8.47 &686 &66\\
ConvNeXt &89.56 &91.08 &87.12 &90.85 &87.47 &79.56 &87.47 &89.24 &82.99 &89.03 &8.80 &880 &91\\
Vision Transformer &94.16 &92.45 &89.36 &92.34 &89.96 &84.84 &89.96 &91.19 &87.00 &91.10 &8.85 &743 &69\\
MLP Mixer &91.66 &90.43 &86.00 &90.27 &88.90 &82.91 &88.90 &89.66 &84.40 &89.56 &8.31 &831 &100\\
Swin Transformer &94.35 &93.47 &91.02 &93.43 &90.80 &84.66 &90.80 &92.12 &87.56 &91.97 &17.09 &1709 &95 \\
\bottomrule
\end{tabular} \end{adjustbox}
\end{table}

\begin{table}[ht]\centering
\caption{Per label results for the pre-trained Swin Transfromer model on the DFC15 dataset.}\label{tab:perclass_dfc15_multilabel}
\scriptsize
\begin{tabular}{lrrrr}\toprule
Label &Precision &Recall &F1 score \\\midrule
impervious &97.28 &98.54 &97.91 \\
water &96.52 &93.72 &95.10 \\
clutter &96.78 &94.26 &95.50 \\
vegetation &93.64 &94.06 &93.85 \\
building &94.36 &90.20 &92.23 \\
tree &90.38 &82.46 &86.24 \\
boat &90.74 &90.74 &90.74 \\
car &91.49 &85.43 &88.36 \\
\bottomrule
\end{tabular}
\end{table}
%

\clearpage

\subsection{Planet UAS}

\begin{figure}[ht] \centering  \includegraphics[width=0.7\linewidth]{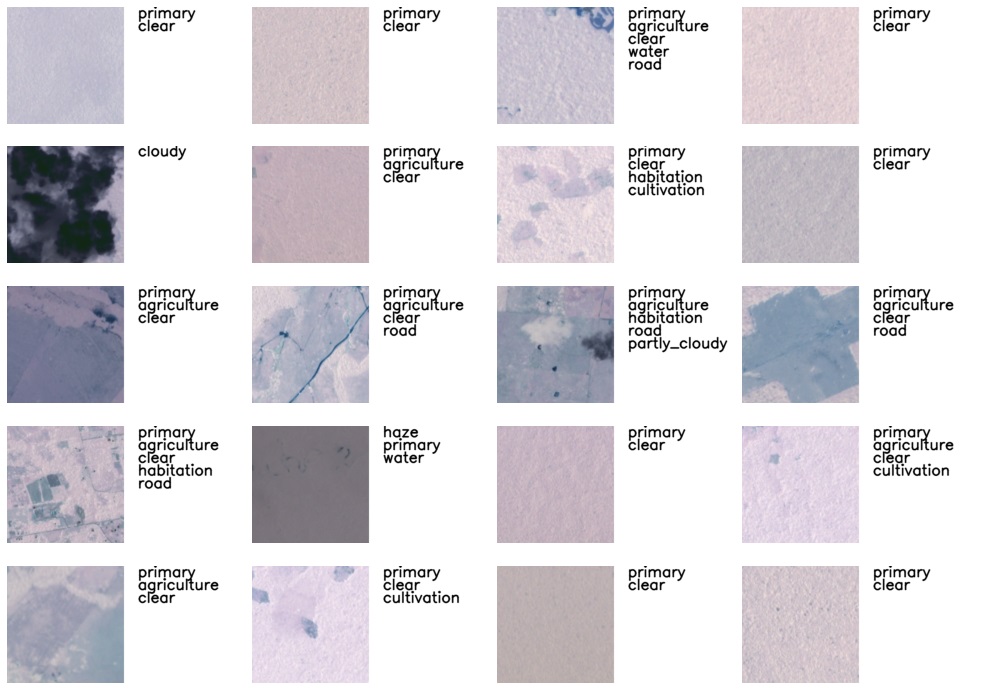}
  \caption{Example images with labels from the Planet UAS dataset.}
  \label{fig:planetuas_multilabel_samples}
\end{figure}

The Planet UAS dataset \citep{webplanetuas} was created by the company, Planet - designer and builder of the world’s largest constellation of Earth-imaging satellites. The aim is to label satellite image chips with atmospheric conditions and various classes of land cover/land use. The dataset is available on Kaggle and is approximately 32 GB worth of data. The data contains 40479 satellite images organized in tiff and jpg files (Figure~\ref{fig:planetuas_multilabel_samples}). The jpg file show the natural light spectrum of the image, whereas the tiff files provide extra information about the infrared features of the satellite image, both with 256x256 pixels resolution. There are a total of 17 different labels with an average of 2.9 labels per image. 

The imagery has a ground-sample distance (GSD) of 3.7m and an orthorectified pixel size of 3m. The data comes from Planet's Flock 2 satellites in both sun-synchronous and ISS orbits and was collected between January 1, 2016 and February 1, 2017. All of the scenes come from the Amazon basin which includes Brazil, Peru, Uruguay, Colombia, Venezuela, Guyana, Bolivia, and Ecuador. There are a total of 17 different labels. Out of those, 4 labels correspond to weather: Clear, Cloudy, Partly Cloudy, Haze. The rest of the (13) labels correspond to land: Habitation, Bare Ground, Cultivation, Agriculture, Blow Down, Conventional Mine, Selective Logging, Slash Burn, Artisanal Mine, Blooming, Primary, Water, and None.

\begin{figure}[ht]
  \centering
  \includegraphics[width=0.6\linewidth]{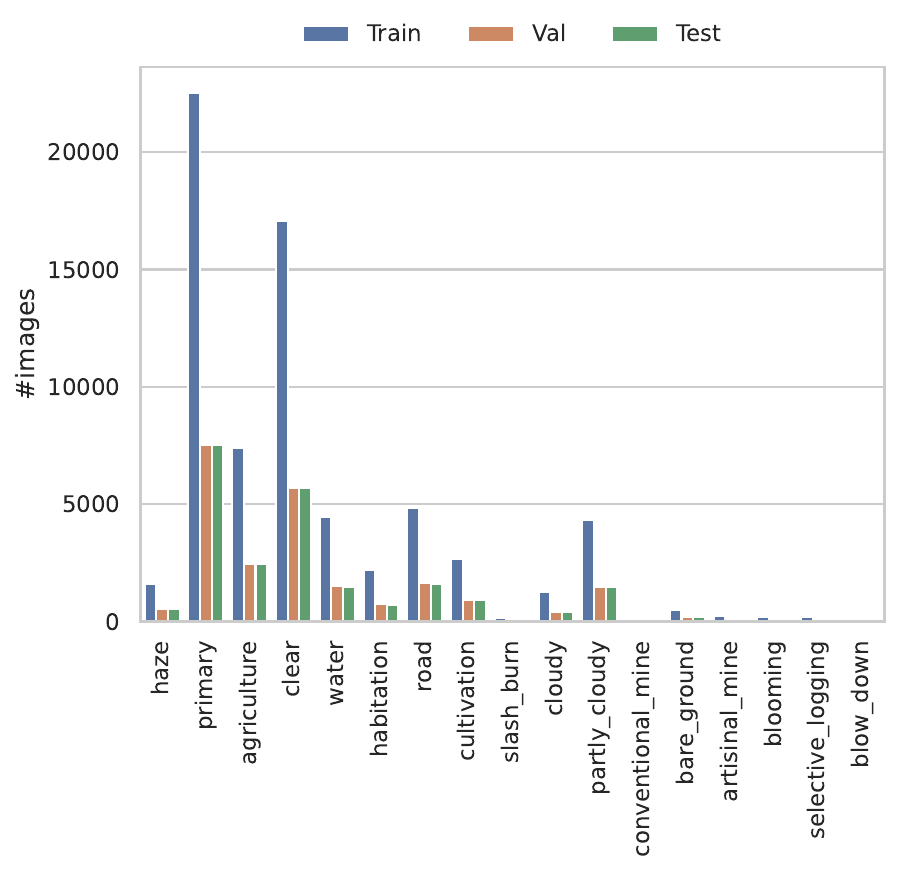}
  \caption{Label distribution for the PlanetUAS dataset.}
  \label{fig:planetuas_multilabel_distribution}
\end{figure}

The dataset only has the train set publicly available and we use that to generate train, test and validation splits (Figure~\ref{fig:planetuas_multilabel_distribution}). 

\begin{figure}[ht]
  \centering
  \includegraphics[width=0.5\linewidth]{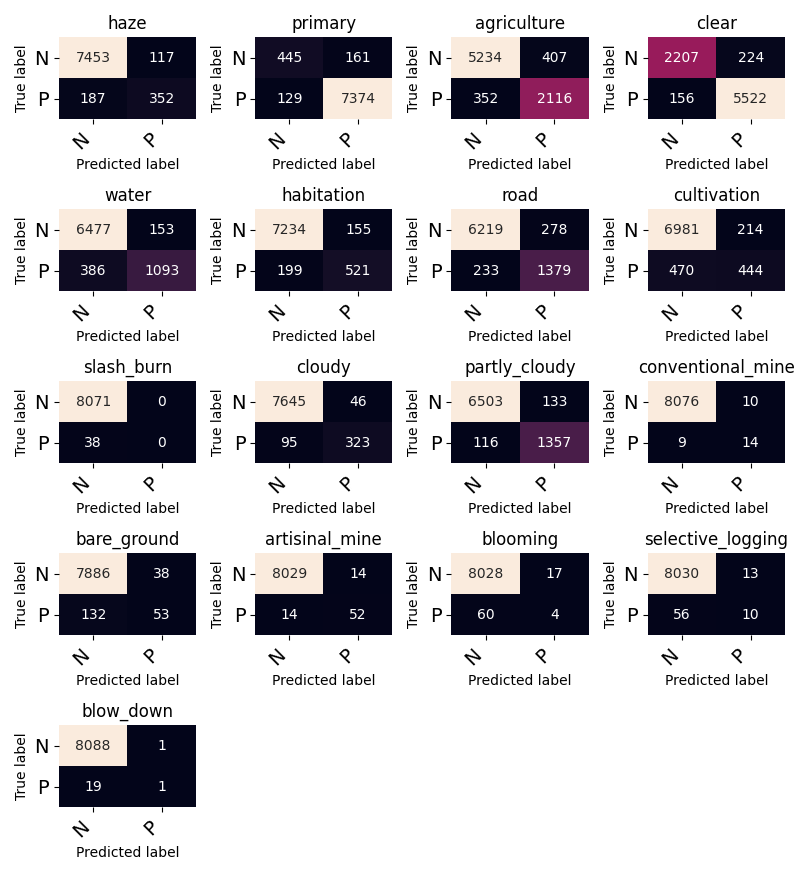}
  \caption{Confusion matrix for the pre-trained Swin Transformer model on the PlanetUAS dataset.}
  \label{fig:planetuas_confusionmatrix}
\end{figure}

Detailed results for all pre-trained models are shown on Table~\ref{tab:pre-trained_planetuas_multilabel} and for all the models learned from scratch are presented on Table~\ref{tab:scratch_planetuas_multilabel}. The best performing model is the pre-trained Swin Transformer model. The results on a class level are show on Table~\ref{tab:perclass_planetuas_multilabel} along with a confusion matrix on Figure~\ref{fig:planetuas_confusionmatrix}.

\begin{table}[ht]\centering
\caption{Detailed results for pre-trained models on the PlanetUAS dataset.}\label{tab:pre-trained_planetuas_multilabel}
\scriptsize \begin{adjustbox}{width=0.75\linewidth}
\begin{tabular}{lrrrrrrrrrrrrrr}\toprule
&\rotatebox{90}{mAP} &\rotatebox{90}{Micro Precision} &\rotatebox{90}{Macro Precision} &\rotatebox{90}{Weighted Precision} &\rotatebox{90}{Micro Recall} &\rotatebox{90}{Macro Recall} &\rotatebox{90}{Weighted Recall} &\rotatebox{90}{Micro F1 score} &\rotatebox{90}{Macro F1 score} &\rotatebox{90}{Weighted F1 score} &\rotatebox{90}{Avg. time / epoch (sec.)} &\rotatebox{90}{Total time (sec.)} &\rotatebox{90}{Best epoch}\\\midrule
AlexNet &64.05 &90.71 &66.56 &89.39 &86.29 &54.39 &86.29 &88.44 &57.73 &87.49 &17.45 &576 &23\\
VGG16 &65.58 &92.09 &64.14 &90.90 &86.88 &55.99 &86.88 &89.41 &59.19 &88.58 &50.38 &1058 &11\\
ResNet50 &65.53 &92.17 &67.64 &90.91 &86.07 &54.98 &86.07 &89.02 &58.72 &87.91 &37.00 &740 &10\\
ResNet152 &64.82 &91.66 &66.67 &90.52 &87.23 &56.03 &87.23 &89.39 &59.47 &88.60 &81.83 &1964 &14\\
DenseNet161 &66.34 &91.75 &73.56 &90.77 &87.42 &55.29 &87.42 &89.53 &59.16 &88.50 &90.40 &1808 &10\\
EfficientNetB0 &64.16 &92.18 &69.45 &90.98 &87.18 &52.66 &87.18 &89.61 &56.02 &88.62 &33.52 &771 &13\\
ConvNeXt &66.45 &91.52 &70.00 &90.47 &87.95 &56.06 &87.95 &89.70 &59.95 &88.92 &59.63 &1431 &14\\
Vision Transformer &66.80 &91.31 &69.63 &90.18 &87.79 &56.11 &87.79 &89.52 &59.95 &88.56 &65.71 &920 &4\\
MLP Mixer &67.33 &92.18 &74.70 &91.30 &86.68 &56.56 &86.68 &89.35 &60.79 &88.59 &45.94 &735 &6\\
Swin Transformer &\textbf{67.84} &91.23 &67.93 &90.33 &88.61 &58.30 &88.61 &89.90 &60.97 &89.27 &180.89 &3256 &8 \\
\bottomrule
\end{tabular} \end{adjustbox}
\end{table}

\begin{table}[ht]\centering
\caption{Detailed results for models trained from scratch on the PlanetUAS dataset.}\label{tab:scratch_planetuas_multilabel}
\scriptsize \begin{adjustbox}{width=0.75\linewidth}
\begin{tabular}{lrrrrrrrrrrrrrr}\toprule
&\rotatebox{90}{mAP} &\rotatebox{90}{Micro Precision} &\rotatebox{90}{Macro Precision} &\rotatebox{90}{Weighted Precision} &\rotatebox{90}{Micro Recall} &\rotatebox{90}{Macro Recall} &\rotatebox{90}{Weighted Recall} &\rotatebox{90}{Micro F1 score} &\rotatebox{90}{Macro F1 score} &\rotatebox{90}{Weighted F1 score} &\rotatebox{90}{Avg. time / epoch (sec.)} &\rotatebox{90}{Total time (sec.)} &\rotatebox{90}{Best epoch}\\\midrule
AlexNet &60.28 &90.32 &67.35 &88.88 &84.81 &51.24 &84.81 &87.48 &54.52 &86.25 &18.65 &1865 &87\\
VGG16 &60.68 &90.39 &60.11 &88.74 &84.97 &50.56 &84.97 &87.60 &53.21 &86.44 &50.68 &2889 &42\\
ResNet50 &64.19 &92.16 &67.02 &90.84 &86.52 &54.31 &86.52 &89.25 &58.47 &88.24 &37.57 &2592 &54\\
ResNet152 &64.96 &91.57 &69.94 &90.42 &86.97 &55.02 &86.97 &89.21 &59.06 &88.28 &80.86 &6792 &69\\
DenseNet161 &64.74 &91.79 &69.52 &90.53 &87.01 &55.20 &87.01 &89.34 &59.12 &88.37 &90.11 &4866 &39\\
EfficientNetB0 &63.87 &91.70 &65.64 &90.55 &87.03 &53.86 &87.03 &89.30 &57.21 &88.40 &33.47 &2711 &66\\
ConvNeXt &61.28 &90.92 &64.25 &89.39 &84.29 &51.55 &84.29 &87.48 &54.68 &86.19 &59.35 &5935 &90\\
Vision Transformer &59.41 &90.35 &60.32 &88.16 &83.12 &47.68 &83.12 &86.58 &51.94 &84.94 &65.52 &4128 &48\\
MLP Mixer &58.55 &89.67 &62.22 &87.58 &82.21 &48.88 &82.21 &85.78 &51.46 &84.06 &45.93 &2572 &41\\
Swin Transformer &\textbf{65.23} &91.53 &66.89 &90.11 &86.34 &54.05 &86.34 &88.86 &57.89 &87.80 &181.83 &16365 &75 \\
\bottomrule
\end{tabular} \end{adjustbox}
\end{table}

\begin{table}[ht]\centering
\caption{Per label results for the pre-trained Swin Transformer model on the PlanetUAS dataset.}\label{tab:perclass_planetuas_multilabel}
\scriptsize 
\begin{tabular}{lrrrr}\toprule
Label &Precision &Recall &F1 score \\\midrule
haze &75.05 &65.31 &69.84 \\
primary &97.86 &98.28 &98.07 \\
agriculture &83.87 &85.74 &84.79 \\
clear &96.10 &97.25 &96.67 \\
water &87.72 &73.90 &80.22 \\
habitation &77.07 &72.36 &74.64 \\
road &83.22 &85.55 &84.37 \\
cultivation &67.48 &48.58 &56.49 \\
slash\_burn &0.00 &0.00 &0.00 \\
cloudy &87.53 &77.27 &82.08 \\
partly\_cloudy &91.07 &92.12 &91.60 \\
conventional\_mine &58.33 &60.87 &59.57 \\
bare\_ground &58.24 &28.65 &38.41 \\
artisinal\_mine &78.79 &78.79 &78.79 \\
blooming &19.05 &6.25 &9.41 \\
selective\_logging &43.48 &15.15 &22.47 \\
blow\_down &50.00 &5.00 &9.09 \\
\bottomrule
\end{tabular} 
\end{table}

\clearpage

\subsection{AID multi-label}

\begin{figure}[ht] \centering  \includegraphics[width=0.7\linewidth]{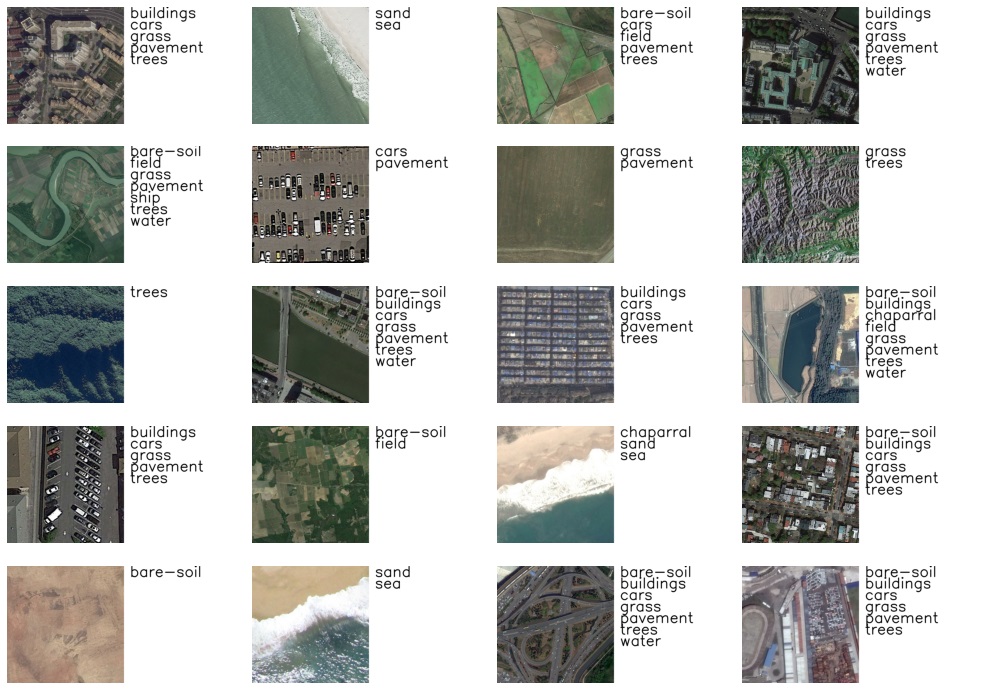}
  \caption{Example images with labels from the AID multilabel dataset.}
  \label{fig:aid_multilabel_samples}
\end{figure}

\begin{figure}[ht]
  \centering
  \includegraphics[width=0.6\linewidth]{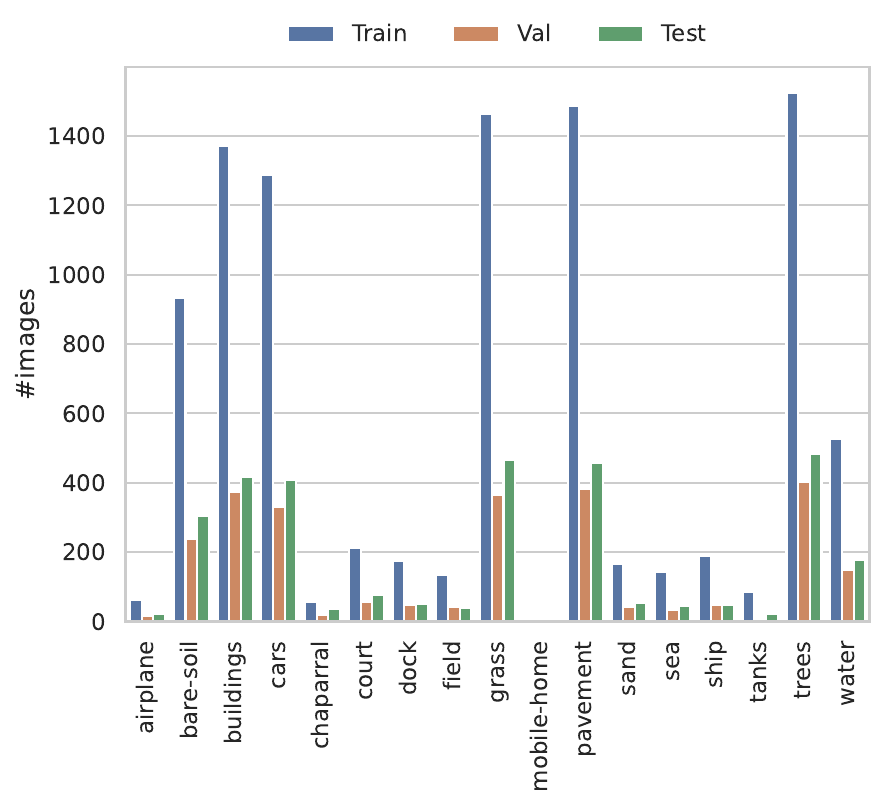}
  \caption{Label distribution for the AID multilabel dataset.}
  \label{fig:aid_multilabel_distribution}
\end{figure}

Hua et al. \citep{hua2020aidmultilabel} extend the AID dataset for multi-label classification. They manually relabeled some images in the AID dataset. With extensive human visual inspections, 3000 aerial images from 30 scenes in the AID dataset were selected and assigned with multiple object labels. The dataset has 17 labels with 5.2 labels per image on average. The labels are: bare soil, airplane, building, car, charparral, court, dock, field, grass, mobile home, pavement, sand, sea, ship, tank, tree and water. The authors provide a proposed train-test split. Figure \ref{fig:aid_multilabel_samples} show some example images from the AID multi-label dataset. The distribution of the labels for the train, validation and test splits is shown in Figure \ref{fig:aid_multilabel_distribution} from which we can observe an imbalanced distribution, some of the labels are heavily populated with images/samples, and some of the labels are with only few images/samples (for example the label mobile-home has only one image in the respective train, validation and test splits).

\begin{figure}[ht]
  \centering
  \includegraphics[width=0.6\linewidth]{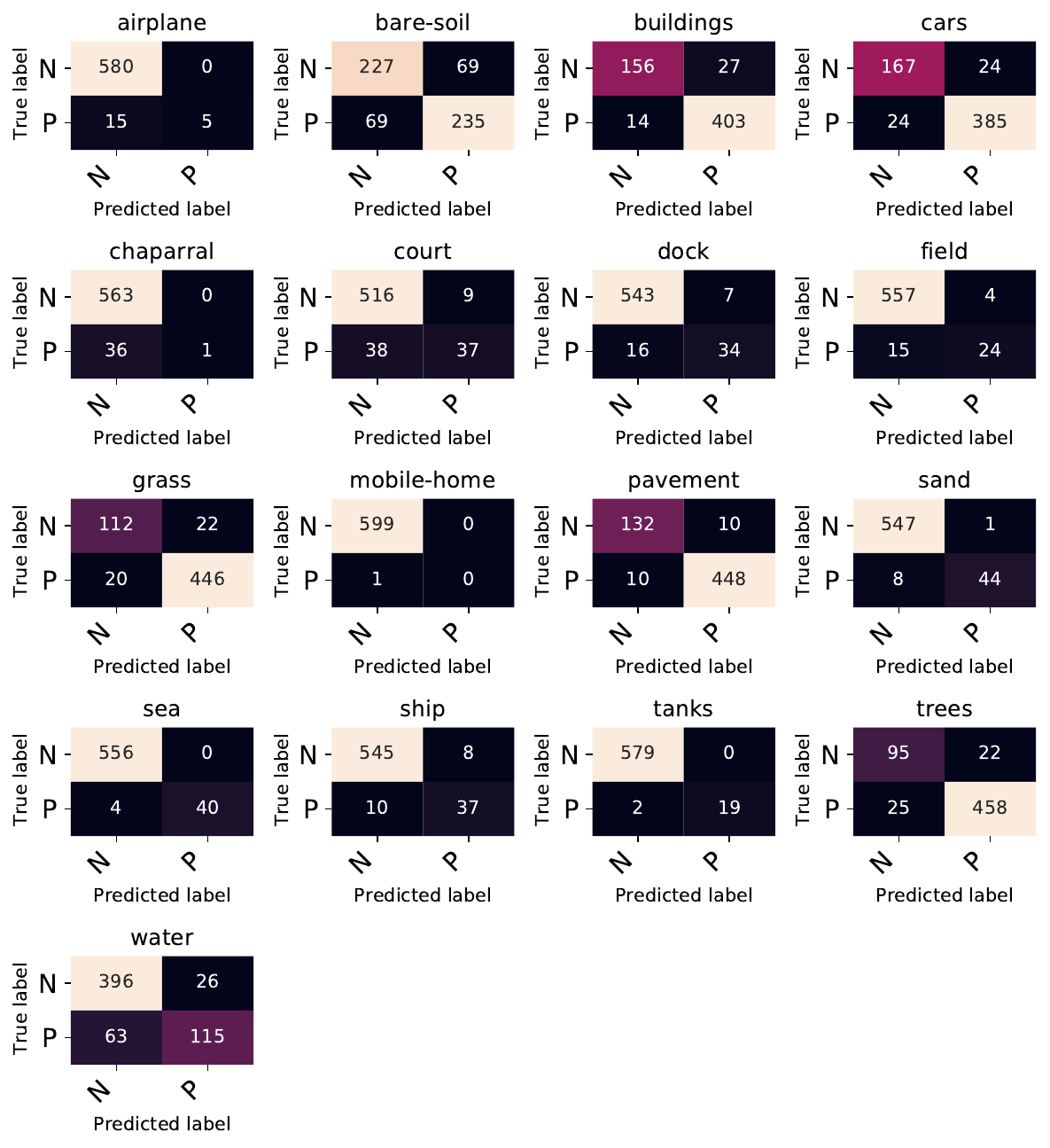}
  \caption{Confusion matrix for the pre-trained ConvNeXt model on the AID multi-label dataset.}
  \label{fig:aid_multilabel_confusionmatrix}
\end{figure}

Detailed results for all pre-trained models are shown on Table~\ref{tab:pre-trained_aid_multilabel} and for all the models learned from scratch are presented on Table~\ref{tab:scratch_aid_multilabel}. The best performing model is the pre-trained ConvNeXt model. The results on a class level are show on Table~\ref{tab:perclass_aid_multilabel} along with a confusion matrix on Figure~\ref{fig:aid_multilabel_confusionmatrix}.

\begin{table}[ht]\centering
\caption{Detailed results for pre-trained models on AID multi-label}\label{tab:pre-trained_aid_multilabel}
\scriptsize \begin{adjustbox}{width=0.75\linewidth}
\begin{tabular}{lrrrrrrrrrrrrrr}\toprule
&\rotatebox{90}{mAP} &\rotatebox{90}{Micro Precision} &\rotatebox{90}{Macro Precision} &\rotatebox{90}{Weighted Precision} &\rotatebox{90}{Micro Recall} &\rotatebox{90}{Macro Recall} &\rotatebox{90}{Weighted Recall} &\rotatebox{90}{Micro F1 score} &\rotatebox{90}{Macro F1 score} &\rotatebox{90}{Weighted F1 score} &\rotatebox{90}{Avg. time / epoch (sec.)} &\rotatebox{90}{Total time (sec.)} &\rotatebox{90}{Best epoch}\\\midrule
AlexNet &75.91 &88.34 &75.10 &87.33 &86.04 &66.15 &86.04 &87.18 &68.61 &86.19 &5.55 &172 &21\\
VGG16 &79.89 &90.12 &76.29 &88.58 &87.62 &67.13 &87.62 &88.85 &70.03 &87.74 &6.33 &190 &20\\
ResNet50 &80.76 &91.36 &79.13 &89.72 &87.68 &68.37 &87.68 &89.48 &72.34 &88.34 &5.94 &190 &22\\
ResNet152 &80.94 &91.92 &80.10 &90.46 &86.62 &64.52 &86.62 &89.19 &69.53 &87.98 &7.97 &239 &20\\
DenseNet161 &81.71 &90.77 &80.12 &89.54 &88.84 &68.22 &88.84 &89.80 &71.80 &88.76 &8.71 &366 &32\\
EfficientNetB0 &78.00 &91.38 &78.79 &89.79 &86.81 &64.22 &86.81 &89.04 &69.40 &87.76 &6.15 &381 &52\\
ConvNeXt &\textbf{82.30} &92.23 &86.10 &92.06 &88.07 &68.96 &88.07 &90.10 &73.01 &89.17 &6.63 &345 &42\\
Vision Transformer &81.54 &93.33 &81.54 &91.76 &87.10 &67.84 &87.10 &90.11 &73.15 &88.96 &6.95 &146 &11\\
MLP Mixer &80.88 &93.09 &85.44 &92.78 &86.88 &64.11 &86.88 &89.87 &69.48 &88.53 &6.35 &165 &16\\
Swin Transformer &82.25 &91.53 &84.73 &91.41 &89.23 &70.89 &89.23 &90.37 &74.04 &89.47 &15.90 &477 &20 \\
\bottomrule
\end{tabular} \end{adjustbox}
\end{table}

\begin{table}[ht]\centering
\caption{Detailed results for models trained from scratch on the AID multi-label dataset.}\label{tab:scratch_aid_multilabel}
\scriptsize \begin{adjustbox}{width=0.75\linewidth}
\begin{tabular}{lrrrrrrrrrrrrrr}\toprule
&\rotatebox{90}{mAP} &\rotatebox{90}{Micro Precision} &\rotatebox{90}{Macro Precision} &\rotatebox{90}{Weighted Precision} &\rotatebox{90}{Micro Recall} &\rotatebox{90}{Macro Recall} &\rotatebox{90}{Weighted Recall} &\rotatebox{90}{Micro F1 score} &\rotatebox{90}{Macro F1 score} &\rotatebox{90}{Weighted F1 score} &\rotatebox{90}{Avg. time / epoch (sec.)} &\rotatebox{90}{Total time (sec.)} &\rotatebox{90}{Best epoch}\\\midrule
AlexNet &68.78 &86.93 &69.98 &85.45 &84.33 &60.45 &84.33 &85.61 &63.48 &84.42 &5.82 &524 &75\\
VGG16 &69.21 &87.03 &66.46 &85.22 &84.42 &62.06 &84.42 &85.71 &63.75 &84.60 &6.28 &490 &63\\
ResNet50 &70.87 &89.52 &74.76 &87.95 &84.04 &59.51 &84.04 &86.69 &64.46 &85.27 &5.74 &379 &51\\
ResNet152 &69.65 &87.95 &76.32 &87.11 &84.49 &58.55 &84.49 &86.18 &62.72 &84.76 &8.08 &477 &44\\
DenseNet161 &71.22 &88.57 &76.27 &87.52 &85.23 &60.19 &85.23 &86.87 &64.09 &85.33 &8.47 &449 &38\\
EfficientNetB0 &\textbf{72.89} &88.51 &71.56 &86.66 &86.42 &64.45 &86.42 &87.45 &67.01 &86.22 &5.94 &398 &52\\
ConvNeXt &65.59 &87.00 &67.56 &85.16 &83.75 &56.43 &83.75 &85.34 &59.44 &83.75 &6.40 &576 &75\\
Vision Transformer &65.58 &85.82 &63.05 &83.61 &83.94 &56.33 &83.94 &84.87 &58.63 &83.37 &6.82 &627 &77\\
MLP Mixer &64.24 &85.72 &66.07 &83.70 &83.46 &56.52 &83.46 &84.58 &59.30 &83.02 &6.41 &506 &64\\
Swin Transformer &69.55 &87.23 &67.48 &85.28 &85.46 &60.09 &85.46 &86.33 &62.57 &84.96 &15.87 &1238 &63 \\
\bottomrule
\end{tabular} \end{adjustbox}
\end{table}

\begin{table}[ht]\centering
\caption{Per label results for the pre-trained ConvNeXt model on the AID multi-label dataset.}\label{tab:perclass_aid_multilabel}
\scriptsize 
\begin{tabular}{lrrrr}\toprule
Label &Precision &Recall &F1 score \\\midrule
airplane &100.00 &25.00 &40.00\\
bare-soil &77.30 &77.30 &77.30\\
buildings &93.72 &96.64 &95.16\\
cars &94.13 &94.13 &94.13\\
chaparral &100.00 &2.70 &5.26\\
court &80.43 &49.33 &61.16\\
dock &82.93 &68.00 &74.73\\
field &85.71 &61.54 &71.64\\
grass &95.30 &95.71 &95.50\\
mobile-home &0.00 &0.00 &0.00\\
pavement &97.82 &97.82 &97.82\\
sand &97.78 &84.62 &90.72\\
sea &100.00 &90.91 &95.24\\
ship &82.22 &78.72 &80.43\\
tanks &100.00 &90.48 &95.00\\
trees &95.42 &94.82 &95.12\\
water &80.99 &64.61 &71.88\\
\bottomrule
\end{tabular} 
\end{table}

\end{document}